\newif\ifislapreprint
\islapreprinttrue
\documentclass{article}

\usepackage[T1]{fontenc}
\usepackage{microtype}
\usepackage{graphicx}
\usepackage{subcaption}
\usepackage{booktabs}
\usepackage{multirow}
\usepackage{array}
\usepackage{enumitem}
\usepackage{xcolor}
\usepackage{hyperref}

\ifdefined\ifislapreprint\else\expandafter\newif\csname ifislapreprint\endcsname\fi
\ifislapreprint
  \usepackage[preprint]{icml2026}
\else
  \usepackage{icml2026}
\fi

\usepackage{amsmath}
\usepackage{amssymb}
\usepackage{mathtools}
\usepackage{amsthm}
\usepackage{bm}
\usepackage{siunitx}
\usepackage{needspace}
\usepackage{placeins}
\usepackage{algorithm}
\usepackage{algorithmic}
\usepackage[capitalize,noabbrev]{cleveref}
\usepackage{xurl}

\theoremstyle{plain}
\newtheorem{theorem}{Theorem}
\newtheorem{proposition}{Proposition}
\newtheorem{lemma}{Lemma}

\theoremstyle{definition}

\newtheorem{assumption}{Assumption}
\crefname{assumption}{Assumption}{Assumptions}
\theoremstyle{remark}

\makeatletter\@secpenalty=0\makeatother
\input{macros}
\isnset{abl-af-ref-halfrange-nut-pct}{6}
\isnset{abl-af-ref-halfrange-p-pct}{4}
\isnset{abl-af-ref-halfrange-u-pct}{2}
\isnset{abl-n-cells-done}{16}
\isnset{abl-ref-all-n-seeds}{4}
\isnset{abl-ref-n-seeds}{4}
\isnset{abl-ref-seed-halfrange-pct}{3.3}
\isnset{abl-s0-pct}{0.2}
\isnset{abl-s1-area}{11}
\isnset{abl-s1-area-abs-pct}{43}
\isnset{abl-s1-n-seeds}{3}
\isnset{abl-s1-native}{0.99}
\isnset{abl-s1-native-abs-pct}{4.0}
\isnset{abl-s1-native-change-pct}{1}
\isnset{abl-s1-native-change-pct1}{0.5}
\isnset{abl-s1-posed}{0.99}
\isnset{abl-s1-posed-abs-pct}{4.0}
\isnset{abl-s2-area}{0.97}
\isnset{abl-s2-area-abs-pct}{3.9}
\isnset{abl-s2-n-seeds}{3}
\isnset{abl-s2-native}{0.96}
\isnset{abl-s2-native-abs-pct}{3.9}
\isnset{abl-s2-native-change-pct}{4}
\isnset{abl-s2-native-change-pct1}{3.6}
\isnset{abl-s2-posed}{21}
\isnset{abl-s2-posed-abs-pct}{83}
\isnset{abl-s3-area}{18}
\isnset{abl-s3-area-abs-pct}{71}
\isnset{abl-s3-n-seeds}{3}
\isnset{abl-s3-native}{18}
\isnset{abl-s3-native-abs-pct}{71}
\isnset{abl-s3-posed}{18}
\isnset{abl-s3-posed-abs-pct}{71}
\isnset{abl-s4-area}{0.98}
\isnset{abl-s4-area-abs-pct}{4.0}
\isnset{abl-s4-n-seeds}{2}
\isnset{abl-s4-native}{0.98}
\isnset{abl-s4-native-abs-pct}{3.9}
\isnset{abl-s4-native-change-pct}{2}
\isnset{abl-s4-native-change-pct1}{2.0}
\isnset{abl-s4-posed}{0.98}
\isnset{abl-s4-posed-abs-pct}{3.9}
\isnset{abl-s4-vs-ref-all-pct}{2}
\isnset{abl-s4-within-ref-all}{1}
\isnset{abl-th0-diff-max-pct}{0.9}
\isnset{abl-th0-mem-gib}{75}
\isnset{abl-th0-n-seeds}{2}
\isnset{abl-th0-native}{1.01}
\isnset{abl-th0-native-abs-pct}{0.4}
\isnset{abl-th0-posed}{1.01}
\isnset{abl-th0-posed-abs-pct}{0.4}
\isnset{abl-th0-ref-mem-gib}{26}
\isnset{abl-th0-step-ratio}{2.7}
\isnset{abl-theta0-ratio-max}{5.1}
\isnset{abl-theta0-ratio-min}{2.0}
\isnset{abl-theta0-ratio-nut}{5.1}
\isnset{abl-theta0-ratio-p}{2.3}
\isnset{abl-theta0-ratio-u}{2.0}
\isnset{abl-theta05-ratio-max}{2.4}
\isnset{abl-theta05-ratio-min}{1.5}
\isnset{abl-theta05-ratio-nut}{2.4}
\isnset{abl-theta05-ratio-p}{1.7}
\isnset{abl-theta05-ratio-u}{1.5}
\isnset{abl-v1-change-nut-pct}{8}
\isnset{abl-v1-change-p-pct}{8}
\isnset{abl-v1-change-u-pct}{10}
\isnset{abl-v1-n-seeds}{2}
\isnset{abl-v1-native}{1.10}
\isnset{abl-v1-native-abs-pct}{0.5}
\isnset{abl-v1-posed}{1.10}
\isnset{abl-v1-posed-abs-pct}{0.5}
\isnset{abl-v1-seed-nut-max}{0.93}
\isnset{abl-v1-seed-nut-min}{0.92}
\isnset{abl-v1-seed-p-max}{1.15}
\isnset{abl-v1-seed-p-min}{1.01}
\isnset{abl-v1-seed-u-max}{1.13}
\isnset{abl-v1-seed-u-min}{1.07}
\isnset{af-aoa-isla-over-globe-max}{1.4}
\isnset{af-aoa-isla-over-globe-min}{1.2}
\isnset{af-aoa-rise-max}{15}
\isnset{af-aoa-rise-min}{4}
\isnset{af-aoa-test-cd-r2-geotransolver}{0.35}
\isnset{af-aoa-test-cd-r2-globe}{0.79}
\isnset{af-aoa-test-cd-r2-isla}{0.65}
\isnset{af-aoa-test-cd-r2-transolver}{-0.07}
\isnset{af-aoa-tgt-over-isla-max}{3.2}
\isnset{af-aoa-tgt-over-isla-min}{2.4}
\isnset{af-aoa-val-cd-r2-geotransolver}{0.79}
\isnset{af-aoa-val-cd-r2-globe}{0.99}
\isnset{af-aoa-val-cd-r2-isla}{0.95}
\isnset{af-aoa-val-cd-r2-transolver}{0.83}
\isnset{af-full-isla-over-globe-max}{2.0}
\isnset{af-full-isla-over-globe-min}{1.0}
\isnset{af-full-tgt-over-isla-max}{2.6}
\isnset{af-full-tgt-over-isla-min}{1.6}
\isnset{ahm-shift-high-drag-max}{3.1}
\isnset{ahm-shift-high-drag-min}{1.6}
\isnset{ahm-shift-image-wake-max}{11.7}
\isnset{ahm-shift-image-wake-min}{3.0}
\isnset{app-airfrans-aoa}{$-2.06^\circ$}
\isnset{app-airfrans-err-pressure-pct}{0.66}
\isnset{app-airfrans-err-velocity-pct}{0.31}
\isnset{app-airfrans-grid-nx}{1{,}000}
\isnset{app-airfrans-grid-ny}{600}
\isnset{app-airfrans-inlet-speed}{44.6}
\isnset{app-airfrans-n-nodes}{100{,}000}
\isnset{app-airfrans-reynolds}{$2.86\times10^{6}$}
\isnset{app-routing-layer-1-n-sharing}{three}
\isnset{app-routing-layer-1-weight-max}{0.34}
\isnset{app-routing-layer-1-weight-min}{0.32}
\isnset{app-routing-layer-2-top-share-pct}{98}
\isnset{app-routing-n-coloured}{eight}
\isnset{app-routing-n-layers}{12}
\isnset{app-routing-n-slices}{128}
\isnset{app-routing-n-slices-layer-1}{1}
\isnset{app-routing-n-slices-layer-10}{124}
\isnset{app-routing-n-slices-layer-11}{126}
\isnset{app-routing-n-slices-layer-12}{124}
\isnset{app-routing-n-slices-layer-2}{1}
\isnset{app-routing-n-slices-layer-3}{5}
\isnset{app-routing-n-slices-layer-4}{33}
\isnset{app-routing-n-slices-layer-5}{12}
\isnset{app-routing-n-slices-layer-6}{33}
\isnset{app-routing-n-slices-layer-7}{65}
\isnset{app-routing-n-slices-layer-8}{114}
\isnset{app-routing-n-slices-layer-9}{70}
\isnset{app-routing-radius-car}{1.3}
\isnset{app-routing-radius-layer-1}{1.3}
\isnset{app-routing-radius-layer-10}{0.6}
\isnset{app-routing-radius-layer-11}{0.6}
\isnset{app-routing-radius-layer-12}{1.1}
\isnset{app-routing-radius-layer-2}{1.3}
\isnset{app-routing-radius-layer-3}{0.6}
\isnset{app-routing-radius-layer-4}{0.4}
\isnset{app-routing-radius-layer-5}{0.6}
\isnset{app-routing-radius-layer-6}{0.4}
\isnset{app-routing-radius-layer-7}{0.8}
\isnset{app-routing-radius-layer-8}{0.7}
\isnset{app-routing-radius-layer-9}{0.6}
\isnset{app-routing-run}{171}
\isnset{app-volume-err-pressure-pct}{4.0}
\isnset{app-volume-err-velocity-pct}{5.5}
\isnset{app-volume-grid-nx}{1{,}200}
\isnset{app-volume-grid-nz}{500}
\isnset{app-volume-n-scored}{200{,}000}
\isnset{app-volume-run}{129}
\isnset{app-volume-slab-mm}{16}
\isnset{aug-available}{1}
\isnset{aug-gt-n-seeds}{2}
\isnset{aug-gt-native-excess-pct}{3.3}
\isnset{aug-gt-native-ratio}{1.03}
\isnset{aug-gt-rot-change-pct}{0.2}
\isnset{aug-gt-unaug-ratio}{31}
\isnset{aug-gt-vs-isla-ci-hi}{1.07}
\isnset{aug-gt-vs-isla-ci-lo}{0.99}
\isnset{aug-gt-vs-isla-pct}{3}
\isnset{aug-gt-vs-isla-ratio}{1.03}
\isnset{aug-plain-n-seeds}{4}
\isnset{aug-rot-change-max-pct}{0.2}
\isnset{aug-t-n-seeds}{2}
\isnset{aug-t-native-excess-pct}{4.5}
\isnset{aug-t-native-ratio}{1.04}
\isnset{aug-t-rot-change-pct}{0.1}
\isnset{aug-t-unaug-ratio}{29}
\isnset{aug-t-vs-isla-ci-hi}{1.12}
\isnset{aug-t-vs-isla-ci-lo}{1.03}
\isnset{aug-t-vs-isla-pct}{8}
\isnset{aug-t-vs-isla-ratio}{1.08}
\isnset{auto-ahmedml-scarce-pct-max}{$+$7}
\isnset{auto-ahmedml-scarce-pct-min}{$-$4}
\isnset{auto-ahmedml-scarce-ratio-max}{1.07}
\isnset{auto-ahmedml-scarce-ratio-min}{0.96}
\isnset{auto-ahmedml-super-scarce-pct-max}{$+$13}
\isnset{auto-ahmedml-super-scarce-pct-min}{$-$7}
\isnset{auto-ahmedml-super-scarce-ratio-max}{1.13}
\isnset{auto-ahmedml-super-scarce-ratio-min}{0.93}
\isnset{auto-drivaerml-scarce-pct-max}{$+$16}
\isnset{auto-drivaerml-scarce-pct-min}{$+$5}
\isnset{auto-drivaerml-scarce-ratio-max}{1.16}
\isnset{auto-drivaerml-scarce-ratio-min}{1.05}
\isnset{auto-full-surf-above-pct}{5}
\isnset{auto-full-surf-below-pct}{12}
\isnset{auto-full-vol-above-pct}{28}
\isnset{auto-full-vol-min-above-pct}{0}
\isnset{auto-full-vol-span}{0\% to 28\% above}
\isnset{auto-scarce-above-max-pct}{16}
\isnset{auto-scarce-below-pct}{4}
\isnset{auto-windsorml-super-scarce-pct-max}{$+$19}
\isnset{auto-windsorml-super-scarce-pct-min}{$+$3}
\isnset{auto-windsorml-super-scarce-ratio-max}{1.19}
\isnset{auto-windsorml-super-scarce-ratio-min}{1.03}
\isnset{auto-wml-full-n-lowest}{3}
\isnset{az-aux-from-repair}{1}
\isnset{az-baseline-n-settings}{3}
\isnset{az-baseline-params-gt}{27.6}
\isnset{az-baseline-params-t}{9.0}
\isnset{az-blobs-above-pct}{-79}
\isnset{az-blobs-epsU-mean-fs}{0.409}
\isnset{az-blobs-epsU-mean-isla}{0.085}
\isnset{az-blobs-epsp-mean-fs}{1.655}
\isnset{az-blobs-epsp-mean-isla}{0.294}
\isnset{az-blobs-ratio-pct}{79}
\isnset{az-cap-blocks-small}{8}
\isnset{az-cap-ctrl-epsp-mean}{0.321}
\isnset{az-cap-epsU-mean-pct}{3}
\isnset{az-cap-epsp-mean-lower-pct}{2}
\isnset{az-cap-mlp-small}{2}
\isnset{az-cap-n-seeds}{2}
\isnset{az-cap-nonbox-ratio}{1.17}
\isnset{az-cap-params-ctrl}{67.5}
\isnset{az-cap-params-small}{28.2}
\isnset{az-cap-sflow-control}{0.193}
\isnset{az-cap-sflow-smaller}{0.193}
\isnset{az-cap-slices-small}{64}
\isnset{az-cap-small-gt-p-ratio}{1.4}
\isnset{az-cap-small-gt-u-ratio}{1.8}
\isnset{az-catalog-hash}{c8e9508e8a9d8291}
\isnset{az-enc-confined-fs}{0.171}
\isnset{az-enc-confined-gtb}{0.117}
\isnset{az-enc-confined-isla}{0.086}
\isnset{az-enc-confined-ratio}{1.4}
\isnset{az-enc-crossflow-fs}{1.019}
\isnset{az-enc-crossflow-gtb}{0.253}
\isnset{az-enc-crossflow-isla}{0.174}
\isnset{az-enc-crossflow-ratio}{1.5}
\isnset{az-enc-dual-fs}{1.586}
\isnset{az-enc-dual-gtb}{0.396}
\isnset{az-enc-dual-isla}{0.164}
\isnset{az-enc-dual-ratio}{2.4}
\isnset{az-enc-duct-fs}{0.538}
\isnset{az-enc-duct-gtb}{0.300}
\isnset{az-enc-duct-isla}{0.136}
\isnset{az-enc-duct-ratio}{2.2}
\isnset{az-enc-ground-fs}{0.085}
\isnset{az-enc-ground-gtb}{0.061}
\isnset{az-enc-ground-isla}{0.047}
\isnset{az-enc-ground-ratio}{1.3}
\isnset{az-enc-hemi-fs}{0.195}
\isnset{az-enc-hemi-gtb}{0.116}
\isnset{az-enc-hemi-isla}{0.057}
\isnset{az-enc-hemi-ratio}{2.0}
\isnset{az-enc-open-fs}{0.049}
\isnset{az-enc-open-gtb}{0.041}
\isnset{az-enc-open-isla}{0.030}
\isnset{az-enc-open-ratio}{1.4}
\isnset{az-enc-ratio-max}{2.4}
\isnset{az-enc-ratio-min}{1.3}
\isnset{az-enc-sphere-fs}{0.041}
\isnset{az-enc-sphere-gtb}{0.035}
\isnset{az-enc-sphere-isla}{0.026}
\isnset{az-enc-sphere-ratio}{1.4}
\isnset{az-full-epsU-mean-best}{0.152}
\isnset{az-full-epsU-mean-fs}{0.440}
\isnset{az-full-epsU-mean-fs-ref}{0.440}
\isnset{az-full-epsU-mean-gt-si}{1.184}
\isnset{az-full-epsU-mean-gtb}{0.152}
\isnset{az-full-epsU-mean-gtb-max}{0.294}
\isnset{az-full-epsU-mean-gtb-min}{0.253}
\isnset{az-full-epsU-mean-gtb-si}{0.475}
\isnset{az-full-epsU-mean-isla}{0.082}
\isnset{az-full-epsU-mean-t-si}{1.202}
\isnset{az-full-epsU-mean-zero}{1.203}
\isnset{az-full-epsp-mean-best}{0.427}
\isnset{az-full-epsp-mean-fs}{1.809}
\isnset{az-full-epsp-mean-gtb-max}{1.292}
\isnset{az-full-epsp-mean-gtb-min}{0.987}
\isnset{az-full-epsp-mean-isla}{0.196}
\isnset{az-full-epsp-tail-all}{0.196}
\isnset{az-full-epsp-tail-n}{5}
\isnset{az-full-epsp-tail-rest}{0.109}
\isnset{az-full-epspn-mean-gtb}{0.434}
\isnset{az-full-epspn-mean-isla}{0.214}
\isnset{az-full-epspn-mean-t}{1.315}
\isnset{az-full-mean-ratio}{1.9}
\isnset{az-full-meanp-ratio}{2.2}
\isnset{az-full-n}{515}
\isnset{az-full-n-train}{2{,}011}
\isnset{az-full-n-val}{290}
\isnset{az-full-rU-gt-si}{0.981}
\isnset{az-full-rU-gtb}{0.180}
\isnset{az-full-rU-mean-isla}{0.110}
\isnset{az-full-rU-t-si}{0.999}
\isnset{az-full-rel-gamma-isla}{0.050}
\isnset{az-full-rel-retheta-isla}{0.103}
\isnset{az-full-rp-gtb}{0.623}
\isnset{az-full-rp-gtb-si}{1.002}
\isnset{az-full-rp-mean-isla}{0.304}
\isnset{az-full-sflow-fs}{1.000}
\isnset{az-full-sflow-gtb}{0.290}
\isnset{az-full-sflow-isla}{0.153}
\isnset{az-full-sflow-joint}{0.123}
\isnset{az-full-sflow-t}{0.679}
\isnset{az-full-sflow-zero}{1.648}
\isnset{az-full-share-gtb}{86}
\isnset{az-full-share-isla}{98}
\isnset{az-full-w90-p-hi}{2.25}
\isnset{az-full-w90-p-lo}{2.10}
\isnset{az-full-w90-p-ratio}{2.17}
\isnset{az-full-w90-u-hi}{1.90}
\isnset{az-full-w90-u-lo}{1.83}
\isnset{az-full-w90-u-ratio}{1.86}
\isnset{az-fulldef-above-pct}{-81}
\isnset{az-fulldef-epsU-mean-isla}{0.082}
\isnset{az-fulldef-epsp-mean-isla}{0.297}
\isnset{az-fulldef-ratio-pct}{81}
\isnset{az-gpuh-gt}{66}
\isnset{az-gpuh-isla}{113}
\isnset{az-gpuh-t}{25}
\isnset{az-gt-tokens-per-boundary}{50{,}000}
\isnset{az-ho-obs-sd-max}{0.26}
\isnset{az-ho-obs-sd-min}{0.14}
\isnset{az-hocur-re-p-ratio}{0.88}
\isnset{az-isla-seeds}{6}
\isnset{az-joint-epsU-mean}{0.080}
\isnset{az-joint-epsUn-mean}{0.110}
\isnset{az-joint-epsp-mean}{0.124}
\isnset{az-joint-mean-ratio}{1.9}
\isnset{az-joint-meanp-ratio}{3.5}
\isnset{az-joint-params}{67.5}
\isnset{az-joint-params-exact}{67{,}493{,}495}
\isnset{az-joint-rU-mean}{0.106}
\isnset{az-joint-relp-cut-pct}{8}
\isnset{az-joint-rp-mean}{0.281}
\isnset{az-joint-seeds}{4}
\isnset{az-joint-share}{98}
\isnset{az-joint-step-ratio}{1.46}
\isnset{az-n-canonical}{2{,}816}
\isnset{az-n-ducts}{161}
\isnset{az-n-released}{3{,}815}
\isnset{az-nonbox-above-pct}{8}
\isnset{az-nonbox-epsU-mean-fs}{0.238}
\isnset{az-nonbox-epsU-mean-isla}{0.257}
\isnset{az-nonbox-epsp-mean-fs}{0.595}
\isnset{az-nonbox-epsp-mean-isla}{0.747}
\isnset{az-nonbox-ratio-pct}{-8}
\isnset{az-pcf-above-pct}{-31}
\isnset{az-pcf-epsU-mean-fs}{0.947}
\isnset{az-pcf-epsU-mean-isla}{0.656}
\isnset{az-pcf-epsp-mean-fs}{3.490}
\isnset{az-pcf-epsp-mean-isla}{2.906}
\isnset{az-pcf-ratio-pct}{31}
\isnset{az-pdi-above-pct}{-13}
\isnset{az-pdi-epsU-mean-fs}{1.659}
\isnset{az-pdi-epsU-mean-isla}{1.450}
\isnset{az-pdi-epsp-mean-fs}{9.134}
\isnset{az-pdi-epsp-mean-isla}{8.379}
\isnset{az-pdi-ratio-pct}{13}
\isnset{az-poolcur-p}{1.174}
\isnset{az-poolcur-p-2dp}{1.17}
\isnset{az-poolcur-p-hi}{1.312}
\isnset{az-poolcur-p-hi-2dp}{1.31}
\isnset{az-poolcur-p-lo}{1.051}
\isnset{az-poolcur-p-lo-2dp}{1.05}
\isnset{az-poolcur-s}{1.344}
\isnset{az-poolcur-s-2dp}{1.34}
\isnset{az-poolcur-s-hi}{1.495}
\isnset{az-poolcur-s-hi-2dp}{1.50}
\isnset{az-poolcur-s-lo}{1.208}
\isnset{az-poolcur-s-lo-2dp}{1.21}
\isnset{az-poolcur-u}{1.503}
\isnset{az-poolcur-u-2dp}{1.50}
\isnset{az-poolcur-u-hi}{1.689}
\isnset{az-poolcur-u-hi-2dp}{1.69}
\isnset{az-poolcur-u-lo}{1.337}
\isnset{az-poolcur-u-lo-2dp}{1.34}
\isnset{az-re-above-pct}{-67}
\isnset{az-re-epsU-mean-fs}{0.379}
\isnset{az-re-epsU-mean-isla}{0.125}
\isnset{az-re-epsp-mean-fs}{1.640}
\isnset{az-re-epsp-mean-isla}{0.576}
\isnset{az-re-ratio-pct}{67}
\isnset{az-release}{v1.2}
\isnset{az-rep-seeds}{6}
\isnset{az-rep-steps}{125{,}500}
\isnset{az-sep-params-exact}{67{,}490{,}930}
\isnset{az-splits-seeds}{2}
\isnset{az-tree-n-cases}{8}
\isnset{az-tree-target-p-th0-max}{29.1}
\isnset{az-tree-target-p-th0-min}{28.4}
\isnset{az-tree-target-p-th1-max}{28.3}
\isnset{az-tree-target-p-th1-min}{27.3}
\isnset{az-tree-th025-p}{1.7}
\isnset{az-tree-th1-p-case-max}{14}
\isnset{az-tree-th1-p-max}{7.3}
\isnset{az-tree-th1-p-min}{7.2}
\isnset{az-tree-th1-u-max}{2.8}
\isnset{az-tree-th1-u-min}{2.5}
\isnset{az-tree-trained025-p-max}{1.6}
\isnset{az-tree-trained025-p-min}{1.4}
\isnset{az-ztc-above-pct}{-75}
\isnset{az-ztc-epsU-mean-fs}{0.466}
\isnset{az-ztc-epsU-mean-isla}{0.117}
\isnset{az-ztc-epsp-mean-fs}{2.004}
\isnset{az-ztc-epsp-mean-isla}{0.362}
\isnset{az-ztc-ratio-pct}{75}
\isnset{bc-arm-isla-above-pct}{11}
\isnset{bc-arm-isla-below-pct}{3}
\isnset{bc-best-steps-max}{10{,}002}
\isnset{bc-best-steps-min}{4{,}502}
\isnset{bc-frozen-isla-max-pct}{72}
\isnset{bc-frozen-isla-min-pct}{22}
\isnset{bc-gt-cut-max}{26}
\isnset{bc-gt-cut-min}{12}
\isnset{bc-sched-ratio}{12.5}
\isnset{bc-sched-steps}{50{,}000}
\isnset{bc-sp-isla-max-pct}{11}
\isnset{bc-sp-isla-min-pct}{9}
\isnset{bc-stop-steps-max}{30{,}502}
\isnset{bc-stop-steps-min}{30{,}002}
\isnset{bc-t-cut-max}{36}
\isnset{bc-t-cut-min}{31}
\isnset{bench-ahmedml-full-n-val}{50}
\isnset{bench-airfrans-full-n-val}{80}
\isnset{bench-auto-members-max}{15}
\isnset{bench-auto-members-min}{4}
\isnset{bench-dml-ss-phase}{test}
\isnset{bench-drivaerml-full-n-val}{34}
\isnset{bench-full-val-note}{test cases}
\isnset{bench-interim-note}{}
\isnset{bench-members-max}{20}
\isnset{bench-members-min}{3}
\isnset{bench-min-train}{7}
\isnset{bench-min-train-ahm}{11}
\isnset{bench-min-train-dml}{11}
\isnset{bench-min-train-wml}{7}
\isnset{bench-n-boards}{27}
\isnset{bench-n-main-rows}{8}
\isnset{bench-n-test}{27}
\isnset{bench-n-val}{27}
\isnset{bench-n-val-nowml}{19}
\isnset{bench-n-val-wml}{8}
\isnset{bench-optimistic-note}{}
\isnset{bench-refit-note}{ \isla{}'s validation rows leave out its networks trained on the validation cases (\cref{app:airfrans-setup}).}
\isnset{bench-retrained-note-ahmedml}{}
\isnset{bench-retrained-note-airfrans}{}
\isnset{bench-retrained-note-auto}{}
\isnset{bench-retrained-note-drivaerml}{}
\isnset{bench-retrained-note-windsorml}{}
\isnset{bench-scarce-lead}{6}
\isnset{bench-scarce-train-dml}{67}
\isnset{bench-ss-field-cut-max}{14}
\isnset{bench-ss-field-cut-max-test}{14}
\isnset{bench-ss-field-cut-max-validation}{13}
\isnset{bench-ss-field-cut-min}{8}
\isnset{bench-ss-field-cut-min-test}{8}
\isnset{bench-ss-field-cut-min-validation}{8}
\isnset{bench-test-rank-globe}{3.19}
\isnset{bench-test-rank-gt}{1.96}
\isnset{bench-test-rank-isla}{2.41}
\isnset{bench-test-rank-t}{2.44}
\isnset{bench-test-wins-globe}{5}
\isnset{bench-test-wins-gt}{13}
\isnset{bench-test-wins-isla}{4}
\isnset{bench-test-wins-t}{5}
\isnset{bench-val-nowml-rank-globe}{3.11}
\isnset{bench-val-nowml-rank-gt}{1.79}
\isnset{bench-val-nowml-rank-isla}{2.42}
\isnset{bench-val-nowml-rank-t}{2.68}
\isnset{bench-val-rank-globe}{3.15}
\isnset{bench-val-rank-gt}{2.07}
\isnset{bench-val-rank-isla}{2.19}
\isnset{bench-val-rank-t}{2.59}
\isnset{bench-val-wins-globe}{5}
\isnset{bench-val-wins-gt}{11}
\isnset{bench-val-wins-isla}{9}
\isnset{bench-val-wins-t}{2}
\isnset{bench-vu-excess-ahmed-max}{30}
\isnset{bench-vu-excess-ahmed-min}{27}
\isnset{bench-vu-excess-dml-max}{28}
\isnset{bench-vu-excess-dml-min}{8}
\isnset{bench-windsorml-full-n-val}{33}
\isnset{canon-aligned-gt-p-pct}{18}
\isnset{canon-aligned-gt-ratio}{1.18}
\isnset{canon-aligned-ratio-max}{1.18}
\isnset{canon-aligned-ratio-min}{1.09}
\isnset{canon-aligned-t-p-pct}{9}
\isnset{canon-aligned-t-ratio}{1.09}
\isnset{canon-aligned-tw-max-pct}{19}
\isnset{canon-aligned-tw-min-pct}{12}
\isnset{canon-raw-gt-ratio}{5.4}
\isnset{canon-raw-ratio-max}{5.8}
\isnset{canon-raw-ratio-min}{5.4}
\isnset{canon-raw-t-ratio}{5.8}
\isnset{canon-tilt-deg}{2.3}
\isnset{car-native-gt-p}{4.0}
\isnset{car-native-gt-wss}{6.2}
\isnset{car-native-isla-p}{3.9}
\isnset{car-native-isla-wss}{6.2}
\isnset{car-native-t-p}{3.8}
\isnset{car-native-t-wss}{6.3}
\isnset{car-pose-angle}{170}
\isnset{car-posed-gt-p}{98}
\isnset{car-posed-gt-wss}{143}
\isnset{car-posed-isla-p}{3.9}
\isnset{car-posed-isla-wss}{6.2}
\isnset{car-posed-t-p}{110}
\isnset{car-posed-t-wss}{187}
\isnset{car-resampled-gt-p}{33}
\isnset{car-resampled-gt-wss}{41}
\isnset{car-resampled-isla-p}{3.8}
\isnset{car-resampled-isla-wss}{6.2}
\isnset{car-resampled-t-p}{31}
\isnset{car-resampled-t-wss}{38}
\isnset{cert-az-isla-f32-chunks-nut}{$4\times10^{-7}$}
\isnset{cert-az-isla-f32-chunks-nut-pct}{0.0}
\isnset{cert-az-isla-f32-chunks-p}{$5\times10^{-7}$}
\isnset{cert-az-isla-f32-chunks-p-pct}{0.0}
\isnset{cert-az-isla-f32-chunks-u}{$4\times10^{-7}$}
\isnset{cert-az-isla-f32-chunks-u-pct}{0.0}
\isnset{cert-az-isla-f32-n}{8}
\isnset{cert-az-isla-f32-perm-nut}{$7\times10^{-7}$}
\isnset{cert-az-isla-f32-perm-nut-pct}{0.0}
\isnset{cert-az-isla-f32-perm-p}{$2\times10^{-6}$}
\isnset{cert-az-isla-f32-perm-p-pct}{0.0}
\isnset{cert-az-isla-f32-perm-u}{$6\times10^{-7}$}
\isnset{cert-az-isla-f32-perm-u-pct}{0.0}
\isnset{cert-az-isla-f32-query-nut}{0}
\isnset{cert-az-isla-f32-query-nut-pct}{0.0}
\isnset{cert-az-isla-f32-query-p}{0}
\isnset{cert-az-isla-f32-query-p-pct}{0.0}
\isnset{cert-az-isla-f32-query-u}{0}
\isnset{cert-az-isla-f32-query-u-pct}{0.0}
\isnset{cert-az-isla-f32-rot-nut}{$10^{-3}$}
\isnset{cert-az-isla-f32-rot-nut-pct}{0.1}
\isnset{cert-az-isla-f32-rot-p}{$2\times10^{-3}$}
\isnset{cert-az-isla-f32-rot-p-pct}{0.2}
\isnset{cert-az-isla-f32-rot-u}{$10^{-3}$}
\isnset{cert-az-isla-f32-rot-u-pct}{0.1}
\isnset{cert-az-isla-f32-rotfrozen-nut}{$3\times10^{-5}$}
\isnset{cert-az-isla-f32-rotfrozen-nut-pct}{0.0}
\isnset{cert-az-isla-f32-rotfrozen-p}{$3\times10^{-4}$}
\isnset{cert-az-isla-f32-rotfrozen-p-pct}{0.0}
\isnset{cert-az-isla-f32-rotfrozen-u}{$10^{-4}$}
\isnset{cert-az-isla-f32-rotfrozen-u-pct}{0.0}
\isnset{cert-az-isla-f32-scale-nut}{$4\times10^{-7}$}
\isnset{cert-az-isla-f32-scale-nut-pct}{0.0}
\isnset{cert-az-isla-f32-scale-p}{$6\times10^{-7}$}
\isnset{cert-az-isla-f32-scale-p-pct}{0.0}
\isnset{cert-az-isla-f32-scale-u}{$4\times10^{-7}$}
\isnset{cert-az-isla-f32-scale-u-pct}{0.0}
\isnset{cert-az-isla-f32-split-nut}{$1.2\times10^{-2}$}
\isnset{cert-az-isla-f32-split-nut-pct}{1.2}
\isnset{cert-az-isla-f32-split-p}{$2.3\times10^{-2}$}
\isnset{cert-az-isla-f32-split-p-pct}{2.3}
\isnset{cert-az-isla-f32-split-u}{$7\times10^{-3}$}
\isnset{cert-az-isla-f32-split-u-pct}{0.7}
\isnset{cert-az-isla-f32-tokendraw-nut}{$1.9\times10^{-2}$}
\isnset{cert-az-isla-f32-tokendraw-nut-pct}{1.9}
\isnset{cert-az-isla-f32-tokendraw-p}{$2.5\times10^{-2}$}
\isnset{cert-az-isla-f32-tokendraw-p-pct}{2.5}
\isnset{cert-az-isla-f32-tokendraw-u}{$1.0\times10^{-2}$}
\isnset{cert-az-isla-f32-tokendraw-u-pct}{1.0}
\isnset{cert-az-isla-f32-ulp-nut}{$9\times10^{-4}$}
\isnset{cert-az-isla-f32-ulp-nut-pct}{0.1}
\isnset{cert-az-isla-f32-ulp-p}{$10^{-3}$}
\isnset{cert-az-isla-f32-ulp-p-pct}{0.1}
\isnset{cert-az-isla-f32-ulp-u}{$8\times10^{-4}$}
\isnset{cert-az-isla-f32-ulp-u-pct}{0.1}
\isnset{cert-az-isla-f32-units-nut}{$2\times10^{-4}$}
\isnset{cert-az-isla-f32-units-nut-pct}{0.0}
\isnset{cert-az-isla-f32-units-p}{$4\times10^{-4}$}
\isnset{cert-az-isla-f32-units-p-pct}{0.0}
\isnset{cert-az-isla-f32-units-u}{$2\times10^{-4}$}
\isnset{cert-az-isla-f32-units-u-pct}{0.0}
\isnset{cert-az-isla-f64-chunks-nut}{$3\times10^{-15}$}
\isnset{cert-az-isla-f64-chunks-nut-pct}{0.0}
\isnset{cert-az-isla-f64-chunks-p}{$2\times10^{-15}$}
\isnset{cert-az-isla-f64-chunks-p-pct}{0.0}
\isnset{cert-az-isla-f64-chunks-u}{$3\times10^{-15}$}
\isnset{cert-az-isla-f64-chunks-u-pct}{0.0}
\isnset{cert-az-isla-f64-n}{8}
\isnset{cert-az-isla-f64-query-nut}{$10^{-15}$}
\isnset{cert-az-isla-f64-query-nut-pct}{0.0}
\isnset{cert-az-isla-f64-query-p}{$10^{-15}$}
\isnset{cert-az-isla-f64-query-p-pct}{0.0}
\isnset{cert-az-isla-f64-query-u}{$10^{-15}$}
\isnset{cert-az-isla-f64-query-u-pct}{0.0}
\isnset{cert-az-isla-f64-rot-nut}{$7\times10^{-10}$}
\isnset{cert-az-isla-f64-rot-nut-pct}{0.0}
\isnset{cert-az-isla-f64-rot-p}{$10^{-9}$}
\isnset{cert-az-isla-f64-rot-p-pct}{0.0}
\isnset{cert-az-isla-f64-rot-u}{$7\times10^{-10}$}
\isnset{cert-az-isla-f64-rot-u-pct}{0.0}
\isnset{cert-az-isla-f64-ulp-nut}{$5\times10^{-10}$}
\isnset{cert-az-isla-f64-ulp-nut-pct}{0.0}
\isnset{cert-az-isla-f64-ulp-p}{$5\times10^{-10}$}
\isnset{cert-az-isla-f64-ulp-p-pct}{0.0}
\isnset{cert-az-isla-f64-ulp-u}{$4\times10^{-10}$}
\isnset{cert-az-isla-f64-ulp-u-pct}{0.0}
\isnset{cert-az-isla-f64-units-nut}{$4\times10^{-10}$}
\isnset{cert-az-isla-f64-units-nut-pct}{0.0}
\isnset{cert-az-isla-f64-units-p}{$5\times10^{-10}$}
\isnset{cert-az-isla-f64-units-p-pct}{0.0}
\isnset{cert-az-isla-f64-units-u}{$3\times10^{-10}$}
\isnset{cert-az-isla-f64-units-u-pct}{0.0}
\isnset{cert-dmls-gt-f32-area-p}{0.29}
\isnset{cert-dmls-gt-f32-area-p-pct}{29.3}
\isnset{cert-dmls-gt-f32-area-wss}{0.35}
\isnset{cert-dmls-gt-f32-area-wss-pct}{34.7}
\isnset{cert-dmls-gt-f32-mirror-p}{1.1}
\isnset{cert-dmls-gt-f32-mirror-p-pct}{111.6}
\isnset{cert-dmls-gt-f32-mirror-wss}{1.3}
\isnset{cert-dmls-gt-f32-mirror-wss-pct}{126.9}
\isnset{cert-dmls-gt-f32-n}{8}
\isnset{cert-dmls-gt-f32-perm-p}{$8\times10^{-7}$}
\isnset{cert-dmls-gt-f32-perm-p-pct}{0.0}
\isnset{cert-dmls-gt-f32-perm-wss}{$10^{-6}$}
\isnset{cert-dmls-gt-f32-perm-wss-pct}{0.0}
\isnset{cert-dmls-gt-f32-redraw-p}{$4\times10^{-3}$}
\isnset{cert-dmls-gt-f32-redraw-p-pct}{0.4}
\isnset{cert-dmls-gt-f32-redraw-wss}{$6\times10^{-3}$}
\isnset{cert-dmls-gt-f32-redraw-wss-pct}{0.6}
\isnset{cert-dmls-gt-f32-rot-p}{1.2}
\isnset{cert-dmls-gt-f32-rot-p-pct}{119.3}
\isnset{cert-dmls-gt-f32-rot-wss}{1.4}
\isnset{cert-dmls-gt-f32-rot-wss-pct}{136.8}
\isnset{cert-dmls-gt-f32-scale-p}{0.62}
\isnset{cert-dmls-gt-f32-scale-p-pct}{62.2}
\isnset{cert-dmls-gt-f32-scale-wss}{0.65}
\isnset{cert-dmls-gt-f32-scale-wss-pct}{64.7}
\isnset{cert-dmls-gt-f32-split-p}{$7\times10^{-4}$}
\isnset{cert-dmls-gt-f32-split-p-pct}{0.1}
\isnset{cert-dmls-gt-f32-split-wss}{$9\times10^{-4}$}
\isnset{cert-dmls-gt-f32-split-wss-pct}{0.1}
\isnset{cert-dmls-gt-f32-ulp-p}{$10^{-3}$}
\isnset{cert-dmls-gt-f32-ulp-p-pct}{0.1}
\isnset{cert-dmls-gt-f32-ulp-wss}{$2\times10^{-3}$}
\isnset{cert-dmls-gt-f32-ulp-wss-pct}{0.2}
\isnset{cert-dmls-gt-f32-units-p}{$6\times10^{-4}$}
\isnset{cert-dmls-gt-f32-units-p-pct}{0.1}
\isnset{cert-dmls-gt-f32-units-wss}{$10^{-3}$}
\isnset{cert-dmls-gt-f32-units-wss-pct}{0.1}
\isnset{cert-dmls-isla-f32-area-p}{$6\times10^{-3}$}
\isnset{cert-dmls-isla-f32-area-p-pct}{0.6}
\isnset{cert-dmls-isla-f32-area-wss}{$1.0\times10^{-2}$}
\isnset{cert-dmls-isla-f32-area-wss-pct}{1.0}
\isnset{cert-dmls-isla-f32-mirror-p}{$6\times10^{-4}$}
\isnset{cert-dmls-isla-f32-mirror-p-pct}{0.1}
\isnset{cert-dmls-isla-f32-mirror-wss}{$2\times10^{-3}$}
\isnset{cert-dmls-isla-f32-mirror-wss-pct}{0.2}
\isnset{cert-dmls-isla-f32-n}{8}
\isnset{cert-dmls-isla-f32-perm-p}{$2\times10^{-6}$}
\isnset{cert-dmls-isla-f32-perm-p-pct}{0.0}
\isnset{cert-dmls-isla-f32-perm-wss}{$4\times10^{-6}$}
\isnset{cert-dmls-isla-f32-perm-wss-pct}{0.0}
\isnset{cert-dmls-isla-f32-redraw-p}{$7\times10^{-3}$}
\isnset{cert-dmls-isla-f32-redraw-p-pct}{0.7}
\isnset{cert-dmls-isla-f32-redraw-wss}{$1.0\times10^{-2}$}
\isnset{cert-dmls-isla-f32-redraw-wss-pct}{1.0}
\isnset{cert-dmls-isla-f32-rot-p}{$6\times10^{-4}$}
\isnset{cert-dmls-isla-f32-rot-p-pct}{0.1}
\isnset{cert-dmls-isla-f32-rot-wss}{$10^{-3}$}
\isnset{cert-dmls-isla-f32-rot-wss-pct}{0.1}
\isnset{cert-dmls-isla-f32-scale-p}{$10^{-6}$}
\isnset{cert-dmls-isla-f32-scale-p-pct}{0.0}
\isnset{cert-dmls-isla-f32-scale-wss}{$3\times10^{-6}$}
\isnset{cert-dmls-isla-f32-scale-wss-pct}{0.0}
\isnset{cert-dmls-isla-f32-split-p}{$4\times10^{-6}$}
\isnset{cert-dmls-isla-f32-split-p-pct}{0.0}
\isnset{cert-dmls-isla-f32-split-wss}{$8\times10^{-6}$}
\isnset{cert-dmls-isla-f32-split-wss-pct}{0.0}
\isnset{cert-dmls-isla-f32-ulp-p}{$5\times10^{-4}$}
\isnset{cert-dmls-isla-f32-ulp-p-pct}{0.1}
\isnset{cert-dmls-isla-f32-ulp-wss}{$10^{-3}$}
\isnset{cert-dmls-isla-f32-ulp-wss-pct}{0.1}
\isnset{cert-dmls-isla-f32-units-p}{$3\times10^{-4}$}
\isnset{cert-dmls-isla-f32-units-p-pct}{0.0}
\isnset{cert-dmls-isla-f32-units-wss}{$8\times10^{-4}$}
\isnset{cert-dmls-isla-f32-units-wss-pct}{0.1}
\isnset{cert-dmls-isla-f64-mirror-p}{$10^{-12}$}
\isnset{cert-dmls-isla-f64-mirror-p-pct}{0.0}
\isnset{cert-dmls-isla-f64-mirror-wss}{$3\times10^{-12}$}
\isnset{cert-dmls-isla-f64-mirror-wss-pct}{0.0}
\isnset{cert-dmls-isla-f64-n}{8}
\isnset{cert-dmls-isla-f64-perm-p}{$3\times10^{-15}$}
\isnset{cert-dmls-isla-f64-perm-p-pct}{0.0}
\isnset{cert-dmls-isla-f64-perm-wss}{$7\times10^{-15}$}
\isnset{cert-dmls-isla-f64-perm-wss-pct}{0.0}
\isnset{cert-dmls-isla-f64-rot-p}{$10^{-12}$}
\isnset{cert-dmls-isla-f64-rot-p-pct}{0.0}
\isnset{cert-dmls-isla-f64-rot-wss}{$3\times10^{-12}$}
\isnset{cert-dmls-isla-f64-rot-wss-pct}{0.0}
\isnset{cert-dmls-isla-f64-scale-p}{$2\times10^{-15}$}
\isnset{cert-dmls-isla-f64-scale-p-pct}{0.0}
\isnset{cert-dmls-isla-f64-scale-wss}{$5\times10^{-15}$}
\isnset{cert-dmls-isla-f64-scale-wss-pct}{0.0}
\isnset{cert-dmls-isla-f64-split-p}{$4\times10^{-15}$}
\isnset{cert-dmls-isla-f64-split-p-pct}{0.0}
\isnset{cert-dmls-isla-f64-split-wss}{$10^{-14}$}
\isnset{cert-dmls-isla-f64-split-wss-pct}{0.0}
\isnset{cert-dmls-isla-f64-ulp-p}{$9\times10^{-13}$}
\isnset{cert-dmls-isla-f64-ulp-p-pct}{0.0}
\isnset{cert-dmls-isla-f64-ulp-wss}{$2\times10^{-12}$}
\isnset{cert-dmls-isla-f64-ulp-wss-pct}{0.0}
\isnset{cert-dmls-isla-f64-units-p}{$3\times10^{-13}$}
\isnset{cert-dmls-isla-f64-units-p-pct}{0.0}
\isnset{cert-dmls-isla-f64-units-wss}{$8\times10^{-13}$}
\isnset{cert-dmls-isla-f64-units-wss-pct}{0.0}
\isnset{cert-dmls-t-f32-area-p}{0.28}
\isnset{cert-dmls-t-f32-area-p-pct}{28.3}
\isnset{cert-dmls-t-f32-area-wss}{0.33}
\isnset{cert-dmls-t-f32-area-wss-pct}{32.5}
\isnset{cert-dmls-t-f32-mirror-p}{1.2}
\isnset{cert-dmls-t-f32-mirror-p-pct}{116.0}
\isnset{cert-dmls-t-f32-mirror-wss}{1.5}
\isnset{cert-dmls-t-f32-mirror-wss-pct}{147.2}
\isnset{cert-dmls-t-f32-n}{8}
\isnset{cert-dmls-t-f32-perm-p}{$5\times10^{-7}$}
\isnset{cert-dmls-t-f32-perm-p-pct}{0.0}
\isnset{cert-dmls-t-f32-perm-wss}{$7\times10^{-7}$}
\isnset{cert-dmls-t-f32-perm-wss-pct}{0.0}
\isnset{cert-dmls-t-f32-redraw-p}{$3\times10^{-3}$}
\isnset{cert-dmls-t-f32-redraw-p-pct}{0.3}
\isnset{cert-dmls-t-f32-redraw-wss}{$4\times10^{-3}$}
\isnset{cert-dmls-t-f32-redraw-wss-pct}{0.4}
\isnset{cert-dmls-t-f32-rot-p}{1.2}
\isnset{cert-dmls-t-f32-rot-p-pct}{121.5}
\isnset{cert-dmls-t-f32-rot-wss}{1.6}
\isnset{cert-dmls-t-f32-rot-wss-pct}{162.3}
\isnset{cert-dmls-t-f32-scale-p}{0.65}
\isnset{cert-dmls-t-f32-scale-p-pct}{64.8}
\isnset{cert-dmls-t-f32-scale-wss}{0.66}
\isnset{cert-dmls-t-f32-scale-wss-pct}{65.8}
\isnset{cert-dmls-t-f32-split-p}{$8\times10^{-4}$}
\isnset{cert-dmls-t-f32-split-p-pct}{0.1}
\isnset{cert-dmls-t-f32-split-wss}{$10^{-3}$}
\isnset{cert-dmls-t-f32-split-wss-pct}{0.1}
\isnset{cert-dmls-t-f32-ulp-p}{$10^{-3}$}
\isnset{cert-dmls-t-f32-ulp-p-pct}{0.1}
\isnset{cert-dmls-t-f32-ulp-wss}{$3\times10^{-3}$}
\isnset{cert-dmls-t-f32-ulp-wss-pct}{0.3}
\isnset{cert-dmls-t-f32-units-p}{$6\times10^{-4}$}
\isnset{cert-dmls-t-f32-units-p-pct}{0.1}
\isnset{cert-dmls-t-f32-units-wss}{$10^{-3}$}
\isnset{cert-dmls-t-f32-units-wss-pct}{0.1}
\isnset{cert-dmlv-isla-f32-chunks-nut}{$9\times10^{-7}$}
\isnset{cert-dmlv-isla-f32-chunks-nut-pct}{0.0}
\isnset{cert-dmlv-isla-f32-chunks-p}{$6\times10^{-7}$}
\isnset{cert-dmlv-isla-f32-chunks-p-pct}{0.0}
\isnset{cert-dmlv-isla-f32-chunks-u}{$9\times10^{-7}$}
\isnset{cert-dmlv-isla-f32-chunks-u-pct}{0.0}
\isnset{cert-dmlv-isla-f32-mirror-nut}{$3\times10^{-3}$}
\isnset{cert-dmlv-isla-f32-mirror-nut-pct}{0.3}
\isnset{cert-dmlv-isla-f32-mirror-p}{$5\times10^{-3}$}
\isnset{cert-dmlv-isla-f32-mirror-p-pct}{0.5}
\isnset{cert-dmlv-isla-f32-mirror-u}{$7\times10^{-3}$}
\isnset{cert-dmlv-isla-f32-mirror-u-pct}{0.7}
\isnset{cert-dmlv-isla-f32-n}{8}
\isnset{cert-dmlv-isla-f32-perm-nut}{$5\times10^{-4}$}
\isnset{cert-dmlv-isla-f32-perm-nut-pct}{0.0}
\isnset{cert-dmlv-isla-f32-perm-p}{$10^{-3}$}
\isnset{cert-dmlv-isla-f32-perm-p-pct}{0.1}
\isnset{cert-dmlv-isla-f32-perm-u}{$10^{-3}$}
\isnset{cert-dmlv-isla-f32-perm-u-pct}{0.1}
\isnset{cert-dmlv-isla-f32-query-nut}{0}
\isnset{cert-dmlv-isla-f32-query-nut-pct}{0.0}
\isnset{cert-dmlv-isla-f32-query-p}{0}
\isnset{cert-dmlv-isla-f32-query-p-pct}{0.0}
\isnset{cert-dmlv-isla-f32-query-u}{0}
\isnset{cert-dmlv-isla-f32-query-u-pct}{0.0}
\isnset{cert-dmlv-isla-f32-rot-nut}{$2\times10^{-3}$}
\isnset{cert-dmlv-isla-f32-rot-nut-pct}{0.2}
\isnset{cert-dmlv-isla-f32-rot-p}{$4\times10^{-3}$}
\isnset{cert-dmlv-isla-f32-rot-p-pct}{0.4}
\isnset{cert-dmlv-isla-f32-rot-u}{$6\times10^{-3}$}
\isnset{cert-dmlv-isla-f32-rot-u-pct}{0.6}
\isnset{cert-dmlv-isla-f32-rotfrozen-nut}{$2\times10^{-4}$}
\isnset{cert-dmlv-isla-f32-rotfrozen-nut-pct}{0.0}
\isnset{cert-dmlv-isla-f32-rotfrozen-p}{$6\times10^{-4}$}
\isnset{cert-dmlv-isla-f32-rotfrozen-p-pct}{0.1}
\isnset{cert-dmlv-isla-f32-rotfrozen-u}{$10^{-3}$}
\isnset{cert-dmlv-isla-f32-rotfrozen-u-pct}{0.1}
\isnset{cert-dmlv-isla-f32-scale-nut}{$10^{-6}$}
\isnset{cert-dmlv-isla-f32-scale-nut-pct}{0.0}
\isnset{cert-dmlv-isla-f32-scale-p}{$2\times10^{-6}$}
\isnset{cert-dmlv-isla-f32-scale-p-pct}{0.0}
\isnset{cert-dmlv-isla-f32-scale-u}{$2\times10^{-6}$}
\isnset{cert-dmlv-isla-f32-scale-u-pct}{0.0}
\isnset{cert-dmlv-isla-f32-split-nut}{$7\times10^{-3}$}
\isnset{cert-dmlv-isla-f32-split-nut-pct}{0.7}
\isnset{cert-dmlv-isla-f32-split-p}{$1.1\times10^{-2}$}
\isnset{cert-dmlv-isla-f32-split-p-pct}{1.1}
\isnset{cert-dmlv-isla-f32-split-u}{$1.4\times10^{-2}$}
\isnset{cert-dmlv-isla-f32-split-u-pct}{1.4}
\isnset{cert-dmlv-isla-f32-tokendraw-nut}{$7\times10^{-3}$}
\isnset{cert-dmlv-isla-f32-tokendraw-nut-pct}{0.7}
\isnset{cert-dmlv-isla-f32-tokendraw-p}{$1.1\times10^{-2}$}
\isnset{cert-dmlv-isla-f32-tokendraw-p-pct}{1.1}
\isnset{cert-dmlv-isla-f32-tokendraw-u}{$1.4\times10^{-2}$}
\isnset{cert-dmlv-isla-f32-tokendraw-u-pct}{1.4}
\isnset{cert-dmlv-isla-f32-ulp-nut}{$2\times10^{-3}$}
\isnset{cert-dmlv-isla-f32-ulp-nut-pct}{0.2}
\isnset{cert-dmlv-isla-f32-ulp-p}{$2.2\times10^{-2}$}
\isnset{cert-dmlv-isla-f32-ulp-p-pct}{2.2}
\isnset{cert-dmlv-isla-f32-ulp-u}{$5\times10^{-3}$}
\isnset{cert-dmlv-isla-f32-ulp-u-pct}{0.5}
\isnset{cert-dmlv-isla-f32-units-nut}{$2\times10^{-3}$}
\isnset{cert-dmlv-isla-f32-units-nut-pct}{0.2}
\isnset{cert-dmlv-isla-f32-units-p}{$3\times10^{-3}$}
\isnset{cert-dmlv-isla-f32-units-p-pct}{0.3}
\isnset{cert-dmlv-isla-f32-units-u}{$5\times10^{-3}$}
\isnset{cert-dmlv-isla-f32-units-u-pct}{0.5}
\isnset{cert-dmlv-isla-f64-chunks-nut}{$2\times10^{-13}$}
\isnset{cert-dmlv-isla-f64-chunks-nut-pct}{0.0}
\isnset{cert-dmlv-isla-f64-chunks-p}{$2\times10^{-13}$}
\isnset{cert-dmlv-isla-f64-chunks-p-pct}{0.0}
\isnset{cert-dmlv-isla-f64-chunks-u}{$2\times10^{-13}$}
\isnset{cert-dmlv-isla-f64-chunks-u-pct}{0.0}
\isnset{cert-dmlv-isla-f64-n}{8}
\isnset{cert-dmlv-isla-f64-perm-nut}{$3\times10^{-13}$}
\isnset{cert-dmlv-isla-f64-perm-nut-pct}{0.0}
\isnset{cert-dmlv-isla-f64-perm-p}{$10^{-12}$}
\isnset{cert-dmlv-isla-f64-perm-p-pct}{0.0}
\isnset{cert-dmlv-isla-f64-perm-u}{$10^{-12}$}
\isnset{cert-dmlv-isla-f64-perm-u-pct}{0.0}
\isnset{cert-dmlv-isla-f64-query-nut}{$2\times10^{-13}$}
\isnset{cert-dmlv-isla-f64-query-nut-pct}{0.0}
\isnset{cert-dmlv-isla-f64-query-p}{$2\times10^{-13}$}
\isnset{cert-dmlv-isla-f64-query-p-pct}{0.0}
\isnset{cert-dmlv-isla-f64-query-u}{$2\times10^{-13}$}
\isnset{cert-dmlv-isla-f64-query-u-pct}{0.0}
\isnset{cert-dmlv-isla-f64-rot-nut}{$3\times10^{-9}$}
\isnset{cert-dmlv-isla-f64-rot-nut-pct}{0.0}
\isnset{cert-dmlv-isla-f64-rot-p}{$2\times10^{-8}$}
\isnset{cert-dmlv-isla-f64-rot-p-pct}{0.0}
\isnset{cert-dmlv-isla-f64-rot-u}{$3\times10^{-8}$}
\isnset{cert-dmlv-isla-f64-rot-u-pct}{0.0}
\isnset{cert-dmlv-isla-f64-split-nut}{$7\times10^{-3}$}
\isnset{cert-dmlv-isla-f64-split-nut-pct}{0.7}
\isnset{cert-dmlv-isla-f64-split-p}{$1.1\times10^{-2}$}
\isnset{cert-dmlv-isla-f64-split-p-pct}{1.1}
\isnset{cert-dmlv-isla-f64-split-u}{$1.4\times10^{-2}$}
\isnset{cert-dmlv-isla-f64-split-u-pct}{1.4}
\isnset{cert-dmlv-isla-f64-tokendraw-nut}{$7\times10^{-3}$}
\isnset{cert-dmlv-isla-f64-tokendraw-nut-pct}{0.7}
\isnset{cert-dmlv-isla-f64-tokendraw-p}{$1.1\times10^{-2}$}
\isnset{cert-dmlv-isla-f64-tokendraw-p-pct}{1.1}
\isnset{cert-dmlv-isla-f64-tokendraw-u}{$1.4\times10^{-2}$}
\isnset{cert-dmlv-isla-f64-tokendraw-u-pct}{1.4}
\isnset{cert-dmlv-isla-f64-ulp-nut}{$3\times10^{-9}$}
\isnset{cert-dmlv-isla-f64-ulp-nut-pct}{0.0}
\isnset{cert-dmlv-isla-f64-ulp-p}{$3\times10^{-8}$}
\isnset{cert-dmlv-isla-f64-ulp-p-pct}{0.0}
\isnset{cert-dmlv-isla-f64-ulp-u}{$3\times10^{-8}$}
\isnset{cert-dmlv-isla-f64-ulp-u-pct}{0.0}
\isnset{cert-dmlv-isla-f64-units-nut}{$3\times10^{-9}$}
\isnset{cert-dmlv-isla-f64-units-nut-pct}{0.0}
\isnset{cert-dmlv-isla-f64-units-p}{$3\times10^{-8}$}
\isnset{cert-dmlv-isla-f64-units-p-pct}{0.0}
\isnset{cert-dmlv-isla-f64-units-u}{$2\times10^{-8}$}
\isnset{cert-dmlv-isla-f64-units-u-pct}{0.0}
\isnset{cert-dmlv-signflips}{710}
\isnset{cert-dmlv-tokens-changed-rot-pct}{9}
\isnset{cert-dmlv-tokens-changed-split-pct}{91}
\isnset{cert-dmlv-ulp-turned-pct}{0.4}
\isnset{cert-f64-query-max}{$2{\times}10^{-13}$}
\isnset{cert-f64-surf-max}{$3{\times}10^{-12}$}
\isnset{cert-n-cases}{8}
\isnset{cert-n-poses}{3}
\isnset{cost-full-az-points-m}{2.2}
\isnset{cost-full-az-s}{4.8}
\isnset{cost-full-volume-gib}{72}
\isnset{cost-full-volume-min}{7}
\isnset{cost-mem-surface}{56}
\isnset{cost-mem-volume}{129}
\isnset{cost-params-gt-m}{9.0}
\isnset{cost-params-t-m}{47.2}
\isnset{cost-step-surface}{0.16}
\isnset{cost-step-volume}{1.06}
\isnset{cost-surf-gt-bf16-gib}{5.1}
\isnset{cost-surf-gt-bf16-s}{0.09}
\isnset{cost-surf-gt-f32-gib}{5.4}
\isnset{cost-surf-gt-f32-s}{0.21}
\isnset{cost-surf-isla-bf16-gib}{2.7}
\isnset{cost-surf-isla-bf16-s}{0.05}
\isnset{cost-surf-isla-eager-bf16-gib}{4.0}
\isnset{cost-surf-isla-eager-bf16-s}{0.08}
\isnset{cost-surf-isla-eager-f32-gib}{6.4}
\isnset{cost-surf-isla-eager-f32-s}{0.37}
\isnset{cost-surf-isla-f32-gib}{3.4}
\isnset{cost-surf-isla-f32-s}{0.32}
\isnset{cost-surf-t-bf16-gib}{5.8}
\isnset{cost-surf-t-bf16-s}{0.14}
\isnset{cost-surf-t-f32-gib}{6.3}
\isnset{cost-surf-t-f32-s}{0.56}
\isnset{cost-vol-10m-bf16-gib}{32}
\isnset{cost-vol-10m-bf16-s}{12}
\isnset{cost-vol-10m-f32-gib}{50}
\isnset{cost-vol-10m-f32-s}{34}
\isnset{cost-vol-boundary-bf16-s}{0.32}
\isnset{cost-vol-boundary-f32-s}{0.58}
\isnset{cost-vol-slope-bf16}{1.2}
\isnset{cost-vol-slope-f32}{3.3}
\isnset{encb-area-ratio-max}{9.0}
\isnset{encb-area-ratio-min}{7.6}
\isnset{encb-aug-area-ratio}{7.0}
\isnset{encb-aug-so3-ratio}{23}
\isnset{encb-aug-stored-gt-ratio}{1.23}
\isnset{encb-aug-stored-t-ratio}{1.13}
\isnset{encb-aug-stored-tw-gt-ratio}{1.15}
\isnset{encb-aug-stored-tw-t-ratio}{1.08}
\isnset{encb-aug-z-ratio-max}{1.001}
\isnset{encb-available}{1}
\isnset{encb-geomonly-ratio-max}{29}
\isnset{encb-geomonly-ratio-min}{27}
\isnset{encb-globe-area-proportional-resampling-p}{9.35}
\isnset{encb-globe-area-proportional-resampling-p-ratio}{1.066}
\isnset{encb-globe-area-proportional-resampling-tw}{10.6}
\isnset{encb-globe-area-proportional-resampling-tw-ratio}{1.096}
\isnset{encb-globe-area-ratio-max}{1.10}
\isnset{encb-globe-area-ratio-min}{1.07}
\isnset{encb-globe-as-stored-p}{8.87}
\isnset{encb-globe-as-stored-tw}{9.67}
\isnset{encb-globe-rot-ratio-max}{1.26}
\isnset{encb-globe-rot-ratio-min}{1.19}
\isnset{encb-globe-rotation-so3-p}{11.1}
\isnset{encb-globe-rotation-so3-p-ratio}{1.255}
\isnset{encb-globe-rotation-so3-tw}{11.8}
\isnset{encb-globe-rotation-so3-tw-ratio}{1.218}
\isnset{encb-globe-rotation-vertical-axis-p}{11.0}
\isnset{encb-globe-rotation-vertical-axis-p-ratio}{1.243}
\isnset{encb-globe-rotation-vertical-axis-tw}{11.5}
\isnset{encb-globe-rotation-vertical-axis-tw-ratio}{1.186}
\isnset{encb-gt-area-proportional-resampling-p}{33.8}
\isnset{encb-gt-area-proportional-resampling-p-ratio}{9.0}
\isnset{encb-gt-area-proportional-resampling-tw}{40.0}
\isnset{encb-gt-area-proportional-resampling-tw-ratio}{6.8}
\isnset{encb-gt-as-stored-p}{3.92}
\isnset{encb-gt-as-stored-tw}{5.93}
\isnset{encb-gt-rotation-so3-p}{122}
\isnset{encb-gt-rotation-so3-p-ratio}{31.0}
\isnset{encb-gt-rotation-so3-tw}{130}
\isnset{encb-gt-rotation-so3-tw-ratio}{21.9}
\isnset{encb-gt-rotation-vertical-axis-p}{104}
\isnset{encb-gt-rotation-vertical-axis-p-ratio}{26.4}
\isnset{encb-gt-rotation-vertical-axis-tw}{125}
\isnset{encb-gt-rotation-vertical-axis-tw-ratio}{21.0}
\isnset{encb-headline-area-pct}{30}
\isnset{encb-headline-native-pct}{4}
\isnset{encb-headline-posed-floor}{100}
\isnset{encb-headline-posed-pct}{120}
\isnset{encb-isla-area-proportional-resampling-p}{3.77}
\isnset{encb-isla-area-proportional-resampling-p-ratio}{0.998}
\isnset{encb-isla-area-proportional-resampling-tw}{6.35}
\isnset{encb-isla-area-proportional-resampling-tw-ratio}{0.999}
\isnset{encb-isla-as-stored-p}{3.91}
\isnset{encb-isla-as-stored-tw}{6.40}
\isnset{encb-isla-rot-dev-pct}{0.2}
\isnset{encb-isla-rotation-so3-p}{3.91}
\isnset{encb-isla-rotation-so3-p-ratio}{1.000}
\isnset{encb-isla-rotation-so3-tw}{6.42}
\isnset{encb-isla-rotation-so3-tw-ratio}{1.002}
\isnset{encb-isla-rotation-vertical-axis-p}{3.91}
\isnset{encb-isla-rotation-vertical-axis-p-ratio}{1.000}
\isnset{encb-isla-rotation-vertical-axis-tw}{6.41}
\isnset{encb-isla-rotation-vertical-axis-tw-ratio}{1.000}
\isnset{encb-n-cases}{34}
\isnset{encb-rot-ratio-max}{31}
\isnset{encb-rot-ratio-min}{26}
\isnset{encb-rot-ratio-tw-max}{22}
\isnset{encb-rot-ratio-tw-min}{18}
\isnset{encb-t-area-proportional-resampling-p}{29.9}
\isnset{encb-t-area-proportional-resampling-p-ratio}{7.6}
\isnset{encb-t-area-proportional-resampling-tw}{39.2}
\isnset{encb-t-area-proportional-resampling-tw-ratio}{6.4}
\isnset{encb-t-as-stored-p}{4.05}
\isnset{encb-t-as-stored-tw}{6.20}
\isnset{encb-t-rotation-so3-p}{110}
\isnset{encb-t-rotation-so3-p-ratio}{27.1}
\isnset{encb-t-rotation-so3-tw}{115}
\isnset{encb-t-rotation-so3-tw-ratio}{18.5}
\isnset{encb-t-rotation-vertical-axis-p}{105}
\isnset{encb-t-rotation-vertical-axis-p-ratio}{26.0}
\isnset{encb-t-rotation-vertical-axis-tw}{112}
\isnset{encb-t-rotation-vertical-axis-tw-ratio}{18.0}
\isnset{encb-uniform-control-max-dev}{0.000}
\isnset{entry-params-af-full-geotransolver}{81~M}
\isnset{entry-params-af-full-globe}{5.5~M}
\isnset{entry-params-af-full-isla}{1.1~B}
\isnset{entry-params-af-full-transolver}{27~M}
\isnset{entry-params-dml-full-globe}{0.07}
\isnset{entry-params-dml-full-gt}{0.72}
\isnset{entry-params-dml-full-isla}{0.61}
\isnset{entry-params-dml-full-t}{1.59}
\isnset{fact-af-length-chords}{305}
\isnset{fact-dml-length-car}{2}
\isnset{fact-dml-length-tunnel}{59.5}
\isnset{fact-tests-files}{31}
\isnset{fact-tests-gpu-passed}{1{,}303}
\isnset{fact-tol-coincident-splitting-exact-read}{$10^{-10}$}
\isnset{fact-tol-measure-rescaling}{$10^{-9}$}
\isnset{fact-tol-permutation}{$10^{-10}$}
\isnset{fact-tol-query-independence}{$10^{-12}$}
\isnset{fact-tol-reflection}{$10^{-11}$}
\isnset{fact-tol-rotation}{$10^{-10}$}
\isnset{fact-tol-translation}{$10^{-10}$}
\isnset{fact-tol-uniform-rescaling}{$10^{-10}$}
\isnset{fact-tree-dev-max-pct}{3.5}
\isnset{fact-tree-dev-min-pct}{1.4}
\isnset{fact-tree-sources}{680}
\isnset{fb-ahmedml-full-test-geotransolver}{87.55}
\isnset{fb-ahmedml-full-test-geotransolver-sp}{3.35}
\isnset{fb-ahmedml-full-test-geotransolver-sw}{5.36}
\isnset{fb-ahmedml-full-test-geotransolver-vp}{3.21}
\isnset{fb-ahmedml-full-test-geotransolver-vu}{1.73}
\isnset{fb-ahmedml-full-test-globe}{84.55}
\isnset{fb-ahmedml-full-test-globe-sp}{3.64}
\isnset{fb-ahmedml-full-test-globe-sw}{5.13}
\isnset{fb-ahmedml-full-test-globe-vp}{3.98}
\isnset{fb-ahmedml-full-test-globe-vu}{2.40}
\isnset{fb-ahmedml-full-test-isla}{87.37}
\isnset{fb-ahmedml-full-test-isla-sp}{2.95}
\isnset{fb-ahmedml-full-test-isla-sw}{4.97}
\isnset{fb-ahmedml-full-test-isla-vp}{3.20}
\isnset{fb-ahmedml-full-test-isla-vu}{2.18}
\isnset{fb-ahmedml-full-test-lead}{0.2}
\isnset{fb-ahmedml-full-test-transolver}{87.29}
\isnset{fb-ahmedml-full-test-transolver-sp}{3.41}
\isnset{fb-ahmedml-full-test-transolver-sw}{5.35}
\isnset{fb-ahmedml-full-test-transolver-vp}{3.23}
\isnset{fb-ahmedml-full-test-transolver-vu}{1.71}
\isnset{fb-ahmedml-full-validation-geotransolver}{88.54}
\isnset{fb-ahmedml-full-validation-geotransolver-sp}{3.34}
\isnset{fb-ahmedml-full-validation-geotransolver-sw}{5.36}
\isnset{fb-ahmedml-full-validation-geotransolver-vp}{3.22}
\isnset{fb-ahmedml-full-validation-geotransolver-vu}{1.79}
\isnset{fb-ahmedml-full-validation-globe}{84.03}
\isnset{fb-ahmedml-full-validation-globe-sp}{3.71}
\isnset{fb-ahmedml-full-validation-globe-sw}{5.26}
\isnset{fb-ahmedml-full-validation-globe-vp}{3.99}
\isnset{fb-ahmedml-full-validation-globe-vu}{2.49}
\isnset{fb-ahmedml-full-validation-isla}{87.80}
\isnset{fb-ahmedml-full-validation-isla-sp}{3.07}
\isnset{fb-ahmedml-full-validation-isla-sw}{5.15}
\isnset{fb-ahmedml-full-validation-isla-vp}{3.55}
\isnset{fb-ahmedml-full-validation-isla-vu}{2.32}
\isnset{fb-ahmedml-full-validation-lead}{0.7}
\isnset{fb-ahmedml-full-validation-transolver}{87.77}
\isnset{fb-ahmedml-full-validation-transolver-sp}{3.59}
\isnset{fb-ahmedml-full-validation-transolver-sw}{5.57}
\isnset{fb-ahmedml-full-validation-transolver-vp}{3.25}
\isnset{fb-ahmedml-full-validation-transolver-vu}{1.80}
\isnset{fb-ahmedml-geometry-test-geotransolver}{85.38}
\isnset{fb-ahmedml-geometry-test-geotransolver-sp}{3.88}
\isnset{fb-ahmedml-geometry-test-geotransolver-sw}{5.44}
\isnset{fb-ahmedml-geometry-test-geotransolver-vp}{3.68}
\isnset{fb-ahmedml-geometry-test-geotransolver-vu}{2.17}
\isnset{fb-ahmedml-geometry-test-globe}{83.71}
\isnset{fb-ahmedml-geometry-test-globe-sp}{3.95}
\isnset{fb-ahmedml-geometry-test-globe-sw}{5.24}
\isnset{fb-ahmedml-geometry-test-globe-vp}{4.55}
\isnset{fb-ahmedml-geometry-test-globe-vu}{2.93}
\isnset{fb-ahmedml-geometry-test-isla}{84.99}
\isnset{fb-ahmedml-geometry-test-isla-sp}{3.46}
\isnset{fb-ahmedml-geometry-test-isla-sw}{5.29}
\isnset{fb-ahmedml-geometry-test-isla-vp}{3.74}
\isnset{fb-ahmedml-geometry-test-isla-vu}{2.60}
\isnset{fb-ahmedml-geometry-test-lead}{0.4}
\isnset{fb-ahmedml-geometry-test-transolver}{85.16}
\isnset{fb-ahmedml-geometry-test-transolver-sp}{3.92}
\isnset{fb-ahmedml-geometry-test-transolver-sw}{5.50}
\isnset{fb-ahmedml-geometry-test-transolver-vp}{3.69}
\isnset{fb-ahmedml-geometry-test-transolver-vu}{2.18}
\isnset{fb-ahmedml-geometry-validation-geotransolver}{87.30}
\isnset{fb-ahmedml-geometry-validation-geotransolver-sp}{3.26}
\isnset{fb-ahmedml-geometry-validation-geotransolver-sw}{5.13}
\isnset{fb-ahmedml-geometry-validation-geotransolver-vp}{3.47}
\isnset{fb-ahmedml-geometry-validation-geotransolver-vu}{1.80}
\isnset{fb-ahmedml-geometry-validation-globe}{80.58}
\isnset{fb-ahmedml-geometry-validation-globe-sp}{3.96}
\isnset{fb-ahmedml-geometry-validation-globe-sw}{5.10}
\isnset{fb-ahmedml-geometry-validation-globe-vp}{4.48}
\isnset{fb-ahmedml-geometry-validation-globe-vu}{2.58}
\isnset{fb-ahmedml-geometry-validation-isla}{86.09}
\isnset{fb-ahmedml-geometry-validation-isla-sp}{3.21}
\isnset{fb-ahmedml-geometry-validation-isla-sw}{5.08}
\isnset{fb-ahmedml-geometry-validation-isla-vp}{3.64}
\isnset{fb-ahmedml-geometry-validation-isla-vu}{2.38}
\isnset{fb-ahmedml-geometry-validation-lead}{1.6}
\isnset{fb-ahmedml-geometry-validation-transolver}{87.73}
\isnset{fb-ahmedml-geometry-validation-transolver-sp}{3.25}
\isnset{fb-ahmedml-geometry-validation-transolver-sw}{5.08}
\isnset{fb-ahmedml-geometry-validation-transolver-vp}{3.42}
\isnset{fb-ahmedml-geometry-validation-transolver-vu}{1.80}
\isnset{fb-ahmedml-high-drag-test-geotransolver}{53.36}
\isnset{fb-ahmedml-high-drag-test-geotransolver-sp}{8.42}
\isnset{fb-ahmedml-high-drag-test-geotransolver-sw}{7.67}
\isnset{fb-ahmedml-high-drag-test-geotransolver-vp}{10.10}
\isnset{fb-ahmedml-high-drag-test-geotransolver-vu}{4.91}
\isnset{fb-ahmedml-high-drag-test-globe}{53.10}
\isnset{fb-ahmedml-high-drag-test-globe-sp}{8.52}
\isnset{fb-ahmedml-high-drag-test-globe-sw}{7.24}
\isnset{fb-ahmedml-high-drag-test-globe-vp}{10.45}
\isnset{fb-ahmedml-high-drag-test-globe-vu}{5.27}
\isnset{fb-ahmedml-high-drag-test-isla}{52.05}
\isnset{fb-ahmedml-high-drag-test-isla-sp}{8.21}
\isnset{fb-ahmedml-high-drag-test-isla-sw}{7.37}
\isnset{fb-ahmedml-high-drag-test-isla-vp}{10.88}
\isnset{fb-ahmedml-high-drag-test-isla-vu}{5.56}
\isnset{fb-ahmedml-high-drag-test-lead}{1.3}
\isnset{fb-ahmedml-high-drag-test-max}{53}
\isnset{fb-ahmedml-high-drag-test-min}{52}
\isnset{fb-ahmedml-high-drag-test-n-models}{4}
\isnset{fb-ahmedml-high-drag-test-transolver}{53.27}
\isnset{fb-ahmedml-high-drag-test-transolver-sp}{8.74}
\isnset{fb-ahmedml-high-drag-test-transolver-sw}{7.64}
\isnset{fb-ahmedml-high-drag-test-transolver-vp}{9.51}
\isnset{fb-ahmedml-high-drag-test-transolver-vu}{4.71}
\isnset{fb-ahmedml-high-drag-validation-geotransolver}{88.21}
\isnset{fb-ahmedml-high-drag-validation-geotransolver-sp}{2.73}
\isnset{fb-ahmedml-high-drag-validation-geotransolver-sw}{4.88}
\isnset{fb-ahmedml-high-drag-validation-geotransolver-vp}{3.21}
\isnset{fb-ahmedml-high-drag-validation-geotransolver-vu}{1.68}
\isnset{fb-ahmedml-high-drag-validation-globe}{87.49}
\isnset{fb-ahmedml-high-drag-validation-globe-sp}{2.77}
\isnset{fb-ahmedml-high-drag-validation-globe-sw}{4.52}
\isnset{fb-ahmedml-high-drag-validation-globe-vp}{3.59}
\isnset{fb-ahmedml-high-drag-validation-globe-vu}{2.15}
\isnset{fb-ahmedml-high-drag-validation-isla}{84.34}
\isnset{fb-ahmedml-high-drag-validation-lead}{3.9}
\isnset{fb-ahmedml-high-drag-validation-max}{88}
\isnset{fb-ahmedml-high-drag-validation-min}{87}
\isnset{fb-ahmedml-high-drag-validation-transolver}{87.37}
\isnset{fb-ahmedml-high-drag-validation-transolver-sp}{2.85}
\isnset{fb-ahmedml-high-drag-validation-transolver-sw}{4.81}
\isnset{fb-ahmedml-high-drag-validation-transolver-vp}{3.22}
\isnset{fb-ahmedml-high-drag-validation-transolver-vu}{1.68}
\isnset{fb-ahmedml-image-wake-test-geotransolver}{11.26}
\isnset{fb-ahmedml-image-wake-test-geotransolver-sp}{22.01}
\isnset{fb-ahmedml-image-wake-test-geotransolver-sw}{15.08}
\isnset{fb-ahmedml-image-wake-test-geotransolver-vp}{25.22}
\isnset{fb-ahmedml-image-wake-test-geotransolver-vu}{11.07}
\isnset{fb-ahmedml-image-wake-test-globe}{12.07}
\isnset{fb-ahmedml-image-wake-test-globe-sp}{21.86}
\isnset{fb-ahmedml-image-wake-test-globe-sw}{15.00}
\isnset{fb-ahmedml-image-wake-test-globe-vp}{23.90}
\isnset{fb-ahmedml-image-wake-test-globe-vu}{10.51}
\isnset{fb-ahmedml-image-wake-test-isla}{9.46}
\isnset{fb-ahmedml-image-wake-test-isla-sp}{23.15}
\isnset{fb-ahmedml-image-wake-test-isla-sw}{16.22}
\isnset{fb-ahmedml-image-wake-test-isla-vp}{26.58}
\isnset{fb-ahmedml-image-wake-test-isla-vu}{11.90}
\isnset{fb-ahmedml-image-wake-test-lead}{2.6}
\isnset{fb-ahmedml-image-wake-test-max}{12}
\isnset{fb-ahmedml-image-wake-test-min}{9}
\isnset{fb-ahmedml-image-wake-test-n-models}{4}
\isnset{fb-ahmedml-image-wake-test-transolver}{10.87}
\isnset{fb-ahmedml-image-wake-test-transolver-sp}{22.10}
\isnset{fb-ahmedml-image-wake-test-transolver-sw}{15.02}
\isnset{fb-ahmedml-image-wake-test-transolver-vp}{25.87}
\isnset{fb-ahmedml-image-wake-test-transolver-vu}{11.42}
\isnset{fb-ahmedml-image-wake-validation-geotransolver}{89.96}
\isnset{fb-ahmedml-image-wake-validation-geotransolver-sp}{2.38}
\isnset{fb-ahmedml-image-wake-validation-geotransolver-sw}{4.96}
\isnset{fb-ahmedml-image-wake-validation-geotransolver-vp}{2.27}
\isnset{fb-ahmedml-image-wake-validation-geotransolver-vu}{1.33}
\isnset{fb-ahmedml-image-wake-validation-globe}{89.22}
\isnset{fb-ahmedml-image-wake-validation-globe-sp}{2.18}
\isnset{fb-ahmedml-image-wake-validation-globe-sw}{4.38}
\isnset{fb-ahmedml-image-wake-validation-globe-vp}{2.77}
\isnset{fb-ahmedml-image-wake-validation-globe-vu}{1.99}
\isnset{fb-ahmedml-image-wake-validation-isla}{89.96}
\isnset{fb-ahmedml-image-wake-validation-isla-sp}{2.04}
\isnset{fb-ahmedml-image-wake-validation-isla-sw}{4.81}
\isnset{fb-ahmedml-image-wake-validation-isla-vp}{2.26}
\isnset{fb-ahmedml-image-wake-validation-isla-vu}{1.89}
\isnset{fb-ahmedml-image-wake-validation-lead}{0.0}
\isnset{fb-ahmedml-image-wake-validation-max}{90}
\isnset{fb-ahmedml-image-wake-validation-min}{89}
\isnset{fb-ahmedml-image-wake-validation-transolver}{89.68}
\isnset{fb-ahmedml-image-wake-validation-transolver-sp}{2.44}
\isnset{fb-ahmedml-image-wake-validation-transolver-sw}{4.98}
\isnset{fb-ahmedml-image-wake-validation-transolver-vp}{2.33}
\isnset{fb-ahmedml-image-wake-validation-transolver-vu}{1.36}
\isnset{fb-ahmedml-low-drag-test-geotransolver}{85.09}
\isnset{fb-ahmedml-low-drag-test-geotransolver-sp}{2.85}
\isnset{fb-ahmedml-low-drag-test-geotransolver-sw}{4.58}
\isnset{fb-ahmedml-low-drag-test-geotransolver-vp}{2.74}
\isnset{fb-ahmedml-low-drag-test-geotransolver-vu}{1.36}
\isnset{fb-ahmedml-low-drag-test-globe}{76.55}
\isnset{fb-ahmedml-low-drag-test-globe-sp}{3.30}
\isnset{fb-ahmedml-low-drag-test-globe-sw}{4.58}
\isnset{fb-ahmedml-low-drag-test-globe-vp}{4.31}
\isnset{fb-ahmedml-low-drag-test-globe-vu}{2.26}
\isnset{fb-ahmedml-low-drag-test-isla}{83.75}
\isnset{fb-ahmedml-low-drag-test-isla-sp}{2.71}
\isnset{fb-ahmedml-low-drag-test-isla-sw}{4.74}
\isnset{fb-ahmedml-low-drag-test-isla-vp}{3.20}
\isnset{fb-ahmedml-low-drag-test-isla-vu}{1.96}
\isnset{fb-ahmedml-low-drag-test-lead}{1.3}
\isnset{fb-ahmedml-low-drag-test-transolver}{82.73}
\isnset{fb-ahmedml-low-drag-test-transolver-sp}{3.01}
\isnset{fb-ahmedml-low-drag-test-transolver-sw}{4.74}
\isnset{fb-ahmedml-low-drag-test-transolver-vp}{2.84}
\isnset{fb-ahmedml-low-drag-test-transolver-vu}{1.40}
\isnset{fb-ahmedml-low-drag-validation-geotransolver}{81.96}
\isnset{fb-ahmedml-low-drag-validation-geotransolver-sp}{3.72}
\isnset{fb-ahmedml-low-drag-validation-geotransolver-sw}{5.46}
\isnset{fb-ahmedml-low-drag-validation-geotransolver-vp}{3.42}
\isnset{fb-ahmedml-low-drag-validation-geotransolver-vu}{1.98}
\isnset{fb-ahmedml-low-drag-validation-globe}{85.73}
\isnset{fb-ahmedml-low-drag-validation-globe-sp}{3.42}
\isnset{fb-ahmedml-low-drag-validation-globe-sw}{4.89}
\isnset{fb-ahmedml-low-drag-validation-globe-vp}{4.06}
\isnset{fb-ahmedml-low-drag-validation-globe-vu}{2.63}
\isnset{fb-ahmedml-low-drag-validation-isla}{78.57}
\isnset{fb-ahmedml-low-drag-validation-lead}{7.2}
\isnset{fb-ahmedml-low-drag-validation-transolver}{79.60}
\isnset{fb-ahmedml-low-drag-validation-transolver-sp}{3.86}
\isnset{fb-ahmedml-low-drag-validation-transolver-sw}{5.64}
\isnset{fb-ahmedml-low-drag-validation-transolver-vp}{3.37}
\isnset{fb-ahmedml-low-drag-validation-transolver-vu}{1.96}
\isnset{fb-ahmedml-medium-test-geotransolver}{83.17}
\isnset{fb-ahmedml-medium-test-geotransolver-sp}{3.73}
\isnset{fb-ahmedml-medium-test-geotransolver-sw}{5.13}
\isnset{fb-ahmedml-medium-test-geotransolver-vp}{4.00}
\isnset{fb-ahmedml-medium-test-geotransolver-vu}{2.16}
\isnset{fb-ahmedml-medium-test-globe}{83.69}
\isnset{fb-ahmedml-medium-test-globe-sp}{3.69}
\isnset{fb-ahmedml-medium-test-globe-sw}{5.18}
\isnset{fb-ahmedml-medium-test-globe-vp}{4.44}
\isnset{fb-ahmedml-medium-test-globe-vu}{2.64}
\isnset{fb-ahmedml-medium-test-isla}{85.77}
\isnset{fb-ahmedml-medium-test-isla-sp}{3.43}
\isnset{fb-ahmedml-medium-test-isla-sw}{5.25}
\isnset{fb-ahmedml-medium-test-isla-vp}{3.57}
\isnset{fb-ahmedml-medium-test-isla-vu}{2.37}
\isnset{fb-ahmedml-medium-test-lead}{2.1}
\isnset{fb-ahmedml-medium-test-transolver}{81.05}
\isnset{fb-ahmedml-medium-test-transolver-sp}{4.08}
\isnset{fb-ahmedml-medium-test-transolver-sw}{5.33}
\isnset{fb-ahmedml-medium-test-transolver-vp}{4.06}
\isnset{fb-ahmedml-medium-test-transolver-vu}{2.16}
\isnset{fb-ahmedml-medium-validation-geotransolver}{87.43}
\isnset{fb-ahmedml-medium-validation-geotransolver-sp}{3.55}
\isnset{fb-ahmedml-medium-validation-geotransolver-sw}{5.37}
\isnset{fb-ahmedml-medium-validation-geotransolver-vp}{3.71}
\isnset{fb-ahmedml-medium-validation-geotransolver-vu}{2.15}
\isnset{fb-ahmedml-medium-validation-globe}{86.11}
\isnset{fb-ahmedml-medium-validation-globe-sp}{3.50}
\isnset{fb-ahmedml-medium-validation-globe-sw}{5.13}
\isnset{fb-ahmedml-medium-validation-globe-vp}{4.21}
\isnset{fb-ahmedml-medium-validation-globe-vu}{2.55}
\isnset{fb-ahmedml-medium-validation-isla}{87.69}
\isnset{fb-ahmedml-medium-validation-isla-sp}{3.13}
\isnset{fb-ahmedml-medium-validation-isla-sw}{5.23}
\isnset{fb-ahmedml-medium-validation-isla-vp}{3.54}
\isnset{fb-ahmedml-medium-validation-isla-vu}{2.39}
\isnset{fb-ahmedml-medium-validation-lead}{0.3}
\isnset{fb-ahmedml-medium-validation-transolver}{87.05}
\isnset{fb-ahmedml-medium-validation-transolver-sp}{3.76}
\isnset{fb-ahmedml-medium-validation-transolver-sw}{5.56}
\isnset{fb-ahmedml-medium-validation-transolver-vp}{3.66}
\isnset{fb-ahmedml-medium-validation-transolver-vu}{2.10}
\isnset{fb-ahmedml-scarce-test-geotransolver}{72.47}
\isnset{fb-ahmedml-scarce-test-geotransolver-sp}{5.01}
\isnset{fb-ahmedml-scarce-test-geotransolver-sw}{5.90}
\isnset{fb-ahmedml-scarce-test-geotransolver-vp}{4.41}
\isnset{fb-ahmedml-scarce-test-geotransolver-vu}{2.48}
\isnset{fb-ahmedml-scarce-test-globe}{82.52}
\isnset{fb-ahmedml-scarce-test-globe-sp}{4.03}
\isnset{fb-ahmedml-scarce-test-globe-sw}{5.20}
\isnset{fb-ahmedml-scarce-test-globe-vp}{4.64}
\isnset{fb-ahmedml-scarce-test-globe-vu}{2.69}
\isnset{fb-ahmedml-scarce-test-isla}{78.13}
\isnset{fb-ahmedml-scarce-test-isla-sp}{4.33}
\isnset{fb-ahmedml-scarce-test-isla-sw}{5.54}
\isnset{fb-ahmedml-scarce-test-isla-vp}{4.11}
\isnset{fb-ahmedml-scarce-test-isla-vu}{2.56}
\isnset{fb-ahmedml-scarce-test-lead}{4.4}
\isnset{fb-ahmedml-scarce-test-transolver}{72.70}
\isnset{fb-ahmedml-scarce-test-transolver-sp}{4.86}
\isnset{fb-ahmedml-scarce-test-transolver-sw}{5.90}
\isnset{fb-ahmedml-scarce-test-transolver-vp}{4.27}
\isnset{fb-ahmedml-scarce-test-transolver-vu}{2.43}
\isnset{fb-ahmedml-scarce-validation-geotransolver}{81.28}
\isnset{fb-ahmedml-scarce-validation-geotransolver-sp}{4.74}
\isnset{fb-ahmedml-scarce-validation-geotransolver-sw}{6.19}
\isnset{fb-ahmedml-scarce-validation-geotransolver-vp}{5.06}
\isnset{fb-ahmedml-scarce-validation-geotransolver-vu}{2.97}
\isnset{fb-ahmedml-scarce-validation-globe}{82.19}
\isnset{fb-ahmedml-scarce-validation-globe-sp}{4.74}
\isnset{fb-ahmedml-scarce-validation-globe-sw}{5.35}
\isnset{fb-ahmedml-scarce-validation-globe-vp}{5.37}
\isnset{fb-ahmedml-scarce-validation-globe-vu}{3.02}
\isnset{fb-ahmedml-scarce-validation-isla}{83.93}
\isnset{fb-ahmedml-scarce-validation-isla-sp}{4.16}
\isnset{fb-ahmedml-scarce-validation-isla-sw}{5.82}
\isnset{fb-ahmedml-scarce-validation-isla-vp}{4.86}
\isnset{fb-ahmedml-scarce-validation-isla-vu}{2.91}
\isnset{fb-ahmedml-scarce-validation-lead}{1.7}
\isnset{fb-ahmedml-scarce-validation-transolver}{81.88}
\isnset{fb-ahmedml-scarce-validation-transolver-sp}{4.78}
\isnset{fb-ahmedml-scarce-validation-transolver-sw}{6.22}
\isnset{fb-ahmedml-scarce-validation-transolver-vp}{5.00}
\isnset{fb-ahmedml-scarce-validation-transolver-vu}{2.90}
\isnset{fb-ahmedml-super-scarce-test-geotransolver}{54.24}
\isnset{fb-ahmedml-super-scarce-test-geotransolver-sp}{8.96}
\isnset{fb-ahmedml-super-scarce-test-geotransolver-sw}{8.58}
\isnset{fb-ahmedml-super-scarce-test-geotransolver-vp}{10.75}
\isnset{fb-ahmedml-super-scarce-test-geotransolver-vu}{5.29}
\isnset{fb-ahmedml-super-scarce-test-globe}{60.52}
\isnset{fb-ahmedml-super-scarce-test-globe-sp}{8.43}
\isnset{fb-ahmedml-super-scarce-test-globe-sw}{7.76}
\isnset{fb-ahmedml-super-scarce-test-globe-vp}{8.68}
\isnset{fb-ahmedml-super-scarce-test-globe-vu}{5.68}
\isnset{fb-ahmedml-super-scarce-test-isla}{62.03}
\isnset{fb-ahmedml-super-scarce-test-isla-sp}{8.49}
\isnset{fb-ahmedml-super-scarce-test-isla-sw}{8.12}
\isnset{fb-ahmedml-super-scarce-test-isla-vp}{9.84}
\isnset{fb-ahmedml-super-scarce-test-isla-vu}{4.80}
\isnset{fb-ahmedml-super-scarce-test-lead}{1.5}
\isnset{fb-ahmedml-super-scarce-test-transolver}{50.04}
\isnset{fb-ahmedml-super-scarce-test-transolver-sp}{9.66}
\isnset{fb-ahmedml-super-scarce-test-transolver-sw}{9.05}
\isnset{fb-ahmedml-super-scarce-test-transolver-vp}{10.27}
\isnset{fb-ahmedml-super-scarce-test-transolver-vu}{5.19}
\isnset{fb-ahmedml-super-scarce-validation-geotransolver}{64.17}
\isnset{fb-ahmedml-super-scarce-validation-geotransolver-sp}{8.25}
\isnset{fb-ahmedml-super-scarce-validation-geotransolver-sw}{7.73}
\isnset{fb-ahmedml-super-scarce-validation-geotransolver-vp}{11.46}
\isnset{fb-ahmedml-super-scarce-validation-geotransolver-vu}{5.31}
\isnset{fb-ahmedml-super-scarce-validation-globe}{62.04}
\isnset{fb-ahmedml-super-scarce-validation-globe-sp}{9.05}
\isnset{fb-ahmedml-super-scarce-validation-globe-sw}{7.33}
\isnset{fb-ahmedml-super-scarce-validation-globe-vp}{11.08}
\isnset{fb-ahmedml-super-scarce-validation-globe-vu}{6.26}
\isnset{fb-ahmedml-super-scarce-validation-isla}{68.85}
\isnset{fb-ahmedml-super-scarce-validation-isla-sp}{8.10}
\isnset{fb-ahmedml-super-scarce-validation-isla-sw}{7.61}
\isnset{fb-ahmedml-super-scarce-validation-isla-vp}{9.43}
\isnset{fb-ahmedml-super-scarce-validation-isla-vu}{4.48}
\isnset{fb-ahmedml-super-scarce-validation-lead}{4.7}
\isnset{fb-ahmedml-super-scarce-validation-transolver}{61.00}
\isnset{fb-ahmedml-super-scarce-validation-transolver-sp}{9.20}
\isnset{fb-ahmedml-super-scarce-validation-transolver-sw}{8.19}
\isnset{fb-ahmedml-super-scarce-validation-transolver-vp}{11.01}
\isnset{fb-ahmedml-super-scarce-validation-transolver-vu}{5.35}
\isnset{fb-airfrans-aoa-extrapolation-isla-n-members}{13}
\isnset{fb-airfrans-aoa-extrapolation-isla-n-refit}{6}
\isnset{fb-airfrans-aoa-extrapolation-test-geotransolver}{63.80}
\isnset{fb-airfrans-aoa-extrapolation-test-geotransolver-sp}{9.29}
\isnset{fb-airfrans-aoa-extrapolation-test-geotransolver-sw}{6.60}
\isnset{fb-airfrans-aoa-extrapolation-test-geotransolver-vp}{13.02}
\isnset{fb-airfrans-aoa-extrapolation-test-geotransolver-vu}{3.91}
\isnset{fb-airfrans-aoa-extrapolation-test-globe}{90.05}
\isnset{fb-airfrans-aoa-extrapolation-test-globe-sp}{1.94}
\isnset{fb-airfrans-aoa-extrapolation-test-globe-sw}{2.14}
\isnset{fb-airfrans-aoa-extrapolation-test-globe-vp}{3.87}
\isnset{fb-airfrans-aoa-extrapolation-test-globe-vu}{0.98}
\isnset{fb-airfrans-aoa-extrapolation-test-isla}{86.05}
\isnset{fb-airfrans-aoa-extrapolation-test-isla-sp}{2.58}
\isnset{fb-airfrans-aoa-extrapolation-test-isla-sw}{2.67}
\isnset{fb-airfrans-aoa-extrapolation-test-isla-vp}{4.55}
\isnset{fb-airfrans-aoa-extrapolation-test-isla-vu}{1.37}
\isnset{fb-airfrans-aoa-extrapolation-test-lead}{4.0}
\isnset{fb-airfrans-aoa-extrapolation-test-transolver}{60.26}
\isnset{fb-airfrans-aoa-extrapolation-test-transolver-sp}{8.31}
\isnset{fb-airfrans-aoa-extrapolation-test-transolver-sw}{6.44}
\isnset{fb-airfrans-aoa-extrapolation-test-transolver-vp}{12.21}
\isnset{fb-airfrans-aoa-extrapolation-test-transolver-vu}{3.84}
\isnset{fb-airfrans-aoa-extrapolation-validation-geotransolver}{93.91}
\isnset{fb-airfrans-aoa-extrapolation-validation-geotransolver-sp}{0.98}
\isnset{fb-airfrans-aoa-extrapolation-validation-geotransolver-sw}{1.27}
\isnset{fb-airfrans-aoa-extrapolation-validation-geotransolver-vp}{1.03}
\isnset{fb-airfrans-aoa-extrapolation-validation-geotransolver-vu}{0.43}
\isnset{fb-airfrans-aoa-extrapolation-validation-globe}{98.98}
\isnset{fb-airfrans-aoa-extrapolation-validation-globe-sp}{0.30}
\isnset{fb-airfrans-aoa-extrapolation-validation-globe-sw}{0.54}
\isnset{fb-airfrans-aoa-extrapolation-validation-globe-vp}{0.25}
\isnset{fb-airfrans-aoa-extrapolation-validation-globe-vu}{0.12}
\isnset{fb-airfrans-aoa-extrapolation-validation-isla}{97.90}
\isnset{fb-airfrans-aoa-extrapolation-validation-isla-sp}{0.52}
\isnset{fb-airfrans-aoa-extrapolation-validation-isla-sw}{0.62}
\isnset{fb-airfrans-aoa-extrapolation-validation-isla-vp}{0.46}
\isnset{fb-airfrans-aoa-extrapolation-validation-isla-vu}{0.21}
\isnset{fb-airfrans-aoa-extrapolation-validation-lead}{1.1}
\isnset{fb-airfrans-aoa-extrapolation-validation-transolver}{94.89}
\isnset{fb-airfrans-aoa-extrapolation-validation-transolver-sp}{0.82}
\isnset{fb-airfrans-aoa-extrapolation-validation-transolver-sw}{1.20}
\isnset{fb-airfrans-aoa-extrapolation-validation-transolver-vp}{0.95}
\isnset{fb-airfrans-aoa-extrapolation-validation-transolver-vu}{0.39}
\isnset{fb-airfrans-full-isla-n-members}{17}
\isnset{fb-airfrans-full-isla-n-refit}{6}
\isnset{fb-airfrans-full-test-geotransolver}{95.57}
\isnset{fb-airfrans-full-test-geotransolver-sp}{1.20}
\isnset{fb-airfrans-full-test-geotransolver-sw}{1.46}
\isnset{fb-airfrans-full-test-geotransolver-vp}{1.09}
\isnset{fb-airfrans-full-test-geotransolver-vu}{0.47}
\isnset{fb-airfrans-full-test-globe}{98.87}
\isnset{fb-airfrans-full-test-globe-sp}{0.34}
\isnset{fb-airfrans-full-test-globe-sw}{0.55}
\isnset{fb-airfrans-full-test-globe-vp}{0.28}
\isnset{fb-airfrans-full-test-globe-vu}{0.15}
\isnset{fb-airfrans-full-test-isla}{97.77}
\isnset{fb-airfrans-full-test-isla-sp}{0.55}
\isnset{fb-airfrans-full-test-isla-sw}{0.57}
\isnset{fb-airfrans-full-test-isla-vp}{0.57}
\isnset{fb-airfrans-full-test-isla-vu}{0.30}
\isnset{fb-airfrans-full-test-lead}{1.1}
\isnset{fb-airfrans-full-test-transolver}{94.96}
\isnset{fb-airfrans-full-test-transolver-sp}{1.17}
\isnset{fb-airfrans-full-test-transolver-sw}{1.76}
\isnset{fb-airfrans-full-test-transolver-vp}{1.14}
\isnset{fb-airfrans-full-test-transolver-vu}{0.53}
\isnset{fb-airfrans-full-validation-geotransolver}{95.93}
\isnset{fb-airfrans-full-validation-geotransolver-sp}{1.34}
\isnset{fb-airfrans-full-validation-geotransolver-sw}{1.52}
\isnset{fb-airfrans-full-validation-geotransolver-vp}{1.27}
\isnset{fb-airfrans-full-validation-geotransolver-vu}{0.50}
\isnset{fb-airfrans-full-validation-globe}{98.87}
\isnset{fb-airfrans-full-validation-globe-sp}{0.37}
\isnset{fb-airfrans-full-validation-globe-sw}{0.57}
\isnset{fb-airfrans-full-validation-globe-vp}{0.34}
\isnset{fb-airfrans-full-validation-globe-vu}{0.16}
\isnset{fb-airfrans-full-validation-isla}{98.27}
\isnset{fb-airfrans-full-validation-isla-sp}{0.57}
\isnset{fb-airfrans-full-validation-isla-sw}{0.61}
\isnset{fb-airfrans-full-validation-isla-vp}{0.57}
\isnset{fb-airfrans-full-validation-isla-vu}{0.27}
\isnset{fb-airfrans-full-validation-lead}{0.6}
\isnset{fb-airfrans-full-validation-transolver}{95.88}
\isnset{fb-airfrans-full-validation-transolver-sp}{1.22}
\isnset{fb-airfrans-full-validation-transolver-sw}{1.84}
\isnset{fb-airfrans-full-validation-transolver-vp}{1.26}
\isnset{fb-airfrans-full-validation-transolver-vu}{0.56}
\isnset{fb-airfrans-scarce-test-geotransolver}{85.80}
\isnset{fb-airfrans-scarce-test-geotransolver-sp}{2.87}
\isnset{fb-airfrans-scarce-test-geotransolver-sw}{2.67}
\isnset{fb-airfrans-scarce-test-geotransolver-vp}{2.91}
\isnset{fb-airfrans-scarce-test-geotransolver-vu}{1.04}
\isnset{fb-airfrans-scarce-test-globe}{96.48}
\isnset{fb-airfrans-scarce-test-globe-sp}{0.86}
\isnset{fb-airfrans-scarce-test-globe-sw}{0.98}
\isnset{fb-airfrans-scarce-test-globe-vp}{0.75}
\isnset{fb-airfrans-scarce-test-globe-vu}{0.32}
\isnset{fb-airfrans-scarce-test-isla}{87.93}
\isnset{fb-airfrans-scarce-test-isla-sp}{1.54}
\isnset{fb-airfrans-scarce-test-isla-sw}{1.70}
\isnset{fb-airfrans-scarce-test-isla-vp}{1.96}
\isnset{fb-airfrans-scarce-test-isla-vu}{0.76}
\isnset{fb-airfrans-scarce-test-lead}{8.6}
\isnset{fb-airfrans-scarce-test-transolver}{83.96}
\isnset{fb-airfrans-scarce-test-transolver-sp}{2.99}
\isnset{fb-airfrans-scarce-test-transolver-sw}{2.84}
\isnset{fb-airfrans-scarce-test-transolver-vp}{3.25}
\isnset{fb-airfrans-scarce-test-transolver-vu}{1.11}
\isnset{fb-airfrans-scarce-validation-geotransolver}{85.49}
\isnset{fb-airfrans-scarce-validation-geotransolver-sp}{3.18}
\isnset{fb-airfrans-scarce-validation-geotransolver-sw}{2.80}
\isnset{fb-airfrans-scarce-validation-geotransolver-vp}{3.46}
\isnset{fb-airfrans-scarce-validation-geotransolver-vu}{1.11}
\isnset{fb-airfrans-scarce-validation-globe}{95.65}
\isnset{fb-airfrans-scarce-validation-globe-sp}{1.03}
\isnset{fb-airfrans-scarce-validation-globe-sw}{1.04}
\isnset{fb-airfrans-scarce-validation-globe-vp}{1.05}
\isnset{fb-airfrans-scarce-validation-globe-vu}{0.38}
\isnset{fb-airfrans-scarce-validation-isla}{91.28}
\isnset{fb-airfrans-scarce-validation-isla-sp}{1.55}
\isnset{fb-airfrans-scarce-validation-isla-sw}{1.71}
\isnset{fb-airfrans-scarce-validation-isla-vp}{2.13}
\isnset{fb-airfrans-scarce-validation-isla-vu}{0.76}
\isnset{fb-airfrans-scarce-validation-lead}{4.4}
\isnset{fb-airfrans-scarce-validation-transolver}{77.17}
\isnset{fb-airfrans-scarce-validation-transolver-sp}{3.46}
\isnset{fb-airfrans-scarce-validation-transolver-sw}{3.04}
\isnset{fb-airfrans-scarce-validation-transolver-vp}{3.91}
\isnset{fb-airfrans-scarce-validation-transolver-vu}{1.24}
\isnset{fb-drivaerml-full-test-geotransolver}{86.22}
\isnset{fb-drivaerml-full-test-geotransolver-sp}{3.25}
\isnset{fb-drivaerml-full-test-geotransolver-sw}{5.33}
\isnset{fb-drivaerml-full-test-geotransolver-vp}{2.99}
\isnset{fb-drivaerml-full-test-geotransolver-vu}{4.43}
\isnset{fb-drivaerml-full-test-globe}{75.79}
\isnset{fb-drivaerml-full-test-globe-sp}{6.79}
\isnset{fb-drivaerml-full-test-globe-sw}{7.87}
\isnset{fb-drivaerml-full-test-globe-vp}{5.61}
\isnset{fb-drivaerml-full-test-globe-vu}{7.15}
\isnset{fb-drivaerml-full-test-isla}{85.33}
\isnset{fb-drivaerml-full-test-isla-sp}{3.36}
\isnset{fb-drivaerml-full-test-isla-sw}{5.58}
\isnset{fb-drivaerml-full-test-isla-vp}{3.23}
\isnset{fb-drivaerml-full-test-isla-vu}{4.80}
\isnset{fb-drivaerml-full-test-lead}{0.9}
\isnset{fb-drivaerml-full-test-transolver}{85.90}
\isnset{fb-drivaerml-full-test-transolver-sp}{3.38}
\isnset{fb-drivaerml-full-test-transolver-sw}{5.53}
\isnset{fb-drivaerml-full-test-transolver-vp}{3.04}
\isnset{fb-drivaerml-full-test-transolver-vu}{4.50}
\isnset{fb-drivaerml-full-validation-geotransolver}{85.67}
\isnset{fb-drivaerml-full-validation-geotransolver-sp}{3.60}
\isnset{fb-drivaerml-full-validation-geotransolver-sw}{5.70}
\isnset{fb-drivaerml-full-validation-geotransolver-vp}{3.11}
\isnset{fb-drivaerml-full-validation-geotransolver-vu}{4.43}
\isnset{fb-drivaerml-full-validation-globe}{75.29}
\isnset{fb-drivaerml-full-validation-globe-sp}{7.08}
\isnset{fb-drivaerml-full-validation-globe-sw}{8.18}
\isnset{fb-drivaerml-full-validation-globe-vp}{5.74}
\isnset{fb-drivaerml-full-validation-globe-vu}{7.14}
\isnset{fb-drivaerml-full-validation-isla}{84.88}
\isnset{fb-drivaerml-full-validation-isla-sp}{3.67}
\isnset{fb-drivaerml-full-validation-isla-sw}{5.98}
\isnset{fb-drivaerml-full-validation-isla-vp}{3.28}
\isnset{fb-drivaerml-full-validation-isla-vu}{4.83}
\isnset{fb-drivaerml-full-validation-lead}{0.8}
\isnset{fb-drivaerml-full-validation-transolver}{85.18}
\isnset{fb-drivaerml-full-validation-transolver-sp}{3.72}
\isnset{fb-drivaerml-full-validation-transolver-sw}{5.92}
\isnset{fb-drivaerml-full-validation-transolver-vp}{3.21}
\isnset{fb-drivaerml-full-validation-transolver-vu}{4.56}
\isnset{fb-drivaerml-geometry-test-geotransolver}{83.36}
\isnset{fb-drivaerml-geometry-test-geotransolver-sp}{4.07}
\isnset{fb-drivaerml-geometry-test-geotransolver-sw}{6.51}
\isnset{fb-drivaerml-geometry-test-geotransolver-vp}{3.64}
\isnset{fb-drivaerml-geometry-test-geotransolver-vu}{5.21}
\isnset{fb-drivaerml-geometry-test-globe}{73.24}
\isnset{fb-drivaerml-geometry-test-globe-sp}{7.48}
\isnset{fb-drivaerml-geometry-test-globe-sw}{9.04}
\isnset{fb-drivaerml-geometry-test-globe-vp}{6.19}
\isnset{fb-drivaerml-geometry-test-globe-vu}{7.92}
\isnset{fb-drivaerml-geometry-test-isla}{81.36}
\isnset{fb-drivaerml-geometry-test-isla-sp}{4.20}
\isnset{fb-drivaerml-geometry-test-isla-sw}{6.85}
\isnset{fb-drivaerml-geometry-test-isla-vp}{4.28}
\isnset{fb-drivaerml-geometry-test-isla-vu}{6.20}
\isnset{fb-drivaerml-geometry-test-lead}{2.0}
\isnset{fb-drivaerml-geometry-test-transolver}{82.33}
\isnset{fb-drivaerml-geometry-test-transolver-sp}{4.26}
\isnset{fb-drivaerml-geometry-test-transolver-sw}{6.74}
\isnset{fb-drivaerml-geometry-test-transolver-vp}{3.94}
\isnset{fb-drivaerml-geometry-test-transolver-vu}{5.59}
\isnset{fb-drivaerml-geometry-validation-geotransolver}{86.53}
\isnset{fb-drivaerml-geometry-validation-geotransolver-sp}{3.14}
\isnset{fb-drivaerml-geometry-validation-geotransolver-sw}{5.22}
\isnset{fb-drivaerml-geometry-validation-geotransolver-vp}{2.99}
\isnset{fb-drivaerml-geometry-validation-geotransolver-vu}{4.37}
\isnset{fb-drivaerml-geometry-validation-globe}{76.40}
\isnset{fb-drivaerml-geometry-validation-globe-sp}{6.62}
\isnset{fb-drivaerml-geometry-validation-globe-sw}{7.71}
\isnset{fb-drivaerml-geometry-validation-globe-vp}{5.56}
\isnset{fb-drivaerml-geometry-validation-globe-vu}{7.07}
\isnset{fb-drivaerml-geometry-validation-isla}{85.26}
\isnset{fb-drivaerml-geometry-validation-isla-sp}{3.30}
\isnset{fb-drivaerml-geometry-validation-isla-sw}{5.55}
\isnset{fb-drivaerml-geometry-validation-isla-vp}{3.31}
\isnset{fb-drivaerml-geometry-validation-isla-vu}{4.91}
\isnset{fb-drivaerml-geometry-validation-lead}{1.3}
\isnset{fb-drivaerml-geometry-validation-transolver}{86.03}
\isnset{fb-drivaerml-geometry-validation-transolver-sp}{3.24}
\isnset{fb-drivaerml-geometry-validation-transolver-sw}{5.34}
\isnset{fb-drivaerml-geometry-validation-transolver-vp}{3.11}
\isnset{fb-drivaerml-geometry-validation-transolver-vu}{4.52}
\isnset{fb-drivaerml-high-drag-test-geotransolver}{82.82}
\isnset{fb-drivaerml-high-drag-test-geotransolver-sp}{4.07}
\isnset{fb-drivaerml-high-drag-test-geotransolver-sw}{6.36}
\isnset{fb-drivaerml-high-drag-test-geotransolver-vp}{3.78}
\isnset{fb-drivaerml-high-drag-test-geotransolver-vu}{5.19}
\isnset{fb-drivaerml-high-drag-test-globe}{71.29}
\isnset{fb-drivaerml-high-drag-test-globe-sp}{7.74}
\isnset{fb-drivaerml-high-drag-test-globe-sw}{9.15}
\isnset{fb-drivaerml-high-drag-test-globe-vp}{6.41}
\isnset{fb-drivaerml-high-drag-test-globe-vu}{7.96}
\isnset{fb-drivaerml-high-drag-test-isla}{81.57}
\isnset{fb-drivaerml-high-drag-test-isla-sp}{4.08}
\isnset{fb-drivaerml-high-drag-test-isla-sw}{6.57}
\isnset{fb-drivaerml-high-drag-test-isla-vp}{4.15}
\isnset{fb-drivaerml-high-drag-test-isla-vu}{5.72}
\isnset{fb-drivaerml-high-drag-test-lead}{1.2}
\isnset{fb-drivaerml-high-drag-test-transolver}{81.87}
\isnset{fb-drivaerml-high-drag-test-transolver-sp}{4.27}
\isnset{fb-drivaerml-high-drag-test-transolver-sw}{6.71}
\isnset{fb-drivaerml-high-drag-test-transolver-vp}{4.01}
\isnset{fb-drivaerml-high-drag-test-transolver-vu}{5.46}
\isnset{fb-drivaerml-high-drag-validation-geotransolver}{86.08}
\isnset{fb-drivaerml-high-drag-validation-geotransolver-sp}{3.46}
\isnset{fb-drivaerml-high-drag-validation-geotransolver-sw}{5.53}
\isnset{fb-drivaerml-high-drag-validation-geotransolver-vp}{2.96}
\isnset{fb-drivaerml-high-drag-validation-geotransolver-vu}{4.33}
\isnset{fb-drivaerml-high-drag-validation-globe}{75.86}
\isnset{fb-drivaerml-high-drag-validation-globe-sp}{6.95}
\isnset{fb-drivaerml-high-drag-validation-globe-sw}{7.91}
\isnset{fb-drivaerml-high-drag-validation-globe-vp}{5.55}
\isnset{fb-drivaerml-high-drag-validation-globe-vu}{7.09}
\isnset{fb-drivaerml-high-drag-validation-isla}{85.27}
\isnset{fb-drivaerml-high-drag-validation-isla-sp}{3.55}
\isnset{fb-drivaerml-high-drag-validation-isla-sw}{5.72}
\isnset{fb-drivaerml-high-drag-validation-isla-vp}{3.13}
\isnset{fb-drivaerml-high-drag-validation-isla-vu}{4.74}
\isnset{fb-drivaerml-high-drag-validation-lead}{0.8}
\isnset{fb-drivaerml-high-drag-validation-transolver}{85.88}
\isnset{fb-drivaerml-high-drag-validation-transolver-sp}{3.50}
\isnset{fb-drivaerml-high-drag-validation-transolver-sw}{5.61}
\isnset{fb-drivaerml-high-drag-validation-transolver-vp}{3.01}
\isnset{fb-drivaerml-high-drag-validation-transolver-vu}{4.39}
\isnset{fb-drivaerml-low-drag-test-geotransolver}{82.96}
\isnset{fb-drivaerml-low-drag-test-geotransolver-sp}{3.88}
\isnset{fb-drivaerml-low-drag-test-geotransolver-sw}{6.21}
\isnset{fb-drivaerml-low-drag-test-geotransolver-vp}{3.91}
\isnset{fb-drivaerml-low-drag-test-geotransolver-vu}{5.49}
\isnset{fb-drivaerml-low-drag-test-globe}{72.86}
\isnset{fb-drivaerml-low-drag-test-globe-sp}{7.31}
\isnset{fb-drivaerml-low-drag-test-globe-sw}{8.68}
\isnset{fb-drivaerml-low-drag-test-globe-vp}{6.41}
\isnset{fb-drivaerml-low-drag-test-globe-vu}{8.24}
\isnset{fb-drivaerml-low-drag-test-isla}{79.31}
\isnset{fb-drivaerml-low-drag-test-isla-sp}{4.33}
\isnset{fb-drivaerml-low-drag-test-isla-sw}{7.07}
\isnset{fb-drivaerml-low-drag-test-isla-vp}{4.91}
\isnset{fb-drivaerml-low-drag-test-isla-vu}{7.04}
\isnset{fb-drivaerml-low-drag-test-lead}{3.7}
\isnset{fb-drivaerml-low-drag-test-transolver}{82.35}
\isnset{fb-drivaerml-low-drag-test-transolver-sp}{4.00}
\isnset{fb-drivaerml-low-drag-test-transolver-sw}{6.41}
\isnset{fb-drivaerml-low-drag-test-transolver-vp}{4.11}
\isnset{fb-drivaerml-low-drag-test-transolver-vu}{5.73}
\isnset{fb-drivaerml-low-drag-validation-geotransolver}{85.72}
\isnset{fb-drivaerml-low-drag-validation-geotransolver-sp}{3.49}
\isnset{fb-drivaerml-low-drag-validation-geotransolver-sw}{5.69}
\isnset{fb-drivaerml-low-drag-validation-geotransolver-vp}{2.99}
\isnset{fb-drivaerml-low-drag-validation-geotransolver-vu}{4.45}
\isnset{fb-drivaerml-low-drag-validation-globe}{76.10}
\isnset{fb-drivaerml-low-drag-validation-globe-sp}{6.82}
\isnset{fb-drivaerml-low-drag-validation-globe-sw}{7.89}
\isnset{fb-drivaerml-low-drag-validation-globe-vp}{5.54}
\isnset{fb-drivaerml-low-drag-validation-globe-vu}{6.95}
\isnset{fb-drivaerml-low-drag-validation-isla}{84.76}
\isnset{fb-drivaerml-low-drag-validation-isla-sp}{3.57}
\isnset{fb-drivaerml-low-drag-validation-isla-sw}{5.90}
\isnset{fb-drivaerml-low-drag-validation-isla-vp}{3.34}
\isnset{fb-drivaerml-low-drag-validation-isla-vu}{4.93}
\isnset{fb-drivaerml-low-drag-validation-lead}{1.0}
\isnset{fb-drivaerml-low-drag-validation-transolver}{85.47}
\isnset{fb-drivaerml-low-drag-validation-transolver-sp}{3.53}
\isnset{fb-drivaerml-low-drag-validation-transolver-sw}{5.75}
\isnset{fb-drivaerml-low-drag-validation-transolver-vp}{3.09}
\isnset{fb-drivaerml-low-drag-validation-transolver-vu}{4.57}
\isnset{fb-drivaerml-medium-test-geotransolver}{83.39}
\isnset{fb-drivaerml-medium-test-geotransolver-sp}{3.95}
\isnset{fb-drivaerml-medium-test-geotransolver-sw}{6.38}
\isnset{fb-drivaerml-medium-test-geotransolver-vp}{3.57}
\isnset{fb-drivaerml-medium-test-geotransolver-vu}{5.24}
\isnset{fb-drivaerml-medium-test-globe}{73.83}
\isnset{fb-drivaerml-medium-test-globe-sp}{7.21}
\isnset{fb-drivaerml-medium-test-globe-sw}{8.60}
\isnset{fb-drivaerml-medium-test-globe-vp}{5.95}
\isnset{fb-drivaerml-medium-test-globe-vu}{7.76}
\isnset{fb-drivaerml-medium-test-isla}{81.71}
\isnset{fb-drivaerml-medium-test-isla-sp}{4.12}
\isnset{fb-drivaerml-medium-test-isla-sw}{6.63}
\isnset{fb-drivaerml-medium-test-isla-vp}{4.11}
\isnset{fb-drivaerml-medium-test-isla-vu}{6.00}
\isnset{fb-drivaerml-medium-test-lead}{1.7}
\isnset{fb-drivaerml-medium-test-transolver}{82.96}
\isnset{fb-drivaerml-medium-test-transolver-sp}{4.01}
\isnset{fb-drivaerml-medium-test-transolver-sw}{6.46}
\isnset{fb-drivaerml-medium-test-transolver-vp}{3.74}
\isnset{fb-drivaerml-medium-test-transolver-vu}{5.41}
\isnset{fb-drivaerml-medium-validation-geotransolver}{82.81}
\isnset{fb-drivaerml-medium-validation-geotransolver-sp}{4.30}
\isnset{fb-drivaerml-medium-validation-geotransolver-sw}{6.77}
\isnset{fb-drivaerml-medium-validation-geotransolver-vp}{3.79}
\isnset{fb-drivaerml-medium-validation-geotransolver-vu}{5.19}
\isnset{fb-drivaerml-medium-validation-globe}{72.89}
\isnset{fb-drivaerml-medium-validation-globe-sp}{7.62}
\isnset{fb-drivaerml-medium-validation-globe-sw}{9.18}
\isnset{fb-drivaerml-medium-validation-globe-vp}{6.18}
\isnset{fb-drivaerml-medium-validation-globe-vu}{7.76}
\isnset{fb-drivaerml-medium-validation-isla}{81.59}
\isnset{fb-drivaerml-medium-validation-isla-sp}{4.41}
\isnset{fb-drivaerml-medium-validation-isla-sw}{7.03}
\isnset{fb-drivaerml-medium-validation-isla-vp}{4.10}
\isnset{fb-drivaerml-medium-validation-isla-vu}{5.74}
\isnset{fb-drivaerml-medium-validation-lead}{1.2}
\isnset{fb-drivaerml-medium-validation-transolver}{82.35}
\isnset{fb-drivaerml-medium-validation-transolver-sp}{4.31}
\isnset{fb-drivaerml-medium-validation-transolver-sw}{6.78}
\isnset{fb-drivaerml-medium-validation-transolver-vp}{3.98}
\isnset{fb-drivaerml-medium-validation-transolver-vu}{5.42}
\isnset{fb-drivaerml-rear-separation-test-geotransolver}{84.67}
\isnset{fb-drivaerml-rear-separation-test-geotransolver-sp}{3.65}
\isnset{fb-drivaerml-rear-separation-test-geotransolver-sw}{5.80}
\isnset{fb-drivaerml-rear-separation-test-geotransolver-vp}{3.35}
\isnset{fb-drivaerml-rear-separation-test-geotransolver-vu}{4.86}
\isnset{fb-drivaerml-rear-separation-test-globe}{74.35}
\isnset{fb-drivaerml-rear-separation-test-globe-sp}{7.06}
\isnset{fb-drivaerml-rear-separation-test-globe-sw}{8.26}
\isnset{fb-drivaerml-rear-separation-test-globe-vp}{5.84}
\isnset{fb-drivaerml-rear-separation-test-globe-vu}{7.53}
\isnset{fb-drivaerml-rear-separation-test-isla}{83.09}
\isnset{fb-drivaerml-rear-separation-test-isla-sp}{3.78}
\isnset{fb-drivaerml-rear-separation-test-isla-sw}{6.13}
\isnset{fb-drivaerml-rear-separation-test-isla-vp}{3.81}
\isnset{fb-drivaerml-rear-separation-test-isla-vu}{5.59}
\isnset{fb-drivaerml-rear-separation-test-lead}{1.6}
\isnset{fb-drivaerml-rear-separation-test-transolver}{83.99}
\isnset{fb-drivaerml-rear-separation-test-transolver-sp}{3.76}
\isnset{fb-drivaerml-rear-separation-test-transolver-sw}{5.95}
\isnset{fb-drivaerml-rear-separation-test-transolver-vp}{3.48}
\isnset{fb-drivaerml-rear-separation-test-transolver-vu}{5.08}
\isnset{fb-drivaerml-rear-separation-validation-geotransolver}{85.78}
\isnset{fb-drivaerml-rear-separation-validation-geotransolver-sp}{3.55}
\isnset{fb-drivaerml-rear-separation-validation-geotransolver-sw}{5.54}
\isnset{fb-drivaerml-rear-separation-validation-geotransolver-vp}{3.14}
\isnset{fb-drivaerml-rear-separation-validation-geotransolver-vu}{4.41}
\isnset{fb-drivaerml-rear-separation-validation-globe}{75.31}
\isnset{fb-drivaerml-rear-separation-validation-globe-sp}{7.11}
\isnset{fb-drivaerml-rear-separation-validation-globe-sw}{8.15}
\isnset{fb-drivaerml-rear-separation-validation-globe-vp}{5.66}
\isnset{fb-drivaerml-rear-separation-validation-globe-vu}{7.13}
\isnset{fb-drivaerml-rear-separation-validation-isla}{84.60}
\isnset{fb-drivaerml-rear-separation-validation-isla-sp}{3.69}
\isnset{fb-drivaerml-rear-separation-validation-isla-sw}{5.88}
\isnset{fb-drivaerml-rear-separation-validation-isla-vp}{3.46}
\isnset{fb-drivaerml-rear-separation-validation-isla-vu}{4.92}
\isnset{fb-drivaerml-rear-separation-validation-lead}{1.2}
\isnset{fb-drivaerml-rear-separation-validation-transolver}{85.44}
\isnset{fb-drivaerml-rear-separation-validation-transolver-sp}{3.58}
\isnset{fb-drivaerml-rear-separation-validation-transolver-sw}{5.63}
\isnset{fb-drivaerml-rear-separation-validation-transolver-vp}{3.23}
\isnset{fb-drivaerml-rear-separation-validation-transolver-vu}{4.55}
\isnset{fb-drivaerml-scarce-test-geotransolver}{79.82}
\isnset{fb-drivaerml-scarce-test-geotransolver-sp}{4.68}
\isnset{fb-drivaerml-scarce-test-geotransolver-sw}{7.29}
\isnset{fb-drivaerml-scarce-test-geotransolver-vp}{4.44}
\isnset{fb-drivaerml-scarce-test-geotransolver-vu}{6.34}
\isnset{fb-drivaerml-scarce-test-globe}{70.31}
\isnset{fb-drivaerml-scarce-test-globe-sp}{7.84}
\isnset{fb-drivaerml-scarce-test-globe-sw}{9.66}
\isnset{fb-drivaerml-scarce-test-globe-vp}{6.65}
\isnset{fb-drivaerml-scarce-test-globe-vu}{8.68}
\isnset{fb-drivaerml-scarce-test-isla}{77.47}
\isnset{fb-drivaerml-scarce-test-isla-sp}{4.97}
\isnset{fb-drivaerml-scarce-test-isla-sw}{7.66}
\isnset{fb-drivaerml-scarce-test-isla-vp}{5.09}
\isnset{fb-drivaerml-scarce-test-isla-vu}{7.33}
\isnset{fb-drivaerml-scarce-test-lead}{2.4}
\isnset{fb-drivaerml-scarce-test-transolver}{78.88}
\isnset{fb-drivaerml-scarce-test-transolver-sp}{4.88}
\isnset{fb-drivaerml-scarce-test-transolver-sw}{7.47}
\isnset{fb-drivaerml-scarce-test-transolver-vp}{4.73}
\isnset{fb-drivaerml-scarce-test-transolver-vu}{6.58}
\isnset{fb-drivaerml-scarce-validation-geotransolver}{79.18}
\isnset{fb-drivaerml-scarce-validation-geotransolver-sp}{5.03}
\isnset{fb-drivaerml-scarce-validation-geotransolver-sw}{7.66}
\isnset{fb-drivaerml-scarce-validation-geotransolver-vp}{4.64}
\isnset{fb-drivaerml-scarce-validation-geotransolver-vu}{6.22}
\isnset{fb-drivaerml-scarce-validation-globe}{69.49}
\isnset{fb-drivaerml-scarce-validation-globe-sp}{8.15}
\isnset{fb-drivaerml-scarce-validation-globe-sw}{10.17}
\isnset{fb-drivaerml-scarce-validation-globe-vp}{6.83}
\isnset{fb-drivaerml-scarce-validation-globe-vu}{8.56}
\isnset{fb-drivaerml-scarce-validation-isla}{77.25}
\isnset{fb-drivaerml-scarce-validation-isla-sp}{5.21}
\isnset{fb-drivaerml-scarce-validation-isla-sw}{7.98}
\isnset{fb-drivaerml-scarce-validation-isla-vp}{5.18}
\isnset{fb-drivaerml-scarce-validation-isla-vu}{7.03}
\isnset{fb-drivaerml-scarce-validation-lead}{1.9}
\isnset{fb-drivaerml-scarce-validation-transolver}{78.36}
\isnset{fb-drivaerml-scarce-validation-transolver-sp}{5.16}
\isnset{fb-drivaerml-scarce-validation-transolver-sw}{7.85}
\isnset{fb-drivaerml-scarce-validation-transolver-vp}{4.96}
\isnset{fb-drivaerml-scarce-validation-transolver-vu}{6.55}
\isnset{fb-drivaerml-super-scarce-test-geotransolver}{42.03}
\isnset{fb-drivaerml-super-scarce-test-geotransolver-sp}{13.34}
\isnset{fb-drivaerml-super-scarce-test-geotransolver-sw}{15.69}
\isnset{fb-drivaerml-super-scarce-test-geotransolver-vp}{12.19}
\isnset{fb-drivaerml-super-scarce-test-geotransolver-vu}{13.87}
\isnset{fb-drivaerml-super-scarce-test-globe}{39.05}
\isnset{fb-drivaerml-super-scarce-test-globe-sp}{15.00}
\isnset{fb-drivaerml-super-scarce-test-globe-sw}{18.21}
\isnset{fb-drivaerml-super-scarce-test-globe-vp}{12.86}
\isnset{fb-drivaerml-super-scarce-test-globe-vu}{14.90}
\isnset{fb-drivaerml-super-scarce-test-isla}{48.17}
\isnset{fb-drivaerml-super-scarce-test-isla-sp}{11.29}
\isnset{fb-drivaerml-super-scarce-test-isla-sw}{14.03}
\isnset{fb-drivaerml-super-scarce-test-isla-vp}{10.45}
\isnset{fb-drivaerml-super-scarce-test-isla-vu}{12.65}
\isnset{fb-drivaerml-super-scarce-test-lead}{5.9}
\isnset{fb-drivaerml-super-scarce-test-second}{42.27}
\isnset{fb-drivaerml-super-scarce-test-transolver}{42.27}
\isnset{fb-drivaerml-super-scarce-test-transolver-sp}{13.09}
\isnset{fb-drivaerml-super-scarce-test-transolver-sw}{15.18}
\isnset{fb-drivaerml-super-scarce-test-transolver-vp}{11.93}
\isnset{fb-drivaerml-super-scarce-test-transolver-vu}{13.72}
\isnset{fb-drivaerml-super-scarce-validation-geotransolver}{40.56}
\isnset{fb-drivaerml-super-scarce-validation-geotransolver-sp}{14.47}
\isnset{fb-drivaerml-super-scarce-validation-geotransolver-sw}{17.38}
\isnset{fb-drivaerml-super-scarce-validation-geotransolver-vp}{12.86}
\isnset{fb-drivaerml-super-scarce-validation-geotransolver-vu}{14.59}
\isnset{fb-drivaerml-super-scarce-validation-globe}{38.10}
\isnset{fb-drivaerml-super-scarce-validation-globe-sp}{16.14}
\isnset{fb-drivaerml-super-scarce-validation-globe-sw}{19.77}
\isnset{fb-drivaerml-super-scarce-validation-globe-vp}{13.98}
\isnset{fb-drivaerml-super-scarce-validation-globe-vu}{15.52}
\isnset{fb-drivaerml-super-scarce-validation-isla}{46.94}
\isnset{fb-drivaerml-super-scarce-validation-isla-sp}{12.29}
\isnset{fb-drivaerml-super-scarce-validation-isla-sw}{15.44}
\isnset{fb-drivaerml-super-scarce-validation-isla-vp}{11.45}
\isnset{fb-drivaerml-super-scarce-validation-isla-vu}{13.23}
\isnset{fb-drivaerml-super-scarce-validation-lead}{6.4}
\isnset{fb-drivaerml-super-scarce-validation-transolver}{40.01}
\isnset{fb-drivaerml-super-scarce-validation-transolver-sp}{14.11}
\isnset{fb-drivaerml-super-scarce-validation-transolver-sw}{16.77}
\isnset{fb-drivaerml-super-scarce-validation-transolver-vp}{12.62}
\isnset{fb-drivaerml-super-scarce-validation-transolver-vu}{14.51}
\isnset{fb-windsorml-full-test-geotransolver}{72.41}
\isnset{fb-windsorml-full-test-geotransolver-sp}{8.77}
\isnset{fb-windsorml-full-test-geotransolver-sw}{6.89}
\isnset{fb-windsorml-full-test-geotransolver-vp}{0.07}
\isnset{fb-windsorml-full-test-geotransolver-vu}{6.82}
\isnset{fb-windsorml-full-test-globe}{67.75}
\isnset{fb-windsorml-full-test-globe-sp}{8.98}
\isnset{fb-windsorml-full-test-globe-sw}{6.94}
\isnset{fb-windsorml-full-test-globe-vp}{0.08}
\isnset{fb-windsorml-full-test-globe-vu}{7.25}
\isnset{fb-windsorml-full-test-isla}{72.66}
\isnset{fb-windsorml-full-test-isla-sp}{8.71}
\isnset{fb-windsorml-full-test-isla-sw}{6.90}
\isnset{fb-windsorml-full-test-isla-vp}{0.07}
\isnset{fb-windsorml-full-test-isla-vu}{6.93}
\isnset{fb-windsorml-full-test-lead}{4.3}
\isnset{fb-windsorml-full-test-transolver}{76.96}
\isnset{fb-windsorml-full-test-transolver-sp}{8.29}
\isnset{fb-windsorml-full-test-transolver-sw}{6.56}
\isnset{fb-windsorml-full-test-transolver-vp}{0.07}
\isnset{fb-windsorml-full-test-transolver-vu}{6.52}
\isnset{fb-windsorml-full-validation-geotransolver}{74.51}
\isnset{fb-windsorml-full-validation-geotransolver-sp}{7.29}
\isnset{fb-windsorml-full-validation-geotransolver-sw}{6.87}
\isnset{fb-windsorml-full-validation-geotransolver-vp}{0.06}
\isnset{fb-windsorml-full-validation-geotransolver-vu}{5.91}
\isnset{fb-windsorml-full-validation-globe}{72.97}
\isnset{fb-windsorml-full-validation-globe-sp}{7.63}
\isnset{fb-windsorml-full-validation-globe-sw}{6.80}
\isnset{fb-windsorml-full-validation-globe-vp}{0.07}
\isnset{fb-windsorml-full-validation-globe-vu}{6.31}
\isnset{fb-windsorml-full-validation-isla}{75.69}
\isnset{fb-windsorml-full-validation-isla-sp}{6.89}
\isnset{fb-windsorml-full-validation-isla-sw}{6.86}
\isnset{fb-windsorml-full-validation-isla-vp}{0.06}
\isnset{fb-windsorml-full-validation-isla-vu}{5.74}
\isnset{fb-windsorml-full-validation-lead}{1.2}
\isnset{fb-windsorml-full-validation-transolver}{73.84}
\isnset{fb-windsorml-full-validation-transolver-sp}{7.40}
\isnset{fb-windsorml-full-validation-transolver-sw}{6.78}
\isnset{fb-windsorml-full-validation-transolver-vp}{0.06}
\isnset{fb-windsorml-full-validation-transolver-vu}{5.93}
\isnset{fb-windsorml-geometry-test-geotransolver}{65.02}
\isnset{fb-windsorml-geometry-test-geotransolver-sp}{10.48}
\isnset{fb-windsorml-geometry-test-geotransolver-sw}{8.76}
\isnset{fb-windsorml-geometry-test-geotransolver-vp}{0.12}
\isnset{fb-windsorml-geometry-test-geotransolver-vu}{8.36}
\isnset{fb-windsorml-geometry-test-globe}{60.47}
\isnset{fb-windsorml-geometry-test-globe-sp}{10.14}
\isnset{fb-windsorml-geometry-test-globe-sw}{8.48}
\isnset{fb-windsorml-geometry-test-globe-vp}{0.13}
\isnset{fb-windsorml-geometry-test-globe-vu}{8.97}
\isnset{fb-windsorml-geometry-test-isla}{65.12}
\isnset{fb-windsorml-geometry-test-isla-sp}{9.85}
\isnset{fb-windsorml-geometry-test-isla-sw}{8.59}
\isnset{fb-windsorml-geometry-test-isla-vp}{0.10}
\isnset{fb-windsorml-geometry-test-isla-vu}{7.67}
\isnset{fb-windsorml-geometry-test-lead}{1.4}
\isnset{fb-windsorml-geometry-test-transolver}{66.50}
\isnset{fb-windsorml-geometry-test-transolver-sp}{9.98}
\isnset{fb-windsorml-geometry-test-transolver-sw}{8.40}
\isnset{fb-windsorml-geometry-test-transolver-vp}{0.10}
\isnset{fb-windsorml-geometry-test-transolver-vu}{7.46}
\isnset{fb-windsorml-geometry-validation-geotransolver}{68.37}
\isnset{fb-windsorml-geometry-validation-geotransolver-sp}{8.26}
\isnset{fb-windsorml-geometry-validation-geotransolver-sw}{7.27}
\isnset{fb-windsorml-geometry-validation-geotransolver-vp}{0.07}
\isnset{fb-windsorml-geometry-validation-geotransolver-vu}{6.35}
\isnset{fb-windsorml-geometry-validation-globe}{68.65}
\isnset{fb-windsorml-geometry-validation-globe-sp}{8.11}
\isnset{fb-windsorml-geometry-validation-globe-sw}{6.88}
\isnset{fb-windsorml-geometry-validation-globe-vp}{0.07}
\isnset{fb-windsorml-geometry-validation-globe-vu}{6.27}
\isnset{fb-windsorml-geometry-validation-isla}{73.29}
\isnset{fb-windsorml-geometry-validation-isla-sp}{8.05}
\isnset{fb-windsorml-geometry-validation-isla-sw}{7.38}
\isnset{fb-windsorml-geometry-validation-isla-vp}{0.06}
\isnset{fb-windsorml-geometry-validation-isla-vu}{5.75}
\isnset{fb-windsorml-geometry-validation-lead}{1.8}
\isnset{fb-windsorml-geometry-validation-transolver}{75.13}
\isnset{fb-windsorml-geometry-validation-transolver-sp}{7.90}
\isnset{fb-windsorml-geometry-validation-transolver-sw}{6.95}
\isnset{fb-windsorml-geometry-validation-transolver-vp}{0.06}
\isnset{fb-windsorml-geometry-validation-transolver-vu}{6.08}
\isnset{fb-windsorml-high-drag-test-geotransolver}{36.67}
\isnset{fb-windsorml-high-drag-test-geotransolver-sp}{15.85}
\isnset{fb-windsorml-high-drag-test-geotransolver-sw}{11.86}
\isnset{fb-windsorml-high-drag-test-geotransolver-vp}{0.21}
\isnset{fb-windsorml-high-drag-test-geotransolver-vu}{14.84}
\isnset{fb-windsorml-high-drag-test-globe}{36.34}
\isnset{fb-windsorml-high-drag-test-globe-sp}{16.01}
\isnset{fb-windsorml-high-drag-test-globe-sw}{11.83}
\isnset{fb-windsorml-high-drag-test-globe-vp}{0.22}
\isnset{fb-windsorml-high-drag-test-globe-vu}{15.40}
\isnset{fb-windsorml-high-drag-test-isla}{35.46}
\isnset{fb-windsorml-high-drag-test-isla-sp}{16.28}
\isnset{fb-windsorml-high-drag-test-isla-sw}{12.12}
\isnset{fb-windsorml-high-drag-test-isla-vp}{0.23}
\isnset{fb-windsorml-high-drag-test-isla-vu}{15.71}
\isnset{fb-windsorml-high-drag-test-lead}{1.2}
\isnset{fb-windsorml-high-drag-test-transolver}{35.91}
\isnset{fb-windsorml-high-drag-test-transolver-sp}{16.43}
\isnset{fb-windsorml-high-drag-test-transolver-sw}{12.09}
\isnset{fb-windsorml-high-drag-test-transolver-vp}{0.21}
\isnset{fb-windsorml-high-drag-test-transolver-vu}{14.98}
\isnset{fb-windsorml-high-drag-validation-geotransolver}{77.13}
\isnset{fb-windsorml-high-drag-validation-geotransolver-sp}{7.07}
\isnset{fb-windsorml-high-drag-validation-geotransolver-sw}{6.02}
\isnset{fb-windsorml-high-drag-validation-geotransolver-vp}{0.06}
\isnset{fb-windsorml-high-drag-validation-geotransolver-vu}{6.21}
\isnset{fb-windsorml-high-drag-validation-globe}{75.77}
\isnset{fb-windsorml-high-drag-validation-globe-sp}{7.30}
\isnset{fb-windsorml-high-drag-validation-globe-sw}{6.36}
\isnset{fb-windsorml-high-drag-validation-globe-vp}{0.06}
\isnset{fb-windsorml-high-drag-validation-globe-vu}{6.10}
\isnset{fb-windsorml-high-drag-validation-isla}{78.36}
\isnset{fb-windsorml-high-drag-validation-isla-sp}{6.78}
\isnset{fb-windsorml-high-drag-validation-isla-sw}{6.14}
\isnset{fb-windsorml-high-drag-validation-isla-vp}{0.05}
\isnset{fb-windsorml-high-drag-validation-isla-vu}{5.81}
\isnset{fb-windsorml-high-drag-validation-lead}{0.7}
\isnset{fb-windsorml-high-drag-validation-transolver}{77.65}
\isnset{fb-windsorml-high-drag-validation-transolver-sp}{7.05}
\isnset{fb-windsorml-high-drag-validation-transolver-sw}{6.13}
\isnset{fb-windsorml-high-drag-validation-transolver-vp}{0.06}
\isnset{fb-windsorml-high-drag-validation-transolver-vu}{6.01}
\isnset{fb-windsorml-image-wake-test-geotransolver}{59.30}
\isnset{fb-windsorml-image-wake-test-geotransolver-sp}{8.31}
\isnset{fb-windsorml-image-wake-test-geotransolver-sw}{7.36}
\isnset{fb-windsorml-image-wake-test-geotransolver-vp}{0.07}
\isnset{fb-windsorml-image-wake-test-geotransolver-vu}{7.41}
\isnset{fb-windsorml-image-wake-test-globe}{59.18}
\isnset{fb-windsorml-image-wake-test-globe-sp}{8.03}
\isnset{fb-windsorml-image-wake-test-globe-sw}{7.26}
\isnset{fb-windsorml-image-wake-test-globe-vp}{0.07}
\isnset{fb-windsorml-image-wake-test-globe-vu}{7.50}
\isnset{fb-windsorml-image-wake-test-isla}{60.77}
\isnset{fb-windsorml-image-wake-test-isla-sp}{8.00}
\isnset{fb-windsorml-image-wake-test-isla-sw}{7.25}
\isnset{fb-windsorml-image-wake-test-isla-vp}{0.06}
\isnset{fb-windsorml-image-wake-test-isla-vu}{7.03}
\isnset{fb-windsorml-image-wake-test-lead}{1.5}
\isnset{fb-windsorml-image-wake-test-transolver}{59.18}
\isnset{fb-windsorml-image-wake-test-transolver-sp}{8.30}
\isnset{fb-windsorml-image-wake-test-transolver-sw}{7.36}
\isnset{fb-windsorml-image-wake-test-transolver-vp}{0.07}
\isnset{fb-windsorml-image-wake-test-transolver-vu}{7.37}
\isnset{fb-windsorml-image-wake-validation-geotransolver}{75.38}
\isnset{fb-windsorml-image-wake-validation-geotransolver-sp}{8.13}
\isnset{fb-windsorml-image-wake-validation-geotransolver-sw}{7.27}
\isnset{fb-windsorml-image-wake-validation-geotransolver-vp}{0.08}
\isnset{fb-windsorml-image-wake-validation-geotransolver-vu}{6.80}
\isnset{fb-windsorml-image-wake-validation-globe}{73.42}
\isnset{fb-windsorml-image-wake-validation-globe-sp}{8.06}
\isnset{fb-windsorml-image-wake-validation-globe-sw}{6.77}
\isnset{fb-windsorml-image-wake-validation-globe-vp}{0.08}
\isnset{fb-windsorml-image-wake-validation-globe-vu}{6.96}
\isnset{fb-windsorml-image-wake-validation-isla}{71.42}
\isnset{fb-windsorml-image-wake-validation-isla-sp}{8.16}
\isnset{fb-windsorml-image-wake-validation-isla-sw}{6.86}
\isnset{fb-windsorml-image-wake-validation-isla-vp}{0.07}
\isnset{fb-windsorml-image-wake-validation-isla-vu}{6.68}
\isnset{fb-windsorml-image-wake-validation-lead}{4.0}
\isnset{fb-windsorml-image-wake-validation-transolver}{73.88}
\isnset{fb-windsorml-image-wake-validation-transolver-sp}{8.29}
\isnset{fb-windsorml-image-wake-validation-transolver-sw}{7.20}
\isnset{fb-windsorml-image-wake-validation-transolver-vp}{0.08}
\isnset{fb-windsorml-image-wake-validation-transolver-vu}{6.76}
\isnset{fb-windsorml-low-drag-test-geotransolver}{67.92}
\isnset{fb-windsorml-low-drag-test-geotransolver-sp}{7.24}
\isnset{fb-windsorml-low-drag-test-geotransolver-sw}{6.58}
\isnset{fb-windsorml-low-drag-test-geotransolver-vp}{0.06}
\isnset{fb-windsorml-low-drag-test-geotransolver-vu}{5.98}
\isnset{fb-windsorml-low-drag-test-globe}{65.53}
\isnset{fb-windsorml-low-drag-test-globe-sp}{7.12}
\isnset{fb-windsorml-low-drag-test-globe-sw}{6.39}
\isnset{fb-windsorml-low-drag-test-globe-vp}{0.06}
\isnset{fb-windsorml-low-drag-test-globe-vu}{6.10}
\isnset{fb-windsorml-low-drag-test-isla}{66.89}
\isnset{fb-windsorml-low-drag-test-isla-sp}{7.21}
\isnset{fb-windsorml-low-drag-test-isla-sw}{6.60}
\isnset{fb-windsorml-low-drag-test-isla-vp}{0.06}
\isnset{fb-windsorml-low-drag-test-isla-vu}{5.80}
\isnset{fb-windsorml-low-drag-test-lead}{4.2}
\isnset{fb-windsorml-low-drag-test-transolver}{71.08}
\isnset{fb-windsorml-low-drag-test-transolver-sp}{7.28}
\isnset{fb-windsorml-low-drag-test-transolver-sw}{6.59}
\isnset{fb-windsorml-low-drag-test-transolver-vp}{0.06}
\isnset{fb-windsorml-low-drag-test-transolver-vu}{5.96}
\isnset{fb-windsorml-low-drag-validation-geotransolver}{64.92}
\isnset{fb-windsorml-low-drag-validation-geotransolver-sp}{7.59}
\isnset{fb-windsorml-low-drag-validation-geotransolver-sw}{7.04}
\isnset{fb-windsorml-low-drag-validation-geotransolver-vp}{0.07}
\isnset{fb-windsorml-low-drag-validation-geotransolver-vu}{6.17}
\isnset{fb-windsorml-low-drag-validation-globe}{64.04}
\isnset{fb-windsorml-low-drag-validation-globe-sp}{7.80}
\isnset{fb-windsorml-low-drag-validation-globe-sw}{7.23}
\isnset{fb-windsorml-low-drag-validation-globe-vp}{0.07}
\isnset{fb-windsorml-low-drag-validation-globe-vu}{6.51}
\isnset{fb-windsorml-low-drag-validation-isla}{66.89}
\isnset{fb-windsorml-low-drag-validation-isla-sp}{7.25}
\isnset{fb-windsorml-low-drag-validation-isla-sw}{7.03}
\isnset{fb-windsorml-low-drag-validation-isla-vp}{0.06}
\isnset{fb-windsorml-low-drag-validation-isla-vu}{6.10}
\isnset{fb-windsorml-low-drag-validation-lead}{1.9}
\isnset{fb-windsorml-low-drag-validation-transolver}{65.03}
\isnset{fb-windsorml-low-drag-validation-transolver-sp}{7.73}
\isnset{fb-windsorml-low-drag-validation-transolver-sw}{7.20}
\isnset{fb-windsorml-low-drag-validation-transolver-vp}{0.06}
\isnset{fb-windsorml-low-drag-validation-transolver-vu}{5.84}
\isnset{fb-windsorml-medium-test-geotransolver}{67.10}
\isnset{fb-windsorml-medium-test-geotransolver-sp}{9.37}
\isnset{fb-windsorml-medium-test-geotransolver-sw}{7.36}
\isnset{fb-windsorml-medium-test-geotransolver-vp}{0.08}
\isnset{fb-windsorml-medium-test-geotransolver-vu}{7.50}
\isnset{fb-windsorml-medium-test-globe}{61.06}
\isnset{fb-windsorml-medium-test-globe-sp}{9.58}
\isnset{fb-windsorml-medium-test-globe-sw}{7.48}
\isnset{fb-windsorml-medium-test-globe-vp}{0.09}
\isnset{fb-windsorml-medium-test-globe-vu}{8.01}
\isnset{fb-windsorml-medium-test-isla}{67.46}
\isnset{fb-windsorml-medium-test-isla-sp}{9.29}
\isnset{fb-windsorml-medium-test-isla-sw}{7.53}
\isnset{fb-windsorml-medium-test-isla-vp}{0.08}
\isnset{fb-windsorml-medium-test-isla-vu}{7.92}
\isnset{fb-windsorml-medium-test-lead}{0.6}
\isnset{fb-windsorml-medium-test-transolver}{68.01}
\isnset{fb-windsorml-medium-test-transolver-sp}{9.64}
\isnset{fb-windsorml-medium-test-transolver-sw}{7.50}
\isnset{fb-windsorml-medium-test-transolver-vp}{0.08}
\isnset{fb-windsorml-medium-test-transolver-vu}{7.41}
\isnset{fb-windsorml-medium-validation-geotransolver}{70.08}
\isnset{fb-windsorml-medium-validation-geotransolver-sp}{7.85}
\isnset{fb-windsorml-medium-validation-geotransolver-sw}{7.05}
\isnset{fb-windsorml-medium-validation-geotransolver-vp}{0.07}
\isnset{fb-windsorml-medium-validation-geotransolver-vu}{6.58}
\isnset{fb-windsorml-medium-validation-globe}{66.25}
\isnset{fb-windsorml-medium-validation-globe-sp}{8.48}
\isnset{fb-windsorml-medium-validation-globe-sw}{7.14}
\isnset{fb-windsorml-medium-validation-globe-vp}{0.08}
\isnset{fb-windsorml-medium-validation-globe-vu}{7.31}
\isnset{fb-windsorml-medium-validation-isla}{71.95}
\isnset{fb-windsorml-medium-validation-isla-sp}{7.41}
\isnset{fb-windsorml-medium-validation-isla-sw}{6.98}
\isnset{fb-windsorml-medium-validation-isla-vp}{0.07}
\isnset{fb-windsorml-medium-validation-isla-vu}{6.40}
\isnset{fb-windsorml-medium-validation-lead}{1.3}
\isnset{fb-windsorml-medium-validation-transolver}{70.69}
\isnset{fb-windsorml-medium-validation-transolver-sp}{7.81}
\isnset{fb-windsorml-medium-validation-transolver-sw}{6.89}
\isnset{fb-windsorml-medium-validation-transolver-vp}{0.07}
\isnset{fb-windsorml-medium-validation-transolver-vu}{6.22}
\isnset{fb-windsorml-scarce-test-geotransolver}{60.20}
\isnset{fb-windsorml-scarce-test-geotransolver-sp}{11.43}
\isnset{fb-windsorml-scarce-test-geotransolver-sw}{9.28}
\isnset{fb-windsorml-scarce-test-geotransolver-vp}{0.10}
\isnset{fb-windsorml-scarce-test-geotransolver-vu}{8.66}
\isnset{fb-windsorml-scarce-test-globe}{59.05}
\isnset{fb-windsorml-scarce-test-globe-sp}{11.03}
\isnset{fb-windsorml-scarce-test-globe-sw}{8.50}
\isnset{fb-windsorml-scarce-test-globe-vp}{0.11}
\isnset{fb-windsorml-scarce-test-globe-vu}{9.12}
\isnset{fb-windsorml-scarce-test-isla}{62.47}
\isnset{fb-windsorml-scarce-test-isla-sp}{10.63}
\isnset{fb-windsorml-scarce-test-isla-sw}{8.90}
\isnset{fb-windsorml-scarce-test-isla-vp}{0.09}
\isnset{fb-windsorml-scarce-test-isla-vu}{8.24}
\isnset{fb-windsorml-scarce-test-lead}{1.6}
\isnset{fb-windsorml-scarce-test-transolver}{64.03}
\isnset{fb-windsorml-scarce-test-transolver-sp}{11.17}
\isnset{fb-windsorml-scarce-test-transolver-sw}{8.95}
\isnset{fb-windsorml-scarce-test-transolver-vp}{0.10}
\isnset{fb-windsorml-scarce-test-transolver-vu}{8.35}
\isnset{fb-windsorml-scarce-validation-geotransolver}{58.29}
\isnset{fb-windsorml-scarce-validation-geotransolver-sp}{9.59}
\isnset{fb-windsorml-scarce-validation-geotransolver-sw}{8.42}
\isnset{fb-windsorml-scarce-validation-geotransolver-vp}{0.10}
\isnset{fb-windsorml-scarce-validation-geotransolver-vu}{7.74}
\isnset{fb-windsorml-scarce-validation-globe}{61.62}
\isnset{fb-windsorml-scarce-validation-globe-sp}{9.30}
\isnset{fb-windsorml-scarce-validation-globe-sw}{7.83}
\isnset{fb-windsorml-scarce-validation-globe-vp}{0.09}
\isnset{fb-windsorml-scarce-validation-globe-vu}{7.70}
\isnset{fb-windsorml-scarce-validation-isla}{61.35}
\isnset{fb-windsorml-scarce-validation-isla-sp}{9.13}
\isnset{fb-windsorml-scarce-validation-isla-sw}{8.25}
\isnset{fb-windsorml-scarce-validation-isla-vp}{0.09}
\isnset{fb-windsorml-scarce-validation-isla-vu}{7.66}
\isnset{fb-windsorml-scarce-validation-lead}{0.3}
\isnset{fb-windsorml-scarce-validation-transolver}{60.93}
\isnset{fb-windsorml-scarce-validation-transolver-sp}{9.33}
\isnset{fb-windsorml-scarce-validation-transolver-sw}{8.22}
\isnset{fb-windsorml-scarce-validation-transolver-vp}{0.10}
\isnset{fb-windsorml-scarce-validation-transolver-vu}{7.66}
\isnset{fb-windsorml-super-scarce-test-geotransolver}{36.44}
\isnset{fb-windsorml-super-scarce-test-geotransolver-sp}{22.45}
\isnset{fb-windsorml-super-scarce-test-geotransolver-sw}{16.06}
\isnset{fb-windsorml-super-scarce-test-geotransolver-vp}{0.22}
\isnset{fb-windsorml-super-scarce-test-geotransolver-vu}{14.70}
\isnset{fb-windsorml-super-scarce-test-globe}{32.43}
\isnset{fb-windsorml-super-scarce-test-globe-sp}{20.30}
\isnset{fb-windsorml-super-scarce-test-globe-sw}{13.49}
\isnset{fb-windsorml-super-scarce-test-globe-vp}{0.21}
\isnset{fb-windsorml-super-scarce-test-globe-vu}{14.95}
\isnset{fb-windsorml-super-scarce-test-isla}{35.79}
\isnset{fb-windsorml-super-scarce-test-isla-sp}{22.10}
\isnset{fb-windsorml-super-scarce-test-isla-sw}{16.02}
\isnset{fb-windsorml-super-scarce-test-isla-vp}{0.21}
\isnset{fb-windsorml-super-scarce-test-isla-vu}{14.83}
\isnset{fb-windsorml-super-scarce-test-lead}{0.7}
\isnset{fb-windsorml-super-scarce-test-transolver}{33.96}
\isnset{fb-windsorml-super-scarce-test-transolver-sp}{23.56}
\isnset{fb-windsorml-super-scarce-test-transolver-sw}{16.59}
\isnset{fb-windsorml-super-scarce-test-transolver-vp}{0.22}
\isnset{fb-windsorml-super-scarce-test-transolver-vu}{13.94}
\isnset{fb-windsorml-super-scarce-validation-geotransolver}{34.78}
\isnset{fb-windsorml-super-scarce-validation-geotransolver-sp}{23.91}
\isnset{fb-windsorml-super-scarce-validation-geotransolver-sw}{17.31}
\isnset{fb-windsorml-super-scarce-validation-geotransolver-vp}{0.23}
\isnset{fb-windsorml-super-scarce-validation-geotransolver-vu}{14.81}
\isnset{fb-windsorml-super-scarce-validation-globe}{32.94}
\isnset{fb-windsorml-super-scarce-validation-globe-sp}{19.94}
\isnset{fb-windsorml-super-scarce-validation-globe-sw}{13.89}
\isnset{fb-windsorml-super-scarce-validation-globe-vp}{0.21}
\isnset{fb-windsorml-super-scarce-validation-globe-vu}{14.37}
\isnset{fb-windsorml-super-scarce-validation-isla}{35.58}
\isnset{fb-windsorml-super-scarce-validation-isla-sp}{23.35}
\isnset{fb-windsorml-super-scarce-validation-isla-sw}{16.93}
\isnset{fb-windsorml-super-scarce-validation-isla-vp}{0.21}
\isnset{fb-windsorml-super-scarce-validation-isla-vu}{14.03}
\isnset{fb-windsorml-super-scarce-validation-lead}{0.8}
\isnset{fb-windsorml-super-scarce-validation-transolver}{32.66}
\isnset{fb-windsorml-super-scarce-validation-transolver-sp}{24.92}
\isnset{fb-windsorml-super-scarce-validation-transolver-sw}{17.80}
\isnset{fb-windsorml-super-scarce-validation-transolver-vp}{0.23}
\isnset{fb-windsorml-super-scarce-validation-transolver-vu}{14.49}
\isnset{fb-windsorml-vp-max}{0.23}
\isnset{fb-windsorml-vp-min}{0.05}
\isnset{globe-drs-area-p-pct}{9.28}
\isnset{globe-drs-area-p-ratio}{2.26}
\isnset{globe-drs-area-tw-pct}{11.8}
\isnset{globe-drs-area-tw-ratio}{1.76}
\isnset{globe-drs-nat-p-pct}{9.10}
\isnset{globe-drs-nat-p-ratio}{2.22}
\isnset{globe-drs-nat-tw-pct}{11.3}
\isnset{globe-drs-nat-tw-ratio}{1.71}
\isnset{globe-drs-posed-over-nat-p}{1.49}
\isnset{globe-drs-posed-over-nat-tw}{1.09}
\isnset{globe-drs-posed-p-pct}{13.6}
\isnset{globe-drs-posed-p-ratio}{3.32}
\isnset{globe-drs-posed-tw-pct}{12.4}
\isnset{globe-drs-posed-tw-ratio}{1.87}
\isnset{hero-area-pct}{20}
\isnset{hero-area-ratio}{5}
\isnset{hero-area-ratio-gt}{3.62}
\isnset{hero-area-ratio-t}{7.25}
\isnset{hero-isla-pct}{4}
\isnset{hero-native-pct}{4}
\isnset{hero-posed-pct}{120}
\isnset{hero-posed-ratio}{30}
\isnset{hero-posed-ratio-gt}{31}
\isnset{hero-posed-ratio-t}{29}
\isnset{inv-af-nut-native-geotransolver}{0.0168}
\isnset{inv-af-nut-native-isla}{0.0158}
\isnset{inv-af-nut-native-transolver}{0.0170}
\isnset{inv-af-nut-posed-geotransolver}{86}
\isnset{inv-af-nut-posed-isla}{1.000}
\isnset{inv-af-nut-posed-transolver}{87}
\isnset{inv-af-p-native-geotransolver}{0.00979}
\isnset{inv-af-p-native-isla}{0.00831}
\isnset{inv-af-p-native-transolver}{0.00888}
\isnset{inv-af-p-posed-geotransolver}{158}
\isnset{inv-af-p-posed-isla}{1.00}
\isnset{inv-af-p-posed-transolver}{134}
\isnset{inv-af-u-native-geotransolver}{0.0355}
\isnset{inv-af-u-native-isla}{0.00411}
\isnset{inv-af-u-native-transolver}{0.0282}
\isnset{inv-af-u-posed-geotransolver}{32}
\isnset{inv-af-u-posed-isla}{1.00}
\isnset{inv-af-u-posed-transolver}{44}
\isnset{inv-area-abs-other-max}{34}
\isnset{inv-area-abs-other-min}{5}
\isnset{inv-area-max}{7.2}
\isnset{inv-area-median}{3.0}
\isnset{inv-area-min}{1.4}
\isnset{inv-djb-d-native-geotransolver}{0.154}
\isnset{inv-djb-d-native-isla}{0.146}
\isnset{inv-djb-d-native-transolver}{0.148}
\isnset{inv-djb-d-posed-geotransolver}{13}
\isnset{inv-djb-d-posed-isla}{1.000}
\isnset{inv-djb-d-posed-transolver}{9.72}
\isnset{inv-djb-vm-native-geotransolver}{0.417}
\isnset{inv-djb-vm-native-isla}{0.379}
\isnset{inv-djb-vm-native-transolver}{0.399}
\isnset{inv-djb-vm-posed-geotransolver}{4.83}
\isnset{inv-djb-vm-posed-isla}{1.000}
\isnset{inv-djb-vm-posed-transolver}{3.19}
\isnset{inv-dmls-area-gt-pct}{14}
\isnset{inv-dmls-area-t-pct}{30}
\isnset{inv-dmls-p-native-geotransolver}{0.0400}
\isnset{inv-dmls-p-native-isla}{0.0402}
\isnset{inv-dmls-p-native-transolver}{0.0413}
\isnset{inv-dmls-p-posed-geotransolver}{31}
\isnset{inv-dmls-p-posed-isla}{1.00}
\isnset{inv-dmls-p-posed-transolver}{29}
\isnset{inv-dmls-p-resampled-geotransolver}{3.62}
\isnset{inv-dmls-p-resampled-isla}{1.00}
\isnset{inv-dmls-p-resampled-transolver}{7.25}
\isnset{inv-dmls-tw-native-geotransolver}{0.0655}
\isnset{inv-dmls-tw-native-isla}{0.0648}
\isnset{inv-dmls-tw-native-transolver}{0.0667}
\isnset{inv-dmls-tw-posed-geotransolver}{20}
\isnset{inv-dmls-tw-posed-isla}{1.00}
\isnset{inv-dmls-tw-posed-transolver}{21}
\isnset{inv-dmls-tw-resampled-geotransolver}{3.30}
\isnset{inv-dmls-tw-resampled-isla}{1.01}
\isnset{inv-dmls-tw-resampled-transolver}{5.65}
\isnset{inv-dmlv-nut-native-geotransolver}{0.0807}
\isnset{inv-dmlv-nut-native-isla}{0.0776}
\isnset{inv-dmlv-nut-native-transolver}{0.0834}
\isnset{inv-dmlv-nut-posed-geotransolver}{24}
\isnset{inv-dmlv-nut-posed-isla}{1.00}
\isnset{inv-dmlv-nut-posed-transolver}{18}
\isnset{inv-dmlv-p-native-geotransolver}{0.0349}
\isnset{inv-dmlv-p-native-isla}{0.0382}
\isnset{inv-dmlv-p-native-transolver}{0.0468}
\isnset{inv-dmlv-p-posed-geotransolver}{24}
\isnset{inv-dmlv-p-posed-isla}{1.000}
\isnset{inv-dmlv-p-posed-transolver}{17}
\isnset{inv-dmlv-u-native-geotransolver}{0.0474}
\isnset{inv-dmlv-u-native-gt-pct}{4.7}
\isnset{inv-dmlv-u-native-isla}{0.0527}
\isnset{inv-dmlv-u-native-transolver}{0.0909}
\isnset{inv-dmlv-u-posed-geotransolver}{24}
\isnset{inv-dmlv-u-posed-isla}{1.00}
\isnset{inv-dmlv-u-posed-transolver}{14}
\isnset{inv-hl-code-p-pct}{3.2}
\isnset{inv-hl-code-tw-pct}{1.6}
\isnset{inv-hl-epochs}{985}
\isnset{inv-hl-p-native-geotransolver}{0.0508}
\isnset{inv-hl-p-native-isla}{0.0539}
\isnset{inv-hl-p-native-transolver}{0.0522}
\isnset{inv-hl-p-posed-geotransolver}{20}
\isnset{inv-hl-p-posed-isla}{1.01}
\isnset{inv-hl-p-posed-transolver}{19}
\isnset{inv-hl-p-resampled-geotransolver}{6.66}
\isnset{inv-hl-p-resampled-isla}{1.00}
\isnset{inv-hl-p-resampled-transolver}{2.29}
\isnset{inv-hl-steps}{125{,}095}
\isnset{inv-hl-tw-native-geotransolver}{0.0855}
\isnset{inv-hl-tw-native-isla}{0.0883}
\isnset{inv-hl-tw-native-transolver}{0.0873}
\isnset{inv-hl-tw-posed-geotransolver}{15}
\isnset{inv-hl-tw-posed-isla}{1.01}
\isnset{inv-hl-tw-posed-transolver}{15}
\isnset{inv-hl-tw-resampled-geotransolver}{4.82}
\isnset{inv-hl-tw-resampled-isla}{1.00}
\isnset{inv-hl-tw-resampled-transolver}{2.37}
\isnset{inv-isla-max-pct}{1.0}
\isnset{inv-isla-posed-max}{1.009}
\isnset{inv-lug-vm-native-geotransolver}{0.00224}
\isnset{inv-lug-vm-native-isla}{0.00193}
\isnset{inv-lug-vm-native-transolver}{0.00198}
\isnset{inv-lug-vm-posed-geotransolver}{415}
\isnset{inv-lug-vm-posed-isla}{1.01}
\isnset{inv-lug-vm-posed-transolver}{504}
\isnset{inv-n-tasks}{8}
\isnset{inv-posed-abs-all-max}{2.0}
\isnset{inv-posed-abs-all-min}{0.8}
\isnset{inv-posed-abs-max}{1.4}
\isnset{inv-posed-abs-min}{1.0}
\isnset{inv-posed-max}{504}
\isnset{inv-posed-max-all}{504}
\isnset{inv-posed-median}{31}
\isnset{inv-posed-min}{10}
\isnset{inv-posed-min-all}{3.2}
\isnset{inv-suv-p-native-geotransolver}{0.0383}
\isnset{inv-suv-p-native-isla}{0.0403}
\isnset{inv-suv-p-native-transolver}{0.0401}
\isnset{inv-suv-p-posed-geotransolver}{37}
\isnset{inv-suv-p-posed-isla}{1.00}
\isnset{inv-suv-p-posed-transolver}{30}
\isnset{inv-suv-p-resampled-geotransolver}{1.37}
\isnset{inv-suv-p-resampled-isla}{0.999}
\isnset{inv-suv-p-resampled-transolver}{1.51}
\isnset{inv-suv-tw-native-geotransolver}{0.0616}
\isnset{inv-suv-tw-native-isla}{0.0628}
\isnset{inv-suv-tw-native-transolver}{0.0626}
\isnset{inv-suv-tw-posed-geotransolver}{19}
\isnset{inv-suv-tw-posed-isla}{1.000}
\isnset{inv-suv-tw-posed-transolver}{18}
\isnset{inv-suv-tw-resampled-geotransolver}{1.23}
\isnset{inv-suv-tw-resampled-isla}{1.000}
\isnset{inv-suv-tw-resampled-transolver}{1.39}
\isnset{inv-t2d-d-native-geotransolver}{0.00327}
\isnset{inv-t2d-d-native-isla}{0.00412}
\isnset{inv-t2d-d-native-transolver}{0.00460}
\isnset{inv-t2d-d-posed-geotransolver}{402}
\isnset{inv-t2d-d-posed-isla}{1.00}
\isnset{inv-t2d-d-posed-transolver}{262}
\isnset{inv-t2d-s-native-geotransolver}{0.00454}
\isnset{inv-t2d-s-native-isla}{0.00517}
\isnset{inv-t2d-s-native-transolver}{0.00615}
\isnset{inv-t2d-s-posed-geotransolver}{241}
\isnset{inv-t2d-s-posed-isla}{1.00}
\isnset{inv-t2d-s-posed-transolver}{155}
\isnset{inv-vs-bench-gt-vu-ratio}{1.1}
\isnset{later-fullrun-mem-new-aerozoo}{105.4}
\isnset{later-fullrun-mem-old-aerozoo}{142.2}
\isnset{later-fullrun-step-new-aerozoo}{0.56}
\isnset{later-fullrun-step-old-aerozoo}{0.66}
\isnset{later-iter-new-aerozoo}{0.56}
\isnset{later-iter-new-airfrans}{0.31}
\isnset{later-iter-new-surface}{0.16}
\isnset{later-iter-new-volume}{1.01}
\isnset{later-iter-old-aerozoo}{0.65}
\isnset{later-iter-old-airfrans}{0.32}
\isnset{later-iter-old-surface}{0.17}
\isnset{later-iter-old-volume}{1.03}
\isnset{later-iter-other-max-pct}{2}
\isnset{later-mem-new-aerozoo}{83.3}
\isnset{later-mem-new-airfrans}{17.6}
\isnset{later-mem-new-surface}{41.7}
\isnset{later-mem-new-volume}{93.3}
\isnset{later-mem-old-aerozoo}{120.5}
\isnset{later-mem-old-airfrans}{25.3}
\isnset{later-mem-old-surface}{55.9}
\isnset{later-mem-old-volume}{129.0}
\isnset{mem-peak-recompute}{105}
\isnset{mem-peak-trained}{142}
\isnset{params-forms-diff-exact}{1{,}214{,}849}
\isnset{params-interior-added-exact}{1{,}214{,}336}
\isnset{params-surface}{66.3}
\isnset{params-surface-exact}{66{,}261{,}600}
\isnset{params-volume}{67.5}
\isnset{params-volume-exact}{67{,}476{,}449}
\isnset{posetie-az-mean-max-pct}{0.05}
\isnset{posetie-az-med-pct}{0.01}
\isnset{posetie-az-n-cases}{128}
\isnset{posetie-az-n-poses}{5}
\isnset{posetie-checked-val}{538}
\isnset{sn-over-entry-max-pct}{9}
\isnset{sn-over-entry-min-pct}{$-$2}
\isnset{splits-ahmedml-full}{400 / 50 / 50}
\isnset{splits-ahmedml-medium}{133 / 50 / 50}
\isnset{splits-ahmedml-ood}{350 / 50 / 100}
\isnset{splits-ahmedml-scarce}{67 / 50 / 50}
\isnset{splits-ahmedml-super-scarce}{11 / 50 / 50}
\isnset{splits-drivaerml-full}{400 / 34 / 50}
\isnset{splits-drivaerml-medium}{133 / 34 / 50}
\isnset{splits-drivaerml-ood}{339 / 48 / 97}
\isnset{splits-drivaerml-scarce}{67 / 34 / 50}
\isnset{splits-drivaerml-super-scarce}{11 / 34 / 50}
\isnset{splits-windsorml-full}{276 / 33 / 35}
\isnset{splits-windsorml-medium}{90 / 33 / 35}
\isnset{splits-windsorml-ood}{239--242 / 34--35 / 67--71}
\isnset{splits-windsorml-scarce}{45 / 33 / 35}
\isnset{splits-windsorml-super-scarce}{7 / 33 / 35}
\isnset{steps-ahmedml-geotransolver-full}{50{,}000}
\isnset{steps-ahmedml-geotransolver-scarce}{50{,}000}
\isnset{steps-ahmedml-geotransolver-super-scarce}{50{,}000}
\isnset{steps-ahmedml-globe-full}{50{,}000}
\isnset{steps-ahmedml-globe-scarce}{50{,}000}
\isnset{steps-ahmedml-globe-super-scarce}{50{,}000}
\isnset{steps-ahmedml-isla-full}{50{,}000}
\isnset{steps-ahmedml-isla-scarce}{50{,}000}
\isnset{steps-ahmedml-isla-super-scarce}{50{,}000}
\isnset{steps-ahmedml-transolver-full}{50{,}000}
\isnset{steps-ahmedml-transolver-scarce}{50{,}000}
\isnset{steps-ahmedml-transolver-super-scarce}{50{,}000}
\isnset{steps-drivaerml-geotransolver-full}{50{,}000}
\isnset{steps-drivaerml-geotransolver-scarce}{50{,}000}
\isnset{steps-drivaerml-geotransolver-super-scarce}{50{,}000}
\isnset{steps-drivaerml-isla-full}{50{,}000}
\isnset{steps-drivaerml-isla-scarce}{50{,}000}
\isnset{steps-drivaerml-isla-super-scarce}{50{,}000}
\isnset{steps-drivaerml-transolver-full}{50{,}000}
\isnset{steps-drivaerml-transolver-scarce}{50{,}000}
\isnset{steps-drivaerml-transolver-super-scarce}{50{,}000}
\isnset{train-gpuh-aerozoo}{105}
\isnset{train-gpuh-airfrans}{38}
\isnset{train-gpuh-surface}{11}
\isnset{train-gpuh-volume}{71}
\isnset{train-mem-aerozoo}{142}
\isnset{train-mem-airfrans}{25}
\isnset{train-mem-floor}{80}
\isnset{train-mem-surface}{56}
\isnset{train-mem-volume}{129}
\isnset{train-step-aerozoo}{0.66}
\isnset{train-step-airfrans}{0.32}
\isnset{train-step-surface}{0.16}
\isnset{train-step-volume}{1.06}
\isnset{traincost-airfrans-gt-gpuh}{27}
\isnset{traincost-airfrans-isla-gpuh}{38}
\isnset{traincost-airfrans-t-gpuh}{36}
\isnset{traincost-surface-gt-gpuh}{28}
\isnset{traincost-surface-isla-gpuh}{11}
\isnset{traincost-surface-t-gpuh}{22}
\isnset{traincost-volume-gt-gpuh}{67}
\isnset{traincost-volume-isla-gpuh}{71}
\isnset{traincost-volume-t-gpuh}{31}

\icmltitlerunning{ISLA: A Coordinate-Free, Measure-Weighted Transformer for Boundary-Driven PDEs}

\begin{document}

\twocolumn[
  \icmltitle{ISLA: A Coordinate-Free, Measure-Weighted Transformer\\
    for Boundary-Driven PDEs}

  \begin{icmlauthorlist}
    \icmlauthor{Peter Sharpe}{nv}
  \end{icmlauthorlist}

  \icmlaffiliation{nv}{NVIDIA, Santa Clara, California, USA}
  \icmlcorrespondingauthor{Peter Sharpe}{psharpe@nvidia.com}

  \icmlkeywords{neural surrogates, partial differential equations, equivariance, discretization
    invariance, transformers, slice attention, computational fluid dynamics, boundary-value problems}

  \vskip 0.3in
]

\printAffiliationsAndNotice{}

\begin{abstract}
Symmetry is a foundational principle of physics: the behavior of a PDE-governed system does not
depend on the coordinate frame, unit system, or mesh used to write it down, and neither should a
surrogate model.
Yet today's leading neural surrogates lack these invariances and fail under simple transformations
of the same problem: for models trained on PDEs of cars in one pose (i.e., translation and
rotation), rotating the full PDE raises the surface-pressure error from
\isn{encb-headline-native-pct}\% to more than \isn{encb-headline-posed-floor}\%, and re-sampling
its boundary raises it to about \isn{encb-headline-area-pct}\%.
We introduce \isla{} (Invariant Slice Attention), a slice-attention transformer for steady,
boundary-driven PDEs.
Because the boundary determines the interior solution, \isla{} encodes the posed boundary once
into learned slices and evaluates arbitrary interior points independently through a
tree-accelerated learned boundary kernel.
Inputs are nondimensionalized in scales computed from the boundary, sums over boundary cells are
area-weighted quadratures that converge under mesh refinement, and geometry enters only as
distances and relative angles to equivariant slice anchors, making the model equivariant to rigid
motions and reflections up to optional token subsampling and tree clustering.
On \aerozoo{}, \isn{az-n-canonical} 3D flows spanning diverse geometries, poses, physical scales,
and boundary conditions, \isla{}'s mean reference-scaled velocity and pressure errors are
\isn{az-full-mean-ratio} and \isn{az-full-meanp-ratio}~times lower than the best baseline's.
On full AirfRANS, \isla{} is second only to an explicit boundary-integral network (\globe{}), with
errors \isn{af-full-tgt-over-isla-min} to \isn{af-full-tgt-over-isla-max}~times lower than the
better of Transolver's and GeoTransolver's in every field.
\end{abstract}

\section{Introduction}
\label{sec:intro}

Neural surrogates now predict the flow around a car in seconds, with errors of a few percent of
the reference simulation \citep{wu2024transolver,alkin2025abupt,adams2025geotransolver}.
The models with the lowest errors on automotive aerodynamics benchmarks are transformers that make global attention
affordable on meshes of $10^5$ to $10^8$ points, often by routing points through a small set of
learned \emph{slices}, which are soft groups of points
\citep{wu2024transolver,adams2025geotransolver}.

\begin{figure*}[t]
  \centering
  \includegraphics[width=\textwidth]{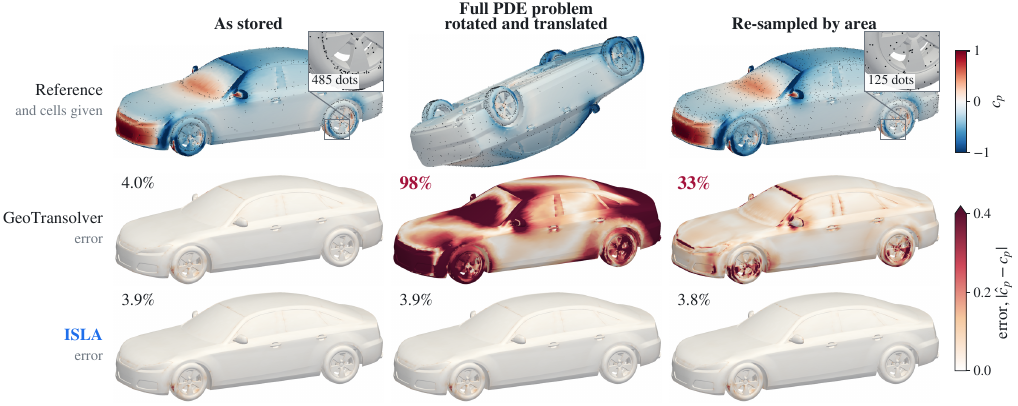}
  \caption{\textbf{\isla{}'s error does not depend on how a problem is presented.}
  One DrivAerML evaluation case; neither network was trained with augmentation.
  \emph{Top:} the reference pressure coefficient $c_p$ as stored, with the full PDE problem
  rotated by \isn{car-pose-angle}$^\circ$ and translated, and re-sampled by area (dots:
  cells given; insets: rear wheel).
  \emph{Below:} each model's error in the car's own frame, with its relative $L^2$ error.}
  \label{fig:hero}
\end{figure*}

A boundary-value problem is a physical statement: a boundary, the conditions imposed on it, and
the properties of the medium.
Writing it down for a computer takes choices that are not part of the physics: an origin and
axes, units, a mesh of the boundary, the cells of that mesh that are kept, and the points at
which the solution is requested.
We call these choices the \emph{presentation} of the problem, and the solution does not depend on
them.
The surrogates with the lowest errors on today's benchmarks do (\cref{fig:hero}), because the
presentation is their input: coordinates enter as features, and sums over points ignore the area
that each point represents.
Transolver and GeoTransolver also pool interior query points into the shared slices, so a
prediction at one point changes when other points are queried (a coupling that anchor-based
cross-attention models such as AB-UPT avoid; \citealp{alkin2025abupt}).
For Transolver and GeoTransolver trained on DrivAerML \citep{ashton2024drivaerml},
rotating its validation cases at random raises the surface-pressure error from
\isn{encb-headline-native-pct}\% to more than \isn{encb-headline-posed-floor}\% (worse than
predicting zero), and re-sampling by area raises it to about \isn{encb-headline-area-pct}\%.

This dependence limits practical use.
A training set fixes one convention: every car points along $+x$, in meters, from one mesher.
A model that depends on the convention fails when the same problem is written in another frame,
sampled by another rule, or stated in other units.
Augmentation helps only with what it samples: networks trained with rotations about the vertical
axis still fail under other rotations and under re-sampling by area, and training with arbitrary
rotations raises the stored-case surface-pressure errors of Transolver and GeoTransolver by
\isn{aug-gt-native-excess-pct} to \isn{aug-t-native-excess-pct}\% (\cref{sec:exp-encoding}).
On \aerozoo{} \citep{sharpe2026aerozoo}, where every case has its own pose (i.e., translation and
rotation), size, speed, and boundary conditions, GeoTransolver needs the inlet conditions and
per-case scales, both added by hand, to improve on predicting the inlet state everywhere.

Exact symmetry can be built into an architecture layer by layer, but at a cost in training time
or accuracy.
Layerwise equivariant networks constrain every linear map and nonlinearity
\citep{weiler20183d,brandstetter2022geometric,geiger2022e3nn,liao2023equiformer,brehmer2023geometric,suk2024labgatr},
and AB-GATr, an equivariant transformer on a subset of the points, takes two to eleven times as
long to train as its non-equivariant counterpart \citep{rygiel2026when}.
\globe{} \citep{sharpe2025globe}, a boundary-integral architecture, is equivariant by
construction and independent of units, and it uses the boundary measure.
It has the lowest errors on two-dimensional airfoil flows, and higher errors than slice transformers
on DrivAerML.
We ask whether the accuracy and linear cost of slice attention can be had together with exact
independence of the frame, the units, and the query set, and with consistency under remeshing.

\textbf{\isla{}} (Invariant Slice Attention) follows three principles, one for each part of the
presentation.
\emph{Principle 1: no learned operation receives a coordinate.}
Each slice has an \emph{anchor}, the mean position and mean normal of its \emph{tokens}
(boundary cells) weighted by measure, and a token's position enters the network only as
distances and relative angles to the anchors.
Lengths are measured in a unit computed from the boundary, and vectors are assembled at the
output from directions that rotate with the body.
\emph{Principle 2: every sum over the boundary is a quadrature.}
Cell measures weight every attention over boundary tokens, so a mesh changes a prediction only
by a quadrature error.
\emph{Principle 3: interior points are read-only}, to give \emph{query independence}.
A point in the domain gets its distance and direction to the boundary and attends to
the slices and, through a tree-accelerated kernel, to every boundary token; it enters no sum, so
its prediction depends on the point and the boundary alone.

\textbf{Contributions.}
First, we introduce \isla{} (\cref{sec:method}), to our knowledge the first slice-attention
surrogate that is exactly equivariant to the Euclidean group.
Second, we prove its invariances (\cref{sec:guarantees}) and show that measure-weighted slice
attention is a ratio of two quadratures that, with every cell a token, converges under
refinement.
Third, we show that \isla{}'s errors are the lowest on \aerozoo{}'s full split, second only to \globe{} on AirfRANS, and change by at most
\isn{inv-isla-max-pct}\% under a change of presentation.
\reviewpreprint{An open-source implementation will be released.}{An open-source implementation of
\isla{} is available in NVIDIA PhysicsNeMo \citep{physicsnemo2023nvidia} as
\texttt{physicsnemo.\allowbreak experimental.\allowbreak models.\allowbreak isla} at
\url{https://github.com/NVIDIA/physicsnemo}.}

\section{What a Solution Cannot Depend On}
\label{sec:problem}

\paragraph{Setting.}
We consider steady boundary-value problems on a domain $\dom \subset \R^d$ whose boundary is a
union of pieces, $\bdry = \bigcup_b \bdry_b$, one per boundary condition:
\begin{equation}
\begin{aligned}
  \mathcal{F}(u, \nabla^{\le k} u;\,\mu) &= 0 &&\text{in }\dom, \\[-2pt]
  \mathcal{B}_b(u, \nabla^{\le k-1} u) &= g_b &&\text{on }\bdry_b .
\end{aligned}
  \label{eq:bvp}
\end{equation}
Here $u$ collects the unknown fields, $\nabla^{\le k} u$ their spatial partial derivatives up to
order $k$, and $g_b$ the data prescribed by condition $b$.
The constants $\mu$ hold material properties and any vector attached to no finite boundary, such
as a far-field velocity on an unbounded domain; a state prescribed at a finite outer boundary
enters through its $g_b$.
Given the \emph{posed problem} $\prob = \bigl(\{\bdry_b, g_b\}_b,\ \mu\bigr)$ alone, a surrogate
approximates $\sol: \prob \mapsto u$ from examples.

\paragraph{A linear model problem.}
For a linear problem with constant coefficients, the field at an interior point is an integral
over the boundary, $u(\vx) = \int_\bdry G(\vx - \vy)\, \psi(\vy) \dd S_\vy$, of a density $\psi$
that the boundary data determine
\citep{hess1967calculation,pozrikidis1992boundary,mclean2000strongly,kress2014linear}.
The formula has three features: every boundary point influences every field point, its
influence is weighted by the surface measure, and it depends on relative positions alone.
\isla{} keeps all three and does not assume linearity.

\paragraph{Presentations.}
A model receives a \emph{presentation} of $\prob$: the coordinates of $\bdry$ in some frame and
units, a mesh of $\bdry$ or a sample of one, and a list of points at which $u$ is requested.
Infinitely many presentations describe one problem.
Assume that the problems of interest are closed under rigid motions, reflections, and changes of
units, each with one solution, and that the equations hold in every frame and system of units
(\cref{ass:physical}).
Then the solution operator
(i) is covariant under rigid motions and reflections,
(ii) rescales each output by the factor that its units dictate,
(iii) depends on $\bdry$ as an oriented surface, and not on how $\bdry$ is meshed or sampled or
on the order of its cells, and
(iv) at each point does not depend on the other query points
(\cref{prop:operator}, stated and proved in \cref{app:proofs}).

The proof uses nothing about the form of $\mathcal{F}$.
We call a surrogate \emph{presentation-independent} if it satisfies exactly (i), (ii), (iv), and
the cell-order part of (iii).
We call it \emph{consistent} if its dependence on the mesh and sample vanishes under refinement
\citep{steinbach2008numerical,sauter2011boundary}; \isla{} is both, under the hypotheses of
\cref{sec:guarantees}.

\section{The ISLA Architecture}
\label{sec:method}

\begin{figure*}[t]
  \centering
  \includegraphics[width=\textwidth]{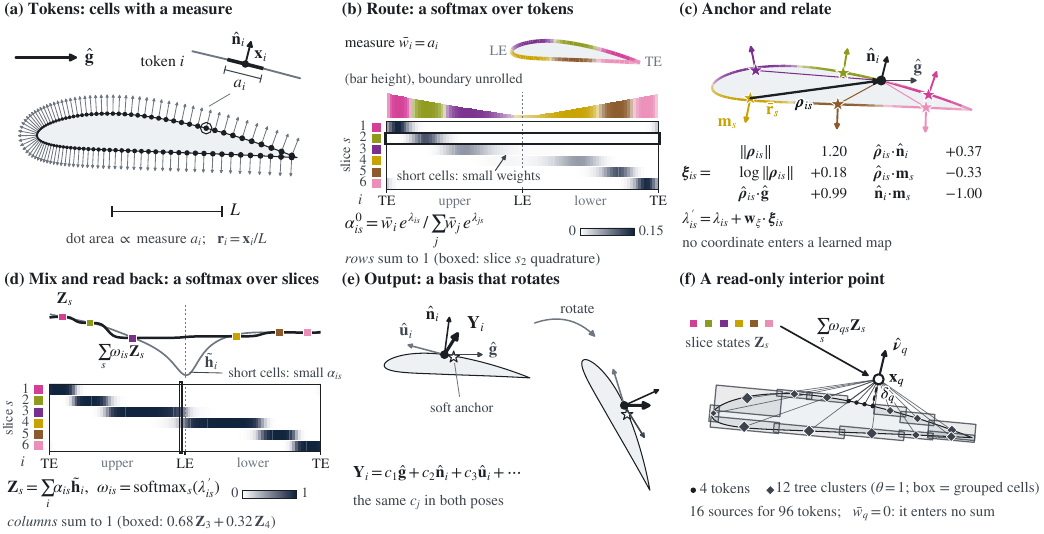}
  \caption{\textbf{One block of \isla{}, piece by piece}, on an airfoil with 96 cells and six
  slices (illustrative scores, not a trained model).
  \textbf{(a)} A token: a cell with centroid, normal, and measure; lengths in units of $L$.
  \textbf{(b)} The routing weights $\alpha^0_{is}$, a softmax over tokens with the measure
  ($\alpha_{is}$ uses the refined scores $\lambda'_{is}$): short leading-edge cells get small
  weights; color: each cell's top-scoring slice.
  \textbf{(c)} Each slice's anchor (star and arrow) and the six scalars ($K = 1$) relating one
  token to it.
  \textbf{(d)} For one channel: token states $\htilde_i$ (gray; here $\hn_i \cdot \hg$), slice
  states $\vZ_s$ (squares), read-back weights $\omega_{is}$, a softmax over slices, and the
  read-back (black).
  \textbf{(e)} A vector output in two poses, with the same coefficients.
  \textbf{(f)} A query point's distance and direction to the boundary, and the clusters and tokens
  of its boundary read at $\theta = 1$.}
  \label{fig:method}
\end{figure*}

\isla{} keeps what makes slice attention effective: a learned soft partition through which every
token exchanges information with every other at linear cost.
It changes what the network receives and how it sums.
\Cref{fig:method} follows one block on an airfoil, and \cref{fig:slices} shows a trained
model's slices on a car (pseudocode in \cref{app:architecture}).

\subsection{Tokens with a Measure and a Scale}
\label{sec:method-tokens}

\paragraph{Inputs.}
A case is one mesh per boundary condition (an inlet, an outlet, a no-slip wall).
A \emph{token} is a cell $i$ with its centroid $\vx_i$, its unit normal $\hn_i$, its measure
$a_i$ (area for 2D boundaries, or length for 1D boundaries), and any data that its condition
prescribes.
The global data are $K$ unit directions $\hg_k$ (such as the inflow direction) and scalars
$\sigma$ (such as a Reynolds number).
A \emph{surface task} predicts fields on the cells of a boundary, and an \emph{interior task}
fields at query points $\vx_q$ in the domain (\cref{sec:method-interior}).
A token's first state is computed from invariants only (the cosines $\hn_i \cdot \hg_k$, norms
and inner products of its condition's data, and $\sigma$) by a per-condition layer; all later
layers are shared (\cref{fig:method}a).

\paragraph{An intrinsic unit of length.}
Coordinates enter only after division by a length that the boundary itself defines, the
root-mean-square distance between two points of the surface,
\begin{equation}
  L^2 \;=\; \frac{1}{A^2} \int_\bdry\!\!\int_\bdry \lVert \vx - \vy \rVert^2 \dd S_\vx \dd S_\vy ,
  \qquad A = \int_\bdry \dd S ,
  \label{eq:length}
\end{equation}
that is, twice the measure-weighted variance of position, evaluated exactly on the mesh in
$O(N)$ operations.
$L$ needs no origin, is unchanged when a cell is split, and scales with the geometry.
A problem may declare which of its data define other scales, such as a speed $U$ from the inlet
and the fluid's density.
Every dimensional input is divided by its scales and every output multiplied by its own (a
pressure by the density times $U^2$).
By the Buckingham $\Pi$ theorem \citep{buckingham1914physically,barenblatt1996scaling}, physical
laws relate dimensionless groups alone; feeding only dimensionless numbers lets the network apply
arbitrary nonlinear maps without breaking unit invariance.
We write $\vr_i = \vx_i / L$.

\paragraph{Measure weights.}
Each token has a weight $w_i$, the surface measure it represents: $w_i = a_i$ when every
cell is a token.
When the boundary has more cells than a budget of $B$ tokens (a car has $10^7$; we use
$B = 2\times10^5$), a cell kept with probability $\pi_i$ has the Horvitz--Thompson weight
$w_i = a_i / \pi_i$ \citep{horvitz1952generalization}.
On interior tasks the model draws the sample itself, with $\pi_i \propto a_i$, systematically
along a space-filling curve \citep{madow1949theory,stevens2004spatially}.
On surface tasks the pipeline draws cells uniformly.
Each condition is normalized by its own measure $A_b$ and given a learned log share $\beta_b$,
so that a far-field boundary's share of every sum does not grow with its radius.
With $\bar w_i = e^{\beta_{b(i)}}\, w_i / A_{b(i)}$, every softmax over tokens is
\begin{equation}
  \wsoftmax_i^{\bar w}\,(\lambda_i) \;=\; \frac{\bar w_i\, e^{\lambda_i}}{\sum_j \bar w_j\, e^{\lambda_j}} ,
  \label{eq:wsoftmax}
\end{equation}
that is, the measure enters as a bias $\log \bar w_i$ on the score
\citep{bolya2023token,yang2026learning}.
A sum $\sum_i \wsoftmax_i^{\bar w}(\lambda_i)\, f_i$ is then a Nadaraya--Watson ratio of two
quadratures \citep{nadaraya1964estimating,watson1964smooth} of the same two surface integrals on
every mesh.
A short cell near a leading edge stands for little surface, so it has little weight
(\cref{fig:method}b), at any mesh density.

\subsection{The Invariant Slice Block}
\label{sec:method-block}

\begin{figure*}[t]
  \centering
  \includegraphics[width=\textwidth]{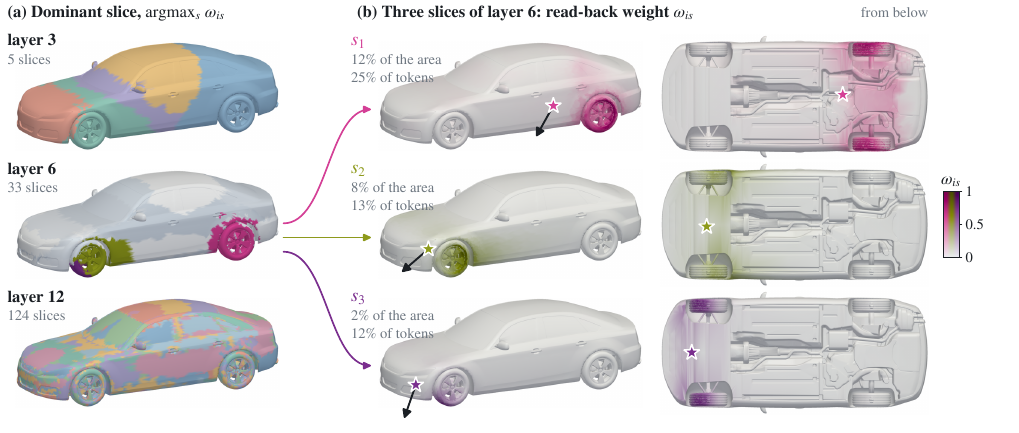}
  \caption{\textbf{The slices of a trained \isla{} on one DrivAerML case.}
  \textbf{(a)} Each surface token colored by its dominant slice, the slice with the largest
  read-back weight $\omega_{is}$, in layers 3, 6, and 12, with the number of dominant slices; pale
  tints only separate slices and are unrelated between layers; the three saturated colors of layer
  6 are the slices of (b).
  \textbf{(b)} The three slices of layer 6 dominant at the most tokens, with their area and token
  shares: $\omega_{is}$ from the side and below, with routing anchor (star)
  and mean normal (arrow).}
  \label{fig:slices}
\end{figure*}

A block updates the token states $\vh_i \in \R^C$ in four steps (\cref{fig:method}b--d), with
the layer norm $\htilde_i = \LN(\vh_i)$.

\paragraph{Route.}
Each of $M$ slices has a prototype $\vp_s \in \R^C$, a learned vector plus a measure-weighted
attention pooling of the case's tokens (\cref{app:prototypes}).
A token is scored by a cosine similarity scaled by a fixed temperature $\tau = 10$,
$\lambda_{is} = \tau\, \htilde_i \cdot \vp_s / (\lVert \htilde_i \rVert\, \lVert \vp_s \rVert)$, and
routed by $\alpha^0_{is} = \wsoftmax_i^{\bar w}(\lambda_{is})$, a softmax over the tokens.
Each slice spreads one unit of weight over the boundary in proportion to each token's surface,
so a slice is a quadrature average over a soft subset.

\paragraph{Anchor.}
From the same weights each slice forms its \emph{anchor}, a position and a facing direction,
\begin{equation}
  \vrbar_s = \sum_i \alpha^0_{is}\, \vr_i ,
  \qquad
  \vm_s = \frac{\sum_i \alpha^0_{is}\, \hn_i}{\bigl\lVert \sum_i \alpha^0_{is}\, \hn_i \bigr\rVert} .
  \label{eq:anchor}
\end{equation}
Both are weighted means, with fixed weights, of vectors that move with the body, so the anchors
move with it and give each layer $M$ adaptive reference points (stars in \cref{fig:method}c).

\paragraph{Relate.}
With $\vrho_{is} = \vr_i - \vrbar_s$ and $\hrho_{is}$ its direction, $5 + K$ scalars relate
token $i$ to anchor $s$,
\begin{equation}
\begin{aligned}
  \vxi_{is} = \bigl[\, &\lVert\vrho_{is}\rVert,\ \log\lVert\vrho_{is}\rVert,\ \hrho_{is}\!\cdot\hg_k,\\
                       &\hrho_{is}\!\cdot\hn_i,\ \hrho_{is}\!\cdot\vm_s,\ \hn_i\!\cdot\vm_s \,\bigr] :
\end{aligned}
\label{eq:invariants}
\end{equation}
how far the anchor is, whether it lies along each global direction, whether the token and the
anchor each lie in front of the other's face, and whether the two faces agree.
Each scalar is a norm or an inner product of vectors that rotate together, so it is unchanged by
every rotation, reflection, and translation.
These scalars are the only positional information that a boundary token receives
(\cref{fig:method}c).
A linear map of them refines the routing,
$\lambda'_{is} = \lambda_{is} + \mathbf{w}_\xi \cdot \vxi_{is}$.

\paragraph{Mix and read back.}
The refined scores give $\alpha_{is} = \wsoftmax^{\bar w}_i(\lambda'_{is})$ over tokens, which
forms the slice states, and $\omega_{is} = \wsoftmax_s(\lambda'_{is})$ over slices (unweighted,
since slices have no measure), which returns them to the tokens.
With biases omitted,
\begin{align}
  \vZ_s &= \textstyle\sum_i \alpha_{is}\, \htilde_i , \notag\\
  \vZ_s &\leftarrow \vZ_s + \MLP_{\mathrm{slice}}(\LN(\vZ_s)) , \label{eq:slice}\\
  \vh_i &\leftarrow \vh_i + W_r \bigl[\, \htilde_i ;\ \textstyle\sum_s \omega_{is} \vZ_s ;\
     W_f \textstyle\sum_s \omega_{is}\, \vxi_{is} \,\bigr] , \label{eq:readback}
\end{align}
followed by a token MLP; the last entry is the token's mean relation to its slices.
The refined scores are thus normalized twice (as in \cref{fig:method}b and d): over tokens with the measure,
so that a slice state is a quadrature average of its tokens, and over slices without it, so that
each token receives a convex mixture of slice states.
Transolver normalizes once, over slices, and pools by soft counts, so its slice states weight
every point equally.
\isla{} omits Transolver's attention among slices, which \citet{hu2026transolver} and
\citet{wen2026does} find unnecessary; a slice state reaches the rest of the body through its
tokens, at the next block's routing.
The cost is $O(N C^2 + N M C)$ per block: linear in the number of tokens, with no pairwise term.

In a trained model the partition refines with depth (\cref{fig:slices}a): a few slices share
the car in a shallow layer and almost all of the $M = 128$ in the last one, while single slices
cover regions such as the rear wheels and the floor between them (\cref{fig:slices}b);
consistent with reflection equivariance (\cref{thm:invariance}), their weights are nearly
left--right symmetric on this nearly symmetric car.

\subsection{Outputs of Any Tensor Rank}
\label{sec:method-heads}

The states are invariant, so a vector or tensor output is assembled at the end, from invariant
coefficients and the token's own directions
\citep{zheng1994theory,ling2016reynolds,villar2021scalars}:
\begin{equation}
  \mathbf{Y}_i = \sum_{j_1 \dots j_r} c_{i, j_1 \dots j_r}\;
  \vd_{i j_1} \otimes \dots \otimes \vd_{i j_r}
  \label{eq:head}
\end{equation}
plus isotropic terms, with $c_i = W_h \LN(\vh_i) + b_h$ and basis vectors $\vd_{ij}$ built from the global directions
$\hg_k$, the normal $\hn_i$, and the direction $\hu_i$ to the token from its \emph{soft
anchor}, a mixture of anchors under one more routing (\cref{app:heads}).
A pressure is one coefficient, and a wall shear stress combines the basis vectors
(\cref{fig:method}e).

\subsection{Read-Only Interior Points}
\label{sec:method-interior}

A query point $\vx_q$ in the domain has no normal and no measure (\cref{fig:method}f).
The \emph{interior form} of \isla{} computes two geometric quantities for it from the
full-resolution surface.
The first is its signed distance $\delta_q$ to the boundary, from an exact closest-point search
whose sign comes from the generalized winding number \citep{jacobson2013robust,barill2018fast}.
The second, the direction $\hnu_q$ in which that distance increases, replaces the normal in
every invariant.

With no measure ($\bar w_q = 0$), a query enters no sum over tokens.
At each layer it performs the read-back of \cref{eq:readback} with the boundary's slice
states and attends to every boundary token $j$ directly (the \emph{boundary read}), as in a
boundary-integral method:
\begin{equation}
  \vh_q \leftarrow \vh_q + W^{o} \sum_{j} \wsoftmax^{\bar w}_j
  \bigl( \kappa(\viota_{qj}) + \langle \mathbf{q}_q, \mathbf{k}_j \rangle \bigr)\, \mathbf{v}_j ,
  \label{eq:read}
\end{equation}
for one head and without biases and layer normalizations (\cref{alg:isla}), where
$\mathbf{k}_j$ and $\mathbf{v}_j$ are keys and values of the boundary states and $\mathbf{q}_q$ is
the query vector of $\vh_q$, scaled by $1/\sqrt{c_k}$ for its $c_k$ channels.
$\kappa$ is a learned kernel of the pair's invariants $\viota_{qj}$: a distance in units of $L$, its
logarithm, the logarithm of its ratio to $\lvert\delta_q\rvert$, and cosines (\cref{app:tree}).
A tree makes this \emph{boundary read} affordable, as in \globe{} \citep{sharpe2025globe} and
panel-clustering and $N$-body codes
\citep{rokhlin1985rapid,barnes1986hierarchical,hackbusch1999sparse}: a distant cluster enters as
one virtual token,
with the measure-weighted means of its members, when its size is below $\theta$ times its
distance and its normals agree, where $\theta$ is the tree's opening angle.
We use $\theta = 1$, about \isn{fact-tree-sources} clusters and tokens per query in place of
$2\times10^5$.
At $\theta = 0$ the read is exact.

Since queries contribute to no sum, the boundary side runs once per case, and a prediction at
$\vx_q$ is a function of $\vx_q$ and the boundary alone, so large fields are evaluated in
chunks.

\paragraph{Configuration.}
\label{sec:method-training}
We use one configuration on every dataset: width $C = 512$, $12$ blocks, and
$M = 128$ slices, with \isn{params-surface}~M parameters (\isn{params-volume}~M for the interior
form).

\section{Guarantees}
\label{sec:guarantees}

The three principles make \isla{} presentation-independent (\cref{sec:problem}) for any values of its
learned parameters and consistent under remeshing (proofs in \cref{app:proofs}).

\begin{theorem}[Exact invariances]
\label{thm:invariance}
In exact arithmetic, with every boundary cell a token and no cell's weight below $10^{-12}$ of
the case's total measure, the outputs of \isla{} are
\emph{(i)} equivariant under rotations, reflections, and translations of the cells, their
normals and data, the global directions, and the query points;
\emph{(ii)} under a rescaling of all lengths and of the units of the declared scales, with data
that declare no units held fixed, multiplied by the factors of their units before target
standardization, with unchanged internal states;
\emph{(iii)} unchanged by a permutation of the cells, by a common rescaling of the measures
given with them, and by replacing a cell with coincident copies that share its measure; and
\emph{(iv)} at each query point, independent of the other query points; this clause needs
neither hypothesis and holds for every $\theta \ge 0$ and every fixed draw of tokens.
For the interior form, (i) to (iii) hold with the exact boundary read ($\theta = 0$) when the
boundaries form a closed surface, at every query point that has a unique closest point on it.
\end{theorem}

The proof is an induction over the layers: a measure-weighted softmax of invariant scores has
invariant weights that sum to one, so the anchors of \cref{eq:anchor} are equivariant, the
relations of \cref{eq:invariants} are invariant, and so is the next state.
Clause (ii) matches \cref{prop:operator}(ii) when every dimensional constant is a declared scale
or enters a global scalar; without a Reynolds number, \isla{} predicts one dimensionless field at
all sizes.

\paragraph{Consistency.}
For remeshing the statement is weaker, convergence in place of invariance
(\cref{prop:quadrature}, \cref{app:consistency}).
Along refining meshes whose cell normals converge to the surface's, every measure-weighted sum
converges to a ratio of surface integrals.
With every cell a token, $\theta = 0$, and no weight floor, the outputs have a mesh-independent
limit; without the measure, it follows the mesher's density.

\paragraph{Approximations.}
Two approximations lie outside the theorem: the tree at $\theta = 1$, which every interior model
but one ablation uses, and, on \aerozoo{} and DrivAerML, a draw of $2\times10^5$ tokens from up to
$10^7$ cells.
Both are laid out in the frame of the boundary's principal axes, which is ambiguous for a
symmetric boundary (\cref{app:thm-proof}).
In float32 a rigid motion changes DrivAerML interior predictions by
\isn{cert-dmlv-isla-f32-rot-nut-pct} to \isn{cert-dmlv-isla-f32-rot-u-pct}\%, and a second token
draw by \isn{cert-dmlv-isla-f32-tokendraw-nut-pct} to \isn{cert-dmlv-isla-f32-tokendraw-u-pct}\%
(\cref{app:certificates}).

\section{Experiments}
\label{sec:experiments}

\paragraph{Setup.}
\label{sec:exp-setup}
\aerozoo{} \citep{sharpe2026aerozoo} contains \isn{az-n-released} time-averaged RANS solutions,
each with its own geometry, in eight enclosure configurations, at Reynolds numbers from $10^3$
to $10^8$; we use the \isn{az-n-canonical} whose averages are converged.
AirfRANS \citep{bonnet2022airfrans} contains RANS solutions of airfoil flows, and DrivAerML
\citep{ashton2024drivaerml}, WindsorML \citep{ashton2024windsorml}, and AhmedML
\citep{ashton2024ahmedml} time-averaged scale-resolving simulations of flows around road vehicles; on these
four we use the splits of FluidsBench \citep{fluidsbench2026fluidsbench}.
We compare \isla{} with Transolver \citep{wu2024transolver} and GeoTransolver
\citep{adams2025geotransolver}, and on AirfRANS and the automotive datasets also with \globe{}
(\cref{app:setup}).

\subsection{AeroZoo}
\label{sec:exp-aerozoo}

\begin{figure*}[t]
  \centering
  \includegraphics[width=\textwidth]{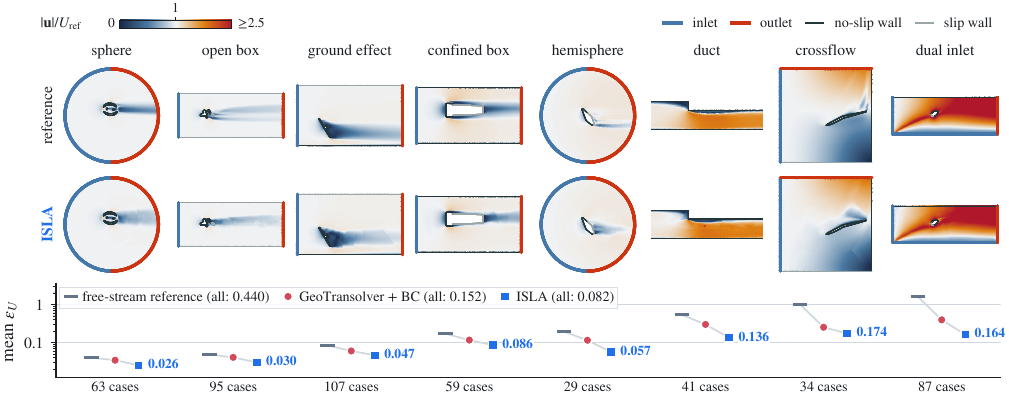}
  \caption{\textbf{\isla{} has a lower velocity error than both baselines in every \aerozoo{} configuration.}
  \emph{Top:} a median-error test case per configuration: the simulated speed on a plane through
  the body and the prediction of one \isla{} trained with learning rate $3.75\times 10^{-4}$ (duct:
  near its contraction).
  \emph{Bottom:} mean $\varepsilon_U$ by configuration for the runs of \cref{tab:aerozoo-main}; \isla{}'s is
  \isn{az-enc-ratio-min}--\isn{az-enc-ratio-max}$\times$ lower than GeoTransolver + BC's.}
  \label{fig:aerozoo}
\end{figure*}

\begin{table}[t]
  \caption{\textbf{\aerozoo{}, \splitname{full}}: means over \isn{az-full-n} test cases and
  \isn{az-rep-seeds} seeds (last row: \isn{az-joint-seeds}); bold: best per column.
  $S_{\mathrm{flow}}$: \aerozoo{}'s overall score, the geometric mean of the four errors over the
  free-stream reference's; near: within $1\,L_{\mathrm{ref}}$ of the body; last column: share of
  cases with $\varepsilon_U$ below the reference's; BC: boundary-condition tokens; flow axis: the
  inlet-to-outlet direction as an input.}
  \label{tab:aerozoo-main}
  \centering
  \scriptsize
  \setlength{\tabcolsep}{1.8pt}
  \begin{tabular}{@{}l r rrrr r@{}}
\toprule
 & Overall $\downarrow$ & \multicolumn{4}{c}{Mean ref.-scaled $\downarrow$} & Share $\uparrow$ \\
\cmidrule(lr){3-6}
Model & $S_{\mathrm{flow}}$ & $\varepsilon_U$ & $\varepsilon_U^{\mathrm{near}}$ & $\varepsilon_p$ & $\varepsilon_p^{\mathrm{near}}$ & $\varepsilon_U <$ ref. \\
\midrule
Free-stream reference & 1.000 & 0.440 & 0.442 & 1.809 & 2.014 & \na \\
Zero predictor & 1.648 & 1.203 & 1.193 & 1.809 & 2.014 & \na \\
\midrule
Transolver + flow axis & 0.679 & 0.306 & 0.309 & 1.212 & 1.315 & 61\% \\
GeoTransolver + BC & 0.290 & 0.152 & 0.179 & 0.427 & 0.434 & 86\% \\
\midrule
ISLA & 0.153 & 0.082 & 0.112 & 0.196 & 0.214 & 98\% \\
ISLA + boundary targets & \textbf{0.123} & \textbf{0.080} & \textbf{0.110} & \textbf{0.124} & \textbf{0.149} & 98\% \\
\bottomrule
\end{tabular}
\end{table}

\aerozoo{} varies the problem (geometry, boundary-condition layout, and speed) and its
presentation (size and pose) together (\cref{fig:aerozoo,tab:aerozoo-main}).
Its $\varepsilon_U$ and $\varepsilon_p$ are volume-weighted root-mean-square errors of
velocity and pressure, divided by the case's reference speed and dynamic pressure
(\cref{app:aerozoo-setup}).
\isla{}'s $S_{\mathrm{flow}}$ is \isn{az-full-sflow-isla} against the best baseline's
\isn{az-full-sflow-gtb}; its mean $\varepsilon_U$ and $\varepsilon_p$ are
\isn{az-full-w90-u-ratio} $[\isn{az-full-w90-u-lo}, \isn{az-full-w90-u-hi}]$ and
\isn{az-full-w90-p-ratio} $[\isn{az-full-w90-p-lo}, \isn{az-full-w90-p-hi}]$ times lower (90\%
Welch intervals, six seeds each; \cref{tab:aerozoo}).
It has a lower $\varepsilon_U$ than the \emph{free-stream reference}, the inlet state
everywhere, on \isn{az-full-share-isla}\% of cases.

As released, Transolver and GeoTransolver have no input for the inlet conditions: in SI units
their mean $\varepsilon_U$ is within 2\% of predicting zero \citep{sharpe2026aerozoo}.
With tokens that carry the boundary conditions (BC), GeoTransolver in SI units is still less
accurate than the free-stream reference \citep{sharpe2026aerozoo}, on cases whose sizes span a
factor of $10^3$ and speeds a factor of 200.
We therefore retrained both baselines with per-case scales at \isla{}'s step budget, each with a
learning rate and schedule chosen on validation by \isla{}'s rule
(\cref{app:aerozoo-setup}); GeoTransolver + BC then has a lower $\varepsilon_U$ than the reference on
\isn{az-full-share-gtb}\% of cases.
An \isla{} of GeoTransolver + BC's size (\isn{az-cap-params-small}~M and
\isn{az-baseline-params-gt}~M parameters, with half the boundary tokens) still has mean
$\varepsilon_U$ and $\varepsilon_p$ \isn{az-cap-small-gt-u-ratio} and
\isn{az-cap-small-gt-p-ratio} times lower (\cref{tab:aerozoo-capacity}).
Also supervising the boundary fields (the last row of \cref{tab:aerozoo-main};
\isn{az-joint-step-ratio} times the step time) lowers \isla{}'s mean $\varepsilon_p$ to
\isn{az-joint-epsp-mean} and $\varepsilon_U$ to \isn{az-joint-epsU-mean}.

\paragraph{Held-out families and layouts.}
With a geometry family held out (blobs, or CAD parts), \isla{}'s mean
$\varepsilon_U$ on cases with a body is \isn{az-blobs-epsU-mean-isla} and
\isn{az-ztc-epsU-mean-isla} (free-stream reference: \isn{az-blobs-epsU-mean-fs} and
\isn{az-ztc-epsU-mean-fs}), and with Reynolds numbers above $10^6$ held out
\isn{az-re-epsU-mean-isla} (free-stream reference: \isn{az-re-epsU-mean-fs}).
With the crossflow or dual-inlet layout held out, it is only \isn{az-pcf-ratio-pct}\% and
\isn{az-pdi-ratio-pct}\% below the free-stream reference's, and with non-box enclosures held out it
exceeds it (\cref{tab:aerozoo-splits}): presentation independence does not extrapolate to unseen
topologies.
Across all seven splits, GeoTransolver + BC's $S_{\mathrm{flow}}$, velocity error, and pressure error are
\isn{az-poolcur-s-2dp} $[\isn{az-poolcur-s-lo-2dp}, \isn{az-poolcur-s-hi-2dp}]$,
\isn{az-poolcur-u-2dp} $[\isn{az-poolcur-u-lo-2dp}, \isn{az-poolcur-u-hi-2dp}]$, and
\isn{az-poolcur-p-2dp} $[\isn{az-poolcur-p-lo-2dp}, \isn{az-poolcur-p-hi-2dp}]$ times those of
\isla{} at $3.75\times10^{-4}$ (geometric means, 95\% intervals; \cref{fig:aerozoo-heldout}).

\subsection{Pose and Re-sampling Tests}
\label{sec:exp-encoding}

We give each trained model the same evaluation problems in another presentation and measure its
error again, on \isn{inv-n-tasks} tasks (\cref{fig:encoding} in \cref{app:encoding-results}): AirfRANS,
DrivAerML, HiLiftAeroML \citep{ashton2026hiliftaeroml}, SHIFT-SUV \citep{luminary2025shiftsuv},
and three solid-mechanics datasets (\cref{app:encoding-setup}).
In the \emph{pose} test, every case, with its global directions and vector targets, is rotated
at random and translated by one body radius.
\emph{Re-sampling by area} gives a model surface cells drawn with probability proportional to
their area instead of the mesher's pattern, and every model's errors are weighted by measure.
These models were trained without augmentation, so the tests isolate invariance.

\begin{table}[t]
  \caption{\textbf{AirfRANS} test cases ($p_s, \tau_w$: surface pressure, wall shear;
  $\mathbf{u}, p_v$: volume velocity, pressure; bold: lowest).}
  \label{tab:airfrans}
  \centering
  \footnotesize
  \setlength{\tabcolsep}{4pt}
  \begin{tabular}{@{}l cccc@{}}
\toprule
 & \multicolumn{4}{c}{Relative $L^2$ error [\%] $\downarrow$} \\
\cmidrule(l){2-5}
Field & Transolver & GeoTransolver & GLOBE & ISLA \\
\midrule
\multicolumn{5}{@{}l}{\splitname{full} (720 training cases)} \\
\quad $p_s$ & 1.17 & 1.20 & \textbf{0.34} & 0.55 \\
\quad $\tau_w$ & 1.76 & 1.46 & \textbf{0.55} & 0.57 \\
\quad $\mathbf{u}$ & 0.53 & 0.47 & \textbf{0.15} & 0.30 \\
\quad $p_v$ & 1.14 & 1.09 & \textbf{0.28} & 0.57 \\
\addlinespace[2pt]
\multicolumn{5}{@{}l}{\splitname{aoa_extrapolation} (724 training cases)} \\
\quad $p_s$ & 8.31 & 9.29 & \textbf{1.94} & 2.58 \\
\quad $\tau_w$ & 6.44 & 6.60 & \textbf{2.14} & 2.67 \\
\quad $\mathbf{u}$ & 3.84 & 3.91 & \textbf{0.98} & 1.37 \\
\quad $p_v$ & 12.21 & 13.02 & \textbf{3.87} & 4.55 \\
\bottomrule
\end{tabular}
\end{table}

\paragraph{All \isn{inv-n-tasks} tasks.}
A rotated case lies outside everything that Transolver and GeoTransolver were trained on, and
their errors rise to \isn{inv-posed-abs-all-min} to \isn{inv-posed-abs-all-max} times that of
predicting zero.
Re-sampling by area changes only the density of the points that they receive, yet it raises
their surface-pressure errors on DrivAerML from \isn{hero-native-pct}\% to
\isn{inv-dmls-area-t-pct}\% and \isn{inv-dmls-area-gt-pct}\%, and on HiLiftAeroML and SHIFT-SUV to
\isn{inv-area-abs-other-min} to \isn{inv-area-abs-other-max}\%.
\isla{}'s error changes by at most \isn{inv-isla-max-pct}\% of its value in every test.

\paragraph{Alignment and augmentation.}
For networks from the DrivAerML ensembles of \cref{sec:exp-bench}, on its
\isn{encb-n-cases} validation cases, rotating the full PDE problem multiplies the surface-pressure
errors of Transolver and GeoTransolver by \isn{encb-rot-ratio-min} to \isn{encb-rot-ratio-max}
and changes \isla{}'s by at most \isn{encb-isla-rot-dev-pct}\% (\cref{tab:encoding-bench}).
Rotating each case into the car's principal axes before prediction leaves their errors at
\isn{canon-raw-t-ratio} and \isn{canon-raw-gt-ratio} times the stored-case values, or
\isn{canon-aligned-t-ratio} and \isn{canon-aligned-gt-ratio} times after one more rotation fitted
on the training cases.
Training with rotations about the vertical axis removes the rise under such rotations but leaves
it at \isn{encb-aug-so3-ratio}$\times$ under
arbitrary rotations (\cref{tab:encoding-bench}); training with arbitrary rotations holds the change
within \isn{aug-rot-change-max-pct}\% and raises stored-case surface-pressure errors by
\isn{aug-t-native-excess-pct}\% and \isn{aug-gt-native-excess-pct}\% (\cref{tab:encoding-aug}).

\subsection{AirfRANS}
\label{sec:exp-airfrans}

On the splits of FluidsBench \citep{fluidsbench2026fluidsbench}, each model's result averages
the predictions of several independently trained networks (\cref{app:airfrans-setup}).
On \splitname{full}, \globe{} has the lowest error in every field
(\cref{tab:airfrans}); \isla{}'s are \isn{af-full-isla-over-globe-min} to
\isn{af-full-isla-over-globe-max} times \globe{}'s and \isn{af-full-tgt-over-isla-min} to
\isn{af-full-tgt-over-isla-max} times lower than the better of Transolver's and GeoTransolver's.
On \splitname{aoa_extrapolation} (unseen angles), every model's errors rise
\isn{af-aoa-rise-min}--\isn{af-aoa-rise-max}$\times$ from validation to test, and \isla{}'s stay
\isn{af-aoa-tgt-over-isla-min} to \isn{af-aoa-tgt-over-isla-max} times lower than the better of
the two.

\subsection{Automotive Aerodynamics Datasets}
\label{sec:exp-bench}

We trained all four models on the FluidsBench splits of DrivAerML, WindsorML, and AhmedML, again
averaging \isn{bench-auto-members-min} to \isn{bench-auto-members-max} networks per result
(\cref{app:protocol}).
Because each dataset sweeps one vehicle family in a fixed pose and boundary layout, full-split
errors are already near saturation for existing architectures \citep{sharpe2026aerozoo}; we
report them for completeness.
On the full splits (\isn{bench-full-val-note}), \isla{}'s surface
errors lie \isn{auto-full-surf-below-pct}\% below to \isn{auto-full-surf-above-pct}\%
above the best other model's, and its volume errors
\isn{auto-full-vol-span} (\cref{tab:bench}).
With \isn{bench-scarce-train-dml} training cases (DrivAerML, AhmedML), errors lie
\isn{auto-scarce-below-pct}\% below to \isn{auto-scarce-above-max-pct}\% above; with DrivAerML's
\isn{bench-min-train-dml} training cases (\isn{bench-dml-ss-phase} errors; every network trained
at least \isn{steps-drivaerml-isla-super-scarce} steps), \isla{}'s errors are
\isn{bench-ss-field-cut-min}--\isn{bench-ss-field-cut-max}\% below the best other model's in all
four fields (\cref{app:bench-all}).

\subsection{Ablations and Cost}
\label{sec:exp-cost}

We remove one component at a time and retrain (\cref{tab:ablation} in \cref{app:ablations}; the
DrivAerML surface task unless stated).
The first two variants keep their accuracy on cases as stored and lose the one invariance that
their principle provides.
With counts in place of measures, the surface-pressure error on cells re-sampled by area is
\isn{abl-s1-area} times higher.
With coordinates as inputs, it is about \isn{abl-s2-native-change-pct}\% lower on stored cases
and \isn{abl-s2-posed} times higher on rotated cases.
Without anchor relations it is \isn{abl-s3-native-abs-pct}\% in every presentation,
\isn{abl-s3-native} times the full model's.
On AirfRANS, dropping the boundary read raises velocity error \isn{abl-v1-change-u-pct}\% and
lowers eddy-viscosity error \isn{abl-v1-change-nut-pct}\%; training at $\theta = 0$
(\isn{abl-th0-step-ratio}$\times$ step time) matches $\theta = 1$ within
\isn{abl-th0-diff-max-pct}\% (\isn{abl-v1-n-seeds} seeds).

\isla{}'s cost is linear in the boundary tokens and the query points (\cref{app:cost}).
One float32 prediction for $2\times10^5$ surface cells of a car takes
\isn{cost-surf-isla-eager-f32-s}~s on one GPU (Transolver: \isn{cost-surf-t-f32-s}~s; 9~M
GeoTransolver: \isn{cost-surf-gt-f32-s}~s).

\section{Related Work}
\label{sec:related}

Slice transformers route mesh points into learned slices
\citep{wu2024transolver,adams2025geotransolver}, while
anchor models compress them into a subset \citep{alkin2024universal,alkin2025abupt}; AB-UPT and
SMART also decouple read-only queries \citep{alkin2025abupt,hagnberger2026smart}, and
\citet{wen2026does} find slice means stable across subsamples drawn by one rule.
Where layerwise equivariant networks constrain intermediate features
\citep{thomas2018tensor,geiger2022e3nn,liao2023equiformer,brehmer2023geometric,rygiel2026when},
\isla{} maps invariant scalars to covariant outputs \citep{villar2021scalars,ling2016reynolds}
with anchors like Invariant Slot Attention \citep{biza2023invariant}.
Its log-measure bias adapts proportional attention \citep{bolya2023token,yang2026learning} per
condition with Horvitz--Thompson draws, and its boundary read builds on panel and
Green's-operator methods \citep{hess1967calculation,schafer2021sparse,boulle2022learning,sharpe2025globe}
(\cref{app:related}).

\section{Limitations and Conclusion}
\label{sec:limitations}

\paragraph{Limitations.}
Presentation independence covers pose, units, and meshing rather than unseen flow topologies:
with a layout or enclosure family held out, \isla{}'s mean error is
\isn{az-pcf-ratio-pct}\% below to \isn{az-nonbox-above-pct}\% above the free-stream reference's
(\cref{sec:exp-aerozoo}).
As a deterministic equivariant map, \isla{} cannot represent a single symmetry-breaking mode such
as an instantaneous bistable wake \citep{grandemange2013turbulent}.
\isla{} depends on where a far field is truncated: its length unit (\cref{eq:length}) scales
with the outer boundary.
Evaluated with the exact read ($\theta = 0$), AirfRANS models trained at $\theta = 1$ have \isn{abl-theta0-ratio-min} to
\isn{abl-theta0-ratio-max} times their error (\cref{app:ablations}).
Three-dimensional interior training uses up to \isn{mem-peak-trained}~GiB per GPU as trained
(\isn{mem-peak-recompute}~GiB with activation recomputation; \cref{app:compute}), and
volume-velocity error is
\isn{bench-vu-excess-dml-min}--\isn{bench-vu-excess-dml-max}\% above GeoTransolver's on
DrivAerML (\isn{bench-scarce-train-dml} or more training cases) and
\isn{bench-vu-excess-ahmed-min}--\isn{bench-vu-excess-ahmed-max}\% on full AhmedML.

\paragraph{Conclusion.}
\isla{} has, by construction, the invariances of a boundary-integral model: a change of
presentation moves its predictions only by rounding, quadrature, interior token sampling, and tree
clustering, and its error by at most \isn{inv-isla-max-pct}\% in every test.
It is second only to \globe{} on full AirfRANS, and on the full automotive splits its surface
errors lie \isn{auto-full-surf-below-pct}\% below to \isn{auto-full-surf-above-pct}\% above the
best other model's.
On \aerozoo{}'s full split, where every case has its own pose, size, and speed, its mean velocity
and pressure errors are \isn{az-full-mean-ratio} and \isn{az-full-meanp-ratio} times lower than
the best baseline's.

\clearpage
\bibliography{isla}
\bibliographystyle{icml2026}

\newpage
\appendix
\onecolumn
\paragraph{Appendix overview.}
\Cref{app:proofs} proves the results of \cref{sec:problem,sec:guarantees}.
\Cref{app:architecture} specifies the architecture of \cref{sec:method} in full and shows the slices
of a trained model (\cref{app:routing}).
\Cref{app:setup} gives the setup of \cref{sec:experiments}: the datasets and the training recipe,
then the protocol of each experiment (\aerozoo{}, the pose and re-sampling tests, AirfRANS, and
the automotive datasets).
\Cref{app:results} gives every result behind \cref{sec:experiments}: \aerozoo{}
(\cref{app:aerozoo-results}), AirfRANS (\cref{app:airfrans-results}), the automotive datasets
(\cref{app:bench-all}), the pose and re-sampling tests (\cref{app:encoding-results}), the measured
invariances (\cref{app:certificates}), cost (\cref{app:cost}), and ablations
(\cref{app:ablations}).
\Cref{app:related} extends the related work of \cref{sec:related}.

\section{Proofs}
\label{app:proofs}

Throughout, a rigid motion $(R, \vt)$ with $R \in \Od$ acts on a posed problem as follows.
It moves every point, $\vx \mapsto R\vx + \vt$.
It rotates every vector datum, $\mathbf{a} \mapsto R\mathbf{a}$, and every rank-2 datum,
$T \mapsto R T R^\top$, and it leaves scalars unchanged.
A pseudotensor takes the further factor $\det R$.
Under a reflection ($\det R = -1$) the orientation of the boundary is carried along, so that a
cell's normal maps as a true vector, $\hn \mapsto R\hn$.
A cell's normal follows its vertex order, so reflecting the coordinates alone would give
$-R\hn$, which exchanges the two sides of the boundary and poses a different problem.

\subsection{What the Solution Operator Cannot Depend On}

\begin{assumption}
\label{ass:physical}
(a) \emph{Well-posedness:} the problems of interest form a class that is closed under rigid
motions, reflections, and changes of units, and each has exactly one solution.
(b) \emph{Frame indifference:} $\mathcal{F}$ and the $\mathcal{B}_b$ commute with every rigid
motion and reflection that acts on $u$, the $g_b$, and $\mu$ by tensor type, and every vector,
tensor, or dimensional constant that they involve is part of $\prob$.
(c) \emph{Dimensional homogeneity:} $\mathcal{F}$ and the $\mathcal{B}_b$ hold in every system of
units.
\end{assumption}

\begin{proposition}[What the solution operator cannot depend on]
\label{prop:operator}
Under \cref{ass:physical}, for every posed problem $\prob$:
\begin{enumerate}[leftmargin=*, itemsep=0pt, topsep=1pt, label=(\roman*)]
  \item \emph{Euclidean covariance.}
  For $R \in \Od$ and a translation $\vt$ acting on the oriented boundary, its data, and the
  tensors in $\mu$ by tensor type,
  $\sol\bigl((R, \vt) \cdot \prob\bigr)(R\vx + \vt) = R^{\otimes r}\, \sol(\prob)(\vx)$ for a
  rank-$r$ output (times $\det R$ if a pseudotensor).
  \item \emph{Unit covariance.}
  Rewriting $\prob$ in other units rescales each output by the factor that its dimensions
  dictate.
  \item \emph{Independence of the discretization.}
  $\sol(\prob)(\vx)$ depends on $\bdry$ as an oriented surface, and not on how $\bdry$ is meshed
  or sampled or on the order of its cells.
  \item \emph{Independence of the query set.}
  $\sol(\prob)(\vx)$ does not depend on the other query points.
\end{enumerate}
\end{proposition}

\paragraph{Proof of \cref{prop:operator}.}

\emph{(i)}
Let $u = \sol(\prob)$, let $\prob' = (R, \vt)\cdot\prob$ be the moved problem, with global data
$\mu'$, and define
$u'(\vx') = R^{\otimes r}\, u\bigl(R^\top(\vx' - \vt)\bigr)$ on the moved domain, componentwise for
each field of rank $r$ (with the factor $\det R$ for a pseudotensor).
By frame indifference, $\mathcal{F}$ and each $\mathcal{B}_b$ commute with the action of
$(R, \vt)$:
$\mathcal{F}(u', \nabla^{\le k} u';\, \mu')$ is the rotated image of
$\mathcal{F}(u, \nabla^{\le k} u;\, \mu) = 0$, and
$\mathcal{B}_b(u', \nabla^{\le k-1} u')$ is the rotated image of
$\mathcal{B}_b(u, \nabla^{\le k-1} u) = g_b$, which is the datum of $\prob'$ on
the moved piece $\bdry_b$.
Hence $u'$ solves $\prob'$, which is again a problem of the class.
By well-posedness it is the only solution, so $\sol(\prob') = u'$.
This is the statement that a symmetry of an equation maps solutions to solutions
\citep{olver1993applications}.

\emph{(ii)}
A change of units multiplies every quantity $q$ by $\prod_j \chi_j^{e_j(q)}$, where
$\chi_j > 0$ rescales the $j$-th base dimension and $e_j(q)$ is the exponent of that dimension
in $q$.
By dimensional homogeneity, if $u$ solves $\prob$ then the rescaled field solves the rescaled
problem, and by well-posedness it is that problem's solution.
The Buckingham $\Pi$ theorem \citep{buckingham1914physically} follows: choosing the $\chi_j$ to normalize a
maximal set of dimensionally independent data of $\prob$ shows that the dimensionless solution
is a function of the remaining dimensionless groups.

\emph{(iii) and (iv)}
$\sol$ is a function of $\prob$, and $\sol(\prob)(\vx)$ of $\prob$ and $\vx$.
A mesh of $\bdry$, a sample of that mesh, an ordering of its cells, and a list of other points
are not part of $\prob$, so the solution cannot depend on them. \qed

\paragraph{What well-posedness excludes.}
Part (a) of \cref{ass:physical} fails where a symmetric problem has several stable solutions,
such as the bistable wake of a bluff body \citep{grandemange2013turbulent}.
There the solution operator is set-valued, and clause (i) of \cref{prop:operator} states only
that a rigid motion maps the solution set of $\prob$ onto that of the moved problem.

\subsection{Proof of \cref{thm:invariance}}
\label{app:thm-proof}

\paragraph{Full statement.}
Consider \isla{} in exact arithmetic, with any parameters for which it is defined, when every
boundary cell is a token and no weight is below its floor.
\begin{enumerate}[leftmargin=*, itemsep=0pt, topsep=1pt, label=(\roman*)]
  \item \emph{Euclidean equivariance.}
  Let $R \in \Od$ and $\vt \in \R^d$ act on the cells, their normals and data, the global
  directions, and the query points, with the orientation of the cells carried along under a
  reflection, as above.
  Every output of rank $r$ is mapped to $R^{\otimes r}$ times itself at the moved points (times
  $\det R$ for an output declared with odd parity).
  \item \emph{Similarity.}
  Multiply every coordinate by $\chi_0 > 0$ and every datum with declared units
  $L^{e_0} \prod_j S_j^{e_j}$, on the boundary or global, by $\chi_0^{e_0} \prod_j \chi_j^{e_j}$,
  with $\chi_j > 0$ for each declared scale $S_j$ (a global value that defines $S_j$ is
  multiplied by $\chi_j$), and leave every datum without declared units as it is.
  Then the internal states are unchanged, and each output, before any standardization, is
  multiplied by the factor of its declared units.
  \item \emph{Measure.}
  The outputs are unchanged by a permutation of the cells, by a common rescaling of the measures
  given with the cells, and by replacing a cell with coincident copies that share its measure.
  \item \emph{Query independence.}
  The prediction at a query point is a function of that point, the boundary, and the global
  data only.
  This clause needs neither of the two hypotheses above: it holds for every opening angle
  $\theta \ge 0$ and, with a token budget, for every fixed draw.
\end{enumerate}
For the interior form, (i) to (iii) hold with the exact boundary read ($\theta = 0$) when the
boundaries form a closed surface, at every query point that has a unique closest point on it.

Write $\mathcal{G}$ for the transformations of clauses (i) to (iii): rigid motions, uniform
rescalings of length $\vx \mapsto \chi_0 \vx$ (under which measures scale by $\chi_0^{d-1}$),
changes of units of the declared scales, a common rescaling of the measures given with the
cells, and permutations of the cells.
A quantity attached to a token is \emph{invariant} if it is unchanged by every element of
$\mathcal{G}$ other than a permutation and is permuted with the cells by a permutation.

\paragraph{Hypotheses.}
The model is defined when no normalization divides by zero, for which a positive gain $\gamma$
in every prototype map (\cref{app:prototypes}) suffices.
Each weight has a floor of $10^{-12}$ times the case's total raw measure over all conditions,
applied before the per-condition normalization, which guards the logarithm, and
the theorem assumes that no weight is below it and that every cell remains a token, before and
after a cell is replaced by copies.
The other floors keep every normalization and logarithm defined, and a clamp on a norm commutes
with $R$.
Boundary data are true tensors of rank at most two.

\begin{lemma}
\label{lem:softmax}
Let $\lambda_i$ be invariant scores and let the weights $\bar w_i$ be invariant up to a common
positive factor.
Then $\alpha_i = \bar w_i e^{\lambda_i} / \sum_j \bar w_j e^{\lambda_j}$ is invariant and
$\sum_i \alpha_i = 1$.
Consequently $\sum_i \alpha_i f_i$ is invariant for invariant $f_i$; it maps to
$R \sum_i \alpha_i \mathbf{a}_i$ for vectors $\mathbf{a}_i \mapsto R\mathbf{a}_i$; and it maps to
$R \sum_i \alpha_i \vr_i + \vt'$ for points $\vr_i \mapsto R\vr_i + \vt'$.
\end{lemma}
\begin{proof}
The common factor cancels in the ratio, so $\alpha_i$ is invariant, and the $\alpha_i$ sum to
one by construction.
A sum of invariant terms $\alpha_i f_i$ is then invariant.
The claim for vectors holds because $R$ is linear, and the claim for points uses
$\sum_i \alpha_i = 1$ to take the translation $\vt'$ out of the sum.
\end{proof}

\paragraph{Inputs.}
The length unit of \cref{eq:length} is a function of pairwise distances and of ratios of
measures.
It is therefore unchanged by rigid motions, permutations, and a common rescaling of the measures,
and it is multiplied by $\chi_0$ under $\vx \mapsto \chi_0 \vx$.
Hence $\vr_i = \vx_i / L$ maps to $R\vr_i + \vt/L$ under a rigid motion and is unchanged by a
rescaling of length.
The normalized weight $\bar w_i = e^{\beta_{b(i)}} w_i / A_{b(i)}$ is a ratio of measures of the
same kind, hence invariant; with a single declared condition the model uses $\bar w_i = w_i$,
which is invariant up to a common factor.
A declared scale is a measure-weighted root-mean-square magnitude of data, or a declared global
value.
A change of its unit multiplies it and every datum that declares it by the same factor, so
dividing each datum by its scales gives numbers that do not change with the units.
The mean direction of a field that defines a scale, which the model adds to the global
directions, is a normalized sum $\sum_i w_i \mathbf{a}_i$, so it maps to $R$ times itself and is
unchanged by the other elements of $\mathcal{G}$.
The invariant input of a token, the vector $\eta_i$ from which its first state is computed (line 7 of
\cref{alg:isla}), consists of inner products of unit vectors that rotate together ($\hn_i$,
$\hg_k$), of norms and inner products of its dimensionless data with those vectors, and of
dimensionless scalars.
All are invariant, including under reflections, because only true vectors enter.
The first state, a per-condition map of the invariant input followed by a shared one, is therefore
invariant.

\paragraph{Induction over the blocks.}
Suppose the states $\vh_i$ entering a block are invariant.
The prototypes are sums of the form of \cref{lem:softmax} over invariant quantities, so they and
the scores $\lambda_{is}$ are invariant.
By the lemma the routing weights $\alpha^0_{is}$ are invariant and sum to one, so the anchors
satisfy $\vrbar_s \mapsto R\vrbar_s + \vt/L$ and $\vm_s \mapsto R\vm_s$.
Then $\vrho_{is} = \vr_i - \vrbar_s \mapsto R\vrho_{is}$: the translation cancels exactly, with no
centering in any input of a learned map.
Every component of $\vxi_{is}$ is the norm of $\vrho_{is}$, a function of that norm, or an inner product of
two of the vectors $\hrho_{is}, \hn_i, \vm_s, \hg_k$ (or a contraction of a boundary vector or
tensor with $\hrho_{is}$), all of which rotate together.
So $\vxi_{is}$ is invariant, and with it $\lambda'_{is}$, $\alpha_{is}$, $\omega_{is}$, $\vZ_s$,
and the updated state.
A permutation of the cells permutes per-token quantities and leaves sums over tokens unchanged.
By induction the final states are invariant.

\paragraph{Outputs.}
The coefficients $c_i$ are an affine map of the normalized final state, hence invariant.
The head's anchors $\vrbar^{\mathrm f}_s$ (\cref{app:heads}) are of the form of \cref{eq:anchor}, and a token's soft
anchor is $\sum_s \omega^{\mathrm f}_{is} \vrbar^{\mathrm f}_s$ with invariant weights that sum to one
over the slices, so it maps to $R(\cdot) + \vt/L$.
The translation cancels in the difference between the token and its soft anchor, so
$\hu_i \mapsto R\hu_i$, and every basis vector, being one of $\hg_k$, $\hn_i$, $\hu_i$ or a
combination of two of them with invariant coefficients, maps as $\vd_{ij} \mapsto R\vd_{ij}$.
The output of \cref{eq:head} therefore maps to $R^{\otimes r}$ times itself; the isotropic terms
(the identity and its products) are fixed by $R^{\otimes r}$.
An odd-parity output replaces exactly one factor by a cross product of two basis directions
(in two dimensions, a perpendicular), which maps as $\det(R)\, R(\cdot)$, giving the extra sign;
an odd output of rank $d$ also has the Levi-Civita tensor, which $R^{\otimes d}$ multiplies by
$\det R$.
Multiplying each output by the scales of its units gives clause (ii).
This proves (i) and (ii), and the permutation and rescaling parts of (iii).

\paragraph{Coincident copies.}
If a cell is replaced by $n$ copies with the same vertices and data and measure $a_i / n$ each,
the copies have equal invariant inputs and, by induction, equal states in every layer, while every sum
$\sum_i \bar w_i e^{\lambda_i} f_i$ is unchanged.
The length unit $L$ is unchanged as well.
Every other cell therefore keeps its output, and each copy has the output of the cell that it
replaces, which proves the last part of (iii).

\paragraph{Queries.}
A query has zero weight, so it contributes to no sum over tokens: the boundary-side states,
prototypes, anchors, and slice states are those of the boundary alone.
Each operation on a query (its embedding, its read-back from the slices, its boundary read, its
token MLP, the head) uses only the query's own state and geometry and boundary-side quantities.
In particular the list of sources of a query in the tree read is a function of the tree and of
that query's position.
This proves (iv) for every $\theta \ge 0$, and, with a token budget, for every fixed start of
the draw, as in evaluation.

\paragraph{The interior form with the exact boundary read.}
The signed distance $\delta_q$ of a query is invariant under rigid motions and scales with
length; where the closest boundary point is unique, the direction $\hnu_q$ is the unit gradient
of the signed distance and maps as $R\hnu_q$.
The sign of $\delta_q$ comes from a winding number, which is exact for a closed surface.
(We take the winding number as evaluated exactly; in three dimensions the implementation uses a
far-field expansion with a measured error of at most $0.03$, where the margin is $1/2$.)
The query's invariant input and its relations to the anchors are then invariant as above.
The pair invariants $\viota_{qj}$ are norms and inner products of
$\vr_q - \vr_j$, $\hn_j$, $\hnu_q$, and $\hg_k$, functions of those norms and of $\delta_q / L$, and
contractions of the boundary vectors with $\hrho$ and $\hnu_q$, and the exact boundary read of \cref{eq:read} is a sum of
the form of \cref{lem:softmax} over all boundary tokens, so the query states remain invariant.
The set of points with more than one closest boundary point (the medial set) has measure zero;
there \isla{} uses the least-norm element of the convex hull of the tied directions, which is
again equivariant wherever the tie is exact. \qed

\paragraph{The approximations.}
Large interior cases use two approximations that the theorem does not cover: a draw of tokens
and a tree.
Both depend on a frame.
With a token budget below the number of cells, tokens are drawn along a Morton curve laid out in
the frame of the boundary's principal axes (signed by third moments), and with $\theta > 0$ the
boundary tokens are grouped by a tree built in the same kind of frame.
The frame moves with the boundary, and both the curve and the tree are laid out in the
boundary's own units.
The draw and the tree are therefore unchanged by rigid motions and rescalings whenever the
principal variances are distinct, no third moment along a principal axis vanishes, and no cell
lies on a threshold of a discrete decision.
Where the frame is degenerate, as for a boundary with equal principal moments or a symmetry that
reverses a principal axis (a mirror plane, two-fold rotation axis, or center of symmetry), a
rigid motion can regroup cells; the prediction then changes by the difference between two
quadratures of the same surface, or between two tree approximations of the same sum (and on an
open boundary, whose winding number is not integer-valued, the BVH far-field approximation of
the winding number can also flip the sign of $\delta_q$ near the threshold $1/2$).

The tree adds an approximation of its own.
A cluster replaces $\sum_{j} \bar w_j e^{\zeta_{qj}} \mathbf{v}_j$ over its members by
$(\sum_j \bar w_j)\, e^{\zeta_{qc}}\, \mathbf{v}_c$ at the cluster's mean position, normal, key, and
value, which is exact when the score $\zeta_{qj}$ of the read (line 37 of \cref{alg:isla}) is
constant over the cluster.
The score depends on learned states as well as on geometry, so $\theta$ limits how much the
geometry varies over a cluster and does not limit how much the learned states vary.
A model should be evaluated at the angle it was trained at.

\subsection{Consistency under Remeshing and Re-sampling}
\label{app:consistency}

\begin{figure}[t]
  \centering
  \includegraphics[width=0.80\textwidth]{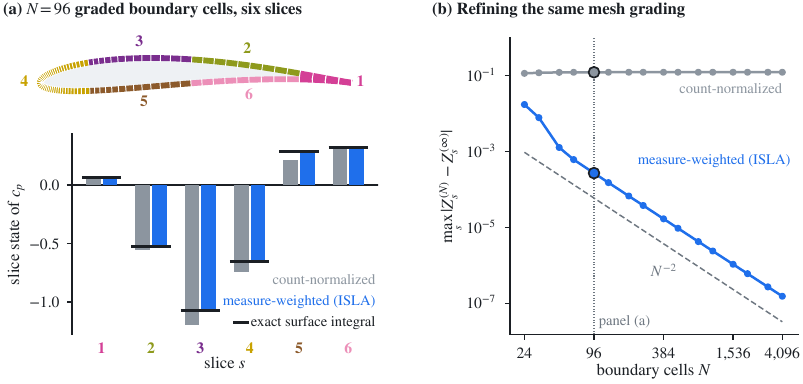}
  \caption{\textbf{Measure-weighted slice pooling converges under mesh refinement;
  count-normalized pooling does not.}
  \textbf{(a)} A Joukowski airfoil in potential flow, discretized into $N = 96$ boundary cells
  that are about nine times shorter at the leading edge, with six soft slices.
  Bars show the slice states of the exact $c_p$ pooled by soft counts (gray) and by cell
  measure (charcoal, \isla{}) beside the exact ratio of surface integrals (crimson ticks).
  \textbf{(b)} Largest error across the six slice states as the same grading law is refined from
  $N = 24$ to $4{,}096$: count-normalized pooling stays near $0.12$, its mesh-density-weighted
  limit, whereas \isla{}'s measure-weighted pooling falls by five decades as $N^{-2}$.}
  \label{fig:mechanisms}
\end{figure}

A new mesh of the same surface is not one of the transformations of \cref{thm:invariance}
(\cref{fig:mechanisms}), so
the statement for remeshing is one of convergence, along sequences of meshes of the following
kind.
Call a sequence of meshes of a compact, oriented, piecewise $C^2$ surface $\bdry$
\emph{admissible} if
\begin{enumerate}[leftmargin=*, itemsep=0pt, topsep=1pt, label=(\alph*)]
  \item every cell has its vertices on the closure of one smooth piece of one condition, on
  which the data are continuous;
  \item the closest-point projection onto that piece is one-to-one on the cells and onto the
  piece, up to sets whose measure tends to zero;
  \item the largest cell diameter $\ell$ tends to zero; and
  \item $\max_i \lVert \hn_i - \vn(\vx_i^\ast) \rVert \to 0$, where $\vn$ is the unit normal of
  $\bdry$ and $\vx_i^\ast$ is the point of the cell's piece closest to the centroid $\vx_i$.
\end{enumerate}
Shape-regular triangulations that satisfy (a) to (c) and are oriented as $\bdry$ satisfy (d)
\citep{dziuk2013finite}; in two dimensions (d) follows from (a) to (c) and the orientation.
Condition (d) is needed: inscribed triangulations of a cylinder can have diameters that tend to
zero and areas that diverge.

\begin{proposition}[Consistency under remeshing and re-sampling]
\label{prop:quadrature}
Let $\bdry$ be compact, oriented, and piecewise $C^2$, and let $\lambda$ and $f$ be bounded on
$\bdry$ and continuous on each smooth piece of each condition.
\begin{enumerate}[leftmargin=*, itemsep=0pt, topsep=1pt, label=(\roman*)]
  \item Along an admissible sequence of meshes,
  $\sum_i \wsoftmax^{\bar w}_i(\lambda(\vx_i^\ast))\, f(\vx_i^\ast) \to
  \int_\bdry e^{\lambda} f \dd\bar S \big/ \int_\bdry e^{\lambda} \dd\bar S$,
  with $\dd\bar S = e^{\beta_b} \dd S / \lvert\bdry_b\rvert$ on $\bdry_b$ (and
  $\dd\bar S = \dd S$ with one declared condition).
  Without the measure, if the empirical distribution of the centroids converges to a density,
  the same sum converges to a ratio of integrals against that density, which depends on the
  mesher.
  \item Let a design keep cell $i$ with probability $\pi_i > 0$ and weight $a_i / \pi_i$.
  For scores and values that do not depend on the sample, the numerator and the denominator of
  the sampled sum are unbiased for their full-mesh values.
  This uses first-order inclusion probabilities only, so it holds for the systematic design of
  \cref{app:draw} with a random start.
  Here $A_b$ is the measure of the full mesh, which the model computes before the draw.
  \item Fix parameters for which the model is defined, with a positive gain $\gamma$ in every
  prototype map, and take the model without the floor on the weights.
  Let $\bdry$ be closed, let the data of cell $i$ be the values at $\vx_i^\ast$ of data that are
  continuous on each smooth piece of each condition, let every boundary cell be a token, and let
  the boundary read be exact ($\theta = 0$).
  Then there is a field $\mathbf{Y}$ on $\bdry$ with
  $\max_i \lVert \mathbf{Y}_i - \mathbf{Y}(\vx_i^\ast) \rVert \to 0$ along every admissible
  sequence, and the output at every fixed query point off $\bdry$ that has a unique closest
  point on $\bdry$ converges.
  The limits depend on $\bdry$, its data, the global data, and the parameters, and not on the
  sequence.
\end{enumerate}
\end{proposition}

\begin{proof}
\emph{(i)}
Let $F$ be bounded on $\bdry$ and continuous on each smooth piece of each condition.
For an admissible sequence, $\sum_{i \in b} a_i \to \lvert\bdry_b\rvert$ and
$\lVert \vx_i - \vx_i^\ast \rVert = O(\ell^2)$ \citep{dziuk2013finite}, so
$\sum_i a_i F(\vx_i^\ast)$ is a Riemann sum and converges to $\int_\bdry F \dd S$.
Apply this on each condition to $F = e^{\lambda} f$ and to $F = e^{\lambda}$, whose integral is
positive; the ratio converges because division is continuous away from zero.
Without the measure, $\frac1N \sum_i F(\vx_i^\ast)$ converges to $\int_\bdry F\, \varrho \dd S$
whenever the empirical distribution of the points converges to a density $\varrho$, and
$\varrho$ is set by the mesher.

\emph{(ii)}
Let $I_i$ indicate that cell $i$ is kept.
Then $\E\bigl[\sum_i I_i (a_i / \pi_i) F_i\bigr] = \sum_i a_i F_i$ for every fixed $F$
\citep{horvitz1952generalization}; this holds for the numerator ($F = e^{\lambda} f$) and the
denominator ($F = e^{\lambda}$).
Their ratio is the H\'ajek estimator \citep{hajek1971comment,sarndal1992model}, which is not
unbiased.
The hypothesis that the scores and values do not depend on the sample holds in the prototype
attention of the first block; elsewhere they are functions of sampled sums.
For the systematic design and every start $u_0 \in [0, 1)$, including the fixed start of
evaluation, the error of each sum is bounded without any assumption of independence.
With $s(t) = a_i F_i / \pi_i$ for $t$ in the interval of cell $i$ on the line on which the cells
are laid, $\sum_i a_i F_i = \int_0^B s$ and the estimate is $\sum_{m=0}^{B-1} s(u_0 + m)$, so
\begin{equation}
  \Bigl\lvert \sum_{i\ \mathrm{kept}} \frac{a_i}{\pi_i} F_i - \sum_i a_i F_i \Bigr\rvert
  \;\le\; \sum_{m=0}^{B-1} \operatorname{osc}_m(s) ,
\end{equation}
where $\operatorname{osc}_m(s)$ is the oscillation of $s$ over the $m$-th unit interval.
Within a condition $a_i / \pi_i$ equals the threshold $a^\ast_b$ of \cref{app:draw} for every cell
with $\pi_i < 1$,
so on those cells $s = a^\ast_b F_i$, and its oscillation over a unit interval is $a^\ast_b$
times that of $F$ over the cells that the interval contains.
The error of each sum, relative to $\sum_i a_i \lvert F_i \rvert$, therefore vanishes when
$e^{\lambda} f$ and $e^{\lambda}$ vary little over those cells in all but a vanishing share of
the intervals, as when the mesh and the budget are refined together.
For $B = N$ every interval is one cell and the error is zero, but the bound does not tend to
zero as $B \to N$ on a fixed mesh.
We claim no rate.

\emph{(iii)}
Every map in a block is continuous: each normalization and each logarithm has a positive floor
on its argument, and the remaining operations are affine maps, GELU, exponentials, and ratios
with positive denominators.
Suppose that at the input of a block there is a bounded field $H: \bdry \to \R^C$, continuous on
each smooth piece of each condition, with $\max_i \lVert \vh_i - H(\vx_i^\ast) \rVert \to 0$ as
$\ell \to 0$.
For the first block $H$ is the embedding of the invariant input, a continuous function of the normal field
and the data, and condition (d) gives the convergence.
By (i) and uniform continuity on bounded sets, every sum over tokens in the block converges to
the corresponding ratio of surface integrals of $H$, the anchors converge, and the updated
states converge to the values of a new field of the same kind at the points $\vx_i^\ast$.
By induction the outputs at the tokens converge to the values of a field $\mathbf{Y}$ determined
by $\bdry$, its data, the global data, and the parameters.
The length unit of the mesh converges to that of $\bdry$ by (i), and the signed distance and
the direction of a fixed query point off the closed surface $\bdry$ converge wherever its
closest point is unique, so the output at the query converges too.
\end{proof}

With a fixed token budget, or with $\theta > 0$, the outputs converge only up to the sampling
error of the draw and the error of the tree, which \cref{app:certificates} measures.
With the floor of $10^{-12}$ of the total weight, the model equals that of clause (iii) on every
mesh whose smallest weight exceeds the floor (fewer than about $10^{12}$ cells of comparable
size), and not in the limit.

\paragraph{Order of convergence under a geometric split.}
If a flat cell $E$ with centroid $\vx_E$ is split into cells $E_1, E_2$ that tile it, then
$a_1 \vx_1 + a_2 \vx_2 = a_E \vx_E$, so for a twice-differentiable summand the first-order terms
cancel and $\sum_m a_m F(\vx_m) - a_E F(\vx_E) = O(a_E \ell^2)$: the token sums are second-order
consistent.
The length unit is exactly unchanged, because \cref{eq:length} is evaluated with each cell's own
second moment (\cref{app:architecture}).
This is a statement about splitting the flat cells of one polyhedral surface; for two meshes of
a curved surface, \cref{prop:quadrature} gives convergence and no rate.

\section{Architecture in Full}
\label{app:architecture}

\begin{figure}[t]
  \centering
  \includegraphics[width=\textwidth]{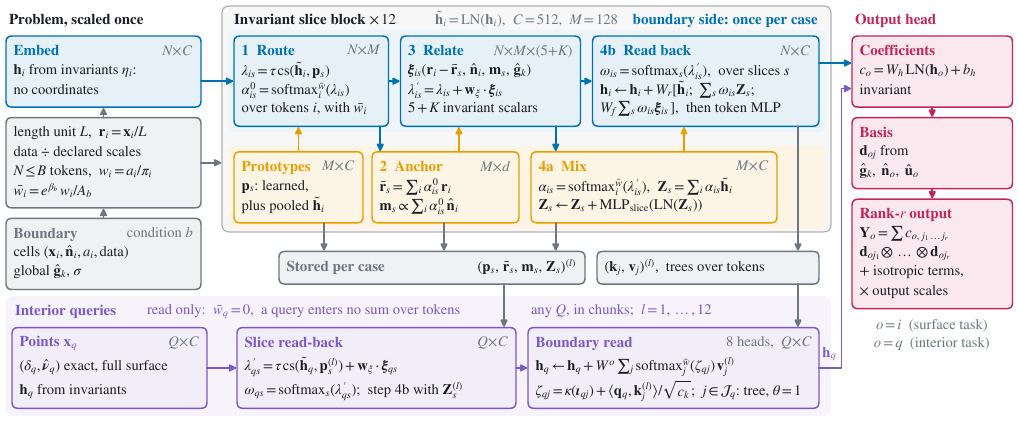}
  \caption{\textbf{The forward pass of \isla{}, with the shape of every tensor.}
  \emph{Left:} the boundary is scaled once, tokens are drawn with Horvitz--Thompson weights, and
  initial states are embedded from invariants without coordinates.
  \emph{Center:} each of 12 blocks passes states from the $N$ boundary tokens (blue) to the $M$
  slices (amber) and back in four steps: route, anchor, relate, and mix and read back.
  \emph{Gray:} the per-block states and token trees stored once per case.
  \emph{Bottom:} interior query points ($\bar w_q = 0$), an independent input, read the stored
  slices and, through the tree, the boundary tokens without contributing to any sum over tokens,
  so arbitrary query sets run in chunks.
  \emph{Right:} the equivariant tensor-basis head.}
  \label{fig:arch-overview}
\end{figure}

This appendix specifies the forward pass in full.
\Cref{fig:arch-overview} maps the dataflow and tensor shapes across the boundary, slice, and
interior-query paths, \cref{alg:isla} gives the computation step by step for boundary data of
rank zero, the subsections
define the pieces that the main text only names, and \cref{app:rank} adds boundary vectors and
tensors.
Throughout, $\LN$ is a layer normalization over the $C$ channels of one token,
$\cs(\mathbf{a}, \mathbf{b}) = \mathbf{a} \cdot \mathbf{b} / (\lVert \mathbf{a} \rVert\, \lVert \mathbf{b} \rVert)$
is the cosine similarity of two vectors, $\wsoftmax^{\bar w}_i$ is the measure-weighted softmax
over tokens of \cref{eq:wsoftmax}, and $\wsoftmax_s$ is an unweighted softmax over slices.
Logarithms and normalizations have small positive floors, relative to the case's own scales,
which the equations omit.

\begin{algorithm}[h]
\caption{The forward pass of \isla{} for one case. The index $i$ runs over boundary tokens, $b$ over conditions, $s$ over slices, $l$ over blocks, $q$ over query points, $j$ over the sources of a boundary read, $h$ over its heads, and $o$ over output tokens. Lines 25 to 39 and the steps marked ``(interior)'' run only for interior tasks.}
\label{alg:isla}
\small
\begin{algorithmic}[1]
\REQUIRE boundary cells $(\vx_i, \hn_i, a_i, \text{data})$ grouped by condition $b$; global directions $\hg_k$ and scalars $\sigma$; query points $\vx_q$ (interior task)
\STATE $L \gets$ root-mean-square pairwise distance of the surface, \cref{eq:length-discrete}
\STATE declared scales from the data; divide every datum by its scale
\STATE $A_b \gets \sum_{i \in b} a_i$
\STATE (interior) \textbf{if} the cells outnumber the budget $B$: draw $B$ tokens, $w_i \gets a_i / \pi_i$ (\cref{app:draw}); \textbf{else} $w_i \gets a_i$
\STATE $\bar w_i \gets e^{\beta_{b(i)}} w_i / A_{b(i)}$
\STATE $\vr_i \gets \vx_i / L$
\STATE $\vh_i \gets W_e\, \mathrm{GELU}(W_{b(i)}\, \eta_i) + b_e$ with invariant input $\eta_i = [1,\ \hn_i\!\cdot\hg_k,\ \text{invariants of the data},\ \sigma]$
\FOR{$l = 1, \dots, 12$}
  \STATE $\htilde_i \gets \LN(\vh_i)$
  \STATE prototypes $\vp_s$ from $\htilde$ (\cref{app:prototypes})
  \STATE $\lambda_{is} \gets \tau \cs(\htilde_i, \vp_s)$
  \STATE $\alpha^0_{is} \gets \wsoftmax^{\bar w}_i(\lambda_{is})$ \hfill (over tokens $i$)
  \STATE $\vrbar_s \gets \sum_i \alpha^0_{is} \vr_i$
  \STATE $\vm_s \gets \operatorname{normalize}\bigl(\sum_i \alpha^0_{is} \hn_i\bigr)$
  \STATE $\vxi_{is} \gets$ relational invariants of $(\vr_i - \vrbar_s,\ \hn_i,\ \vm_s,\ \hg_k)$, \cref{eq:invariants}
  \STATE $\lambda'_{is} \gets \lambda_{is} + \mathbf{w}_\xi\!\cdot\vxi_{is}$
  \STATE $\alpha_{is} \gets \wsoftmax^{\bar w}_i(\lambda'_{is})$ \hfill (over tokens $i$)
  \STATE $\omega_{is} \gets \wsoftmax_s(\lambda'_{is})$ \hfill (over slices $s$)
  \STATE $\vZ_s \gets \sum_i \alpha_{is} \htilde_i$
  \STATE $\vZ_s \gets \vZ_s + \MLP_{\mathrm{slice}}(\LN(\vZ_s))$
  \STATE $\vh_i \gets \vh_i + W_r[\htilde_i;\ \sum_s \omega_{is} \vZ_s;\ W_f \sum_s \omega_{is} \vxi_{is} + b_f] + b_r$
  \STATE $\vh_i \gets \vh_i + \MLP(\LN(\vh_i))$
  \STATE (interior) store $(\vp, \vrbar, \vm, \vZ)^{(l)}$ and the tokens' keys and values $\mathbf{k}_j^{(l)}, \mathbf{v}_j^{(l)}$
\ENDFOR
\STATE $(\delta_q, \hnu_q) \gets$ exact signed distance and direction to the full surface (\cref{app:geometry})
\STATE $\vr_q \gets \vx_q / L$
\STATE $\vh_q \gets W_e\, \mathrm{GELU}(W_{\mathrm{in}}[\hnu_q\!\cdot\hg_k,\ \sigma,\ \delta_q/L,\ \asinh(10^6 \delta_q / L)] + b_{\mathrm{in}}) + b_e$
\STATE sources $\mathcal{J}_q \gets$ tokens and accepted clusters of the tree (\cref{app:tree})
\STATE $\vphi_{qj} \gets \MLP_{\mathrm{basis}}(\viota_{qj})$ for $j \in \mathcal{J}_q$
\FOR{$l = 1, \dots, 12$}
  \STATE $\htilde_q \gets \LN(\vh_q)$
  \STATE $\vxi_{qs} \gets$ relational invariants of $(\vr_q - \vrbar^{(l)}_s,\ \hnu_q,\ \vm^{(l)}_s,\ \hg_k)$
  \STATE $\lambda'_{qs} \gets \tau \cs(\htilde_q, \vp^{(l)}_s) + \mathbf{w}_\xi\!\cdot\vxi_{qs}$
  \STATE $\omega_{qs} \gets \wsoftmax_s(\lambda'_{qs})$ \hfill (over slices $s$)
  \STATE $\vh_q \gets \vh_q + W_r[\htilde_q;\ \sum_s \omega_{qs} \vZ^{(l)}_s;\ W_f \sum_s \omega_{qs} \vxi_{qs} + b_f] + b_r$
  \STATE $\vh_q \gets \vh_q + \MLP(\LN(\vh_q))$
  \STATE $\zeta_{qjh} \gets \mathbf{w}^{\kappa}_h\!\cdot\vphi_{qj} + \langle \mathbf{q}_{qh}, \mathbf{k}^{(l)}_{jh} \rangle / \sqrt{c_k}$ for each head $h$, with $\mathbf{q}_{qh} = W^{\mathrm q}_h \LN(\vh_q)$
  \STATE $\vh_q \gets \vh_q + W^{o}\bigl[\sum_{j \in \mathcal{J}_q} \wsoftmax^{\bar w}_j(\zeta_{qjh})\, \mathbf{v}^{(l)}_{jh}\bigr]_{h = 1, \dots, 8}$
\ENDFOR
\STATE output tokens $o$: the cells of the requested boundaries (surface task) or the queries (interior task)
\STATE $c_o \gets W_h \LN(\vh_o) + b_h$
\STATE $\hu_o \gets$ direction from the token's soft anchor (\cref{app:heads})
\STATE $\mathbf{Y}_o \gets \sum_{j_1 \dots j_r} c_{o, j_1 \dots j_r}\, \vd_{o j_1} \otimes \dots \otimes \vd_{o j_r} + \text{isotropic terms}$, per declared field
\STATE multiply each output by the scales of its units
\end{algorithmic}
\end{algorithm}

\subsection{The Length Unit and the Scales}
\label{app:length}

For a mesh with cells (segments or triangles) of measure $a_i$, centroid $\vx_i$, and vertices
$\vx_{iv}$, $v = 0, \dots, n$ with $n = d - 1$, let $A = \sum_j a_j$ and
$\bar\vx = \sum_i a_i \vx_i / A$.
The length unit of \cref{eq:length}, the root-mean-square distance between two points of the
surface, is evaluated exactly for the piecewise-flat surface by
\begin{equation}
  L^2 = \frac{2}{A} \sum_i a_i \bigl( \lVert \vx_i - \bar\vx \rVert^2 + J_i \bigr),
  \qquad
  J_i = \frac{1}{(n+1)(n+2)} \sum_{v=0}^{n} \lVert \vx_{iv} - \vx_i \rVert^2 ,
  \label{eq:length-discrete}
\end{equation}
where $J_i$ is the second moment of cell $i$ about its own centroid per unit measure.
The identity $\frac{1}{A^2}\iint \lVert \vx - \vy \rVert^2 = 2 \operatorname{Var}(\vx)$ gives the
$O(N)$ form, and the term $J_i$ makes $L$ exactly unchanged when a cell is split into cells that
tile it.
$L$ is $\sqrt 2$ times the radius of gyration of the surface.
On the surface tasks of this paper the cells given to the model are the data pipeline's sample
with Horvitz--Thompson measures, so there $L$ and the $A_b$ are estimates from that sample.
The mean position $\bar\vx$ appears inside this one scalar and in no input of a learned map.

The other scales come from the declarations of a problem's fields.
The declaration of a field, an input or an output, may state its units as a monomial in named
scales, and may state that the field defines one of them.
A scale defined by a boundary vector field, such as an inlet's prescribed velocity, is its
measure-weighted root-mean-square magnitude over the cells that carry it, and the
measure-weighted mean direction of that field becomes one of the global directions $\hg_k$.
A scale may also be given directly as a global value, such as a density.
Scales are computed in float64 from the full-resolution boundary, before any token is drawn.
Each input with units is divided by the monomial of its scales and each output is multiplied by
its own.
A Reynolds number, where a problem has one, is an ordinary global scalar formed from the
problem's own scales and viscosity; the model itself names no physical quantity.
In this paper \aerozoo{} declares a velocity and a density scale, and AirfRANS a velocity scale.
The automotive datasets hold one flow condition each and declare none: their data pipeline makes the
targets dimensionless with the dataset's free-stream constants, and only $L$ depends on the size
of the geometry.
A global vector contributes only its direction, because the outputs do not scale as any fixed
power of its magnitude; a magnitude that matters enters as a scale or as a scalar.

\subsection{Drawing Boundary Tokens}
\label{app:draw}

When an interior task's boundary has more cells than the budget $B$, tokens are drawn as follows.
\begin{enumerate}[leftmargin=*, itemsep=1pt, topsep=2pt]
  \item \emph{Shares.} The budget is divided among the conditions by water filling: equal shares,
  except that a condition with fewer cells than its share is kept whole and its unused share is
  passed to the others.
  \item \emph{Inclusion probabilities.} Within a condition with share $B_b$, cell $i$ is kept
  with probability $\pi_i = \min(1, a_i / a^\ast_b)$, where the threshold $a^\ast_b$ makes
  $\sum_i \pi_i = B_b$.
  \item \emph{Order.} Within each condition, the cells are sorted along a Morton (Z-order) curve
  \citep{morton1966computer} of their centroids, and the conditions are laid one after another.
  The curve lies in the frame of the boundary's measure-weighted principal axes, each axis
  signed so that the third moment along it is positive, and in units of the boundary's own
  extent.
  The frame moves with the boundary, so in exact arithmetic the same cells are kept in every pose and at every scale,
  except where two principal variances are equal, a third moment vanishes (as for a
  mirror-symmetric boundary), or a centroid lies on a grid line of the curve.
  In float32, rounding moves cells across such lines: a rigid motion changes
  \isn{cert-dmlv-tokens-changed-rot-pct}\% of the drawn tokens on DrivAerML
  (\cref{app:certificates}).
  \item \emph{Systematic draw.} The cells are laid end to end as intervals of length $\pi_i$, and
  the cells that contain the points $u_0, u_0 + 1, u_0 + 2, \dots$ are kept
  \citep{madow1949theory}, with the start $u_0$ uniform on $[0, 1)$ in training and
  $u_0 = 1/2$ in evaluation, so that inference is deterministic.
  \item \emph{Weights.} A kept cell has the weight $w_i = a_i / \pi_i$.
\end{enumerate}
The number of tokens is exactly $\min(N, B)$.
In training every cell is kept with probability exactly $\pi_i$, so every sum
$\sum_i w_i F_i$ with $F$ independent of the draw is an unbiased estimate of the sum over all
cells, and it is exact for a condition that is kept whole.
Ordering along a space-filling curve spreads the sample over the surface
\citep{stevens2004spatially}.
Since large cells are kept with certainty and small cells in proportion to their area, the
tokens of a finely meshed region do not crowd out the rest of the boundary.

A surface task uses every cell that it is given, with the measures that it is given.
On the surface tasks of this paper the data pipeline draws 200,000 cells of the body uniformly,
multiplies their measures by the inverse of the sampling fraction, and passes the tunnel's
boundaries whole.

\subsection{Prototypes Adapted to the Case}
\label{app:prototypes}

Each block has learned base prototypes $\mathbf{w}_s \in \R^C$, $s = 1, \dots, M$, and adapts them
to the case by one measure-weighted attention,
\begin{align}
  \Delta_s &= \sum_i \wsoftmax^{\bar w}_i\Bigl(\tfrac{1}{\sqrt C}\, \mathbf{w}_s \cdot W^{\mathrm{pk}} \htilde_i\Bigr)\, W^{\mathrm{pv}} \htilde_i , \\
  \Delta^{\mathrm c}_s &= \Delta_s - \frac1M \sum_{t=1}^M \Delta_t , \\
  \bar\gamma &= \gamma \operatorname{rms}(\mathbf{w}) , \qquad
  \vp_s = \mathbf{w}_s + \frac{\bar\gamma}{\operatorname{rms}(\Delta^{\mathrm c}) + \bar\gamma}\, \Delta^{\mathrm c}_s ,
\end{align}
where $\Delta_s$ is the summary of the case that the attention forms for slice $s$, rms is the
root mean square over all $M C$ components of its argument, and $\gamma$ is a learned gain,
initialized to $0.1$.
Centering removes the part of the summaries that is common to all slices, and for $\gamma > 0$
the last factor keeps the root-mean-square size of the adaptation below a fraction $\gamma$ of
that of the base prototypes.
The cosine-similarity score is bounded, $\lambda_{is} \in [-\tau, \tau]$, so the first routing
cannot sharpen without limit as the weights grow.
The refinement $\mathbf{w}_\xi \cdot \vxi_{is}$ adds a learned linear function of the invariants,
without a bias.
It contains the distance and its logarithm, so it is not bounded.

\subsection{Boundary Data of Rank One and Two}
\label{app:rank}

A condition may prescribe vectors $\mathbf{a}_p$ on its cells, such as an inlet's velocity, and
a rank-2 tensor $T$.
With the frame vectors $\mathbf{f}_0 = \hn_i$ and $\mathbf{f}_a = \hg_a$, a token's invariant input contains
the invariants of these data: $\mathbf{a}_p \cdot \mathbf{f}_a$, $\lVert \mathbf{a}_p \rVert$, $\mathbf{a}_p \cdot \mathbf{a}_{p'}$,
$\mathbf{f}_a^\top T\, \mathbf{f}_b$, the trace of $T$, and its Frobenius norm.
Each condition also maps its vectors (and the vectors $T \mathbf{f}_a$) onto a set of vector
channels shared by all conditions, and its tensors onto shared tensor channels, by learned
scalar matrices initialized to the identity.
A combination of vectors with scalar weights commutes with every rotation and reflection, so
the channels $V_{ip}$ and $T_i$ remain true vectors and tensors.
They enter each block through the contractions $\hrho_{is} \cdot V_{ip}$ and
$\hrho_{is}^\top T_i\, \hrho_{is}$, which join the relational invariants of
\cref{eq:invariants}, and the boundary read through $\hrho \cdot V_{jp}$ and
$\hnu_q \cdot V_{jp}$, which join the pair invariants of \cref{app:tree}.
Query points have zeros in these channels.
On the DrivAerML tasks, which have no boundary data and one global direction, the relational
invariants number six.
The model accepts global scalars and unit directions only, so a problem whose global data hold
a rank-2 material tensor or a pseudovector, such as a rotation rate, cannot be posed to it in
full.
The interior form accepts boundary data of rank zero and one only.

\subsection{Output Basis}
\label{app:heads}

The basis of \cref{eq:head} contains the direction $\hu_o$ to each output token $o$ from a soft
anchor of its own.
To form it, one more routing map, with its own prototypes $\vp^{\mathrm f}_s$ and without the
geometric refinement, is applied to the final states; its prototypes are adapted to the case as
in \cref{app:prototypes}, with the final states $\vh_i$ in place of $\htilde_i$:
\begin{align}
  \alpha^{\mathrm f}_{is} &= \wsoftmax^{\bar w}_i\bigl(\tau \cs(\vh_i, \vp^{\mathrm f}_s)\bigr),
  &
  \vrbar^{\mathrm f}_s &= \sum_i \alpha^{\mathrm f}_{is} \vr_i, \\
  \omega^{\mathrm f}_{os} &= \wsoftmax_s\bigl(\tau \cs(\vh_o, \vp^{\mathrm f}_s)\bigr),
  &
  \hu_o &= \frac{\vr_o - \sum_s \omega^{\mathrm f}_{os} \vrbar^{\mathrm f}_s}{\bigl\lVert \vr_o - \sum_s \omega^{\mathrm f}_{os} \vrbar^{\mathrm f}_s \bigr\rVert} .
\end{align}
The soft anchor of token $o$ is $\sum_s \omega^{\mathrm f}_{os} \vrbar^{\mathrm f}_s$, and $\hu_o$
is the direction from it to the token.
The anchors $\vrbar^{\mathrm f}_s$ are formed by the boundary tokens alone.
With $K$ global directions the true basis vectors are
\begin{equation}
  \vd_o = \bigl( \hg_1, \dots, \hg_K,\ \hn_o,\ \hu_o,\ e(\hu_o, \hn_o),\ e(\hu_o, \hg_1), \dots, e(\hu_o, \hg_K) \bigr),
  \qquad e(\hu, \mathbf{a}) = -\bigl(\mathbf{a} - (\mathbf{a}\cdot\hu)\, \hu\bigr) ,
\end{equation}
$m = 2K + 3$ in all, where $e(\hu, \mathbf{a})$ is minus the part of $\mathbf{a}$ that is
perpendicular to $\hu$.
For a query, $\hnu_q$ takes the place of $\hn_o$.
An even-parity output of rank $r$ is the contraction of $m^r$ coefficients with the products
$\vd_{oj_1} \otimes \dots \otimes \vd_{oj_r}$, plus the isotropic tensors of that rank (the
identity at rank 2).
A symmetric output uses one coefficient per unordered multi-index.
An odd-parity output replaces exactly one factor by a pseudovector: in three dimensions a cross
product of two of $\hg_k, \hn_o, \hu_o$, and in two dimensions the perpendicular of one of them.
An odd output of rank $d$ also has the Levi-Civita tensor.
With one global direction in three dimensions a vector uses 5 coefficients, a symmetric rank-2
tensor 16, and a pseudovector 3.

The span of the true basis is that of $\{\hg_k, \hn_o, \hu_o\}$, of dimension at most
$\min(d, K + 2)$.
With at least one global direction it is all of $\R^3$ except where the three directions are
coplanar.
With none ($K = 0$) in three dimensions, every even vector output lies in the plane of $\hn_o$
and $\hu_o$.
For general vector outputs such a problem needs a declared direction, or a boundary vector
field that defines a scale, whose mean direction then becomes a global direction.

\subsection{Exact Geometry of a Query Point}
\label{app:geometry}

A query point receives two geometric quantities: its signed distance $\delta_q$ to the surface,
and the direction $\hnu_q$ in which that distance increases.
The surface is the union of all cells of all given boundaries at full resolution, never the
drawn tokens.
For a query $\vx_q$ an exact point-to-simplex search (on a bounding-volume hierarchy in three
dimensions) finds the closest point $\vx_q^\ast$ of the surface, and
\begin{equation}
  \delta_q = \pm \lVert \vx_q - \vx_q^\ast \rVert ,
  \qquad
  \hnu_q = \sign(\delta_q)\, \frac{\vx_q - \vx_q^\ast}{\lVert \vx_q - \vx_q^\ast \rVert} ,
\end{equation}
with the sign positive on the side that the cells' normals point to.
The sign is taken from the generalized winding number of the oriented cells
\citep{jacobson2013robust}, evaluated with a hierarchical far-field expansion in three
dimensions \citep{barill2018fast}.
It is therefore defined at folds of the surface, where cells meet back to back and a sign taken
from the nearest cell's normal would depend on rounding.
For a closed surface the sign is exact; behind an isolated open sheet it is approximate near
the level at which the winding number changes sign.
Where the closest point lies inside a cell, $\hnu_q$ is that cell's normal, including at a
point on the surface itself.
At a point with several closest features, such as the mid-plane of a thin plate, the gradient
of the distance is not defined.
There \isla{} takes the least-norm point of the convex hull of the
tied directions (the midpoint of two, zero where they cancel), which is a function of the tied
features alone and not of the order of the mesh.
Both $\delta_q$ and $\hnu_q$ are functions of the surface: they are the same for every draw of
the tokens and are unchanged when a cell is split.
This is why \isla{} takes no stored signed-distance field.
The channel $\asinh(10^6\, \delta_q / L)$ is logarithmic for distances above $10^{-6} L$, so that
the viscous sublayer and the far wake are both resolved, and linear through the wall.

\subsection{The Boundary Read and Its Tree}
\label{app:tree}

The boundary read of \cref{eq:read} scores a query $q$ and a source $j$ by a learned kernel of
the pair's invariants plus an inner product of the query's vector and the source's key.
For a source with position $\vr_j$ and normal $\hn_j$, let $\vrho = \vr_q - \vr_j$.
The pair invariants are
\begin{equation}
  \viota_{qj} = \bigl[\, \lVert\vrho\rVert,\ \log\lVert\vrho\rVert,\ \log(\lVert\vrho\rVert L / \lvert\delta_q\rvert),\
  \hrho\cdot\hn_j,\ \hrho\cdot\hnu_q,\ \hnu_q\cdot\hn_j,\ \hrho\cdot\hg_k,\ \hn_j\cdot\hg_k \,\bigr],
\end{equation}
together with the contractions of any boundary vector data with $\hrho$ and $\hnu_q$.
A small MLP maps them once per forward pass to 16 basis functions $\vphi_{qj}$, which all layers
share.
Each layer and each of its 8 heads has its own kernel, the linear combination
$\mathbf{w}^{\kappa}_h \cdot \vphi_{qj}$, and its own keys and values of the boundary tokens'
states (4 key and 8 value channels per head).
The output projection $W^{o}$ of each layer is initialized to zero.

The tokens of each condition are first grouped by the face of a cube that their normal points to
in the boundary's principal frame.
Each group forms a binary radix tree over a Morton order of its tokens
\citep{karras2012maximizing}, with leaves of at most four
tokens.
A node $c$ stores the sum $\bar w_c$ of its tokens' weights and the measure-weighted means of
their positions, normals (renormalized to unit length), keys, and values.
A query traverses each tree from the root and accepts a node as one virtual token when both
\begin{align}
  \operatorname{diam}(c) &< \theta\, \operatorname{dist}(\vx_q, c) , \\
  \bigl\lVert \textstyle\sum_{j \in c} \bar w_j \hn_j \bigr\rVert / \bar w_c &> 1 - \theta^2 / 2 ,
\end{align}
that is, under the $\mathcal{H}$-matrix admissibility criterion
\citep{hackbusch1999sparse,sauter2011boundary} together with normal coherence;
otherwise it opens the node, down to individual tokens.
Here $\operatorname{diam}(c)$ is the diagonal of the box of the node's centroids, widened by a
bound on rounding so that no tie of the criterion is decided by rounding, and
$\operatorname{dist}(\vx_q, c)$ the distance from the query to that box.
Within one tree all normals point to the same face of the cube, so in three dimensions the
second condition always holds for $\theta > 0.92$, and at $\theta = 1$ only the first is active.
Each query's list covers every token exactly once, as itself or inside one accepted node.
Each query traverses the trees independently: queries never form clusters with one another.
At $\theta = 1$ a query of a DrivAerML case has about \isn{fact-tree-sources} sources for $2\times10^5$ tokens.
Only zeroth moments are stored, so the approximation is that of a monopole expansion whose
``charges'' are learned states.
In training no gradient reaches the pair invariants (the geometry is data), and a model is
trained and evaluated at one $\theta$.
With randomly initialized weights in float64, the outputs at $\theta = 1$ deviate from those of
the exact boundary read by \isn{fact-tree-dev-min-pct} to \isn{fact-tree-dev-max-pct}\% on two synthetic
geometries, a sphere and a random point cloud.
Trained models depend on their angle more strongly: evaluated with the exact boundary read, AirfRANS
models trained at $\theta = 1$ have \isn{abl-theta0-ratio-min} to \isn{abl-theta0-ratio-max}
times their error (\cref{app:ablations}).

\subsection{Sizes and Training}
\label{app:sizes}

\Cref{tab:hyper} lists the reference configuration, and \cref{tab:recipe} the training recipe of
the pose and re-sampling tests; the optimizers are Muon \citep{jordan2024muon} and AdamW
\citep{loshchilov2019decoupled}.
Each block has \num{5449991} parameters, of which the token MLP and the slice MLP account for
77\%.
The interior form adds the query input layer, the kernel basis, and 12 boundary reads
(\isn{params-interior-added-exact} parameters in all), and its head has one more scalar output
channel, the eddy viscosity (512 weights and a bias), so the two totals of \cref{tab:hyper}
differ by \isn{params-forms-diff-exact}.
Under bfloat16 autocast, the query geometry, the tree, and the anchors' mean normals are computed
in float32, and so is the head when scales or output statistics are declared; the anchor
positions follow autocast.

\begin{table}[h]
  \caption{The reference configuration of \isla{}, used on every dataset in this paper (its
  training recipe: \cref{tab:recipe}).}
  \label{tab:hyper}
  \centering
  \small
  \begin{tabular}{@{}ll@{}}
    \toprule
    Setting & Value \\
    \midrule
    Width $C$, blocks, slices $M$ & 512, 12, 128 \\
    MLP expansion; activation & 4; GELU \\
    Routing temperature $\tau$ & 10 \\
    Boundary token budget $B$ (interior) & 200,000 \\
    Boundary read & 8 heads, 16 kernel basis functions, opening angle $\theta = 1$, leaves of 4 \\
    Parameters & \isn{params-surface-exact} (surface form); \isn{params-volume-exact} (interior; \isn{az-sep-params-exact} on \aerozoo{}) \\
    \bottomrule
  \end{tabular}
\end{table}

\paragraph{Cost.}
Per block, the boundary side costs about $12.5\, N C^2 + 5\, N M C$ multiply-adds for $N$ tokens:
the dense maps dominate, the slice terms are a tenth of the total, and there is no term in
$N^2$.
A query costs $(2.5 + 8) C^2 + 2 M C$ per layer for its slice read and MLPs, plus 224
multiply-adds per source for the boundary read.
The boundary side runs once per case, and the query side is linear in the number of queries
with no interaction among them.

\FloatBarrier
\subsection{The Slices of a Trained Model}
\label{app:routing}

\begin{figure*}[!ht]
  \centering
  \includegraphics[width=\textwidth]{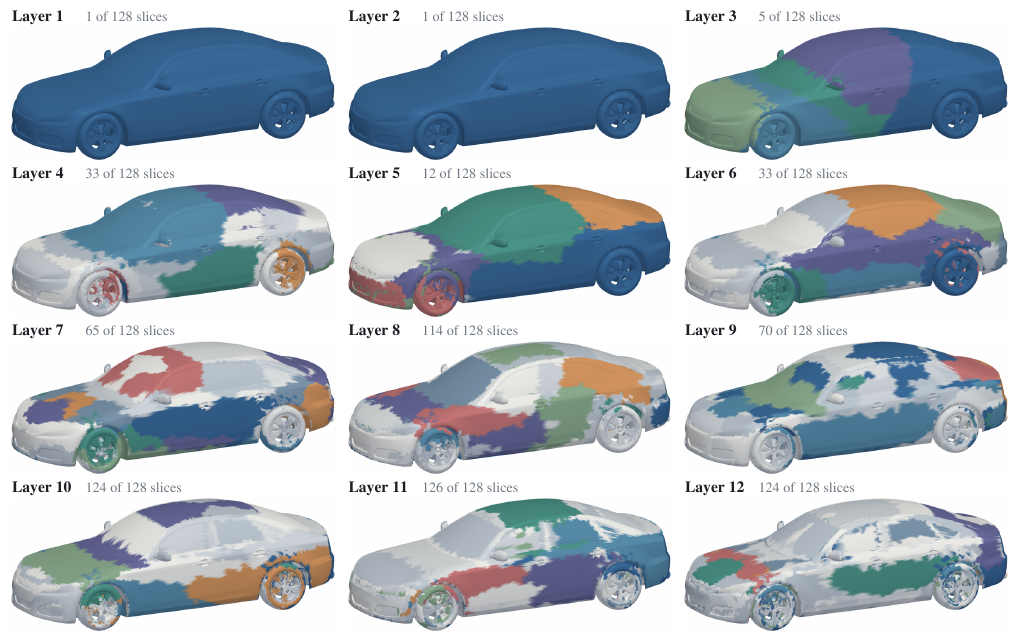}
  \caption{\textbf{What the slices of a trained model hold.}
  All \isn{app-routing-n-layers} layers of a trained \isla{} on one DrivAerML case (case
  \isn{app-routing-run} of the dataset); \cref{fig:slices} shows layers 3, 6, and 12 larger.
  Each surface point has the color of the slice that receives the largest share of its weight.
  In each panel the \isn{app-routing-n-coloured} slices that hold the most area have colors of
  their own, and all other slices are shades of gray; colors are unrelated between panels.
  The number in each panel counts the slices that hold some point.
  From layer 3 to layer 12 the slices go from broad bands to compact patches to sets scattered
  over the car (\cref{app:routing}).
  Anchors are not drawn: an anchor is a mean over its slice's tokens and need not lie in or near
  the slice's territory.}
  \label{fig:routing}
\end{figure*}

The slices of \cref{fig:routing} change character with depth.
The first two layers pool the whole car: in layer 1,
\isn{app-routing-layer-1-n-sharing} slices share every point almost equally (weights between
\isn{app-routing-layer-1-weight-min} and \isn{app-routing-layer-1-weight-max}), and in layer 2
one slice holds \isn{app-routing-layer-2-top-share-pct}\% of the weight.
Layer 3 has \isn{app-routing-n-slices-layer-3} broad bands, and the middle layers have tens of
compact patches, which begin to recur in separate places by layer 9.
In the last layer the slices are many and each is scattered over the car: there a slice groups
tokens by their state and no longer by their place.
The mean radius of a slice's territory (the root-mean-square distance of its points from
their centroid on one side of the car, averaged over slices by area) is
\isn{app-routing-radius-layer-6}\,m in layer 6 and \isn{app-routing-radius-layer-12}\,m in
layer 12, compared with \isn{app-routing-radius-car}\,m for that side of the car as a whole.
No coordinate enters the routing, only invariant relations to the anchors and the learned
state.

\FloatBarrier
\section{Experimental Setup}
\label{app:setup}

AirfRANS and DrivAerML appear in two sets of experiments.
The pose and re-sampling tests (\cref{app:encoding-setup}) train one network per model; the
benchmark comparisons (\cref{app:airfrans-setup,app:protocol}) average several networks per model
on the FluidsBench splits of \cref{tab:splits}.
\Cref{tab:recipe} gives the training recipe of the pose and re-sampling tests, and
\cref{app:aerozoo-setup,app:protocol} state how the recipes of \aerozoo{} and the automotive
datasets differ from it.

\begin{table}[h]
  \caption{The training recipe of the pose and re-sampling tests (\cref{app:encoding-setup}).
  Parentheses: the learning rate and length that differ on \aerozoo{} and on two tasks;
  \aerozoo{}'s loss is in \cref{app:aerozoo-setup}.}
  \label{tab:recipe}
  \centering
  \small
  \begin{tabular}{@{}ll@{}}
    \toprule
    Setting & Value \\
    \midrule
    Optimizer & Muon for matrices, AdamW for all other parameters; weight decay $10^{-4}$ \\
    Learning rate & $3.75\times10^{-4}$ (\aerozoo{} $1.125\times10^{-3}$); 5-epoch warm-up, cosine to $10^{-6}$ \\
    Length & 500 epochs (\aerozoo{} 250, HiLiftAeroML \isn{inv-hl-epochs}, DeepJEB++ 40) \\
    Batch & four cases per step, one on each of four GPUs \\
    Loss & Huber (threshold 1) on standardized targets, measure-weighted on surface tasks \\
    Points per case and step & 200,000 surface cells; or 200,000 tokens and 200,000 query points \\
    Precision & bfloat16 autocast in training; float32 in evaluation \\
    Checkpoint & the final one \\
    Augmentation, gradient clipping, weight averaging & none \\
    \bottomrule
  \end{tabular}
\end{table}

\FloatBarrier
\subsection{Datasets}

\begin{table}[h]
  \caption{The datasets. ``Cases'' counts the cases with released fields (the dataset papers
  describe 500 for DrivAerML and 355 for WindsorML), and the cell counts are per case, from the
  released files.
  DrivAerML stores polygons: split into triangles for the interior form of \isla{}, the surface
  of a car has about $1.7\times10^7$ cells.}
  \label{tab:datasets}
  \centering
  \small
  \setlength{\tabcolsep}{4pt}
  \begin{tabular}{@{}llllll@{}}
    \toprule
     & & & & \multicolumn{2}{c}{Cells per case} \\
    \cmidrule(l){5-6}
    Dataset & Cases & What varies & Simulation & Surface & Volume \\
    \midrule
    \aerozoo{} \citep{sharpe2026aerozoo} & 3,815 & shape, enclosure, size, speed, & unsteady RANS, & 0.43 M & 5.0 M \\
     & & $\Rey$ ($10^3$ to $10^8$), orientation & transition SST & (median) & (median) \\
    AirfRANS \citep{bonnet2022airfrans} & 1,000 & airfoil, angle of attack, $\Rey$ & RANS (2-D) & about 1,000 & about 179,000 \\
    DrivAerML \citep{ashton2024drivaerml} & 484 & car shape (16 parameters) & hybrid RANS-LES & 7.4 to 10.1 M & about 142 M \\
    WindsorML \citep{ashton2024windsorml} & 350 & Windsor-body shape (7 parameters) & wall-modeled LES & 2.0 to 2.6 M & 271 to 330 M \\
    AhmedML \citep{ashton2024ahmedml} & 500 & Ahmed-body shape (6 parameters) & hybrid RANS-LES & 0.7 to 1.5 M & 20 to 23 M \\
    \bottomrule
  \end{tabular}
\end{table}

\Cref{tab:datasets} lists the datasets.
\aerozoo{} varies the shape (seven geometry families, no body used twice) and the enclosure, and
with it the layout of the boundary conditions (eight configurations; \isn{az-n-ducts} of the
cases are ducts without a body).
It also varies the size (0.02 to 20~m), the inlet speed (0.25 to 50~m/s), the Reynolds number,
and the orientation of every case, which is random.
Its inputs are the four boundary meshes of a case, by condition type (inlet, outlet, no-slip,
slip), the values that each prescribes, and the fluid's viscosity and density
\citep{sharpe2026aerozoo}.
\aerozoo{} calls a solved and released case a sample; we say ``case'' for every dataset and
keep ``sample'' for a draw of cells.
AirfRANS varies airfoil geometry, inlet direction, and speed in two dimensions, while the three
road-vehicle datasets (DrivAerML, WindsorML, and AhmedML) hold one flow condition each, in one
orientation, from one mesher, so their cases differ in shape only.
The closed-surface hypothesis of \cref{thm:invariance} concerns the interior form only.
It holds on AirfRANS, whose airfoil and far field are closed curves, and on Tensile2d and the
bracket lug, whose boundaries are a closed curve and a watertight surface.
On \aerozoo{} release \isn{az-release} and DrivAerML small gaps remain between the stored boundary pieces (on DrivAerML,
at the tire contact patches), so there the winding number that signs $\delta_q$
(\cref{app:geometry}) is close to an integer but not exactly an integer.

\FloatBarrier
\subsection{AeroZoo}
\label{app:aerozoo-setup}

\paragraph{Data.}
We use the interior task of \aerozoo{} release
\isn{az-release}\reviewpreprint{}{ (catalog SHA-256 \texttt{\isn{az-catalog-hash}\dots})}
\citep{sharpe2026aerozoo}, on its \splitname{full} split and
on six of its splits that hold out a region (\cref{tab:aerozoo-splits}).
Training, validation, and test use the canonical cases of a split, those whose time averages
are converged (\isn{az-n-canonical} of the \isn{az-n-released}); on \splitname{full} these are
\isn{az-full-n-train} training, \isn{az-full-n-val} validation, and \isn{az-full-n} test cases.
The flag that marks a case as canonical is computed from the solution and is not an input.
As \aerozoo{} specifies, the two splits that hold out a geometry family are scored on test
cases with a body.

\paragraph{Metrics.}
With cell volumes $V_i$, reference speed $U_{\mathrm{ref}}$, and reference dynamic pressure
$q_{\mathrm{ref}}$ of a case,
\begin{equation}
  \varepsilon_U = \frac{1}{U_{\mathrm{ref}}}
    \Bigl(\frac{\sum_i V_i \lVert \hat{\vu}_i - \vu_i \rVert^2}{\sum_i V_i}\Bigr)^{1/2},
  \qquad
  \varepsilon_p = \frac{1}{q_{\mathrm{ref}}}
    \Bigl(\frac{\sum_i V_i (\hat p_i - p_i)^2}{\sum_i V_i}\Bigr)^{1/2},
\end{equation}
over every interior cell \citep{sharpe2026aerozoo}.
The reference scales and the cell volumes are used by the scorer only.
Results are means over the canonical test cases; the means are dominated by
crossflow and dual-inlet cases.
The free-stream reference predicts the mean prescribed inlet state everywhere and calibrates the
metrics case by case; the zero predictor predicts zero velocity and gauge pressure.
We also report the share of cases on which a model's $\varepsilon_U$ is below the
reference's.

\paragraph{\isla{} on \aerozoo{}.}
The four conditions are declared by name: the inlet with its prescribed velocity and turbulence
quantities, the outlet with its pressure and reverse-flow values, and the no-slip and slip walls
without data.
The inlet velocity defines the velocity scale and the one global direction, the density is a
declared scale, and the length unit is that of \cref{eq:length}.
The only global scalar is $\log_{10}(U L / \nu)$ minus its mean over the training cases, so
the length by which coordinates are scaled enters the model only through a Reynolds number
built on it.
Other dimensionless inputs and outputs are divided by constants fitted on the training
cases.
\isla{} predicts all seven fields of the task (velocity, pressure, and five turbulence
quantities).
Training uses 200,000 interior points and 200,000 boundary tokens per case, for 250 epochs with
5 epochs of linear warm-up and a cosine decay to $10^{-6}$, a per-case relative $L^2$ loss per
field in which velocity has weight 3, four cases per step, and the final checkpoint.
The interior-only \isla{} has three sets of \aerozoo{} runs.
The interior-only runs of \cref{tab:aerozoo-main} use a peak learning rate of $1.125\times 10^{-3}$, three times
the rate of \cref{tab:recipe}, chosen on validation by the rule of the baselines below
(\isn{az-isla-seeds} seeds, \splitname{full} only).
The held-out runs of \cref{tab:aerozoo-splits} and \cref{fig:aerozoo-heldout} use
$3.75\times 10^{-4}$ with the same schedule (\isn{az-splits-seeds} seeds on every split,
\splitname{full} included).
The size runs of \cref{tab:aerozoo-capacity} use $3.75\times 10^{-4}$ with a step schedule that
divides the learning rate by 10 after epochs 100 and 200 (\isn{az-cap-n-seeds} seeds).
For scoring, the model predicts every interior cell in float32, with the one deterministic draw
of boundary tokens that evaluation uses.

The row ``\isla{} + boundary targets'' of \cref{tab:aerozoo-main} also supervises the boundary.
Each step adds 200,000 boundary faces, queried $10^{-6} L$ off the face on the fluid side, and
supervises boundary pressure, boundary velocity, and wall shear stress where no Dirichlet
condition fixes them, with three more output channels (\isn{az-joint-params}~M parameters).
This costs \isn{az-joint-step-ratio} times the interior-only step time and leaves interior
inference unchanged.
Its learning rate, $1.125\times 10^{-3}$, was chosen over $3.75\times 10^{-4}$ on the validation part
of \splitname{full} (\isn{az-joint-seeds} seeds each).

\paragraph{Baselines on \aerozoo{}.}
\Cref{tab:aerozoo-main,tab:aerozoo} hold our runs of Transolver
(\isn{az-baseline-params-t}~M parameters) and GeoTransolver (\isn{az-baseline-params-gt}~M), both
with 12 layers of width 256.
Every model trains on the training part of \splitname{full}
(\isn{az-full-n-train} cases) for 250 epochs on four GPUs with one case per GPU, so
\isn{az-rep-steps} optimizer steps, with 200,000 interior points per case per epoch, bfloat16
autocast, and the final checkpoint.
Each interior-only model's schedule and peak learning rate were chosen on the validation part of
\splitname{full} by one rule.
Six settings ran at seed 42: a step schedule, or warm-up and cosine, at one third, one, and three
times the model's reference rate.
The two best by the mean of $\ln \varepsilon_U$ and $\ln \varepsilon_p$ and the model's default
setting got a second seed, the lowest two-seed mean was chosen, and seeds 42 to 47 then ran at
that setting.
Cosine at the reference rate ($10^{-3}$) was chosen for both baselines, and cosine at three
times the reference rate for \isla{}.
The baselines take per-case scales: coordinates are divided by a length computed from the
posed problem, velocities by the inlet's speed, and the global input is a standardized
Reynolds number.
Both take each point's signed distance to the no-slip body and its normal; GeoTransolver also
takes tokens of all four boundaries with their condition types and prescribed values, and
Transolver the flow axis from the inlet to the outlet.

\FloatBarrier
\subsection{Pose and Re-sampling Tests}
\label{app:encoding-setup}

\paragraph{Models.}
The tests of \cref{sec:exp-encoding} use single models, each trained on a dataset's stored
presentation without augmentation: Transolver with 16 layers of width 512 (47~M parameters);
GeoTransolver with 12 layers of width 1,024 on the three surface flow tasks (143~M parameters,
trained with 100,000 cells) and width 256 elsewhere (9~M parameters; 28~M with its local
features on the DrivAerML volume task); and \isla{} with the configuration of \cref{tab:hyper}
and the recipe of \cref{tab:recipe}.
\Cref{fig:hero} and Tables~\ref{tab:certificates}, \ref{tab:cost-surface}, and~\ref{tab:traincost} use the 9~M width-256
GeoTransolver on the surface task.
The baselines use the same optimizer, batch, number of epochs, warm-up, and cosine schedule as
\isla{} (GeoTransolver uses a step schedule on the bracket lug and, in float32 with tuned field
weights, on DrivAerML volume), a learning rate of $10^{-3}$, and bfloat16.
They take the coordinates of the sampled cells or points, centered on the case, the normals,
the unit free-stream direction, and for volume points the stored signed distance and its
gradient.
\isla{} takes the named boundaries of the whole domain: on DrivAerML, HiLiftAeroML, and SHIFT-SUV
it therefore receives the tunnel or far-field boundaries as well as the body, which the baselines
do not.
Each model has four seeds on the three surface flow tasks and on DrivAerML volume (and \isla{}
on Tensile2d), and two seeds elsewhere.
All models are evaluated in float32 at their final checkpoint.

\paragraph{Data and metric.}
On AirfRANS we use the 800 training and 200 evaluation cases of its \splitname{full} task, with
100,000 points per case.
On DrivAerML we use 435 training and 48 evaluation cases (one of its 484 cases is in neither set),
with 200,000 cells or points per case (Transolver is trained with 100,000 and evaluated with
200,000).
We also report the tests on HiLiftAeroML \citep{ashton2026hiliftaeroml} and SHIFT-SUV
\citep{luminary2025shiftsuv}, and on three solid-mechanics datasets to which the same models
apply unchanged: Tensile2d \citep{casenave2025plaid}, the bracket lug of
\citet{liu2025geometry}, and DeepJEB++ \citep{yoo2026deepjeb}.
On HiLiftAeroML (510 training and 180 evaluation cases), all three models train for
\isn{inv-hl-epochs} epochs (\isn{inv-hl-steps} optimizer steps) on one newer code snapshot; a
four-seed control with \isla{}'s earlier snapshot on the same surfaces differs by
\isn{inv-hl-code-p-pct}\% in pressure and \isn{inv-hl-code-tw-pct}\% in wall shear, both within
four-seed noise, so any code effect is unresolved.
\Cref{fig:encoding} shows one field per task: velocity on the two interior flow tasks, pressure
on the three surface tasks, stress on Tensile2d and the lug, and displacement on DeepJEB++.
\Cref{tab:encoding} gives every field.
The metric is the relative $L^2$ error of a field over the scored cells or points of a case,
weighted by cell measure on surfaces and unweighted over the sampled points of an interior
task, averaged over cases and then over seeds.
Every model is scored on the same cells of each case.
The tests compare each model with itself in another presentation; the baselines here are single
networks rather than the ensembles of \cref{sec:exp-bench} (\cref{tab:encoding-bench} repeats the
DrivAerML tests on networks from those ensembles).

\paragraph{Pose.}
Each case receives one rotation, uniform on $\mathrm{SO}(3)$ (in the plane for two-dimensional
problems), and a translation by one root-mean-square radius of its points, both fixed by the
case's identifier.
Positions, normals, global vectors, and vector and tensor targets are transformed together, and
predictions are compared with the transformed targets.
The data pipelines of the automotive, wing, and DeepJEB++ datasets center each case on the mean of its
points, which removes the translation before any model receives it; on AirfRANS, Tensile2d, and
the bracket lug the translation reaches every model.

\paragraph{Re-sampling by area.}
From a pool of $10^6$ cells drawn uniformly, each cell is kept independently with probability
proportional to its area, for an expected 200,000 cells, and is weighted by its measure divided by its
inclusion probability.
\isla{} uses the cells with these measures; the baselines use the same cells' centroids and
normals.
The metric uses the same weights for every model.
This test applies where a surface has more cells than a model receives (DrivAerML, HiLiftAeroML,
SHIFT-SUV).

\paragraph{Benchmark networks.}
\Cref{tab:encoding-bench} repeats the DrivAerML surface tests with one network of each family's
DrivAerML \splitname{full} entry, on its \isn{encb-n-cases} validation cases.
In each rotated row each case has one pose, the same for every model; every pipeline centers the
case, so no model receives the translation.
The rows of the top block down to the canonical frames are scored on every cell of the stored mesh
(the two indented rows apply the vehicle's principal-axis canonical frame at inference, without
and with a fixed alignment rotation fitted on the training cases), and the last two on one sample
of about $10^6$ cells per case drawn in proportion to area, with Horvitz--Thompson weights, the
same cells for every model.
\isla{}'s network there is its interior form, a joint surface and volume network, which draws its
own tokens in proportion to area (\cref{app:draw}).
The networks of Transolver, GeoTransolver, and \globe{} are at the checkpoints of their entries,
chosen on these cases; \isla{}'s is its final checkpoint.
The bottom block retrains one surface network each of Transolver and GeoTransolver with the
pipeline's vertical-axis rotation and $\pm 1\,$m horizontal translation (one seed), scored in the
same way.

\FloatBarrier
\subsection{AirfRANS}
\label{app:airfrans-setup}

On AirfRANS we use the three splits and node-level evaluator defined by FluidsBench
\citep{fluidsbench2026fluidsbench} (\cref{tab:splits}): \splitname{full} and \splitname{scarce},
with the same 200 test cases, and \splitname{aoa_extrapolation}, whose test angles of attack lie
outside the training range.
AirfRANS defines 800 official training cases on \splitname{full} and 804 on
\splitname{aoa_extrapolation}, with no separate validation list; we hold out 80 of them for
validation (the same 80 on \splitname{scarce}, whose 200 training cases do not contain them, and
80 of its own on \splitname{aoa_extrapolation}).
Each model's result averages the predictions of several independently trained networks
(\cref{tab:lineups}): on \splitname{full}, Transolver and GeoTransolver blend two kinds of network
by wall distance, \globe{} averages five networks of 1.1~M parameters, and \isla{} 17 of 67~M.
On \splitname{full} and \splitname{aoa_extrapolation}, every network trains on the remaining 720
or 724 cases, except \isn{fb-airfrans-full-isla-n-refit} of \isla{}'s networks on each split, which
train on all 800 or 804 with a fixed schedule and the final checkpoint; \isla{}'s validation rows in \cref{tab:bench-airfrans} leave
those out, and no other model trains on the 80 validation cases.
Each model predicts velocity and pressure at every mesh node of each test case, and the evaluator
derives the wall quantities and integrates the drag and lift coefficients ($R^2$ over cases).
\isla{} takes the airfoil and the free stream and predicts a wall-scaled velocity, Transolver and
GeoTransolver predict through a form that vanishes on the wall, and \globe{} predicts a
disturbance of the free stream.

\FloatBarrier
\subsection{Automotive Aerodynamics Datasets}
\label{app:protocol}

On DrivAerML, WindsorML, and AhmedML we use the splits and evaluators defined by FluidsBench
\citep{fluidsbench2026fluidsbench}.
We report the evaluator's field errors and $R^2$ values.

\paragraph{Splits.}
Each road-vehicle dataset has eight splits (\cref{tab:splits}): four share one validation and one
test list and differ in the number of training cases (\splitname{full}, \splitname{medium},
\splitname{scarce}, and \splitname{super_scarce}), and four hold out a region for testing: the
cases farthest from the others in shape (\splitname{geometry}), the highest or lowest drag
(\splitname{high_drag}, \splitname{low_drag}), and the largest wake or rear separation, measured
on published images (\splitname{rear_separation} on DrivAerML, \splitname{image_wake} on
WindsorML and AhmedML).
On AhmedML the validation cases of these four splits are drawn from the training-side
population, so validation errors there do not show the shift of the test cases.

\begin{table}[h]
  \caption{Splits of AirfRANS and the automotive aerodynamics datasets (defined by FluidsBench;
  \citealp{fluidsbench2026fluidsbench}): numbers of training / validation / test cases.
  The last row gives the counts of each out-of-distribution split (one on AirfRANS, four on each
  automotive dataset).
  Dash: the dataset has no such split.}
  \label{tab:splits}
  \centering
  \small
  \begin{tabular}{@{}lllll@{}}
    \toprule
     & \multicolumn{4}{c}{Cases (train / validation / test)} \\
    \cmidrule(l){2-5}
    Split & AirfRANS & DrivAerML & WindsorML & AhmedML \\
    \midrule
    \splitname{full} & 720 / 80 / 200 & \isn{splits-drivaerml-full} & \isn{splits-windsorml-full} & \isn{splits-ahmedml-full} \\
    \splitname{medium} & \na & \isn{splits-drivaerml-medium} & \isn{splits-windsorml-medium} & \isn{splits-ahmedml-medium} \\
    \splitname{scarce} & 200 / 80 / 200 & \isn{splits-drivaerml-scarce} & \isn{splits-windsorml-scarce} & \isn{splits-ahmedml-scarce} \\
    \splitname{super_scarce} & \na & \isn{splits-drivaerml-super-scarce} & \isn{splits-windsorml-super-scarce} & \isn{splits-ahmedml-super-scarce} \\
    out of distribution & 724 / 80 / 196 (one) & \isn{splits-drivaerml-ood} & \isn{splits-windsorml-ood} & \isn{splits-ahmedml-ood} \\
    \bottomrule
  \end{tabular}
\end{table}

\paragraph{Entries and metrics.}
An entry predicts pressure and wall shear stress at every surface cell and velocity and pressure
at every volume cell of the reference mesh of each test case; force coefficients are integrated by
the evaluator from the predicted fields.
All entries are trained from scratch on the official training cases of their split.
For each case the evaluator computes the relative $L^2$ error of each field over all evaluated
entities, weighted by face area on the surface and equally per cell in the volume, and averages
it over cases, alongside the $R^2$ of drag and lift.

\paragraph{Model families.}
We trained all four families ourselves, using the released PhysicsNeMo implementations of
Transolver, GeoTransolver, and \globe{} \citep{physicsnemo2023nvidia} and the \isla{}
architecture of \cref{tab:hyper}.
Each model's result, its \emph{entry}, is the equal-weight mean of independently trained
networks, taken separately for the surface and volume (\cref{tab:lineups}), with the number of networks, the recipe, and the
inputs set per model and dataset:
\begin{enumerate}[leftmargin=*, itemsep=1pt, topsep=2pt]
  \item \emph{Size.} On DrivAerML one \isla{} network predicts the surface and the volume; on its
  full training set the entries hold \isn{entry-params-dml-full-t} (Transolver, 15 networks),
  \isn{entry-params-dml-full-gt} (GeoTransolver, 15), and \isn{entry-params-dml-full-isla}
  (\isla{}, 9) billion parameters.
  On WindsorML and AhmedML every family trains separate surface and volume networks, and
  \isla{}'s networks and entries are the largest.
  \item \emph{Inputs.} Transolver and GeoTransolver take the coordinates and normals of the
  sampled cells, the free-stream direction, and for volume points a stored signed distance to
  the body.
  \globe{} takes the body and, on DrivAerML and WindsorML, patches of the floor and tunnel as
  boundary sources.
  \isla{} takes the named boundaries of the whole domain (the body and the tunnel's inlet,
  outlet, floor, and walls) and no coordinates, computing boundary distance itself
  (\cref{app:geometry}).
  \item \emph{Loss.} \isla{} trains on a per-case relative $L^2$ loss, and the baselines on a
  Huber loss of standardized fields.
  \item \emph{Budget.} Every network of all four families trains for at least
  \isn{steps-drivaerml-isla-full} optimizer steps with a cosine schedule (matched in steps rather
  than sampled points per step; \cref{tab:lineups} gives the steps and, for DrivAerML and AhmedML,
  the checkpoints).
  \item \emph{Augmentation and precision.} No family was trained with rotations or translations.
  Transolver and GeoTransolver are trained in float32 on DrivAerML and in bfloat16 elsewhere;
  \globe{} and \isla{} are trained in bfloat16.
\end{enumerate}

\begin{table}[h]
  \caption{The networks behind each model's result on AirfRANS and the automotive aerodynamics
  datasets (\cref{tab:airfrans,tab:bench}); where results differ between the splits of a dataset,
  the table lists the values that occur ($S$, $V$: surface and volume networks). Parameters are
  per network, and the total sums them over
  the networks of the \splitname{full} result. Learning rates are peak values, and optimizer steps
  are those scheduled. Checkpoint: the saved one with the lowest validation error (``best
  saved'') or the final one. A dash: does not apply.}
  \label{tab:lineups}
  \centering
  \footnotesize
  \setlength{\tabcolsep}{3pt}
  \begin{tabular}{@{}>{\raggedright\arraybackslash}p{2.5cm}>{\raggedright\arraybackslash}p{3.0cm}>{\raggedright\arraybackslash}p{4.0cm}>{\raggedright\arraybackslash}p{2.7cm}>{\raggedright\arraybackslash}p{3.5cm}@{}}
    \toprule
    Setting & Transolver & GeoTransolver & \globe{} & \isla{} \\
    \midrule
    \multicolumn{5}{@{}l}{\emph{AirfRANS} (\splitname{full}, \splitname{scarce}, \splitname{aoa_extrapolation})} \\
    Networks & 3, 15, 20 & 9, 9, 9 & 5, 7, 4 & 17, 6, 13 \\
    Parameters & 9 M & 9 M and larger & 1.1 M (two of 0.3 M) & 67 M \\
    Total parameters, \splitname{full} & \isn{entry-params-af-full-transolver} & $\ge$\,\isn{entry-params-af-full-geotransolver} & \isn{entry-params-af-full-globe} & \isn{entry-params-af-full-isla} \\
    \addlinespace[3pt]
    \multicolumn{5}{@{}l}{\emph{DrivAerML}} \\
    Networks (surface + volume) & $5{+}5$ (\splitname{full}: $5{+}10$) & $5{+}5$ (\splitname{full}: $10{+}5$) & $3{+}3$ & 5 (\splitname{full}: 9), each a joint surface and volume network \\
    Layers $\times$ width & $16 \times 768$ & $12 \times 768$ or $12 \times 512$ ($S$), $12 \times 256$ ($V$) & \na & $12 \times 512$ \\
    Parameters & 106 M & 80 M or 36 M ($S$), 28 M ($V$) & 12 M & 67 M \\
    Training points & 100,000 & 200,000 & 50,000 & 200,000 ($S$) and 200,000 ($V$) \\
    Learning rate & $10^{-3}$ & $10^{-3}$ & $3\times10^{-3}$ & $7.5\times10^{-4}$ \\
    Optimizer steps & \isn{steps-drivaerml-transolver-full} on every split & \isn{steps-drivaerml-geotransolver-full} on every split & $\ge 50{,}000$ on every split & \isn{steps-drivaerml-isla-full} on every split \\
    Checkpoint & best saved & best saved & best saved & final \\
    \addlinespace[3pt]
    \multicolumn{5}{@{}l}{\emph{WindsorML}} \\
    Networks (surface + volume) & $6{+}6$ (\splitname{super_scarce}: $3{+}3$) & $3{+}3$ & $3{+}3$ & 2 to 5 ($S$), 2 to 5 ($V$) \\
    Parameters & 7 M & 18 M & 4 M & 66 M ($S$), 67 M ($V$) \\
    Training points & 200,000 & 200,000 & 40,000 & 200,000 \\
    Learning rate & $10^{-3}$ or $2\times10^{-3}$ & $3\times10^{-3}$ & $3\times10^{-3}$ & $3.75\times10^{-4}$ ($S$), $7.5\times10^{-4}$ ($V$) \\
    Optimizer steps & $\ge 50{,}000$ on every split & $\ge 50{,}000$ on every split & $\ge 50{,}000$ on every split & $\ge 50{,}000$ on every split \\
    \addlinespace[3pt]
    \multicolumn{5}{@{}l}{\emph{AhmedML}} \\
    Networks (surface + volume) & $3{+}3$ or $3{+}4$ & $3{+}3$ or $3{+}4$ & $3{+}3$, $4{+}3$, or $3{+}4$ & $2{+}2$ to $3{+}8$ \\
    Parameters & 47 M & 36 M ($S$), 28 M ($V$) & 1 M & 66 M ($S$), 67 M ($V$) \\
    Training points & 100,000 & 200,000 ($S$), 100,000 ($V$) & 80,000 & 200,000 \\
    Learning rate & $10^{-3}$ & $10^{-3}$ & $10^{-2}$ & $3.75\times10^{-4}$ ($S$), $7.5\times10^{-4}$ ($V$) \\
    Optimizer steps & \isn{steps-ahmedml-transolver-full} on every split & \isn{steps-ahmedml-geotransolver-full} on every split & \isn{steps-ahmedml-globe-full} on every split & \isn{steps-ahmedml-isla-full} on every split \\
    Checkpoint & final & final & final & final \\
    \bottomrule
  \end{tabular}
\end{table}

\paragraph{Dataset notes.}
\emph{WindsorML.}
WindsorML stores an absolute volume pressure of about 37~kPa, so the relative error of volume
pressure is only \isn{fb-windsorml-vp-min} to \isn{fb-windsorml-vp-max}\% for every model.
Every WindsorML result uses FluidsBench's corrected split lists, which leave out one case of each
of six pairs with identical geometry.\isn{bench-retrained-note-windsorml}

\emph{AhmedML.}
On \splitname{image_wake}, the field errors of all four families on test cases are
\isn{ahm-shift-image-wake-min} to \isn{ahm-shift-image-wake-max} times those on validation cases,
and on \splitname{high_drag} those of Transolver, GeoTransolver, and \globe{} are
\isn{ahm-shift-high-drag-min} to \isn{ahm-shift-high-drag-max} times; this is the shift that those
splits are built to contain.
\isla{}'s \splitname{full} entry has three surface and eight volume networks, the latter
trained with five variants of the recipe; its validation entries on \splitname{high_drag} and
\splitname{low_drag} were not evaluated per field.

\FloatBarrier
\section{Additional Results}
\label{app:results}

Where a table bolds a best value, it bolds the single best printed value of each group that its
caption names, and a tie at the printed precision is not bold; \cref{tab:bench} instead bolds every
ratio below 1.

\FloatBarrier
\subsection{AeroZoo: Every Model and Split}
\label{app:aerozoo-results}

\Cref{tab:aerozoo} gives every model on \splitname{full}, and \cref{tab:aerozoo-enclosure} their
$S_{\mathrm{flow}}$ and mean $\varepsilon_U$ by enclosure configuration.
The mean pressure error of every model is set by a few cases: \isla{}'s mean
$\varepsilon_p$ is \isn{az-full-epsp-mean-isla}, and
the \isn{az-full-epsp-tail-n} cases with the largest $\varepsilon_p$ account for nearly half of it
(\isn{az-full-epsp-tail-rest} without them).
\isla{}'s mean $\varepsilon_p$ on \splitname{full} therefore depends on its run set
(\cref{app:aerozoo-setup}): \isn{az-full-epsp-mean-isla} for the runs of
\cref{tab:aerozoo-main,tab:aerozoo}, \isn{az-fulldef-epsp-mean-isla} for the held-out runs
(\cref{tab:aerozoo-splits}), and \isn{az-cap-ctrl-epsp-mean} for the size runs
(\cref{tab:aerozoo-capacity}).
At the learning rate of \cref{tab:aerozoo-main} ($1.125\times 10^{-3}$), supervising boundary targets as well lowers
mean $\varepsilon_p$ to \isn{az-joint-epsp-mean}, with relative $L^2$
pressure error \isn{az-joint-relp-cut-pct}\% lower than the interior-only model's.
\isla{} also predicts the five turbulence fields of the task (\cref{tab:aerozoo-aux}).
For the intermittency and the transition Reynolds number we report its mean relative $L^2$
errors, \isn{az-full-rel-gamma-isla} and \isn{az-full-rel-retheta-isla}, in place of
\aerozoo{}'s own metrics.

\begin{figure}[htb]
  \centering
  \includegraphics{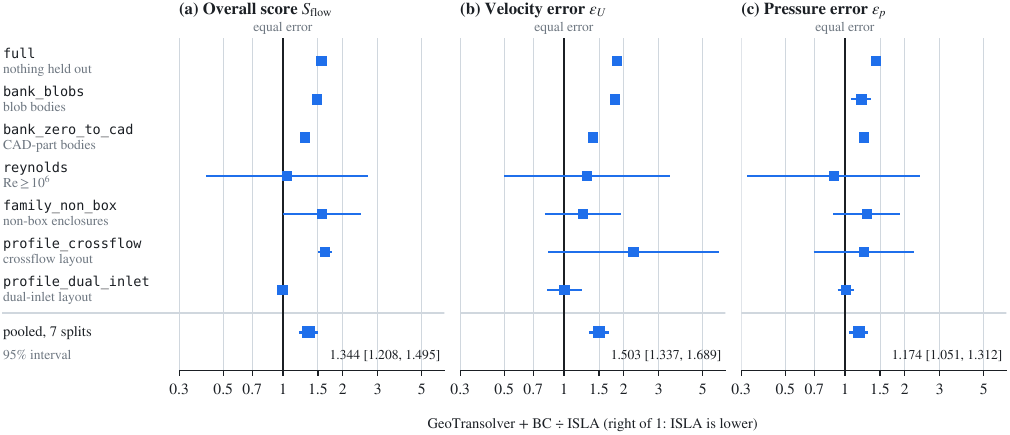}
  \caption{\textbf{GeoTransolver + BC's overall score and errors over \isla{}'s on the test part
  of each \aerozoo{} split.}
  \isla{} at $3.75\times 10^{-4}$, two seeds per split, compared with GeoTransolver + BC's
  \isn{az-rep-seeds} seeds on \splitname{full} and three seeds per held-out split (the runs of
  \cref{tab:aerozoo-splits}).
  Each ratio divides geometric means over seeds, of $S_{\mathrm{flow}}$ or of the mean error over
  canonical test cases; bars are two-sample Welch 90\% intervals on per-seed log values, and the
  bottom row is the equal-weight ratio pooled over the seven splits with its 95\%
  Welch--Satterthwaite interval.
  Both models share the training parts, steps, scorer, and test cases;
  GeoTransolver + BC's schedule is chosen per split on validation, which favors it.
  On \splitname{family_non_box} both models' errors exceed the free stream's, on
  \splitname{profile_crossflow} GeoTransolver + BC's velocity and pressure errors do, and on
  \splitname{profile_dual_inlet} both remain far from the target ($\varepsilon_p$ above 8).
  As in \cref{tab:aerozoo-splits}, the geometry-family splits are scored on cases with a body.}
  \label{fig:aerozoo-heldout}
\end{figure}

\Cref{tab:aerozoo-splits} gives GeoTransolver + BC and \isla{} on every \aerozoo{} split that we
trained, with \isla{} at $3.75\times 10^{-4}$ on every split.
With a geometry family held out, \isla{}'s mean velocity error on cases with a body is
\isn{az-blobs-epsU-mean-isla} (blobs) and \isn{az-ztc-epsU-mean-isla} (CAD parts), compared
with \isn{az-fulldef-epsU-mean-isla} over all test cases of \splitname{full}.
With the Reynolds numbers above $10^6$ held out, the mean velocity error
(\isn{az-re-epsU-mean-isla}) is three times lower than the reference's
(\isn{az-re-epsU-mean-fs}).
With a layout of the boundary conditions held out (\splitname{profile_crossflow},
\splitname{profile_dual_inlet}), \isla{}'s mean $\varepsilon_U$ is \isn{az-pcf-ratio-pct} and
\isn{az-pdi-ratio-pct}\% below the reference's.
With every non-box enclosure held out, \isla{} is less accurate than the reference in both
fields.
\Cref{fig:aerozoo-heldout} compares GeoTransolver + BC, with its schedule chosen per split on that
split's validation cases, with these \isla{} runs.
On \splitname{full}, \splitname{bank_blobs}, and \splitname{bank_zero_to_cad}, \isla{} has a
lower $S_{\mathrm{flow}}$ and lower errors in both fields, and on \splitname{profile_crossflow} a
lower $S_{\mathrm{flow}}$, under a two-sample Welch 90\% interval on per-seed log values.
The remaining held-out pairs are inconclusive across seeds: on \splitname{family_non_box},
\splitname{reynolds}, and \splitname{profile_crossflow}, the per-seed log-error standard
deviations reach \isn{az-ho-obs-sd-min} to \isn{az-ho-obs-sd-max}.
Pooled with equal weight across \splitname{full} and the six held-out splits, GeoTransolver + BC's
geometric-mean $S_{\mathrm{flow}}$ is \isn{az-poolcur-s} times \isla{}'s (95\%
Welch--Satterthwaite interval $[\isn{az-poolcur-s-lo}, \isn{az-poolcur-s-hi}]$), its velocity error
\isn{az-poolcur-u} times ($[\isn{az-poolcur-u-lo}, \isn{az-poolcur-u-hi}]$), and its pressure
error \isn{az-poolcur-p} times ($[\isn{az-poolcur-p-lo}, \isn{az-poolcur-p-hi}]$).
\begin{table}[h]
  \caption{\textbf{\aerozoo{}, \splitname{full} split} (\isn{az-full-n} canonical test cases):
  our runs at one step budget (\isn{az-rep-steps} optimizer steps), each interior-only model with
  its schedule and learning rate chosen on validation, and the boundary-target \isla{} row at the
  rate chosen between $3.75\times 10^{-4}$ and $1.125\times 10^{-3}$ on validation
  (\cref{app:aerozoo-setup}); the baselines use per-case scales; means of \isn{az-rep-seeds} seeds
  (\isn{az-joint-seeds} for \isla{} + boundary targets).
  $S_{\mathrm{flow}}$, near, BC, flow axis, and the last column: as in \cref{tab:aerozoo-main}.
  $r_U$, $r_p$: relative $L^2$ errors, with every interior point counted equally.
  ``\isla{} + boundary targets'' also supervises boundary targets in training.
  Bold: the best value in each column.
  Dash: no value.}
  \label{tab:aerozoo}
  \centering
  \small
  \setlength{\tabcolsep}{6pt}
  \begin{tabular}{@{}lr r rrrr rr r@{}}
\toprule
 & & Overall $\downarrow$ & \multicolumn{4}{c}{Mean ref.-scaled $\downarrow$} & \multicolumn{2}{c}{Mean rel.\ $L^2$ $\downarrow$} & Share $\uparrow$ \\
\cmidrule(lr){4-7}\cmidrule(lr){8-9}
Model & Parameters & $S_{\mathrm{flow}}$ & $\varepsilon_U$ & $\varepsilon_U^{\mathrm{near}}$ & $\varepsilon_p$ & $\varepsilon_p^{\mathrm{near}}$ & $r_U$ & $r_p$ & $\varepsilon_U <$ ref. \\
\midrule
Free-stream reference & \na & 1.000 & 0.440 & 0.442 & 1.809 & 2.014 & 0.389 & 1.000 & \na \\
Zero predictor & \na & 1.648 & 1.203 & 1.193 & 1.809 & 2.014 & 1.000 & 1.000 & \na \\
\midrule
Transolver + flow axis & 9.0 M & 0.679 & 0.306 & 0.309 & 1.212 & 1.315 & 0.276 & 1.419 & 61\% \\
GeoTransolver + BC & 27.6 M & 0.290 & 0.152 & 0.179 & 0.427 & 0.434 & 0.180 & 0.623 & 86\% \\
ISLA & 67.5 M & 0.153 & 0.082 & 0.112 & 0.196 & 0.214 & 0.110 & 0.304 & 98\% \\
ISLA + boundary targets & 67.5 M & \textbf{0.123} & \textbf{0.080} & \textbf{0.110} & \textbf{0.124} & \textbf{0.149} & \textbf{0.106} & \textbf{0.281} & 98\% \\
\bottomrule
\end{tabular}
\end{table}

\begin{table}[h]
  \caption{\aerozoo{}, \splitname{full}: $S_{\mathrm{flow}}$ and mean $\varepsilon_U$ by enclosure
  configuration (the latter are the numbers behind \cref{fig:aerozoo}), for the models of
  \cref{tab:aerozoo-main} (\isn{az-rep-seeds} seeds each, \isn{az-joint-seeds} for \isla{} +
  boundary targets).
  $S_{\mathrm{flow}}$ is taken against the free-stream reference on the cases of each
  configuration, so the reference scores 1.
  BC, flow axis: as in \cref{tab:aerozoo-main}.
  Bold: the lowest value in each column of each block.}
  \label{tab:aerozoo-enclosure}
  \centering
  \small
  \setlength{\tabcolsep}{3.5pt}
  \begin{tabular}{@{}lrrrrrrrr@{}}
\toprule
Model & Sphere & Open box & Ground effect & Confined box & Hemisphere & Duct & Crossflow & Dual inlet \\
\quad (\emph{test cases}) & (63) & (95) & (107) & (59) & (29) & (41) & (34) & (87) \\
\midrule
\multicolumn{9}{@{}l}{\emph{Overall score $S_{\mathrm{flow}}$ $\downarrow$}} \\
Transolver + flow axis & 0.877 & 3.110 & 2.166 & 0.967 & 0.847 & 0.930 & 0.553 & 0.603 \\
GeoTransolver + BC & 0.726 & 0.720 & 0.699 & 0.630 & 0.637 & 0.598 & 0.276 & 0.222 \\
ISLA & 0.510 & 0.495 & 0.476 & 0.431 & 0.377 & 0.306 & 0.168 & 0.100 \\
ISLA + boundary targets & \textbf{0.485} & \textbf{0.481} & \textbf{0.452} & \textbf{0.416} & \textbf{0.355} & \textbf{0.285} & \textbf{0.143} & \textbf{0.075} \\
\midrule
\multicolumn{9}{@{}l}{\emph{Mean reference-scaled velocity error $\varepsilon_U$ $\downarrow$}} \\
Free-stream reference & 0.041 & 0.049 & 0.085 & 0.171 & 0.195 & 0.538 & 1.019 & 1.586 \\
Transolver + flow axis & 0.049 & 0.108 & 0.108 & 0.145 & 0.212 & 0.391 & 0.566 & 0.954 \\
GeoTransolver + BC & 0.035 & 0.041 & 0.061 & 0.117 & 0.116 & 0.300 & 0.253 & 0.396 \\
ISLA & 0.026 & 0.030 & 0.047 & 0.086 & 0.057 & 0.136 & \textbf{0.174} & 0.164 \\
ISLA + boundary targets & \textbf{0.025} & 0.030 & \textbf{0.045} & \textbf{0.084} & \textbf{0.054} & \textbf{0.129} & 0.186 & \textbf{0.157} \\
\bottomrule
\end{tabular}
\end{table}

\begin{table}[h]
  \caption{\aerozoo{}, \splitname{full}: mean errors of three turbulence fields, in the units
  that \aerozoo{} defines ($U_{\mathrm{ref}}^2$, $U_{\mathrm{ref}}/L_{\mathrm{ref}}$, and
  $U_{\mathrm{ref}} L_{\mathrm{ref}}$).
  The errors of \aerozoo{}'s free-stream reference \citep{sharpe2026aerozoo} give their scale.
  Both models are the runs of \cref{tab:aerozoo-main} (\isn{az-rep-seeds} seeds each; BC: as there).
  Bold: the lowest error in each column.}
  \label{tab:aerozoo-aux}
  \centering
  \small
  \begin{tabular}{@{}lrrr@{}}
\toprule
 & \multicolumn{3}{c}{Mean ref.-scaled error $\downarrow$} \\
\cmidrule(l){2-4}
Model & $\varepsilon_k$ & $\varepsilon_\omega$ & $\varepsilon_{\nu_t}$ \\
\midrule
Free-stream reference & 0.011 & 2030 & 0.0029 \\
GeoTransolver + BC & 0.0089 & 2000 & 0.0012 \\
ISLA & \textbf{0.0070} & \textbf{1960} & \textbf{0.00070} \\
\bottomrule
\end{tabular}
\end{table}

\begin{table}[h]
  \caption{\textbf{GeoTransolver + BC and \isla{} on every \aerozoo{} split}: the
  \splitname{full} split and six splits that hold a region out of training, with the free-stream
  reference on the same canonical test cases.
  \isla{}: trained at $3.75\times 10^{-4}$ with warm-up and cosine, mean of
  \isn{az-splits-seeds} seeds on every split.
  GeoTransolver + BC: its schedule chosen per split on validation, mean of \isn{az-rep-seeds}
  seeds on \splitname{full} and three on each held-out split.
  $S_{\mathrm{flow}}$ and BC: as in \cref{tab:aerozoo-main}, with $S_{\mathrm{flow}}$ against the
  reference on the same cases.
  Bold: the lower of the two models' values in each split.
  The two splits that hold out a geometry family are scored on cases with a body.}
  \label{tab:aerozoo-splits}
  \centering
  \scriptsize
  \setlength{\tabcolsep}{5pt}
  \begin{tabular}{@{}llrr rr rrr rrr@{}}
\toprule
 & & \multicolumn{2}{c}{Cases} & \multicolumn{2}{c}{Free-stream reference} & \multicolumn{3}{c}{GeoTransolver + BC} & \multicolumn{3}{c}{\isla{} at $3.75\times 10^{-4}$} \\
\cmidrule(lr){3-4}\cmidrule(lr){5-6}\cmidrule(lr){7-9}\cmidrule(l){10-12}
 & & & & \multicolumn{2}{c}{Mean ref.-scaled $\downarrow$} & Overall $\downarrow$ & \multicolumn{2}{c}{Mean ref.-scaled $\downarrow$} & Overall $\downarrow$ & \multicolumn{2}{c}{Mean ref.-scaled $\downarrow$} \\
\cmidrule(lr){5-6}\cmidrule(lr){8-9}\cmidrule(l){11-12}
Split & Held out of training & train & test & $\varepsilon_U$ & $\varepsilon_p$ & $S_{\mathrm{flow}}$ & $\varepsilon_U$ & $\varepsilon_p$ & $S_{\mathrm{flow}}$ & $\varepsilon_U$ & $\varepsilon_p$ \\
\midrule
\splitname{full} & nothing & 2{,}011 & 515 & 0.440 & 1.809 & 0.290 & 0.152 & 0.427 & \textbf{0.186} & \textbf{0.082} & \textbf{0.297} \\
\splitname{bank_blobs} & blob bodies & 1{,}662 & 872 & 0.409 & 1.655 & 0.303 & 0.154 & 0.355 & \textbf{0.205} & \textbf{0.085} & \textbf{0.294} \\
\splitname{bank_zero_to_cad} & CAD-part bodies & 2{,}250 & 224 & 0.466 & 2.004 & 0.304 & 0.164 & 0.452 & \textbf{0.236} & \textbf{0.117} & \textbf{0.362} \\
\splitname{reynolds} & $\Rey \ge 10^6$ & 1{,}577 & 1{,}001 & 0.379 & 1.640 & 0.364 & 0.161 & \textbf{0.498} & \textbf{0.352} & \textbf{0.125} & 0.576 \\
\splitname{family_non_box} & non-box enclosures & 1{,}850 & 692 & 0.238 & 0.595 & 1.745 & 0.328 & 0.979 & \textbf{1.085} & \textbf{0.257} & \textbf{0.747} \\
\splitname{profile_crossflow} & crossflow layout & 2{,}298 & 173 & 0.947 & 3.490 & 1.243 & 1.450 & 3.605 & \textbf{0.773} & \textbf{0.656} & \textbf{2.906} \\
\splitname{profile_dual_inlet} & dual-inlet layout & 2{,}044 & 459 & 1.659 & 9.134 & \textbf{0.904} & 1.456 & 8.463 & 0.909 & \textbf{1.450} & \textbf{8.379} \\
\bottomrule
\end{tabular}
\end{table}

\Cref{tab:aerozoo-capacity} compares an \isla{} of about GeoTransolver + BC's size with one of the
reference size, both trained on the same step schedule (\cref{app:aerozoo-setup}).
The smaller one has \isn{az-cap-blocks-small} blocks, \isn{az-cap-slices-small} slices, an MLP
ratio of \isn{az-cap-mlp-small}, and $10^5$ boundary tokens per case in place of $2\times10^5$, so it
reduces the token budget as well as the parameters.
On \splitname{full} its mean $\varepsilon_U$ is \isn{az-cap-epsU-mean-pct}\% higher than the larger model's
and its mean $\varepsilon_p$ \isn{az-cap-epsp-mean-lower-pct}\% lower, and both have an
$S_{\mathrm{flow}}$ of \isn{az-cap-sflow-smaller}.
With every non-box enclosure held out (\splitname{family_non_box}), its mean $\varepsilon_U$ is
\isn{az-cap-nonbox-ratio} times the larger model's.
\begin{table}[!htb]
  \caption{\textbf{An \isla{} of 28.2 M parameters and one of 67.5 M} on \aerozoo{}'s
  \splitname{full} split (\isn{az-full-n} canonical test cases), both trained at
  $3.75\times 10^{-4}$ on the same step schedule, \isn{az-cap-n-seeds} seeds each.
  The smaller model also uses half the MLP width and half the boundary tokens per case.
  $S_{\mathrm{flow}}$ and the last column: as in \cref{tab:aerozoo-main}.}
  \label{tab:aerozoo-capacity}
  \centering
  \small
  \setlength{\tabcolsep}{4pt}
  \begin{tabular}{@{}lrrrrr@{}}
\toprule
 & & Overall $\downarrow$ & \multicolumn{2}{c}{Mean ref.-scaled $\downarrow$} & Share $\uparrow$ \\
\cmidrule(lr){4-5}
\isla{}, step schedule & Parameters & $S_{\mathrm{flow}}$ & $\varepsilon_U$ & $\varepsilon_p$ & $\varepsilon_U <$ ref. \\
\midrule
12 blocks, 128 slices & 67.5 M & 0.193 & 0.083 & 0.321 & 98\% \\
8 blocks, 64 slices & 28.2 M & 0.193 & 0.085 & 0.316 & 98\% \\
\bottomrule
\end{tabular}
\end{table}

\FloatBarrier
\subsection{AirfRANS}
\label{app:airfrans-results}

\Cref{tab:bench-airfrans} gives every field error and $R^2$ value on all three AirfRANS splits
(defined by FluidsBench; \citealp{fluidsbench2026fluidsbench}), on validation and test cases.
Bold marks the lowest error and the highest $R^2$ among the four families, and a dash a value
that does not exist.

\begin{table}[h]
  \caption{\textbf{AirfRANS}: every split (training cases in parentheses), on validation and test
  cases.
  $n$: evaluated cases.
  $p_s$ and $\tau_w$: surface pressure and wall shear stress; $\mathbf{u}$ and $p_v$: volume
  velocity and pressure; $C_d$ and $C_l$: the $R^2$ of the drag and lift coefficients.
  Bold: the lowest error and the highest $R^2$ among the four families where all four are
  evaluated.
  Dash: not evaluated.\isn{bench-refit-note}\isn{bench-retrained-note-airfrans}}
  \label{tab:bench-airfrans}
  \centering
  \scriptsize
  \setlength{\tabcolsep}{4pt}
  \begin{tabular}{@{}llr l rrrr rr@{}}
\toprule
 & & & & \multicolumn{4}{c}{Relative $L^2$ error [\%] $\downarrow$} & \multicolumn{2}{c}{Force $R^2$ $\uparrow$} \\
\cmidrule(lr){5-8}\cmidrule(l){9-10}
Split (train) & Cases & $n$ & Model & $p_s$ & $\tau_w$ & $\mathbf{u}$ & $p_v$ & $C_d$ & $C_l$ \\
\midrule
\splitname{full} (720) & validation & 80 & Transolver & 1.22 & 1.84 & 0.56 & 1.26 & 0.974 & 1.000 \\
 & &  & GeoTransolver & 1.34 & 1.52 & 0.50 & 1.27 & 0.970 & 1.000 \\
 & &  & GLOBE & \textbf{0.37} & \textbf{0.57} & \textbf{0.16} & \textbf{0.34} & \textbf{0.997} & 1.000 \\
 & &  & ISLA & 0.57 & 0.61 & 0.27 & 0.57 & 0.993 & 1.000 \\
\addlinespace[1pt]
\splitname{full} (720) & test & 200 & Transolver & 1.17 & 1.76 & 0.53 & 1.14 & 0.899 & 1.000 \\
 & &  & GeoTransolver & 1.20 & 1.46 & 0.47 & 1.09 & 0.924 & 1.000 \\
 & &  & GLOBE & \textbf{0.34} & \textbf{0.55} & \textbf{0.15} & \textbf{0.28} & \textbf{0.991} & 1.000 \\
 & &  & ISLA & 0.55 & 0.57 & 0.30 & 0.57 & 0.958 & 1.000 \\
\addlinespace[1pt]
\midrule
\splitname{scarce} (200) & validation & 80 & Transolver & 3.46 & 3.04 & 1.24 & 3.91 & 0.100 & 0.999 \\
 & &  & GeoTransolver & 3.18 & 2.80 & 1.11 & 3.46 & 0.594 & 0.999 \\
 & &  & GLOBE & \textbf{1.03} & \textbf{1.04} & \textbf{0.38} & \textbf{1.05} & \textbf{0.896} & 1.000 \\
 & &  & ISLA & 1.55 & 1.71 & 0.76 & 2.13 & 0.743 & 1.000 \\
\addlinespace[1pt]
\splitname{scarce} (200) & test & 200 & Transolver & 2.99 & 2.84 & 1.11 & 3.25 & 0.470 & 0.999 \\
 & &  & GeoTransolver & 2.87 & 2.67 & 1.04 & 2.91 & 0.557 & 0.999 \\
 & &  & GLOBE & \textbf{0.86} & \textbf{0.98} & \textbf{0.32} & \textbf{0.75} & \textbf{0.919} & \textbf{1.000} \\
 & &  & ISLA & 1.54 & 1.70 & 0.76 & 1.96 & 0.509 & 0.999 \\
\addlinespace[1pt]
\midrule
\splitname{aoa_extrapolation} (724) & validation & 80 & Transolver & 0.82 & 1.20 & 0.39 & 0.95 & 0.830 & 1.000 \\
 & &  & GeoTransolver & 0.98 & 1.27 & 0.43 & 1.03 & 0.786 & 1.000 \\
 & &  & GLOBE & \textbf{0.30} & \textbf{0.54} & \textbf{0.12} & \textbf{0.25} & \textbf{0.992} & 1.000 \\
 & &  & ISLA & 0.52 & 0.62 & 0.21 & 0.46 & 0.953 & 1.000 \\
\addlinespace[1pt]
\splitname{aoa_extrapolation} (724) & test & 196 & Transolver & 8.31 & 6.44 & 3.84 & 12.21 & -0.067 & 0.998 \\
 & &  & GeoTransolver & 9.29 & 6.60 & 3.91 & 13.02 & 0.355 & 0.996 \\
 & &  & GLOBE & \textbf{1.94} & \textbf{2.14} & \textbf{0.98} & \textbf{3.87} & \textbf{0.793} & 1.000 \\
 & &  & ISLA & 2.58 & 2.67 & 1.37 & 4.55 & 0.652 & 1.000 \\
\addlinespace[1pt]
\bottomrule
\end{tabular}
\end{table}

\begin{figure*}[!ht]
  \centering
  \includegraphics[width=0.97\textwidth]{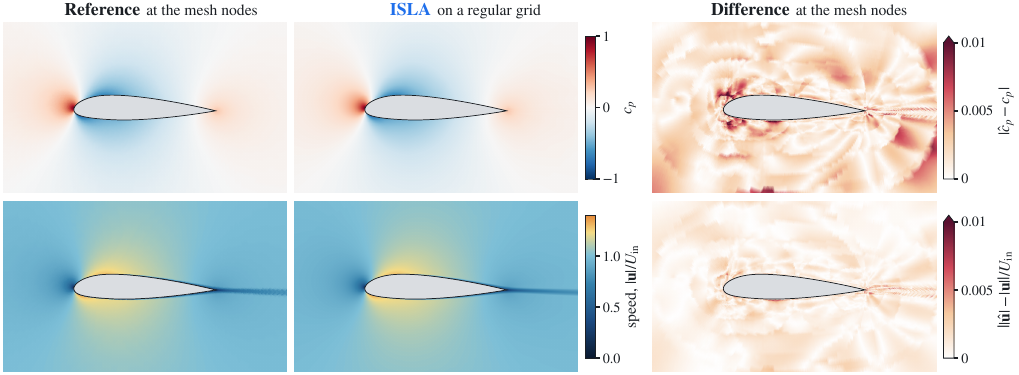}
  \caption{\textbf{An AirfRANS evaluation case.}
  \emph{Top:} pressure coefficient, formed with the inlet dynamic pressure.
  \emph{Bottom:} velocity magnitude over the inlet speed.
  \emph{Left:} the reference, which exists only at mesh nodes, so it is drawn as a triangulation
  of the \isn{app-airfrans-n-nodes} scored nodes of this case and never across the airfoil; the
  nodes are a random subset of the mesh, which makes the thin wake look jagged.
  \emph{Middle:} \isla{}, evaluated on a regular grid of \isn{app-airfrans-grid-nx} by
  \isn{app-airfrans-grid-ny} points.
  \emph{Right:} the difference, taken at the scored nodes, with its own and much finer color
  scale.
  The case has an inlet speed of \isn{app-airfrans-inlet-speed}\,m/s, an angle of attack of
  \isn{app-airfrans-aoa}, and a Reynolds number of \isn{app-airfrans-reynolds}.
  Over its scored nodes, the relative $L^2$ error is \isn{app-airfrans-err-pressure-pct}\% for
  pressure and \isn{app-airfrans-err-velocity-pct}\% for velocity.}
  \label{fig:airfrans}
\end{figure*}

\Cref{fig:airfrans} shows \isla{}'s interior prediction on one AirfRANS evaluation case on a regular
grid in place of the cells of the mesh.

\FloatBarrier
\subsection{Automotive Aerodynamics Datasets}
\label{app:bench-all}

\begin{figure}[h]
  \centering
  \includegraphics[width=0.62\textwidth]{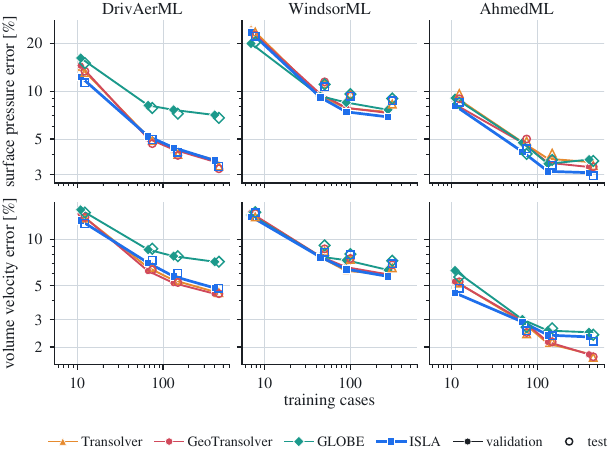}
  \caption{\textbf{Surface-pressure and volume-velocity errors as functions of the number of
  training cases} on the three automotive datasets (relative $L^2$, logarithmic axes).
  Lines and filled markers: validation cases.
  Open markers: test cases.}
  \label{fig:scarce}
\end{figure}

\begin{figure*}[!ht]
  \centering
  \includegraphics[width=\textwidth]{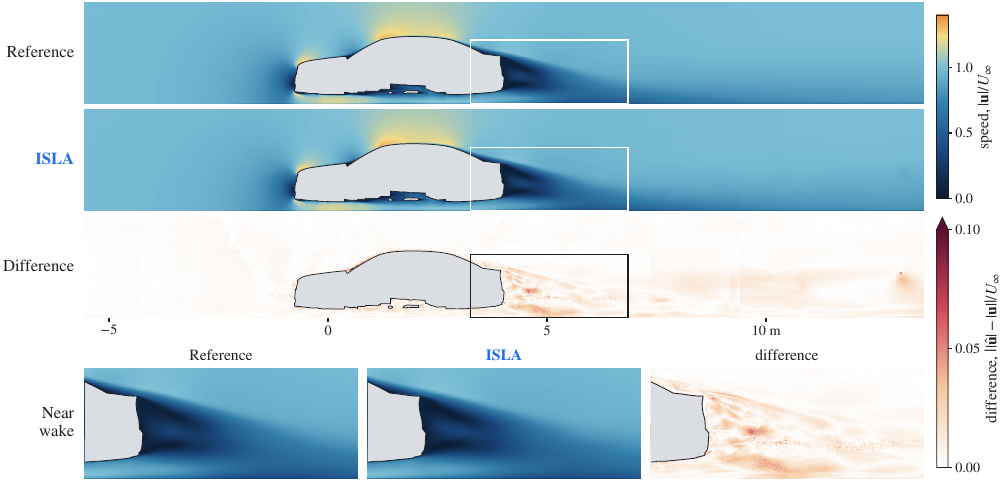}
  \caption{\textbf{An interior field on a plane.}
  \isla{} predicts volume fields from the boundary surfaces alone (the car, the ground, and the
  tunnel), at any requested points.
  Here the points are a regular grid of \isn{app-volume-grid-nx} by \isn{app-volume-grid-nz} on
  the symmetry plane of one DrivAerML evaluation case (case \isn{app-volume-run} of the dataset), in place of
  the cells of its mesh.
  \emph{Rows:} velocity magnitude over the free-stream speed in the reference (the stored cells
  within \isn{app-volume-slab-mm}\,mm of the plane, each grid point showing its nearest cell) and
  from \isla{}, and the magnitude of their difference at the stored cells.
  The car's section is gray, and the ticks give the streamwise coordinate.
  \emph{Bottom:} the same three fields in the near wake (the outlined window).
  Over the \isn{app-volume-n-scored} scored points of this case, the relative $L^2$ error is
  \isn{app-volume-err-velocity-pct}\% for velocity and \isn{app-volume-err-pressure-pct}\% for
  pressure.}
  \label{fig:volume-plane}
\end{figure*}

\begin{figure*}[!ht]
  \centering
  \includegraphics[width=\textwidth]{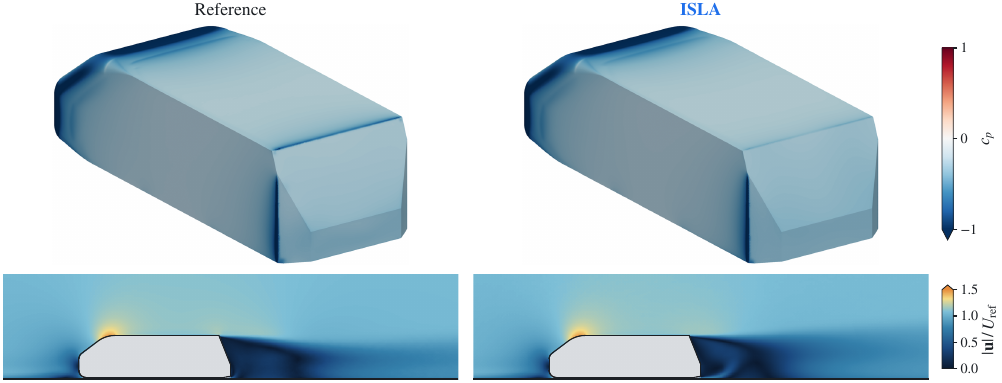}
  \caption{\textbf{WindsorML fields.}
  What the ensemble predicts on a case it was not trained on: case~8 of WindsorML, a
  validation case of the \splitname{full} split.
  \emph{Top:} the surface pressure coefficient $c_p$ on the body, seen from behind.
  \emph{Bottom:} the velocity magnitude over the reference speed $U_{\mathrm{ref}}$ (40~m/s) on
  the body's center plane $y = 0$, with the body's outline; the window is three body lengths
  wide.
  The inflow is yawed by about 2.5~degrees, so this plane is not a symmetry plane of the flow.
  \emph{Left:} the dataset's time-averaged simulation.
  \emph{Right:} the \isla{} ensemble, the equal-weight mean of five surface networks and
  of five volume networks (\cref{app:protocol}).
  \isla{} predicts both fields from the boundary alone.
  The two columns share one color scale in each row.}
  \label{fig:windsorml}
\end{figure*}

\begin{figure*}[!ht]
  \centering
  \includegraphics[width=\textwidth]{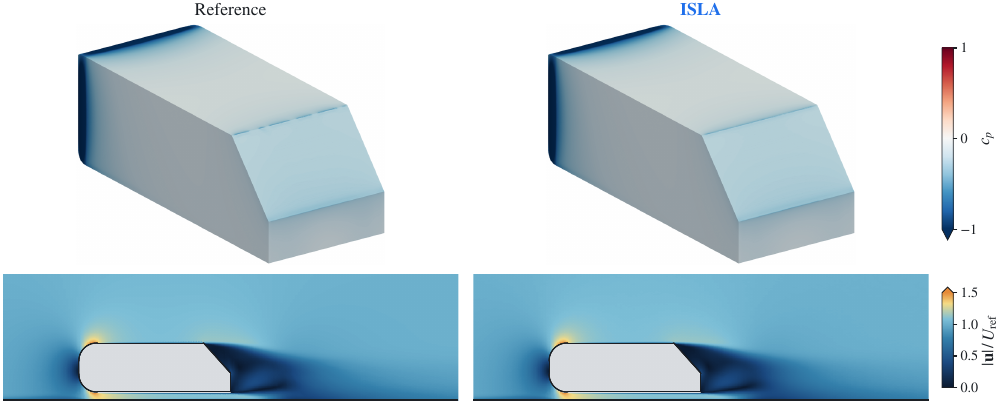}
  \caption{\textbf{AhmedML fields.}
  What the ensemble predicts on a case it was not trained on: case~24 of AhmedML, a
  validation case of the \splitname{full} split.
  \emph{Top:} the surface pressure coefficient $c_p$ on the body, seen from behind, so that the
  slant and the base are in view.
  \emph{Bottom:} the velocity magnitude over the inlet speed $U_{\mathrm{ref}}$ on the center
  plane $y = 0$, with the body's outline; the window is three body lengths wide.
  \emph{Left:} the dataset's time-averaged simulation.
  \emph{Right:} the \isla{} ensemble, the equal-weight mean of three surface networks and
  of eight volume networks (\cref{app:protocol}).
  \isla{} predicts both fields from the boundary alone.
  The two columns share one color scale in each row.}
  \label{fig:ahmedml}
\end{figure*}

\Cref{tab:bench} compares \isla{} with the lowest of the other three families on the main splits of
DrivAerML, WindsorML, and AhmedML (using the splits defined by FluidsBench;
\citealp{fluidsbench2026fluidsbench}), and
\cref{tab:bench-drivaerml,tab:bench-windsorml,tab:bench-ahmedml} give every field error and $R^2$
value for every split.
With WindsorML's \isn{bench-min-train-wml} training cases, \isla{}'s test errors differ from the lowest other's by
\isn{auto-windsorml-super-scarce-pct-min} to \isn{auto-windsorml-super-scarce-pct-max}\%, and
with AhmedML's \isn{bench-min-train-ahm} by \isn{auto-ahmedml-super-scarce-pct-min} to
\isn{auto-ahmedml-super-scarce-pct-max}\%.
Every split of the three datasets, the out-of-distribution splits included, has
test errors for all four families.
\Cref{fig:scarce} plots the surface-pressure and volume-velocity errors as functions of the number
of training cases on the three automotive datasets.
\Cref{fig:volume-plane} shows an interior field of DrivAerML on a plane, and
\cref{fig:windsorml,fig:ahmedml} the fields of the ensembles on WindsorML and AhmedML.

\begin{table}[h]
  \caption{\textbf{Automotive aerodynamics datasets}: \isla{}'s relative $L^2$ error divided by
  the lowest of Transolver's, GeoTransolver's, and \globe{}'s, for each field ($p_s$ and $\tau_w$:
  surface pressure and wall shear stress; $\mathbf{u}$ and $p_v$: velocity and pressure in the
  volume), on test cases (\cref{app:protocol}).
  Rows: the full and the smallest training split of each dataset, and the 67-case
  \splitname{scarce} splits of DrivAerML and AhmedML (\cref{sec:exp-bench}).
  Bold, below 1: \isla{}'s error is the lowest; three decimals where two would print 1.00.
  Every error is in \cref{tab:bench-drivaerml,tab:bench-windsorml,tab:bench-ahmedml}.\isn{bench-interim-note}\isn{bench-retrained-note-auto}}
  \label{tab:bench}
  \centering
  \small
  \setlength{\tabcolsep}{6.5pt}
  \begin{tabular}{@{}llc cccc@{}}
\toprule
 & & & \multicolumn{4}{c}{Rel.\ $L^2$ error: \isla{} $\div$ lowest other $\downarrow$} \\
\cmidrule(l){4-7}
Dataset & Split (train) & Cases & $p_s$ & $\tau_w$ & $\mathbf{u}$ & $p_v$ \\
\midrule
DrivAerML & \splitname{full} (400) & test & 1.03 & 1.05 & 1.08 & 1.08 \\
 & \splitname{scarce} (67) & test & 1.06 & 1.05 & 1.16 & 1.15 \\
 & \splitname{super_scarce} (11) & test & \textbf{0.86} & \textbf{0.92} & \textbf{0.92} & \textbf{0.88} \\
\addlinespace[2pt]
WindsorML & \splitname{full} (276) & test & 1.05 & 1.05 & 1.06 & 1.06 \\
 & \splitname{super_scarce} (7) & test & 1.09 & 1.19 & 1.06 & 1.03 \\
\addlinespace[2pt]
AhmedML & \splitname{full} (400) & test & \textbf{0.88} & \textbf{0.97} & 1.28 & \textbf{0.996} \\
 & \splitname{scarce} (67) & test & 1.07 & 1.06 & 1.05 & \textbf{0.96} \\
 & \splitname{super_scarce} (11) & test & 1.01 & 1.05 & \textbf{0.93} & 1.13 \\
\bottomrule
\end{tabular}
\end{table}

\begin{table}[p]
  \caption{DrivAerML: every split (training cases in parentheses). Columns, bold, and dashes as in
  \cref{tab:bench-airfrans}.\isn{bench-interim-note}\isn{bench-optimistic-note}\isn{bench-retrained-note-drivaerml}}
  \label{tab:bench-drivaerml}
  \centering
  \scriptsize
  \setlength{\tabcolsep}{4pt}
  \begin{tabular}{@{}llr l rrrr rr@{}}
\toprule
 & & & & \multicolumn{4}{c}{Relative $L^2$ error [\%] $\downarrow$} & \multicolumn{2}{c}{Force $R^2$ $\uparrow$} \\
\cmidrule(lr){5-8}\cmidrule(l){9-10}
Split (train) & Cases & $n$ & Model & $p_s$ & $\tau_w$ & $\mathbf{u}$ & $p_v$ & $C_d$ & $C_l$ \\
\midrule
\splitname{full} (400) & validation & 34 & Transolver & 3.72 & 5.92 & 4.56 & 3.21 & 0.990 & 0.992 \\
 & &  & GeoTransolver & \textbf{3.60} & \textbf{5.70} & \textbf{4.43} & \textbf{3.11} & 0.992 & 0.992 \\
 & &  & GLOBE & 7.08 & 8.18 & 7.14 & 5.74 & 0.965 & 0.983 \\
 & &  & ISLA & 3.67 & 5.98 & 4.83 & 3.28 & \textbf{0.993} & \textbf{0.993} \\
\addlinespace[1pt]
\splitname{full} (400) & test & 50 & Transolver & 3.38 & 5.53 & 4.50 & 3.04 & \textbf{0.989} & 0.985 \\
 & &  & GeoTransolver & \textbf{3.25} & \textbf{5.33} & \textbf{4.43} & \textbf{2.99} & 0.988 & 0.985 \\
 & &  & GLOBE & 6.79 & 7.87 & 7.15 & 5.61 & 0.965 & 0.970 \\
 & &  & ISLA & 3.36 & 5.58 & 4.80 & 3.23 & 0.988 & 0.982 \\
\addlinespace[1pt]
\midrule
\splitname{medium} (133) & validation & 34 & Transolver & 4.31 & 6.78 & 5.42 & 3.98 & 0.982 & 0.982 \\
 & &  & GeoTransolver & \textbf{4.30} & \textbf{6.77} & \textbf{5.19} & \textbf{3.79} & \textbf{0.984} & 0.981 \\
 & &  & GLOBE & 7.62 & 9.18 & 7.76 & 6.18 & 0.959 & 0.962 \\
 & &  & ISLA & 4.41 & 7.03 & 5.74 & 4.10 & 0.980 & 0.982 \\
\addlinespace[1pt]
\splitname{medium} (133) & test & 50 & Transolver & 4.01 & 6.46 & 5.41 & 3.74 & 0.982 & 0.968 \\
 & &  & GeoTransolver & \textbf{3.95} & \textbf{6.38} & \textbf{5.24} & \textbf{3.57} & 0.982 & 0.968 \\
 & &  & GLOBE & 7.21 & 8.60 & 7.76 & 5.95 & 0.964 & 0.956 \\
 & &  & ISLA & 4.12 & 6.63 & 6.00 & 4.11 & 0.980 & 0.967 \\
\addlinespace[1pt]
\midrule
\splitname{scarce} (67) & validation & 34 & Transolver & 5.16 & 7.85 & 6.55 & 4.96 & \textbf{0.969} & 0.948 \\
 & &  & GeoTransolver & \textbf{5.03} & \textbf{7.66} & \textbf{6.22} & \textbf{4.64} & 0.962 & \textbf{0.957} \\
 & &  & GLOBE & 8.15 & 10.17 & 8.56 & 6.83 & 0.918 & 0.933 \\
 & &  & ISLA & 5.21 & 7.98 & 7.03 & 5.18 & 0.955 & 0.945 \\
\addlinespace[1pt]
\splitname{scarce} (67) & test & 50 & Transolver & 4.88 & 7.47 & 6.58 & 4.73 & 0.971 & 0.918 \\
 & &  & GeoTransolver & \textbf{4.68} & \textbf{7.29} & \textbf{6.34} & \textbf{4.44} & \textbf{0.973} & \textbf{0.934} \\
 & &  & GLOBE & 7.84 & 9.66 & 8.68 & 6.65 & 0.937 & 0.918 \\
 & &  & ISLA & 4.97 & 7.66 & 7.33 & 5.09 & 0.968 & 0.917 \\
\addlinespace[1pt]
\midrule
\splitname{super_scarce} (11) & validation & 34 & Transolver & 14.11 & 16.77 & 14.51 & 12.62 & 0.500 & 0.188 \\
 & &  & GeoTransolver & 14.47 & 17.38 & 14.59 & 12.86 & 0.546 & 0.334 \\
 & &  & GLOBE & 16.14 & 19.77 & 15.52 & 13.98 & 0.694 & 0.207 \\
 & &  & ISLA & \textbf{12.29} & \textbf{15.44} & \textbf{13.23} & \textbf{11.45} & \textbf{0.743} & \textbf{0.530} \\
\addlinespace[1pt]
\splitname{super_scarce} (11) & test & 50 & Transolver & 13.09 & 15.18 & 13.72 & 11.93 & 0.577 & -0.006 \\
 & &  & GeoTransolver & 13.34 & 15.69 & 13.87 & 12.19 & 0.588 & 0.039 \\
 & &  & GLOBE & 15.00 & 18.21 & 14.90 & 12.86 & \textbf{0.739} & -0.258 \\
 & &  & ISLA & \textbf{11.29} & \textbf{14.03} & \textbf{12.65} & \textbf{10.45} & 0.729 & \textbf{0.186} \\
\addlinespace[1pt]
\midrule
\splitname{geometry} (339) & validation & 48 & Transolver & 3.24 & 5.34 & 4.52 & 3.11 & 0.992 & 0.978 \\
 & &  & GeoTransolver & \textbf{3.14} & \textbf{5.22} & \textbf{4.37} & \textbf{2.99} & \textbf{0.994} & \textbf{0.983} \\
 & &  & GLOBE & 6.62 & 7.71 & 7.07 & 5.56 & 0.982 & 0.969 \\
 & &  & ISLA & 3.30 & 5.55 & 4.91 & 3.31 & 0.992 & 0.980 \\
\addlinespace[1pt]
\splitname{geometry} (339) & test & 97 & Transolver & 4.26 & 6.74 & 5.59 & 3.94 & 0.985 & 0.986 \\
 & &  & GeoTransolver & \textbf{4.07} & \textbf{6.51} & \textbf{5.21} & \textbf{3.64} & 0.987 & \textbf{0.991} \\
 & &  & GLOBE & 7.48 & 9.04 & 7.92 & 6.19 & 0.970 & 0.983 \\
 & &  & ISLA & 4.20 & 6.85 & 6.20 & 4.28 & 0.987 & 0.989 \\
\addlinespace[1pt]
\midrule
\splitname{high_drag} (339) & validation & 48 & Transolver & 3.50 & 5.61 & 4.39 & 3.01 & 0.989 & \textbf{0.988} \\
 & &  & GeoTransolver & \textbf{3.46} & \textbf{5.53} & \textbf{4.33} & \textbf{2.96} & 0.990 & 0.985 \\
 & &  & GLOBE & 6.95 & 7.91 & 7.09 & 5.55 & 0.974 & 0.977 \\
 & &  & ISLA & 3.55 & 5.72 & 4.74 & 3.13 & \textbf{0.991} & 0.986 \\
\addlinespace[1pt]
\splitname{high_drag} (339) & test & 97 & Transolver & 4.27 & 6.71 & 5.46 & 4.01 & 0.961 & 0.964 \\
 & &  & GeoTransolver & \textbf{4.07} & \textbf{6.36} & \textbf{5.19} & \textbf{3.78} & 0.961 & 0.974 \\
 & &  & GLOBE & 7.74 & 9.15 & 7.96 & 6.41 & 0.886 & 0.963 \\
 & &  & ISLA & 4.08 & 6.57 & 5.72 & 4.15 & 0.943 & \textbf{0.975} \\
\addlinespace[1pt]
\midrule
\splitname{low_drag} (339) & validation & 48 & Transolver & 3.53 & 5.75 & 4.57 & 3.09 & 0.990 & \textbf{0.984} \\
 & &  & GeoTransolver & \textbf{3.49} & \textbf{5.69} & \textbf{4.45} & \textbf{2.99} & 0.988 & 0.983 \\
 & &  & GLOBE & 6.82 & 7.89 & 6.95 & 5.54 & 0.972 & 0.969 \\
 & &  & ISLA & 3.57 & 5.90 & 4.93 & 3.34 & \textbf{0.991} & 0.983 \\
\addlinespace[1pt]
\splitname{low_drag} (339) & test & 97 & Transolver & 4.00 & 6.41 & 5.73 & 4.11 & \textbf{0.986} & 0.972 \\
 & &  & GeoTransolver & \textbf{3.88} & \textbf{6.21} & \textbf{5.49} & \textbf{3.91} & 0.981 & \textbf{0.975} \\
 & &  & GLOBE & 7.31 & 8.68 & 8.24 & 6.41 & 0.972 & 0.953 \\
 & &  & ISLA & 4.33 & 7.07 & 7.04 & 4.91 & 0.974 & 0.974 \\
\addlinespace[1pt]
\midrule
\splitname{rear_separation} (339) & validation & 48 & Transolver & 3.58 & 5.63 & 4.55 & 3.23 & 0.991 & 0.983 \\
 & &  & GeoTransolver & \textbf{3.55} & \textbf{5.54} & \textbf{4.41} & \textbf{3.14} & \textbf{0.992} & 0.985 \\
 & &  & GLOBE & 7.11 & 8.15 & 7.13 & 5.66 & 0.967 & 0.968 \\
 & &  & ISLA & 3.69 & 5.88 & 4.92 & 3.46 & 0.991 & \textbf{0.986} \\
\addlinespace[1pt]
\splitname{rear_separation} (339) & test & 97 & Transolver & 3.76 & 5.95 & 5.08 & 3.48 & 0.985 & 0.958 \\
 & &  & GeoTransolver & \textbf{3.65} & \textbf{5.80} & \textbf{4.86} & \textbf{3.35} & 0.988 & \textbf{0.969} \\
 & &  & GLOBE & 7.06 & 8.26 & 7.53 & 5.84 & 0.959 & 0.939 \\
 & &  & ISLA & 3.78 & 6.13 & 5.59 & 3.81 & 0.988 & 0.960 \\
\addlinespace[1pt]
\bottomrule
\end{tabular}
\end{table}

\begin{table}[p]
  \caption{WindsorML: every split (training cases in parentheses; \cref{app:protocol}).
  Columns, bold, and dashes as in \cref{tab:bench-airfrans}.}
  \label{tab:bench-windsorml}
  \centering
  \scriptsize
  \setlength{\tabcolsep}{4pt}
  \begin{tabular}{@{}llr l rrrr rr@{}}
\toprule
 & & & & \multicolumn{4}{c}{Relative $L^2$ error [\%] $\downarrow$} & \multicolumn{2}{c}{Force $R^2$ $\uparrow$} \\
\cmidrule(lr){5-8}\cmidrule(l){9-10}
Split (train) & Cases & $n$ & Model & $p_s$ & $\tau_w$ & $\mathbf{u}$ & $p_v$ & $C_d$ & $C_l$ \\
\midrule
\splitname{full} (276) & validation & 33 & Transolver & 7.40 & \textbf{6.78} & 5.93 & 0.063 & 0.568 & 0.939 \\
 & &  & GeoTransolver & 7.29 & 6.87 & 5.91 & 0.062 & 0.602 & 0.946 \\
 & &  & GLOBE & 7.63 & 6.80 & 6.31 & 0.066 & 0.561 & 0.943 \\
 & &  & ISLA & \textbf{6.89} & 6.86 & \textbf{5.74} & \textbf{0.059} & \textbf{0.629} & \textbf{0.954} \\
\addlinespace[1pt]
\splitname{full} (276) & test & 35 & Transolver & \textbf{8.29} & \textbf{6.56} & \textbf{6.52} & \textbf{0.070} & \textbf{0.870} & \textbf{0.968} \\
 & &  & GeoTransolver & 8.77 & 6.89 & 6.82 & 0.074 & 0.691 & 0.907 \\
 & &  & GLOBE & 8.98 & 6.94 & 7.25 & 0.079 & 0.462 & 0.880 \\
 & &  & ISLA & 8.71 & 6.90 & 6.93 & 0.075 & 0.716 & 0.908 \\
\addlinespace[1pt]
\midrule
\splitname{medium} (90) & validation & 33 & Transolver & 7.81 & \textbf{6.89} & \textbf{6.22} & 0.070 & 0.440 & 0.916 \\
 & &  & GeoTransolver & 7.85 & 7.05 & 6.58 & 0.074 & 0.441 & 0.914 \\
 & &  & GLOBE & 8.48 & 7.14 & 7.31 & 0.081 & 0.311 & 0.902 \\
 & &  & ISLA & \textbf{7.41} & 6.98 & 6.40 & \textbf{0.069} & \textbf{0.504} & \textbf{0.925} \\
\addlinespace[1pt]
\splitname{medium} (90) & test & 35 & Transolver & 9.64 & 7.50 & \textbf{7.41} & 0.082 & \textbf{0.583} & 0.854 \\
 & &  & GeoTransolver & 9.37 & \textbf{7.36} & 7.50 & 0.082 & 0.496 & \textbf{0.874} \\
 & &  & GLOBE & 9.58 & 7.48 & 8.01 & 0.090 & 0.202 & 0.808 \\
 & &  & ISLA & \textbf{9.29} & 7.53 & 7.92 & 0.084 & 0.573 & 0.857 \\
\addlinespace[1pt]
\midrule
\splitname{scarce} (45) & validation & 33 & Transolver & 9.33 & 8.22 & 7.66 & 0.095 & 0.124 & 0.875 \\
 & &  & GeoTransolver & 9.59 & 8.42 & 7.74 & 0.098 & 0.005 & 0.848 \\
 & &  & GLOBE & 9.30 & \textbf{7.83} & 7.70 & 0.089 & \textbf{0.148} & \textbf{0.883} \\
 & &  & ISLA & \textbf{9.13} & 8.25 & 7.66 & \textbf{0.086} & 0.127 & 0.872 \\
\addlinespace[1pt]
\splitname{scarce} (45) & test & 35 & Transolver & 11.17 & 8.95 & 8.35 & 0.100 & \textbf{0.596} & \textbf{0.819} \\
 & &  & GeoTransolver & 11.43 & 9.28 & 8.66 & 0.104 & 0.439 & 0.781 \\
 & &  & GLOBE & 11.03 & \textbf{8.50} & 9.12 & 0.107 & 0.346 & 0.775 \\
 & &  & ISLA & \textbf{10.63} & 8.90 & \textbf{8.24} & \textbf{0.091} & 0.466 & 0.785 \\
\addlinespace[1pt]
\midrule
\splitname{super_scarce} (7) & validation & 33 & Transolver & 24.92 & 17.80 & 14.49 & 0.227 & -5.389 & 0.347 \\
 & &  & GeoTransolver & 23.91 & 17.31 & 14.81 & 0.231 & -9.210 & \textbf{0.487} \\
 & &  & GLOBE & \textbf{19.94} & \textbf{13.89} & 14.37 & \textbf{0.205} & \textbf{-4.967} & -0.015 \\
 & &  & ISLA & 23.35 & 16.93 & \textbf{14.03} & 0.208 & -7.547 & 0.476 \\
\addlinespace[1pt]
\splitname{super_scarce} (7) & test & 35 & Transolver & 23.56 & 16.59 & \textbf{13.94} & 0.215 & -3.645 & 0.342 \\
 & &  & GeoTransolver & 22.45 & 16.06 & 14.70 & 0.224 & -5.187 & \textbf{0.571} \\
 & &  & GLOBE & \textbf{20.30} & \textbf{13.49} & 14.95 & \textbf{0.208} & \textbf{-1.724} & -0.181 \\
 & &  & ISLA & 22.10 & 16.02 & 14.83 & 0.214 & -2.286 & 0.509 \\
\addlinespace[1pt]
\midrule
\splitname{geometry} (239) & validation & 34 & Transolver & \textbf{7.90} & 6.95 & 6.08 & 0.062 & \textbf{0.728} & \textbf{0.923} \\
 & &  & GeoTransolver & 8.26 & 7.27 & 6.35 & 0.068 & 0.386 & 0.852 \\
 & &  & GLOBE & 8.11 & \textbf{6.88} & 6.27 & 0.067 & 0.366 & 0.859 \\
 & &  & ISLA & 8.05 & 7.38 & \textbf{5.75} & \textbf{0.057} & 0.625 & 0.889 \\
\addlinespace[1pt]
\splitname{geometry} (239) & test & 71 & Transolver & 9.98 & \textbf{8.40} & \textbf{7.46} & 0.104 & 0.523 & 0.860 \\
 & &  & GeoTransolver & 10.48 & 8.76 & 8.36 & 0.119 & \textbf{0.570} & \textbf{0.863} \\
 & &  & GLOBE & 10.14 & 8.48 & 8.97 & 0.128 & 0.285 & 0.862 \\
 & &  & ISLA & \textbf{9.85} & 8.59 & 7.67 & \textbf{0.103} & 0.477 & 0.831 \\
\addlinespace[1pt]
\midrule
\splitname{high_drag} (239) & validation & 35 & Transolver & 7.05 & 6.13 & 6.01 & 0.056 & 0.763 & 0.969 \\
 & &  & GeoTransolver & 7.07 & \textbf{6.02} & 6.21 & 0.058 & 0.743 & 0.973 \\
 & &  & GLOBE & 7.30 & 6.36 & 6.10 & 0.057 & 0.696 & 0.941 \\
 & &  & ISLA & \textbf{6.78} & 6.14 & \textbf{5.81} & \textbf{0.053} & \textbf{0.768} & 0.973 \\
\addlinespace[1pt]
\splitname{high_drag} (239) & test & 70 & Transolver & 16.43 & 12.09 & 14.98 & 0.213 & -1.259 & 0.355 \\
 & &  & GeoTransolver & \textbf{15.85} & 11.86 & \textbf{14.84} & \textbf{0.211} & \textbf{-1.072} & \textbf{0.404} \\
 & &  & GLOBE & 16.01 & \textbf{11.83} & 15.40 & 0.221 & -1.179 & 0.401 \\
 & &  & ISLA & 16.28 & 12.12 & 15.71 & 0.227 & -1.164 & 0.352 \\
\addlinespace[1pt]
\midrule
\splitname{low_drag} (242) & validation & 35 & Transolver & 7.73 & 7.20 & \textbf{5.84} & 0.065 & -0.085 & 0.945 \\
 & &  & GeoTransolver & 7.59 & 7.04 & 6.17 & 0.068 & 0.002 & 0.956 \\
 & &  & GLOBE & 7.80 & 7.23 & 6.51 & 0.072 & -0.272 & 0.955 \\
 & &  & ISLA & \textbf{7.25} & \textbf{7.03} & 6.10 & \textbf{0.064} & \textbf{0.095} & \textbf{0.960} \\
\addlinespace[1pt]
\splitname{low_drag} (242) & test & 67 & Transolver & 7.28 & 6.59 & 5.96 & 0.063 & \textbf{0.379} & 0.930 \\
 & &  & GeoTransolver & 7.24 & 6.58 & 5.98 & 0.063 & 0.162 & \textbf{0.937} \\
 & &  & GLOBE & \textbf{7.12} & \textbf{6.39} & 6.10 & 0.063 & -0.316 & 0.933 \\
 & &  & ISLA & 7.21 & 6.60 & \textbf{5.80} & \textbf{0.060} & 0.089 & 0.918 \\
\addlinespace[1pt]
\midrule
\splitname{image_wake} (239) & validation & 34 & Transolver & 8.29 & 7.20 & 6.76 & 0.078 & 0.816 & 0.826 \\
 & &  & GeoTransolver & 8.13 & 7.27 & 6.80 & 0.079 & \textbf{0.899} & \textbf{0.852} \\
 & &  & GLOBE & \textbf{8.06} & \textbf{6.77} & 6.96 & 0.077 & 0.762 & 0.846 \\
 & &  & ISLA & 8.16 & 6.86 & \textbf{6.68} & \textbf{0.074} & 0.664 & 0.790 \\
\addlinespace[1pt]
\splitname{image_wake} (239) & test & 71 & Transolver & 8.30 & 7.36 & 7.37 & 0.066 & -0.092 & 0.677 \\
 & &  & GeoTransolver & 8.31 & 7.36 & 7.41 & 0.065 & -0.165 & 0.703 \\
 & &  & GLOBE & 8.03 & 7.26 & 7.50 & 0.068 & -0.603 & 0.670 \\
 & &  & ISLA & \textbf{8.00} & \textbf{7.25} & \textbf{7.03} & \textbf{0.061} & \textbf{-0.034} & \textbf{0.738} \\
\addlinespace[1pt]
\bottomrule
\end{tabular}
\end{table}

\begin{table}[p]
  \caption{AhmedML: every split (training cases in parentheses). Columns, bold, and dashes as in
  \cref{tab:bench-airfrans}.\isn{bench-retrained-note-ahmedml}}
  \label{tab:bench-ahmedml}
  \centering
  \scriptsize
  \setlength{\tabcolsep}{4pt}
  \begin{tabular}{@{}llr l rrrr rr@{}}
\toprule
 & & & & \multicolumn{4}{c}{Relative $L^2$ error [\%] $\downarrow$} & \multicolumn{2}{c}{Force $R^2$ $\uparrow$} \\
\cmidrule(lr){5-8}\cmidrule(l){9-10}
Split (train) & Cases & $n$ & Model & $p_s$ & $\tau_w$ & $\mathbf{u}$ & $p_v$ & $C_d$ & $C_l$ \\
\midrule
\splitname{full} (400) & validation & 50 & Transolver & 3.59 & 5.57 & 1.80 & 3.25 & 0.961 & 0.951 \\
 & &  & GeoTransolver & 3.34 & 5.36 & \textbf{1.79} & \textbf{3.22} & 0.963 & \textbf{0.975} \\
 & &  & GLOBE & 3.71 & 5.26 & 2.49 & 3.99 & 0.843 & 0.893 \\
 & &  & ISLA & \textbf{3.07} & \textbf{5.15} & 2.32 & 3.55 & \textbf{0.968} & 0.960 \\
\addlinespace[1pt]
\splitname{full} (400) & test & 50 & Transolver & 3.41 & 5.35 & \textbf{1.71} & 3.23 & 0.921 & 0.919 \\
 & &  & GeoTransolver & 3.35 & 5.36 & 1.73 & 3.21 & \textbf{0.932} & \textbf{0.925} \\
 & &  & GLOBE & 3.64 & 5.13 & 2.40 & 3.98 & 0.861 & 0.896 \\
 & &  & ISLA & \textbf{2.95} & \textbf{4.97} & 2.18 & \textbf{3.20} & 0.927 & 0.915 \\
\addlinespace[1pt]
\midrule
\splitname{medium} (133) & validation & 50 & Transolver & 3.76 & 5.56 & \textbf{2.10} & 3.66 & 0.974 & 0.947 \\
 & &  & GeoTransolver & 3.55 & 5.37 & 2.15 & 3.71 & \textbf{0.982} & 0.954 \\
 & &  & GLOBE & 3.50 & \textbf{5.13} & 2.55 & 4.21 & 0.935 & 0.953 \\
 & &  & ISLA & \textbf{3.13} & 5.23 & 2.39 & \textbf{3.54} & 0.976 & \textbf{0.959} \\
\addlinespace[1pt]
\splitname{medium} (133) & test & 50 & Transolver & 4.08 & 5.33 & 2.16 & 4.06 & 0.698 & 0.823 \\
 & &  & GeoTransolver & 3.73 & \textbf{5.13} & 2.16 & 4.00 & 0.775 & 0.869 \\
 & &  & GLOBE & 3.69 & 5.18 & 2.64 & 4.44 & 0.866 & 0.881 \\
 & &  & ISLA & \textbf{3.43} & 5.25 & 2.37 & \textbf{3.57} & \textbf{0.892} & \textbf{0.919} \\
\addlinespace[1pt]
\midrule
\splitname{scarce} (67) & validation & 50 & Transolver & 4.78 & 6.22 & \textbf{2.90} & 5.00 & 0.892 & 0.907 \\
 & &  & GeoTransolver & 4.74 & 6.19 & 2.97 & 5.06 & 0.877 & 0.897 \\
 & &  & GLOBE & 4.74 & \textbf{5.35} & 3.02 & 5.37 & 0.902 & 0.902 \\
 & &  & ISLA & \textbf{4.16} & 5.82 & 2.91 & \textbf{4.86} & \textbf{0.941} & \textbf{0.941} \\
\addlinespace[1pt]
\splitname{scarce} (67) & test & 50 & Transolver & 4.86 & 5.90 & \textbf{2.43} & 4.27 & 0.416 & 0.584 \\
 & &  & GeoTransolver & 5.01 & 5.90 & 2.48 & 4.41 & 0.424 & 0.578 \\
 & &  & GLOBE & \textbf{4.03} & \textbf{5.20} & 2.69 & 4.64 & \textbf{0.841} & \textbf{0.855} \\
 & &  & ISLA & 4.33 & 5.54 & 2.56 & \textbf{4.11} & 0.631 & 0.720 \\
\addlinespace[1pt]
\midrule
\splitname{super_scarce} (11) & validation & 50 & Transolver & 9.20 & 8.19 & 5.35 & 11.01 & 0.560 & 0.754 \\
 & &  & GeoTransolver & 8.25 & 7.73 & 5.31 & 11.46 & 0.694 & 0.756 \\
 & &  & GLOBE & 9.05 & \textbf{7.33} & 6.26 & 11.08 & 0.721 & 0.675 \\
 & &  & ISLA & \textbf{8.10} & 7.61 & \textbf{4.48} & \textbf{9.43} & \textbf{0.753} & \textbf{0.798} \\
\addlinespace[1pt]
\splitname{super_scarce} (11) & test & 50 & Transolver & 9.66 & 9.05 & 5.19 & 10.27 & 0.096 & 0.370 \\
 & &  & GeoTransolver & 8.96 & 8.58 & 5.29 & 10.75 & 0.326 & 0.384 \\
 & &  & GLOBE & \textbf{8.43} & \textbf{7.76} & 5.68 & \textbf{8.68} & 0.532 & 0.499 \\
 & &  & ISLA & 8.49 & 8.12 & \textbf{4.80} & 9.84 & \textbf{0.550} & \textbf{0.604} \\
\addlinespace[1pt]
\midrule
\splitname{geometry} (350) & validation & 50 & Transolver & 3.25 & 5.08 & 1.80 & \textbf{3.42} & \textbf{0.933} & \textbf{0.951} \\
 & &  & GeoTransolver & 3.26 & 5.13 & 1.80 & 3.47 & 0.919 & 0.936 \\
 & &  & GLOBE & 3.96 & 5.10 & 2.58 & 4.48 & 0.727 & 0.814 \\
 & &  & ISLA & \textbf{3.21} & 5.08 & 2.38 & 3.64 & 0.903 & 0.921 \\
\addlinespace[1pt]
\splitname{geometry} (350) & test & 100 & Transolver & 3.92 & 5.50 & 2.18 & 3.69 & 0.938 & 0.868 \\
 & &  & GeoTransolver & 3.88 & 5.44 & \textbf{2.17} & \textbf{3.68} & \textbf{0.941} & 0.876 \\
 & &  & GLOBE & 3.95 & \textbf{5.24} & 2.93 & 4.55 & 0.940 & 0.870 \\
 & &  & ISLA & \textbf{3.46} & 5.29 & 2.60 & 3.74 & 0.921 & \textbf{0.882} \\
\addlinespace[1pt]
\midrule
\splitname{high_drag} (350) & validation & 50 & Transolver & 2.85 & 4.81 & 1.68 & 3.22 & 0.852 & 0.947 \\
 & &  & GeoTransolver & 2.73 & 4.88 & 1.68 & 3.21 & 0.890 & 0.963 \\
 & &  & GLOBE & 2.77 & 4.52 & 2.15 & 3.59 & 0.891 & 0.947 \\
 & &  & ISLA & \na & \na & \na & \na & \na & \na \\
\addlinespace[1pt]
\splitname{high_drag} (350) & test & 100 & Transolver & 8.74 & 7.64 & \textbf{4.71} & \textbf{9.51} & -1.518 & 0.569 \\
 & &  & GeoTransolver & 8.42 & 7.67 & 4.91 & 10.10 & -1.389 & \textbf{0.622} \\
 & &  & GLOBE & 8.52 & \textbf{7.24} & 5.27 & 10.45 & -1.412 & 0.610 \\
 & &  & ISLA & \textbf{8.21} & 7.37 & 5.56 & 10.88 & \textbf{-1.334} & 0.615 \\
\addlinespace[1pt]
\midrule
\splitname{low_drag} (350) & validation & 50 & Transolver & 3.86 & 5.64 & 1.96 & 3.37 & 0.552 & 0.806 \\
 & &  & GeoTransolver & 3.72 & 5.46 & 1.98 & 3.42 & 0.669 & 0.846 \\
 & &  & GLOBE & 3.42 & 4.89 & 2.63 & 4.06 & 0.914 & 0.938 \\
 & &  & ISLA & \na & \na & \na & \na & \na & \na \\
\addlinespace[1pt]
\splitname{low_drag} (350) & test & 100 & Transolver & 3.01 & 4.74 & 1.40 & 2.84 & 0.534 & 0.913 \\
 & &  & GeoTransolver & 2.85 & 4.58 & \textbf{1.36} & \textbf{2.74} & 0.643 & \textbf{0.943} \\
 & &  & GLOBE & 3.30 & 4.58 & 2.26 & 4.31 & 0.308 & 0.884 \\
 & &  & ISLA & \textbf{2.71} & 4.74 & 1.96 & 3.20 & \textbf{0.651} & 0.919 \\
\addlinespace[1pt]
\midrule
\splitname{image_wake} (350) & validation & 50 & Transolver & 2.44 & 4.98 & 1.36 & 2.33 & 0.959 & 0.869 \\
 & &  & GeoTransolver & 2.38 & 4.96 & \textbf{1.33} & 2.27 & 0.959 & 0.880 \\
 & &  & GLOBE & 2.18 & \textbf{4.38} & 1.99 & 2.77 & 0.957 & 0.880 \\
 & &  & ISLA & \textbf{2.04} & 4.81 & 1.89 & \textbf{2.26} & \textbf{0.968} & \textbf{0.899} \\
\addlinespace[1pt]
\splitname{image_wake} (350) & test & 100 & Transolver & 22.10 & 15.02 & 11.42 & 25.87 & -0.552 & -10.153 \\
 & &  & GeoTransolver & 22.01 & 15.08 & 11.07 & 25.22 & \textbf{-0.535} & -9.917 \\
 & &  & GLOBE & \textbf{21.86} & \textbf{15.00} & \textbf{10.51} & \textbf{23.90} & -0.613 & \textbf{-9.607} \\
 & &  & ISLA & 23.15 & 16.22 & 11.90 & 26.58 & -0.543 & -10.476 \\
\addlinespace[1pt]
\bottomrule
\end{tabular}
\end{table}

\FloatBarrier
\paragraph{One network per family.}
\Cref{tab:single} gives the field errors of one network of each family, the first seed of its
entry, beside those of the entry, on validation cases (\cref{tab:lineups} lists network sizes and
schedules).
A single network's errors differ from its entry's by \isn{sn-over-entry-min-pct}\% to
$+$\isn{sn-over-entry-max-pct}\%.

\paragraph{Schedule check.}
On DrivAerML's \isn{bench-min-train-dml}-case split, single networks of Transolver and
GeoTransolver from a 4,000-step recipe are trained again on a schedule \isn{bc-sched-ratio} times
longer (\isn{bc-sched-steps} steps scheduled, as for \isla{}'s networks; stopped at
\isn{bc-stop-steps-min} to \isn{bc-stop-steps-max} steps by that recipe's learning-rate rule, with
their best checkpoints at \isn{bc-best-steps-min} to \isn{bc-best-steps-max}).
Compared with controls trained beside them at the original length, their four field errors on
validation cases fall by \isn{bc-t-cut-min} to \isn{bc-t-cut-max}\% and \isn{bc-gt-cut-min} to
\isn{bc-gt-cut-max}\%, each with a 90\% case-bootstrap interval below 1.
Their errors then lie between \isn{bc-arm-isla-below-pct}\% below and
\isn{bc-arm-isla-above-pct}\% above those of \isla{}'s single network, compared with
\isn{bc-frozen-isla-min-pct} to \isn{bc-frozen-isla-max-pct}\% above at the original length, with
surface pressure still \isn{bc-sp-isla-min-pct} to \isn{bc-sp-isla-max-pct}\% higher (one seed
each; the longer-schedule networks' checkpoints are chosen on these validation cases, and \isla{}
uses its final checkpoint).
The entries of every family train at least \isn{bc-sched-steps} steps per network
(\cref{tab:lineups}).

\begin{table*}[h]
  \caption{\textbf{One network per family} beside the family's entry: relative $L^2$ errors in
  percent on validation cases, for the splits where one network of every model's ensemble is
  evaluated (the first seed, at its final checkpoint).
  GLOBE's single network is not shown: it was predicted in chunks of neighboring cells, and its
  entry in random chunks, as in training.\isn{bench-interim-note}}
  \label{tab:single}
  \centering
  \small
  \setlength{\tabcolsep}{2.9pt}
  \begin{tabular}{@{}lll ccc cccc@{}}
\toprule
 & & & \multicolumn{7}{c}{Relative $L^2$ error [\%] $\downarrow$} \\
\cmidrule(l){4-10}
 & & & \multicolumn{3}{c}{One network} & \multicolumn{4}{c}{Entry (ensemble)} \\
\cmidrule(lr){4-6}\cmidrule(l){7-10}
Dataset & Split (train) & Field & Transolver & GeoTransolver & ISLA & Transolver & GeoTransolver & GLOBE & ISLA \\
\midrule
AhmedML & \splitname{full} (400) & $p_s$ & 3.72 & 3.43 & 3.27 & 3.59 & 3.34 & 3.71 & 3.07 \\
 &  & $\tau_w$ & 5.76 & 5.56 & 5.35 & 5.57 & 5.36 & 5.26 & 5.15 \\
 &  & $\mathbf{u}$ & 1.78 & 1.88 & 2.50 & 1.80 & 1.79 & 2.49 & 2.32 \\
 &  & $p_v$ & 3.20 & 3.42 & 3.87 & 3.25 & 3.22 & 3.99 & 3.55 \\
\bottomrule
\end{tabular}
\end{table*}

\FloatBarrier
\subsection{Pose and Re-sampling Tests: Every Field}
\label{app:encoding-results}

\begin{figure}[h]
  \centering
  \includegraphics{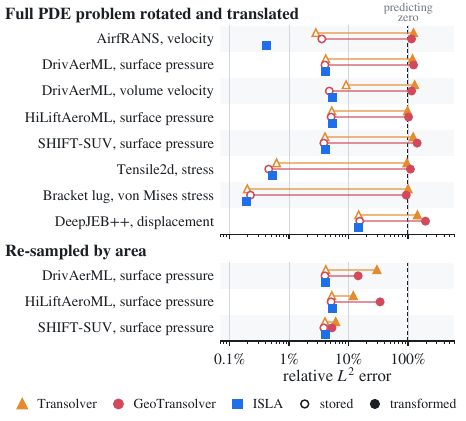}
  \caption{\textbf{Changing how a problem is presented raises the baselines' errors and leaves
  \isla{}'s unchanged.}
  Each model's error on the evaluation cases as stored (open markers) and on the same cases
  rotated and translated (top) or re-sampled by area (bottom); means over 2 to 4 seeds
  (\cref{app:encoding-setup}).
  The pipelines of the automotive, wing, and DeepJEB++ datasets center each case, so there the top
  panel tests rotation alone.}
  \label{fig:encoding}
\end{figure}

\Cref{tab:encoding} gives the numbers behind \cref{fig:encoding}, with every field of every
task.
A rotated case lies outside the training set of a model that takes coordinates, and the model's
error on it is then between \isn{inv-posed-abs-all-min} and \isn{inv-posed-abs-all-max} times that of
predicting zero.
The ratio in \cref{tab:encoding} is therefore of the order of one over the model's error as
stored, and it is largest on the tasks that are fitted best (the bracket lug, to 0.2\%).

\makeatletter
\ifnum\csname isn@encb-available\endcsname=1
\Cref{tab:encoding-bench} repeats the DrivAerML surface tests with one network from each model's
DrivAerML \splitname{full} ensemble, on its validation cases.

\paragraph{Cause of the rise.}
A rotation changes two inputs of Transolver and GeoTransolver at once: the geometry and the
free-stream velocity, which is the same in every training case.
The geometry causes the rise: with the free-stream input left as stored, the pressure error is
still \isn{encb-geomonly-ratio-min} to \isn{encb-geomonly-ratio-max} times that on the stored
cases.

\paragraph{Canonical frames.}
Moving each rotated case at inference into the vehicle's area-weighted principal-axis frame (signs
fixed by third central moments and right-handedness) lowers the pressure ratio only to
\isn{canon-raw-t-ratio} and \isn{canon-raw-gt-ratio}, because a car's principal axes tilt by
about \isn{canon-tilt-deg}$^\circ$ from the stored frame.
Adding one fixed alignment rotation fitted on the training cases reduces the residual rise to
\isn{canon-aligned-t-p-pct} and \isn{canon-aligned-gt-p-pct}\% in pressure and
\isn{canon-aligned-tw-min-pct} to \isn{canon-aligned-tw-max-pct}\% in wall shear (inference
only; a canonical frame does not help with re-sampling, units, query sets, or symmetric bodies).

\paragraph{Re-sampling.}
Re-sampling by area multiplies their pressure errors by \isn{encb-area-ratio-min} to
\isn{encb-area-ratio-max}, while as many cells drawn uniformly leave them unchanged, so the rise
comes from how the cells are drawn and not from how many there are.

\paragraph{GLOBE.}
\globe{} is equivariant by construction, yet the errors of its network rise by a factor of
\isn{encb-globe-rot-ratio-min} to \isn{encb-globe-rot-ratio-max} under rotation and
\isn{encb-globe-area-ratio-min} to \isn{encb-globe-area-ratio-max} under re-sampling.
We have not identified the cause of either rise.
A separate comparison trains two \globe{} and two \isla{} networks for 54,000 steps on the
DrivAerML surface task of \cref{app:encoding-setup} (200,000 cells, cosine schedule) and scores
them on its 48 evaluation cases.
\globe{}'s pressure and wall-shear errors are \isn{globe-drs-nat-p-pct} and
\isn{globe-drs-nat-tw-pct}\% on the stored cases, \isn{globe-drs-nat-p-ratio} and
\isn{globe-drs-nat-tw-ratio} times \isla{}'s.
Rotation and translation multiply them by \isn{globe-drs-posed-over-nat-p} and
\isn{globe-drs-posed-over-nat-tw}, to \isn{globe-drs-posed-p-ratio} and
\isn{globe-drs-posed-tw-ratio} times \isla{}'s (a few rotated cases fail in pressure), and under
re-sampling by area they are \isn{globe-drs-area-p-ratio} and \isn{globe-drs-area-tw-ratio}
times \isla{}'s.

\paragraph{Vertical-axis augmentation.}
The bottom block of \cref{tab:encoding-bench} retrains one surface network each of Transolver
and GeoTransolver with the pipeline's vertical-axis rotation and $\pm 1\,$m horizontal
translation enabled (one seed; one random pose per four-GPU step; stopped near epoch 300 by the
learning-rate rule).
That augmentation removes the rise under vertical-axis rotation (pressure ratio at most
\isn{encb-aug-z-ratio-max}), leaves arbitrary rotation at \isn{encb-aug-so3-ratio} times the
stored error and re-sampling by area at \isn{encb-aug-area-ratio} times the stored error on the
same sample, and multiplies stored-case errors by \isn{encb-aug-stored-t-ratio} and
\isn{encb-aug-stored-gt-ratio} in pressure (\isn{encb-aug-stored-tw-t-ratio} and
\isn{encb-aug-stored-tw-gt-ratio} in wall shear).
\fi
\makeatother

\makeatletter
\ifnum\csname isn@aug-available\endcsname=1
\Cref{tab:encoding-aug} adds Transolver and GeoTransolver trained on rotations drawn uniformly
over all orientations, as in the rotated test, with the pipelines and schedules of the
unaugmented seeds (\isn{aug-t-n-seeds} seeds per augmented model).
Their surface-pressure errors change by \isn{aug-t-rot-change-pct} and
\isn{aug-gt-rot-change-pct}\% under rotation, and are \isn{aug-t-native-excess-pct} and
\isn{aug-gt-native-excess-pct}\% higher on the stored cases than without augmentation.
On the rotated cases they are \isn{aug-t-vs-isla-ratio} times \isla{}'s (90\% interval
\isn{aug-t-vs-isla-ci-lo} to \isn{aug-t-vs-isla-ci-hi}) and \isn{aug-gt-vs-isla-ratio} times
(\isn{aug-gt-vs-isla-ci-lo} to \isn{aug-gt-vs-isla-ci-hi}).
\fi
\makeatother

\begin{table}[h]
  \caption{\textbf{Rotation multiplies the baselines' errors by \isn{inv-posed-min-all} to
  \isn{inv-posed-max-all} and \isla{}'s by at most \isn{inv-isla-posed-max}.}
  Pose and re-sampling tests on every task and field: each model's relative $L^2$ error in percent
  on the cases as stored, and its error on the rotated and translated cases and on the cases
  re-sampled by area, divided by that error (seed means, float32; \cref{app:encoding-setup}).
  Re-sampling by area applies only to the three tasks whose surfaces have more cells than a model
  receives (dash elsewhere).
  Trans.: Transolver; GeoT.: GeoTransolver.}
  \label{tab:encoding}
  \centering
  \small
  \setlength{\tabcolsep}{3pt}
  \begin{tabular}{@{}ll rrr rrr rrr@{}}
\toprule
 & & \multicolumn{3}{c}{Rel.\ $L^2$ error, stored [\%] $\downarrow$} & \multicolumn{3}{c}{Rotated, translated $\div$ stored $\downarrow$} & \multicolumn{3}{c}{Re-sampled by area $\div$ stored $\downarrow$} \\
\cmidrule(lr){3-5}\cmidrule(lr){6-8}\cmidrule(l){9-11}
Task & Field & Trans. & GeoT. & \isla{} & Trans. & GeoT. & \isla{} & Trans. & GeoT. & \isla{} \\
\midrule
AirfRANS & pressure & 0.888 & 0.979 & 0.831 & 134 & 158 & 1.004 & \na & \na & \na \\
 & velocity & 2.82 & 3.55 & 0.411 & 44 & 32 & 1.004 & \na & \na & \na \\
 & eddy viscosity & 1.70 & 1.68 & 1.58 & 87 & 86 & 1.000 & \na & \na & \na \\
\addlinespace[1pt]
DrivAerML surface & pressure & 4.13 & 4.00 & 4.02 & 29 & 31 & 1.000 & 7.2 & 3.6 & 1.005 \\
 & wall shear & 6.67 & 6.55 & 6.48 & 21 & 20 & 1.000 & 5.6 & 3.3 & 1.010 \\
\addlinespace[1pt]
DrivAerML volume & pressure & 4.68 & 3.49 & 3.82 & 17 & 24 & 1.000 & \na & \na & \na \\
 & velocity & 9.09 & 4.74 & 5.27 & 14 & 24 & 1.000 & \na & \na & \na \\
 & eddy viscosity & 8.34 & 8.07 & 7.76 & 18 & 24 & 1.001 & \na & \na & \na \\
\addlinespace[1pt]
HiLiftAeroML surface & pressure & 5.22 & 5.08 & 5.39 & 19 & 20 & 1.009 & 2.3 & 6.7 & 1.002 \\
 & wall shear & 8.73 & 8.55 & 8.83 & 15 & 15 & 1.006 & 2.4 & 4.8 & 1.002 \\
\addlinespace[1pt]
SHIFT-SUV surface & pressure & 4.01 & 3.83 & 4.03 & 30 & 37 & 1.000 & 1.51 & 1.37 & 0.999 \\
 & wall shear & 6.26 & 6.16 & 6.28 & 18 & 19 & 1.000 & 1.39 & 1.23 & 1.000 \\
\addlinespace[1pt]
Tensile2d & stress & 0.615 & 0.454 & 0.517 & 155 & 241 & 1.004 & \na & \na & \na \\
 & displacement & 0.460 & 0.327 & 0.412 & 262 & 402 & 1.003 & \na & \na & \na \\
\addlinespace[1pt]
Bracket lug & von Mises stress & 0.198 & 0.224 & 0.193 & 504 & 415 & 1.005 & \na & \na & \na \\
\addlinespace[1pt]
DeepJEB++ surface & von Mises stress & 39.9 & 41.7 & 37.9 & 3.2 & 4.8 & 1.000 & \na & \na & \na \\
 & displacement & 14.8 & 15.4 & 14.6 & 9.7 & 13 & 1.000 & \na & \na & \na \\
\addlinespace[1pt]
\bottomrule
\end{tabular}
\end{table}

\makeatletter
\ifnum\csname isn@encb-available\endcsname=1
\begin{table}[h]
  \caption{\textbf{On DrivAerML's \isn{encb-n-cases} validation cases, rotation multiplies Transolver's
  and GeoTransolver's pressure errors by \isn{encb-rot-ratio-min} to \isn{encb-rot-ratio-max}, and no
  inference-time frame or vertical-axis augmentation removes the rise.}
  Relative $L^2$ error in percent of one network of each family's DrivAerML \splitname{full} entry
  ($p$: surface pressure; $\boldsymbol{\tau}_w$: wall shear stress); protocol in
  \cref{app:encoding-setup}.
  \isla{}'s network draws its own tokens in proportion to area, so for it the last row of the top
  block is a second draw of the same kind.}
  \label{tab:encoding-bench}
  \centering
  \small
  \setlength{\tabcolsep}{5pt}
  \begin{tabular}{@{}lrrrrrrrr@{}}
\toprule
 & \multicolumn{8}{c}{Relative $L^2$ error [\%] $\downarrow$} \\
\cmidrule(l){2-9}
 & \multicolumn{2}{c}{Transolver} & \multicolumn{2}{c}{GeoTransolver} & \multicolumn{2}{c}{\globe{}} & \multicolumn{2}{c}{\isla{}} \\
\cmidrule(lr){2-3}\cmidrule(lr){4-5}\cmidrule(lr){6-7}\cmidrule(lr){8-9}
Presentation & $p$ & $\boldsymbol{\tau}_w$ & $p$ & $\boldsymbol{\tau}_w$ & $p$ & $\boldsymbol{\tau}_w$ & $p$ & $\boldsymbol{\tau}_w$ \\
\midrule
As stored & 4.05 & 6.20 & 3.92 & 5.93 & 8.87 & 9.67 & 3.91 & 6.40 \\
Rotated about the vertical axis & 105 & 112 & 104 & 125 & 11.0 & 11.5 & 3.91 & 6.41 \\
Rotated on $\mathrm{SO}(3)$ & 110 & 115 & 122 & 130 & 11.1 & 11.8 & 3.91 & 6.42 \\
\quad in principal-axis frame & 23.3 & 31.5 & 21.2 & 28.8 & \na & \na & \na & \na \\
\quad in aligned principal-axis frame & 4.40 & 6.91 & 4.61 & 7.08 & \na & \na & \na & \na \\
\addlinespace[2pt]
As stored, on the sample & 3.91 & 6.15 & 3.78 & 5.88 & 8.77 & 9.64 & 3.78 & 6.36 \\
Re-sampled by area, on the sample & 29.9 & 39.2 & 33.8 & 40.0 & 9.35 & 10.6 & 3.77 & 6.35 \\
\midrule
\multicolumn{9}{@{}l}{\emph{Retrained with vertical-axis rotation and translation (1 seed)}} \\
As stored & 4.59 & 6.67 & 4.82 & 6.84 & \na & \na & \na & \na \\
Rotated about the vertical axis & 4.59 & 6.66 & 4.83 & 6.84 & \na & \na & \na & \na \\
Rotated on $\mathrm{SO}(3)$ & 107 & 98.1 & 113 & 100 & \na & \na & \na & \na \\
\addlinespace[2pt]
As stored, on the sample & 4.45 & 6.62 & 4.69 & 6.79 & \na & \na & \na & \na \\
Re-sampled by area, on the sample & 31.4 & 41.4 & 32.7 & 42.8 & \na & \na & \na & \na \\
\bottomrule
\end{tabular}
\end{table}
\fi
\makeatother

\makeatletter
\ifnum\csname isn@aug-available\endcsname=1
\begin{table}[h]
  \caption{\textbf{Rotation augmentation} on the DrivAerML surface task of the pose and re-sampling tests:
  relative $L^2$ error in percent on the evaluation cases as stored and rotated, and the ratio of
  the two (geometric mean over \isn{aug-plain-n-seeds} seeds, \isn{aug-t-n-seeds} for the models
  trained on rotations).
  The two augmented models are trained on rotations drawn uniformly over all orientations, as in
  the rotated test; the other models are trained on cases as stored.}
  \label{tab:encoding-aug}
  \centering
  \small
  \setlength{\tabcolsep}{4pt}
  \begin{tabular}{@{}l ccc ccc@{}}
\toprule
 & \multicolumn{3}{c}{Surface pressure (rel.\ $L^2$) $\downarrow$} & \multicolumn{3}{c}{Wall shear stress (rel.\ $L^2$) $\downarrow$} \\
\cmidrule(lr){2-4}\cmidrule(l){5-7}
Model & Stored [\%] & Rotated [\%] & Rot.\ $\div$ st. & Stored [\%] & Rotated [\%] & Rot.\ $\div$ st. \\
\midrule
Transolver & 4.13 & 118 & 29 & 6.67 & 141 & 21 \\
Transolver, trained on rotations & 4.32 & 4.32 & 1.001 & 6.99 & 6.99 & 1.001 \\
GeoTransolver & 4.00 & 123 & 31 & 6.55 & 133 & 20 \\
GeoTransolver, trained on rotations & 4.13 & 4.14 & 1.002 & 6.75 & 6.74 & 0.999 \\
\isla{} & 4.02 & 4.02 & 1.000 & 6.48 & 6.48 & 1.000 \\
\bottomrule
\end{tabular}
\end{table}
\fi
\makeatother

\FloatBarrier
\subsection{Measured Invariances}
\label{app:certificates}

\Cref{tab:certificates} gives the relative change
$\lVert \operatorname{back}(M(T x)) - M(x) \rVert / \lVert M(x) \rVert$ of a trained model
$M$'s own float32 prediction when the same problem $x$ is given in another presentation $T x$.
The norm is the measure-weighted $L^2$ norm over the evaluated cells or points, and ``back''
undoes $T$ on the output: vectors are rotated back, rows are put back in order, and units are
converted back.
Entries are medians over \isn{cert-n-cases} evaluation cases of DrivAerML or test cases of
\aerozoo{}, with \isn{cert-n-poses} poses per case for rigid motions.
$T$ acts on a case as the dataset stores it, before the data pipeline; the pipeline centers
each DrivAerML case, so no model receives a pure translation there.
With weights, inputs, and pipeline cast to float64, every exact invariance of
\cref{thm:invariance} holds to at most \isn{cert-f64-surf-max} on the surface task, and query
independence to at most \isn{cert-f64-query-max} in the interior.

\begin{table}[h]
  \caption{\textbf{Measured invariances in float32}: relative change of each trained model's
  own prediction under each change of presentation ($p$: pressure; $\boldsymbol{\tau}_w$: wall
  shear stress; $\mathbf{U}$: velocity; $\nu_t$: eddy viscosity), as medians over \isn{cert-n-cases} cases, where $1$ is
  a change as large as the prediction.
  The last two rows test query independence.
  Dash: not measured.}
  \label{tab:certificates}
  \centering
  \small
  \setlength{\tabcolsep}{3.8pt}
  \begin{tabular}{@{}l cc cc cc@{}}
\toprule
 & \multicolumn{6}{c}{Relative $L^2$ change of prediction $\downarrow$} \\
\cmidrule(l){2-7}
\multicolumn{7}{@{}l}{\emph{Surface task (DrivAerML, float32)}} \\
 & \multicolumn{2}{c}{Transolver} & \multicolumn{2}{c}{GeoTransolver} & \multicolumn{2}{c}{\isla{}} \\
\cmidrule(lr){2-3}\cmidrule(lr){4-5}\cmidrule(l){6-7}
Change of presentation & $p$ & $\boldsymbol{\tau}_w$ & $p$ & $\boldsymbol{\tau}_w$ & $p$ & $\boldsymbol{\tau}_w$ \\
\midrule
Rotation and translation & 1.2 & 1.6 & 1.2 & 1.4 & $6{\times}10^{-4}$ & $10^{-3}$ \\
\quad rotation alone & 1.1 & 1.6 & 1.1 & 1.4 & $4{\times}10^{-4}$ & $10^{-3}$ \\
\quad translation alone & $9{\times}10^{-4}$ & $2{\times}10^{-3}$ & $8{\times}10^{-4}$ & $2{\times}10^{-3}$ & $4{\times}10^{-4}$ & $10^{-3}$ \\
Reflection & 1.2 & 1.5 & 1.1 & 1.3 & $6{\times}10^{-4}$ & $2{\times}10^{-3}$ \\
Lengths in millimeters, all inputs converted & $6{\times}10^{-4}$ & $10^{-3}$ & $6{\times}10^{-4}$ & $10^{-3}$ & $3{\times}10^{-4}$ & $8{\times}10^{-4}$ \\
Size $\times 2$ at the same $\Rey$ & 0.65 & 0.66 & 0.62 & 0.65 & $10^{-6}$ & $3{\times}10^{-6}$ \\
Order of the cells & $5{\times}10^{-7}$ & $7{\times}10^{-7}$ & $8{\times}10^{-7}$ & $10^{-6}$ & $2{\times}10^{-6}$ & $4{\times}10^{-6}$ \\
Every cell split in two coincident halves & $8{\times}10^{-4}$ & $10^{-3}$ & $7{\times}10^{-4}$ & $9{\times}10^{-4}$ & $4{\times}10^{-6}$ & $8{\times}10^{-6}$ \\
Cells drawn by area (on shared cells) & 0.28 & 0.33 & 0.29 & 0.35 & $6{\times}10^{-3}$ & 0.010 \\
\quad redrawn by the same law (on shared cells) & $3{\times}10^{-3}$ & $4{\times}10^{-3}$ & $4{\times}10^{-3}$ & $6{\times}10^{-3}$ & $7{\times}10^{-3}$ & 0.010 \\
Last-bit move of every coordinate & $10^{-3}$ & $3{\times}10^{-3}$ & $10^{-3}$ & $2{\times}10^{-3}$ & $5{\times}10^{-4}$ & $10^{-3}$ \\
\midrule
\multicolumn{7}{@{}l}{\emph{Interior tasks (\isla{}, float32)}} \\
 & \multicolumn{2}{c}{\aerozoo{}} & \multicolumn{3}{c}{DrivAerML, interior} & \\
\cmidrule(lr){2-3}\cmidrule(lr){4-6}
Change of presentation & $\mathbf{U}$ & $p$ & $\mathbf{U}$ & $p$ & $\nu_t$ & \\
\midrule
Rotation and translation & $10^{-3}$ & $2{\times}10^{-3}$ & $6{\times}10^{-3}$ & $4{\times}10^{-3}$ & $2{\times}10^{-3}$ &  \\
\quad with the token draw held fixed & $10^{-4}$ & $3{\times}10^{-4}$ & $10^{-3}$ & $6{\times}10^{-4}$ & $2{\times}10^{-4}$ &  \\
Reflection & \na & \na & $7{\times}10^{-3}$ & $5{\times}10^{-3}$ & $3{\times}10^{-3}$ &  \\
Lengths in millimeters, all inputs converted & $2{\times}10^{-4}$ & $4{\times}10^{-4}$ & $5{\times}10^{-3}$ & $3{\times}10^{-3}$ & $2{\times}10^{-3}$ &  \\
Size $\times 2$ at the same $\Rey$ & $4{\times}10^{-7}$ & $6{\times}10^{-7}$ & $2{\times}10^{-6}$ & $2{\times}10^{-6}$ & $10^{-6}$ &  \\
Order of the cells and of the query points & $6{\times}10^{-7}$ & $2{\times}10^{-6}$ & $10^{-3}$ & $10^{-3}$ & $5{\times}10^{-4}$ &  \\
Every cell split in two coincident halves & $7{\times}10^{-3}$ & 0.023 & 0.014 & 0.011 & $7{\times}10^{-3}$ &  \\
Another token draw of the same budget & 0.010 & 0.025 & 0.014 & 0.011 & $7{\times}10^{-3}$ &  \\
Last-bit move of every coordinate & $8{\times}10^{-4}$ & $10^{-3}$ & $5{\times}10^{-3}$ & 0.022 & $2{\times}10^{-3}$ &  \\
10,000 points alone or among 200,000 & 0 & 0 & 0 & 0 & 0 &  \\
10,000 points in ten chunks & $4{\times}10^{-7}$ & $5{\times}10^{-7}$ & $9{\times}10^{-7}$ & $6{\times}10^{-7}$ & $9{\times}10^{-7}$ &  \\
\bottomrule
\end{tabular}
\end{table}

\paragraph{The surface task.}
In float32 a rigid motion moves \isla{}'s pressure by \isn{cert-dmls-isla-f32-rot-p} and its wall shear
stress by \isn{cert-dmls-isla-f32-rot-wss}.
A move of every coordinate by one unit in its last place, at random up or down, moves them by
as much (\isn{cert-dmls-isla-f32-ulp-p} and \isn{cert-dmls-isla-f32-ulp-wss}), and a rescaling
that is exact in binary arithmetic (doubling) by \isn{cert-dmls-isla-f32-scale-p}.
The float32 rows therefore measure the rounding of the transformed input: where cells of less
than a millimeter lie at coordinates of meters, their normals and areas change in the fourth
digit.
Transolver and GeoTransolver respond to the last-place move by \isn{cert-dmls-t-f32-ulp-p} and
\isn{cert-dmls-gt-f32-ulp-p}.
A conversion of units in which every dimensional input and the dataset's reference length are
converted together leaves the baselines unchanged to the same rounding, because their data
pipeline divides by that reference length.
Doubling the geometry at a fixed Reynolds number, with the dataset's constants left as they
are, changes their predictions by \isn{cert-dmls-t-f32-scale-p} and
\isn{cert-dmls-gt-f32-scale-p}.

When the cells are drawn by another law, the predictions can be compared on the cells that both
draws contain.
Between two draws from the dataset's own law that share half of their cells, the predictions on
the shared cells differ by \isn{cert-dmls-t-f32-redraw-p} for Transolver,
\isn{cert-dmls-gt-f32-redraw-p} for GeoTransolver, and \isn{cert-dmls-isla-f32-redraw-p} for
\isla{}.
No model is exactly consistent between two samples, and \isla{} is somewhat less so than the
baselines.
Between a draw from the dataset's law and a draw by area, the differences are
\isn{cert-dmls-t-f32-area-p}, \isn{cert-dmls-gt-f32-area-p}, and
\isn{cert-dmls-isla-f32-area-p}: \isla{}'s is no larger than between two draws from the
dataset's law.

\paragraph{The interior tasks.}
Query independence is exact: a point predicted alone, or in other chunks, receives the same
value to rounding.
The other rows include the model's own draw of $2\times10^5$ boundary tokens.
On DrivAerML, a rigid motion in float32 changes \isn{cert-dmlv-tokens-changed-rot-pct}\% of the
drawn tokens, through the rounding of the frame in which the cells are ordered, and the
predicted velocity moves by \isn{cert-dmlv-isla-f32-rot-u}; with the draw held at that of the
first presentation it moves by \isn{cert-dmlv-isla-f32-rotfrozen-u}.
A last-bit move of every coordinate moves the predicted pressure by
\isn{cert-dmlv-isla-f32-ulp-p}, more than a rigid motion does: it flips the computed side of the
boundary for about \isn{cert-dmlv-signflips} of the 200,000 query points and turns the direction
to the boundary by more than 0.1~rad at \isn{cert-dmlv-ulp-turned-pct}\% of them.
Splitting every cell changes which tokens are drawn (\isn{cert-dmlv-tokens-changed-split-pct}\%
of them), because the budget is shared among the conditions by cell count, and moves the
velocity by \isn{cert-dmlv-isla-f32-split-u}, in float64 as in float32.
This is the difference between two draws of the same budget
(\isn{cert-dmlv-isla-f32-tokendraw-u}), which is the sampling error of $2\times10^5$ tokens for
$1.7\times10^7$ cells.
With a fixed budget this error does not vanish under refinement, so the interior form as run is
consistent only up to it (\cref{prop:quadrature}).
On \aerozoo{}, whose boundaries are smaller, a rigid motion changes few tokens and moves the
velocity by \isn{cert-az-isla-f32-rot-u}; across all \isn{posetie-checked-val} evaluation cases
(\isn{az-full-n-val} of \aerozoo{} \splitname{full}, 200 of AirfRANS, and 48 of DrivAerML;
\cref{app:encoding-setup}), zero cases hit an exact frame tie in float64, and on
\isn{posetie-az-n-cases} near-degenerate \aerozoo{} validation cases across
\isn{posetie-az-n-poses} float32 poses, mean $\varepsilon_U$ and $\varepsilon_p$ change by at
most \isn{posetie-az-mean-max-pct}\% (median per case \isn{posetie-az-med-pct}\%).

\FloatBarrier
\subsection{Cost and Scaling}
\label{app:cost}

\begin{figure}[h]
  \centering
  \includegraphics[width=\textwidth]{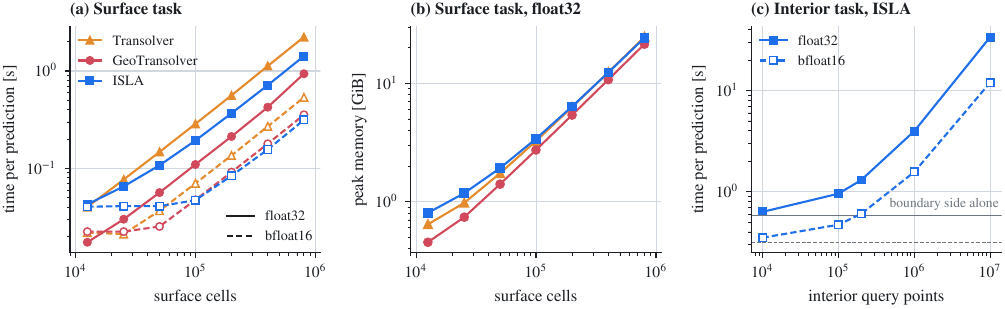}
  \caption{\textbf{Cost of one prediction as a function of problem size}, on one GPU for one
  DrivAerML case.
  \textbf{(a, b)} The surface task, with every model uncompiled.
  \textbf{(c)} \isla{}'s interior task with $2\times10^5$ boundary tokens: the boundary side is a
  fixed cost per case, and the rest is linear in the number of query points.}
  \label{fig:cost}
\end{figure}

\Cref{fig:cost} and \cref{tab:cost-surface,tab:cost-interior} give the cost of a prediction.
Time and memory are linear in the number of surface cells for all three models above about
$5\times10^4$ cells.
Uncompiled, \isla{}'s time lies between those of GeoTransolver and Transolver in float32 and
below both in bfloat16 (\cref{tab:cost-surface,app:encoding-setup}).
For the interior task, the boundary side (the token draw, the 12 blocks, and the tree)
takes \isn{cost-vol-boundary-f32-s}~s in float32 and \isn{cost-vol-boundary-bf16-s}~s in
bfloat16, once per case, and each million query points adds \isn{cost-vol-slope-f32}~s or
\isn{cost-vol-slope-bf16}~s.
Query points are processed in chunks, so memory grows slowly: $10^7$ points take
\isn{cost-vol-10m-f32-gib}~GiB.
In a separate run, $10^8$ points of one DrivAerML case, whose volume mesh has
$1.4\times10^8$ cells, were predicted in \isn{cost-full-volume-min}~minutes at
\isn{cost-full-volume-gib}~GiB, and all \isn{cost-full-az-points-m}~million cells of an
\aerozoo{} case in \isn{cost-full-az-s}~s.
The paragraph ``Training cost'' below gives the cost of training.

\begin{table}[h]
  \caption{Time and memory of one prediction for 200,000 surface cells of a DrivAerML car on one
  GPU (median of five calls; peak allocated memory).
  The baselines are the Transolver of \cref{sec:exp-encoding} and the 9~M width-256 GeoTransolver of \cref{fig:hero} (\cref{app:encoding-setup}), uncompiled, as released.
  \isla{} is shown uncompiled and with each of its blocks compiled, as it is trained and
  evaluated.}
  \label{tab:cost-surface}
  \centering
  \small
  \begin{tabular}{@{}lr rr rr@{}}
\toprule
 & & \multicolumn{2}{c}{Inference, float32} & \multicolumn{2}{c}{Inference, bfloat16} \\
\cmidrule(lr){3-4}\cmidrule(l){5-6}
Model & Parameters & time [s] $\downarrow$ & peak mem.\ [GiB] $\downarrow$ & time [s] $\downarrow$ & peak mem.\ [GiB] $\downarrow$ \\
\midrule
Transolver & 47.2 M & 0.561 & 6.3 & 0.136 & 5.8 \\
GeoTransolver & 9.0 M & 0.213 & 5.4 & 0.092 & 5.1 \\
\isla{} & 66.3 M & 0.368 & 6.4 & 0.084 & 4.0 \\
\isla{}, blocks compiled & 66.3 M & 0.321 & 3.4 & 0.050 & 2.7 \\
\bottomrule
\end{tabular}
\end{table}

\begin{table}[h]
  \caption{Time and memory of \isla{}'s interior prediction for one DrivAerML case on one GPU, by
  the number of query points ($2\times10^5$ boundary tokens drawn from $1.7\times10^7$ boundary
  cells; blocks compiled).}
  \label{tab:cost-interior}
  \centering
  \small
  \begin{tabular}{@{}r rr rr@{}}
\toprule
 & \multicolumn{2}{c}{Inference, float32} & \multicolumn{2}{c}{Inference, bfloat16} \\
\cmidrule(lr){2-3}\cmidrule(l){4-5}
Query points & time [s] $\downarrow$ & peak mem.\ [GiB] $\downarrow$ & time [s] $\downarrow$ & peak mem.\ [GiB] $\downarrow$ \\
\midrule
$10^{4}$ & 0.63 & 16 & 0.35 & 16 \\
$10^{5}$ & 0.96 & 16 & 0.47 & 16 \\
$10^{6}$ & 3.98 & 48 & 1.57 & 31 \\
$10^{7}$ & 33.75 & 50 & 11.99 & 32 \\
\bottomrule
\end{tabular}
\end{table}

\FloatBarrier
\paragraph{Training cost.}
\label{app:compute}
All models were trained on nodes with four \gpuname{}, with one case per GPU per step on the
three-dimensional datasets.
On \aerozoo{} \splitname{full}, the runs of \cref{tab:aerozoo-main} take \isn{az-gpuh-t}
(Transolver + flow axis), \isn{az-gpuh-gt} (GeoTransolver + BC), and \isn{az-gpuh-isla} (\isla{}) GPU-hours per
run (the elapsed time of every job of a run times its four GPUs, which also covers start-up,
validation, and checkpoints); per case and step, GeoTransolver + BC takes a
uniform draw of \isn{az-gt-tokens-per-boundary} faces per boundary condition as tokens, Transolver
none, and \isla{} $2\times10^5$ boundary tokens.
One \isla{} model there takes \isn{train-step-aerozoo}~s per step and \isn{train-mem-aerozoo}~GiB
per GPU.
\Cref{tab:traincost} gives the training cost of single models on AirfRANS and DrivAerML
(\cref{app:encoding-setup}).
Recomputing some activations of the compiled blocks in the backward pass, instead of storing them
as the runs of this paper do, lowers the peak allocated memory per GPU on \aerozoo{} from
\isn{later-mem-old-aerozoo} to \isn{later-mem-new-aerozoo}~GiB over the second and third epochs and
the time per iteration from \isn{later-iter-old-aerozoo} to \isn{later-iter-new-aerozoo}~s; over a
full 250-epoch \aerozoo{} run, it lowers the largest-epoch peak from
\isn{later-fullrun-mem-old-aerozoo} to \isn{later-fullrun-mem-new-aerozoo}~GiB and the median step
time from \isn{later-fullrun-step-old-aerozoo} to \isn{later-fullrun-step-new-aerozoo}~s.
Over epochs 2 and 3, it lowers peak memory from \isn{later-mem-old-airfrans} to
\isn{later-mem-new-airfrans}~GiB on AirfRANS, from \isn{later-mem-old-surface} to
\isn{later-mem-new-surface}~GiB on the DrivAerML surface, and from \isn{later-mem-old-volume} to
\isn{later-mem-new-volume}~GiB on its volume, while iteration time changes by at most
\isn{later-iter-other-max-pct}\% on these tasks.
The joint surface-and-volume DrivAerML networks of \cref{app:protocol} cost more.

\begin{table}[h]
  \caption{Training cost of single models (\cref{app:encoding-setup}; GeoTransolver at width 256):
  time per optimizer step, peak allocated memory per GPU, and GPU-hours per run.
  \isla{}'s blocks are compiled; the baselines run uncompiled, as released.
  Step time and memory are measured on runs of the same models with another learning-rate
  schedule.
  The models differ in size (\cref{app:encoding-setup}).}
  \label{tab:traincost}
  \centering
  \small
  \begin{tabular}{@{}llrrr@{}}
\toprule
Task & Model & Step time [s] $\downarrow$ & Peak memory [GiB] $\downarrow$ & GPU-hours / run $\downarrow$ \\
\midrule
AirfRANS & Transolver & 0.30 & 62 & 36 \\
 & GeoTransolver & 0.21 & 39 & 27 \\
 & \isla{} & 0.32 & 25 & 38 \\
\addlinespace[2pt]
DrivAerML surface & Transolver & 0.27 & 62 & 22 \\
 & GeoTransolver & 0.39 & 78 & 28 \\
 & \isla{} & 0.16 & 56 & 11 \\
\addlinespace[2pt]
DrivAerML volume & Transolver & 0.27 & 62 & 31 \\
 & GeoTransolver & 0.98 & 118 & 67 \\
 & \isla{} & 1.06 & 129 & 71 \\
\bottomrule
\end{tabular}
\end{table}

\FloatBarrier
\subsection{Ablations}
\label{app:ablations}

For \cref{tab:ablation} we remove one component at a time and retrain the reference configuration
with nothing else changed, on the DrivAerML surface task of the pose and re-sampling tests, and on AirfRANS
for the boundary read (\isn{abl-s1-n-seeds} seeds for the first three variants and
\isn{abl-s4-n-seeds} for the last three).
On DrivAerML each ratio divides by the mean of the reference's \isn{abl-ref-n-seeds} seeds, which
lie within \isn{abl-ref-seed-halfrange-pct}\% of it; a fresh run of its first seed differs from
that seed by \isn{abl-s0-pct}\%.

\emph{Counts in place of measures}: within each boundary condition, every cell is given the
condition's mean measure, so that sums over tokens weight cells equally, as a model without
quadrature weights does; the loss and the metric keep the true measures.
The error on cells as stored is \isn{abl-s1-native-change-pct1}\% lower, well inside the spread
of the reference's seeds.
On cells re-sampled by area it is \isn{abl-s1-area-abs-pct}\%, \isn{abl-s1-area} times the
reference's.

\emph{Coordinates as inputs}: the three coordinates of each cell's centroid, centered and
scaled as the baselines' pipeline does, are added to the token's input.
The error on cases as stored is \isn{abl-s2-native-change-pct1}\% lower than the reference's
(\isn{abl-s2-n-seeds} seeds), a difference near the spread of the reference's seeds.
On rotated cases it is \isn{abl-s2-posed-abs-pct}\%.

\emph{No relations to anchors}: the relational invariants of \cref{eq:invariants} are set to
zero in every block, which removes all positional information from the boundary tokens.
The error is \isn{abl-s3-native-abs-pct}\% in every presentation.

\emph{Body alone as the boundary}: the boundaries of the tunnel are removed from the input, so
the model receives the car's surface alone, and its length unit (\cref{eq:length}) becomes that
of the car.
The error is \isn{abl-s4-native-change-pct}\% lower than the reference's in every presentation, and
both seeds of this variant lie within the range of the reference's seeds.
On this task the tunnel's boundaries do not lower the error.

\emph{No boundary read}: the read of \cref{eq:read} is removed from the interior form,
which leaves the slices as a query's only access to the boundary.
On AirfRANS the velocity error is \isn{abl-v1-change-u-pct}\% higher than the reference's and the
pressure error \isn{abl-v1-change-p-pct}\% higher, and the eddy-viscosity error is
\isn{abl-v1-change-nut-pct}\% lower.
The reference's two seeds lie within \isn{abl-af-ref-halfrange-u-pct}\%,
\isn{abl-af-ref-halfrange-p-pct}\%, and \isn{abl-af-ref-halfrange-nut-pct}\% of their mean in
these three fields.
Both seeds of the variant lie outside that spread in velocity (ratios \isn{abl-v1-seed-u-min}
and \isn{abl-v1-seed-u-max}) and in eddy viscosity (\isn{abl-v1-seed-nut-min} and
\isn{abl-v1-seed-nut-max}), but not in pressure (\isn{abl-v1-seed-p-min} and
\isn{abl-v1-seed-p-max}).

\emph{Exact boundary read ($\theta = 0$)}: the AirfRANS interior model is trained and evaluated
with the exact boundary read ($\theta = 0$) in place of the $\theta = 1$ tree (\isn{abl-th0-n-seeds}
seeds).
Its errors differ from the $\theta = 1$ reference's by at most \isn{abl-th0-diff-max-pct}\%
across pressure, velocity, and eddy viscosity (well within the reference's two-seed spread), at
\isn{abl-th0-step-ratio} times the training step time and \isn{abl-th0-mem-gib} vs.\
\isn{abl-th0-ref-mem-gib}~GiB per GPU.

\paragraph{The opening angle at inference.}
The trained weights depend on the tree.
The two AirfRANS models of the pose and re-sampling tests, trained at $\theta = 1$, have
\isn{abl-theta05-ratio-min} to \isn{abl-theta05-ratio-max} times their error when evaluated at
$\theta = 0.5$, and \isn{abl-theta0-ratio-min} to \isn{abl-theta0-ratio-max} times with the
exact boundary read ($\theta = 0$): \isn{abl-theta0-ratio-p} times in pressure, \isn{abl-theta0-ratio-u}
in velocity, and \isn{abl-theta0-ratio-nut} in eddy viscosity.
On \aerozoo{} (\isn{az-tree-n-cases} validation cases of \splitname{full}, $2\times10^5$
points each), the two \isla{} models trained at $\theta = 1$ deviate from their own exact
boundary read ($\theta = 0$) by \isn{az-tree-th1-u-min} and \isn{az-tree-th1-u-max}\% in
velocity and \isn{az-tree-th1-p-min} and \isn{az-tree-th1-p-max}\% in pressure (up to
\isn{az-tree-th1-p-case-max}\% on a single case, and \isn{az-tree-th025-p}\% at $\theta = 0.25$; models trained and evaluated at
$\theta = 0.25$ deviate by \isn{az-tree-trained025-p-min} and
\isn{az-tree-trained025-p-max}\%), while their pressure error relative to the targets moves only
from
\isn{az-tree-target-p-th1-min} and \isn{az-tree-target-p-th1-max}\% at $\theta = 1$ to
\isn{az-tree-target-p-th0-min} and \isn{az-tree-target-p-th0-max}\% at $\theta = 0$.
A model must be evaluated at the angle at which it was trained, and the guarantees of the exact
boundary read do not transfer to a model trained at another angle.

\begin{table}[h]
  \caption{\textbf{Ablations.}
  Error of a variant divided by the error of \isla{} on cases as stored, for cases as stored,
  rotated, and re-sampled by area (seed means; DrivAerML surface pressure, and AirfRANS velocity
  for the last three rows).
  On DrivAerML the reference is the mean of \isn{abl-ref-n-seeds} seeds of \isla{}.
  Dash: no value.}
  \label{tab:ablation}
  \centering
  \small
  \setlength{\tabcolsep}{6pt}
  \begin{tabular}{@{}lccc@{}}
\toprule
 & \multicolumn{3}{c}{Rel.\ $L^2$ error $\div$ \isla{} (stored) $\downarrow$} \\
\cmidrule(l){2-4}
Variant & As stored & Rotated & Re-sampled \\
\midrule
\isla{} & 1.00 & 1.00 & 1.00 \\
Counts in place of measures & 0.99 & 0.99 & 11 \\
Coordinates as inputs & 0.96 & 21 & 0.97 \\
No relations to anchors & 18 & 18 & 18 \\
Body alone as the boundary & 0.98 & 0.98 & 0.98 \\
\midrule
\isla{} (AirfRANS) & 1.00 & 1.00 & \na \\
No boundary read & 1.10 & 1.10 & \na \\
Exact boundary read ($\theta = 0$) & 1.01 & 1.01 & \na \\
\bottomrule
\end{tabular}
\end{table}

\FloatBarrier
\section{Extended Related Work}
\label{app:related}

\paragraph{Transformers for PDEs on general geometry.}
Operator learning on sensors and graphs \citep{lu2021learning,pfaff2021learning} came before
transformers.
Transformers reach industrial meshes by compressing points into a small latent set: supernodes
and anchor tokens \citep{alkin2024universal,alkin2025abupt}, or learned slices
\citep{wu2024transolver,adams2025geotransolver}.
For slices, \citet{wen2026does} read the slice average as a quadrature, and with
\citet{hu2026transolver} show that Transolver does not need its attention among slices.
\isla{} has none.
These models take coordinates as features, and their sums over points or anchors carry no
measure, so their limit follows the density of the points that they are given.
\isla{} keeps slice routing and changes its inputs, which are invariants in place of
coordinates, and its weights, which are surface measures.

\paragraph{Symmetry.}
An invariance that a model lacks must come from training examples, at a cost in data that is
strict for linear models \citep{elesedy2021provably} and known exactly for kernel regression
\citep{tahmasebi2023exact}.
Two classical routes to geometric symmetry \citep{bronstein2021geometric} constrain the layers or
the inputs.
The first constrains every layer through steerable or Clifford representations
\citep{thomas2018tensor,weiler20183d,fuchs2020se3,satorras2021en,geiger2022e3nn,liao2023equiformer,brehmer2023geometric},
and it has been applied to PDEs and flow on surfaces and in volumes
\citep{brandstetter2022geometric,suk2024labgatr,rygiel2026when}.
The closest such model to \isla{} in purpose is AB-GATr \citep{rygiel2026when}, an equivariant
transformer on a subset of the points with read-only queries, which its authors train in two to
eleven times the time of its non-equivariant counterpart.
Even with specialized tensor-product kernels \citep{geiger2022e3nn,nvidia2025cuequivariance},
carrying steerable representations through every layer adds memory and compute at industrial token
counts.
\isla{} takes the second route: standard scalar matrix multiplications run throughout the trunk,
invariant scalars go in, and invariant coefficients times covariant vectors come out at the head,
following the classical representation of isotropic tensor functions
\citep{pope1975more,zheng1994theory,ling2016reynolds,villar2021scalars}.

The invariants are formed relative to anchors, which have two close relatives.
In Invariant Slot Attention \citep{biza2023invariant}, each slot of an image model has a
position and a scale, and inputs are expressed in the slot's frame.
In equivariant neural fields \citep{wessels2025grounding}, pose-conditioned latents are related
to query points through pair invariants of the kind that \citet{bekkers2024fast} use for
message passing.
\isla{} relates tokens to anchors by rotation-invariant scalars, uses the mean normal as the
anchor's orientation, and forms anchors with the surface measure.
No learned map receives a frame, unlike in frame averaging and learned canonicalization
\citep{puny2022frame,kaba2023equivariance}.
A canonical frame cannot be both continuous and equivariant \citep{dym2024equivariant}; \isla{}
uses a principal-axis frame only to order cells for its token draw and its tree
(\cref{app:draw,app:tree}), and a rigid motion then changes its float32 DrivAerML interior
predictions by \isn{cert-dmlv-isla-f32-rot-nut} (eddy viscosity) to \isn{cert-dmlv-isla-f32-rot-u}
(velocity) in relative $L^2$ (\cref{app:certificates}).

Invariants of distances and angles are not a complete set \citep{pozdnyakov2020incompleteness},
so we claim no universality.
Nor do we claim that built-in symmetry always helps: whether it is better than augmentation
depends on the data \citep{brehmer2025does,rygiel2026when}, and \cref{sec:experiments} adds
evidence from both regimes.

\paragraph{Units and similarity.}
Dimensional analysis and similarity scaling \citep{buckingham1914physically,barenblatt1996scaling}
have been built into learning by nondimensional features
\citep{bakarji2022dimensionally,villar2023dimensionless}, and scale equivariance into mesh
networks through the local mesh spacing \citep{horie2024graph}.
\globe{} is invariant to units when the user supplies reference lengths \citep{sharpe2025globe}.
\isla{} instead computes its length unit from the boundary measure, so no reference length is
an input.
That unit, $\sqrt2$ times the radius of gyration of the surface, is the size measure of
statistical shape analysis.

\paragraph{Discretization.}
Neural operators are defined to converge under refinement \citep{kovachki2023neural}, but
analyses and measurements show that the property often fails
\citep{bartolucci2023representation,gao2025discretization,sakarvadia2026false}.
DiffusionNet tests agreement between meshes of one surface \citep{sharp2022diffusionnet}.
For current surrogates, sensitivity to the sampling of the inputs has been observed for anchor
tokens \citep{alkin2025abupt}.
\citet{wen2026does} find that the error of a trained Transolver has a relative standard
deviation of at most 0.18\% over six random subsamples drawn by one law; the tests of
\cref{sec:exp-encoding} draw by another law.

\isla{} makes every sum over the boundary a quadrature, an idea with several precedents.
With surface measures, a weighted softmax takes the self-normalized kernel-regression form of
\citet{nadaraya1964estimating} and \citet{watson1964smooth}; reading unweighted attention as a
quadrature of an integral operator goes back to \citet{cao2021choose}.
The log-measure bias that we use is the proportional attention of token merging
\citep{bolya2023token}.
\citet{yang2026learning} apply it, with surface area as the weight, to a softmax over surface
points into latent tokens, and compare it with a variant without the weight under non-uniform
re-sampling.
We apply the bias to the routing softmax of slice attention with one normalization per boundary
condition, and draw tokens inside the forward pass as a H\'ajek ratio estimator with
Horvitz--Thompson weights from a systematic design
\citep{madow1949theory,horvitz1952generalization,hajek1971comment,stevens2004spatially}.

\paragraph{From the boundary to the field.}
The interior form of \isla{} draws on three lines of work.
The first is boundary integral equations and aerodynamic panel methods: representation formulas
and Steklov--Poincar\'e (Calder\'on) operators for elliptic systems and viscous flows
\citep{pozrikidis1992boundary,mclean2000strongly,steinbach2008numerical,kress2014linear}, panel
and boundary-element discretizations \citep{hess1967calculation,katz2001low,sauter2011boundary},
rigorous sample-complexity results for learning elliptic Green's operators from their
off-diagonal low-rank structure \citep{schafer2021sparse,boulle2022learning,boulle2023elliptic},
and neural boundary-integral surrogates \citep{lin2023bigreennet,wang2024beno}.
\globe{} \citep{sharpe2025globe} superposes learned kernels from boundary faces, with a tree for
distant faces, and \isla{}'s boundary read follows it.
\isla{} adds learned slices for global mixing and exact distances to the boundary for
near-wall resolution, which \globe{} omits by design.
The second is decoding at independent query points, which is standard
\citep{lu2021learning,jaegle2022perceiver}.
Queries that receive a state and contribute to none are used by AB-UPT
\citep{alkin2025abupt} and SMART \citep{hagnberger2026smart}.
\isla{}'s queries receive states that the boundary alone forms, and also attend to every
boundary token.
The third is the hierarchical summary of distant sources, from elliptic-integral treecodes,
$N$-body methods, $\mathcal{H}$-matrices, and parallel Morton BVHs
\citep{rokhlin1985rapid,barnes1986hierarchical,greengard1987fast,hackbusch1999sparse,karras2012maximizing}
to hierarchical transformers \citep{zhdanov2025erwin}.

\end{document}